\documentclass[journal]{IEEEtran}

\usepackage[utf8]{inputenc}
\usepackage[T1]{fontenc}

\usepackage[numbers]{natbib}

\usepackage{amsmath, amssymb, amsfonts}

\usepackage{graphicx}
\usepackage{subcaption}
\usepackage{float}
\usepackage{placeins}

\usepackage[numbers]{natbib}

\usepackage[utf8]{inputenc} % allow utf-8 input
\usepackage[T1]{fontenc}    % use 8-bit T1 fonts

\usepackage{hyperref}       % hyperlinks
\usepackage{url}            % simple URL typesetting
\usepackage{booktabs}       % professional-quality tables
\usepackage{amsfonts}       % blackboard math symbols
\usepackage{nicefrac}       % compact symbols for 1/2, etc.
\usepackage{microtype}      % microtypography

\usepackage{amsmath} 
\usepackage{multirow}
\usepackage{placeins}
\usepackage{amssymb}
\usepackage{float}
\usepackage{color, soul}
\usepackage{enumitem}
\usepackage{graphicx}
\usepackage{subcaption}
\usepackage[T1]{fontenc}
\usepackage{booktabs}
\usepackage{multicol}
\usepackage{subcaption}
\usepackage[dvipsnames,table,xcdraw]{xcolor}
\usepackage[utf8]{inputenc}
\usepackage{tcolorbox}
\newcommand{\red}[1]{\textcolor{black}{#1}}
\newcommand{\blue}[1]{\textcolor{blue}{#1}}
\usepackage{array}
\newcommand{\chec}[1]{\textcolor{brown}{#1}}

\graphicspath{{./Pictures/}}
\newcommand{\rcomment}[1]{\textcolor{blue}{\textbf{\textit{Comment:}} #1}}

\usepackage[normalem]{ulem}
\definecolor{darkgreen}{rgb}{0,.4,0}
\definecolor{darkcyan}{rgb}{0,.4,.4}
\newcommand{\REMOVE}[1]%
{{\color{black}\sout{#1}}}
\newcommand{\ADD}[1]{{\color{blue}{#1}}}
\newcommand{\REPLACE}[2]{\REMOVE{#1}{\color{blue}{#2}}}
\newcommand{\COMMENT}[1]%
{{\color{darkgreen}\textbf{{Editor: }} {#1}}}
\newcommand{\tani}[1]{{\color{green}{#1}}}
\begin{document}
\title{HyperCap: Hyperspectral Land Cover Captioning Dataset for Vision Language Models}

% % The paper headers
% \markboth{IEEE Geoscience and Remote Sensing Magazine, November~2025}%
% {Rachamalla \MakeLowercase{\textit{et al.}}: SM-HAD}

\author{
Aryan Das\IEEEauthorrefmark{1}, 
Tanishq Rachamalla\IEEEauthorrefmark{1}, 
Pravendra Singh, 
Koushik Biswas, 
Vinay Kumar Verma, \\
Salvador Garcia,~%\IEEEmembership{Senior~Member,~IEEE,} 
Antonio Plaza,~\IEEEmembership{Fellow,~IEEE,} and 
Swalpa Kumar Roy,~\IEEEmembership{Senior~Member,~IEEE}
\thanks{The Work funded by Consejería de Economía, Ciencia y Agenda Digital of Junta de Extremadura and the European Regional Development Fund (ERDF) of the European Union, grant GR24035 (ayudas a grupos de investigación, Junta de Extremadura, Fondo Europeo de Desarrollo Regional, GR24035). \textit{(Corresponding authors: Antonio Plaza and Swalpa Kumar Roy.)}}
\thanks{A. Das is with the Department of Computer Science and Engineering, 
Vellore Institute of Technology, Bhopal, Madhya Pradesh 466114, India (e-mail: aryan.das2021@vitbhopal.ac.in).}
\thanks{T. Rachamalla is with the Department of Information Technology, Siddhartha Academy of Higher Education, Vijayawada, Andhra Pradesh 521108, India (e-mail: tanishqrachamalla12@gmail.com).}
\thanks{P. Singh is with the Department of Computer Science and Engineering, 
Indian Institute of Technology, Roorkee, Uttarakhand 247667, India (e-mail: pravendra.singh@cs.iitr.ac.in).}
\thanks{K. Biswas is with the Department of Computer Science and Engineering, Indraprastha Institute of Information Technology Delhi, New Delhi 110020, Delhi, India (e-mail: koushikb@iiitd.ac.in).}
\thanks{V. K. Verma is with the Department of Computer Science and Engineering, Indian Institute of Technology, Kanpur, Uttar Pradesh 208016, India (e-mail: vinayugc@gmail.com).}
\thanks{S. Garcia is with the Department of Computer Science and Artificial Intelligence, University of Granada, Granada 18071, Spain (e-mail: salvagl@decsai.ugr.es).}
\thanks{A. Plaza is with the Hyperspectral Computing Laboratory, Department of Computers and Communications, University of Extremadura, Cáceres E-10003, Spain (e-mail: aplaza@unex.es).}
\thanks{S. K. Roy is with the Department of Computer Science and Engineering, Tezpur University, Assam 784028, India (e-mail: swalpa@tezu.ernet.in).}

\vspace{-10mm}
\thanks{*Equal contribution,}

}

\maketitle

% \begin{center}
% \textit{Accepted for publication in IEEE Geoscience and Remote Sensing Magazine (GRSM), 2025.}
% \end{center}

\vspace{-2mm}

\begin{abstract}
We introduce HyperCap, the first large-scale hyperspectral captioning dataset designed to enhance model performance and effectiveness in remote sensing applications. Unlike traditional hyperspectral imaging (HSI) benchmarks, HyperCap integrates spectral data with pixel-wise textual annotations, enabling deeper semantic understanding. This dataset enhances model performance in tasks like classification and feature extraction, providing a valuable resource for advanced remote sensing applications. HyperCap is constructed from four benchmark datasets and annotated through a hybrid approach combining automated and manual methods to ensure accuracy and consistency. Empirical evaluations using state-of-the-art encoders and diverse fusion techniques demonstrate significant improvements in classification performance. These results underscore the potential of vision-language learning in HSI and position HyperCap as a foundational dataset for future research in the field. The code and dataset are available at \href{https://github.com/arya-domain/HyperCap}{https://github.com/arya-domain/HyperCap}.

\end{abstract}


\section{Introduction} \label{sec:intro}
\begin{figure*}[t]
    % \vspace{-5mm}
    \centering
    \begin{subfigure}{0.49\textwidth}
        \centering
        % \fbox{
        \includegraphics[height=0.80\textwidth, width=0.80\textwidth, trim=105 120 105 105, clip]{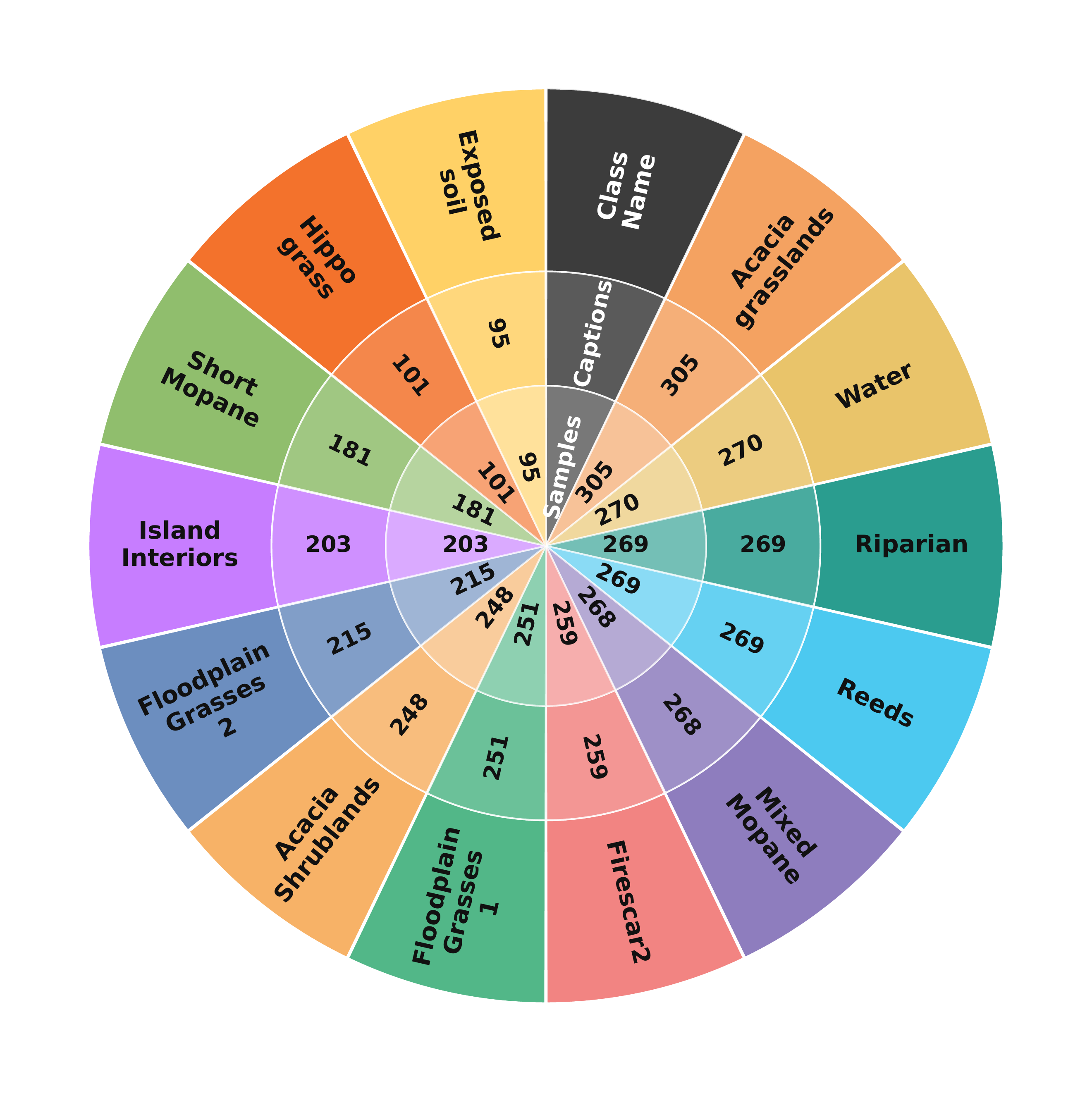}%}
        \caption{}
        \label{subfig:sun_vis_a}
    \end{subfigure}
    % \hfill
    \begin{subfigure}{0.49\textwidth}
        \centering
        \includegraphics[height=0.80\textwidth, width=0.80\textwidth, trim=105 120 105 105, clip]{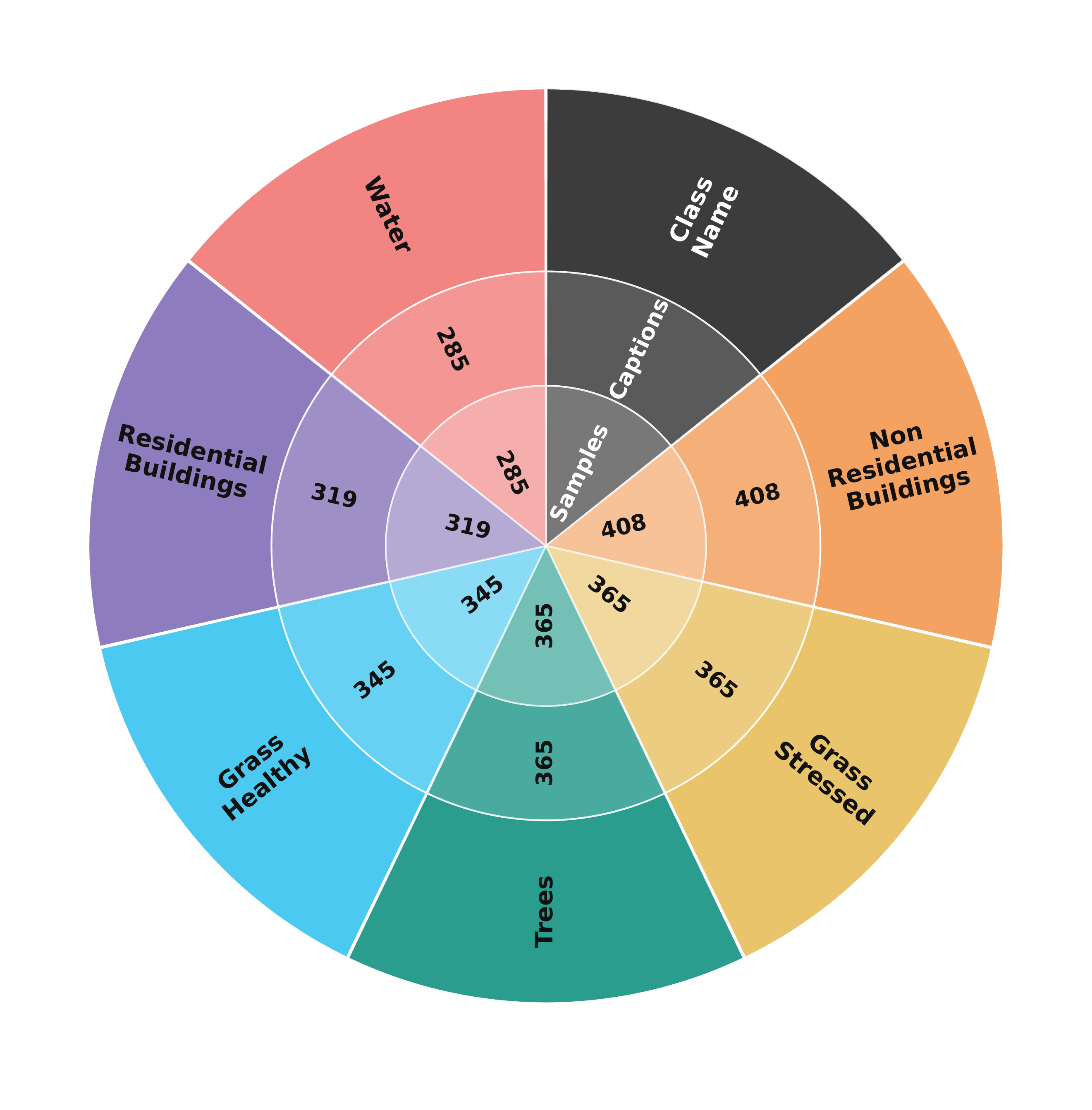}
        \caption{}
        \label{subfig:sun_vis_b}
    \end{subfigure}
    \begin{subfigure}{0.49\textwidth}
        \centering
        \includegraphics[height=0.80\textwidth, width=0.80\textwidth, trim=105 120 105 105, clip]{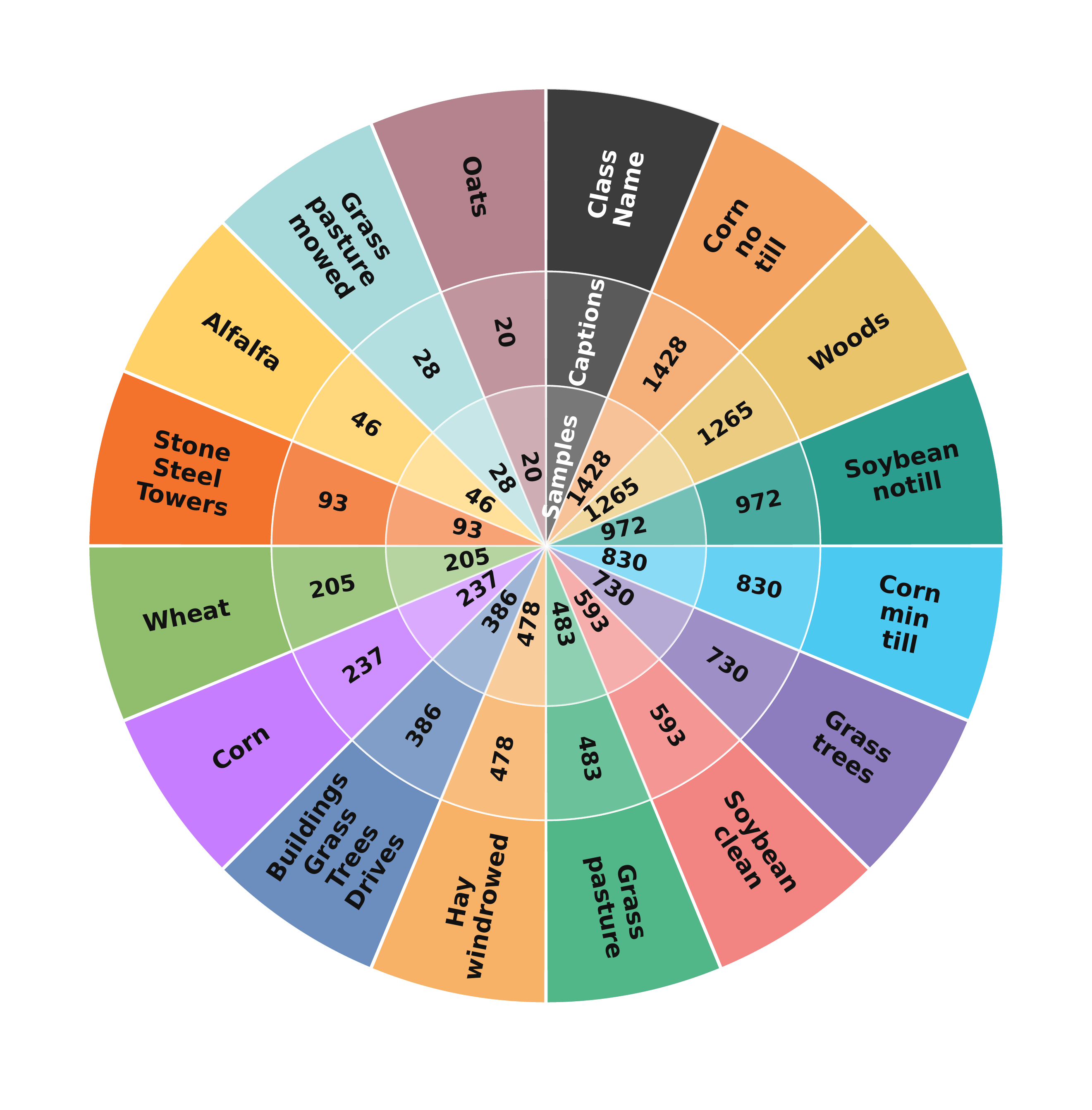}
        \caption{}
        \label{subfig:sun_vis_c}
    \end{subfigure}
    % \hfill
    \begin{subfigure}{0.49\textwidth}
        \centering
        \includegraphics[height=0.80\textwidth, width=0.80\textwidth, trim=105 120 105 105, clip]{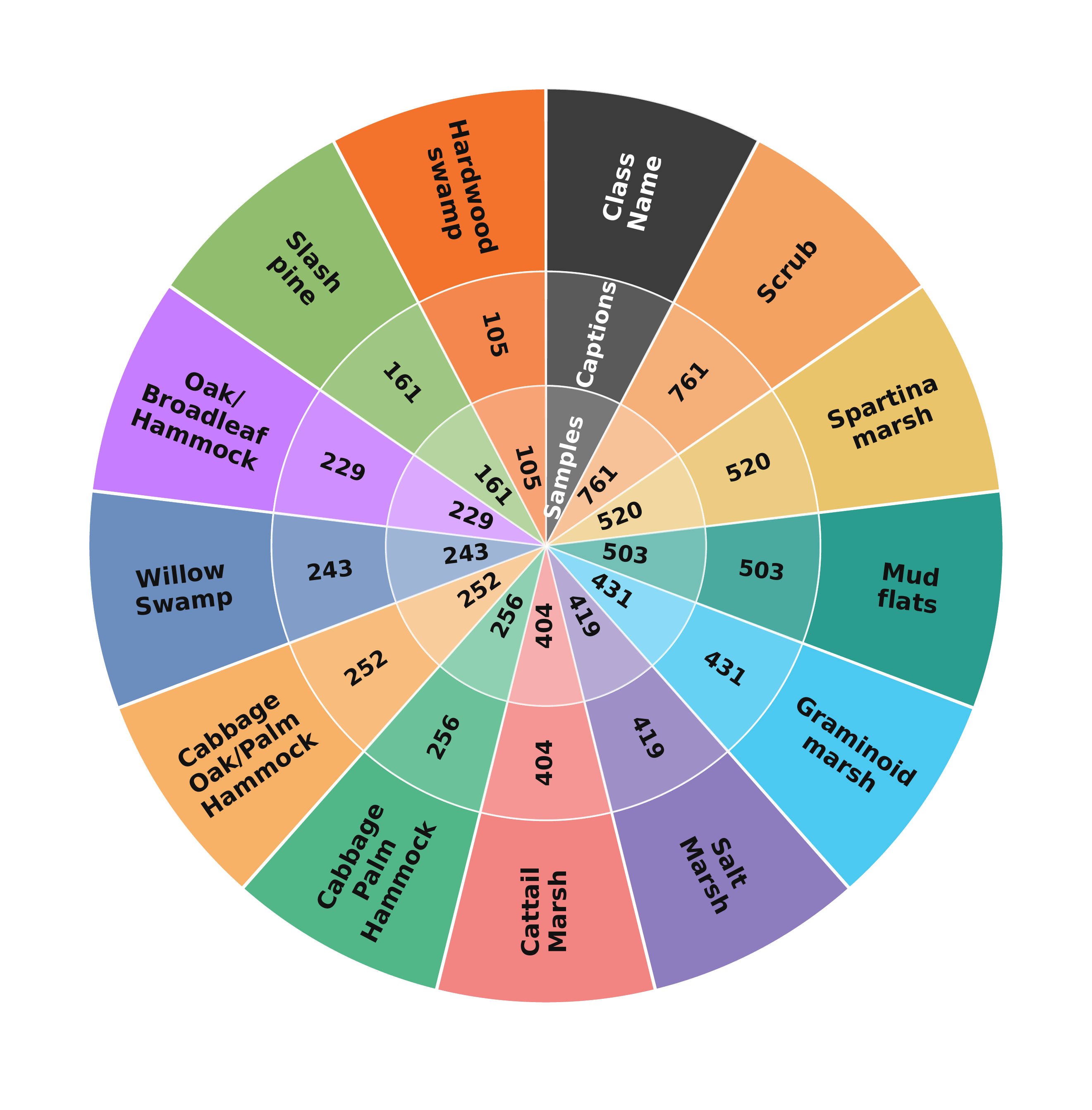}
        \caption{}
        \label{subfig:sun_vis_d}
    \end{subfigure}
    %\vspace{+3mm}
    \caption{Quantitative per-class sample and caption distribution across (a)~Botswana, (b)~Houston13, (c)~Indian Pines, and (d)~Kennedy Space Centre, visualized as pie charts.}
    \label{fig:sunburst_visualizations}
    \vspace{-3mm}
\end{figure*}
%%%%%%%%%%%%
%%%%%%%%%%%%%%%%%%%%%%%%%%%%%%%%%%%%%%%%%%%%%%%%%%%%%%%%%%%%%%%%%%%%%%%%%%%%%%%%%%
\begin{figure*}[h]
    % \vspace{-5mm}
    \centering
    \begin{subfigure}{0.48\textwidth}
        \centering
        \includegraphics[height=0.40\textwidth, width=0.8\textwidth]{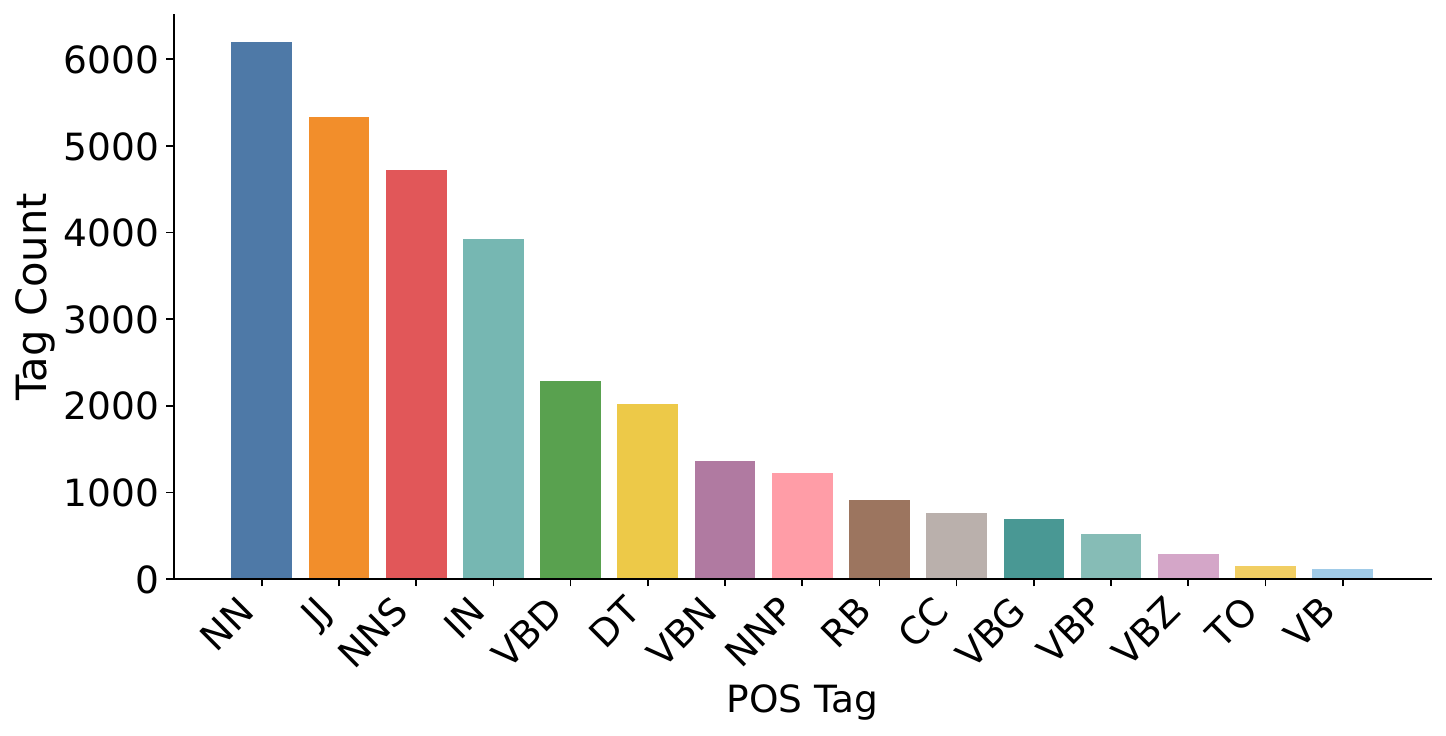}
        \caption{}
        \label{subfig:pos_vis_a}
    \end{subfigure}
    % \hfill
    \begin{subfigure}{0.48\textwidth}
        \centering
        \includegraphics[height=0.40\textwidth, width=0.8\textwidth]{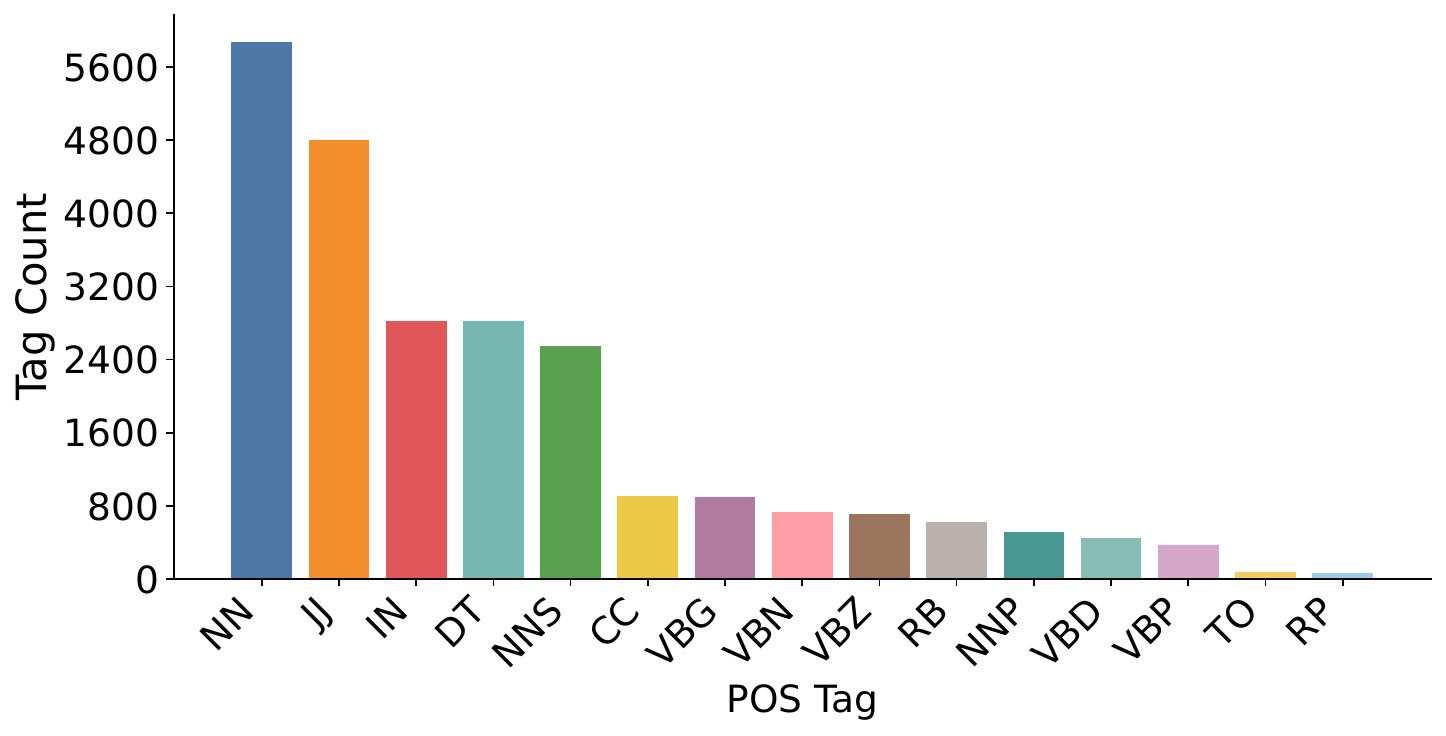}
        \caption{}
        \label{subfig:pos_vis_b}
    \end{subfigure}
    % \hfill
    \begin{subfigure}{0.48\textwidth}
        \centering
        \includegraphics[height=0.40\textwidth, width=0.8\textwidth]{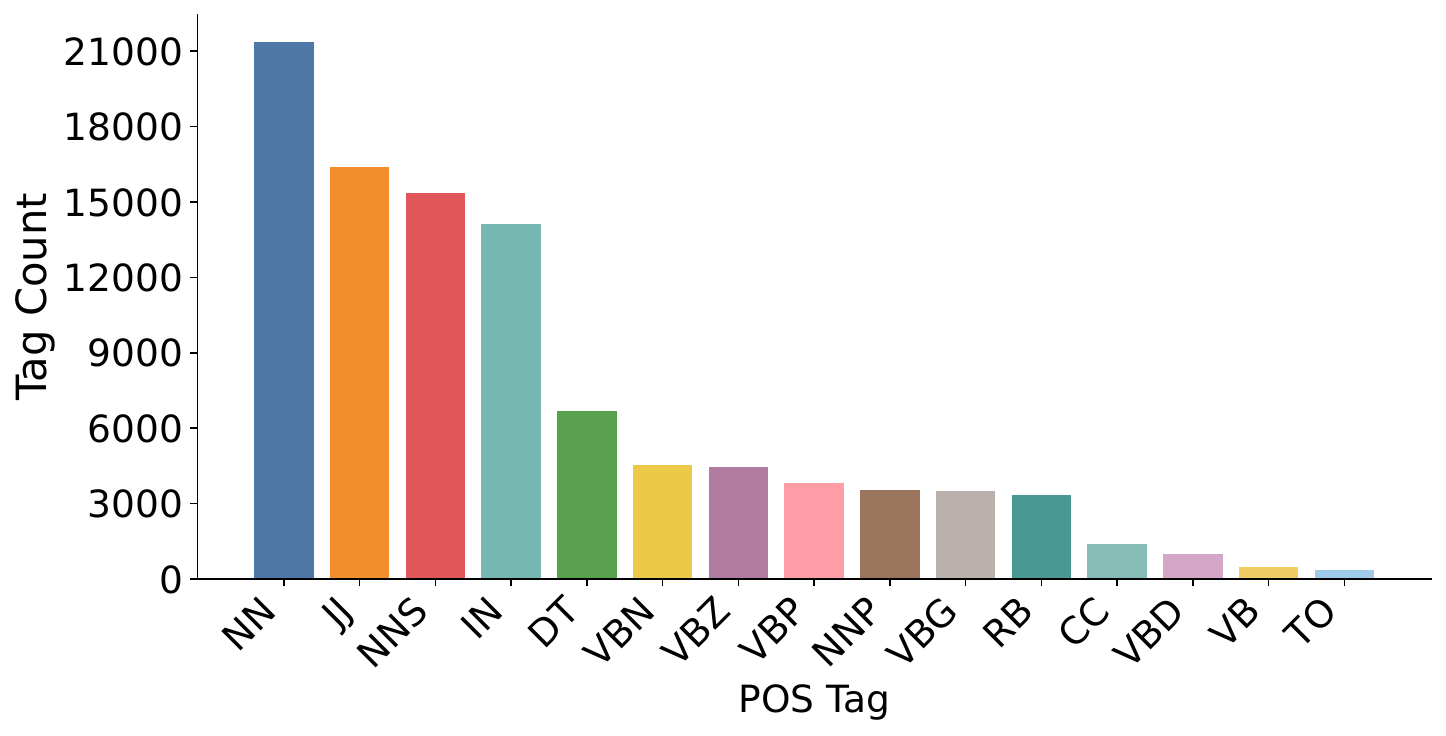}
        \caption{}
        \label{subfig:pos_vis_c}
    \end{subfigure}
    % \hfill
    \begin{subfigure}{0.48\textwidth}
        \centering
        \includegraphics[height=0.40\textwidth, width=0.8\textwidth]{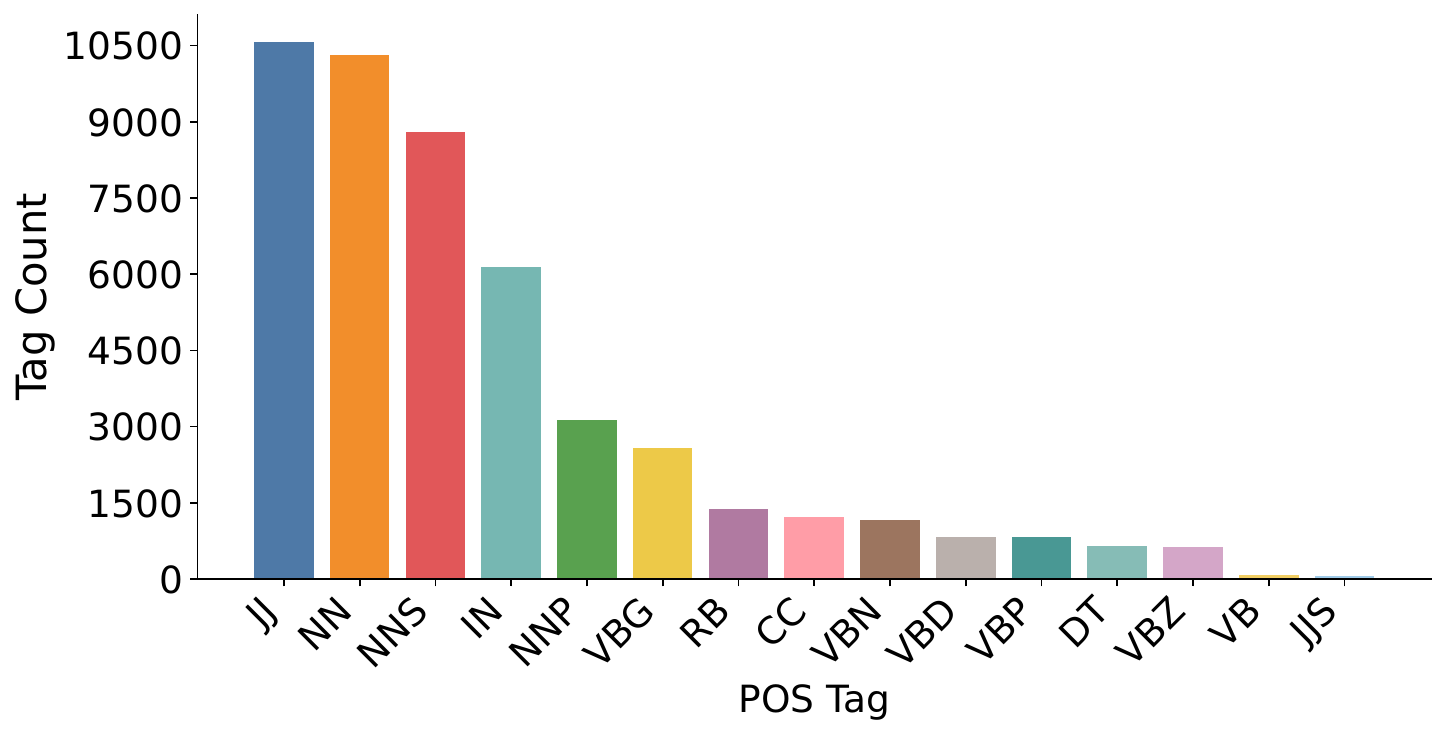}
        \caption{}
        \label{subfig:pos_vis_d}
    \end{subfigure}
    %\vspace{+3mm}
    \caption{Part-of-speech tag frequency distribution in captions across (a)~Botswana, (b)~Houston13, (c)~Indian Pines, and (d)~Kennedy Space Center.}
    \label{fig:pos_visualizations}
    \vspace{-5mm}
\end{figure*}
%%%%%%%%%%%%%%%%%%%%%%%%%%%%%%%%%%%%%%%%%%%%%%%%%%%%%%%%%%%%%%%%%%%%%%%%%%%%%%%%%%%%%%%%%%%%%%

Hyperspectral Imaging (HSI) has evolved as a transformative technology in remote sensing, precision agriculture, environmental monitoring, and medical diagnostics~\cite{1,2,3}. HSI encapsulates reflectance data over hundreds of contiguous wavelengths, unlike conventional imaging methods that record information in a few spectral bands~\cite{4}. Applications include vegetation health assessment, mineral prospecting, and pollution detection, which are highly dependent on this fine-grained spectral resolution, which enables exceptional material discrimination and land cover classification~\cite{5}.
Recent developments in deep learning have greatly improved hyperspectral imagery through Convolutional Neural Networks (CNNs)~\cite{6, roy2020attention}, Generative Adversarial Networks~(GANs)~\cite{zhu2018generative, roy2021generative}, Morphological Convolutional Neural Networks~\cite{roy2021morphological} and Transformer-based architectures especially ~\cite{7}. These models outperform standard machine learning methods using spectral-spatial correlation to achieve state-of-the-art classification performance~\cite{8}. However, key obstacles hinder AI-driven HSI classification: limited semantic understanding, lack of large-scale labelled data, and high computational cost of processing high-dimensional data cubes~\cite{9}.

Despite high accuracy, deep networks lack transparency, raising concerns in high-stakes domains like disaster response, precision agriculture, and urban planning, where expert validation and regulatory compliance demand semantic understanding. The absence of human-interpretable logic hinders trust and real-world adoption. Another key limitation is the scarcity of large-scale labelled hyperspectral datasets. The annotation process is costly and labor-intensive, leading to limited and imbalanced datasets that affect generalization. To address this, researchers explore Self-Supervised Learning (SSL) \cite{10} and Semi-Supervised Learning (Semi-SL) \cite{11} to leverage unlabeled data for representation learning.  However, existing hyperspectral datasets are primarily designed for pixel-wise categorization and lack natural language annotations.
Language awareness in hyperspectral remote sensing is essential for improving semantic understanding, aiding decision-making, and enhancing domain generalization \cite{12}. Unlike conventional domain adaptation, where models access both source and target domains, Domain Generalization (DG) requires learning from labeled source data without exposure to the target domain \cite{Liu_2023_CVPR}. Recent DG approaches, such as adversarial transformation networks and progressive domain expansion, have focused on visual-level domain-invariant representation learning. However, incorporating language into remote sensing has gained traction, enabling tasks like image captioning, classification, and retrieval \cite{Kuckreja_2024_CVPR}. Techniques like topic-sensitive word embedding and recurrent attention mechanisms have been explored for generating meaningful descriptions \cite{9515452}.

Despite significant advancements, HSI classification lacks textual annotations that capture semantic land cover information, limiting its generalization ability \cite{9527083}, while existing HSI captioning datasets remain constrained by limitations in scale, granularity, and annotation diversity \cite{9400386}. In particular, no existing data set fully captures HSI images with detailed captions at the pixel level \cite{8672156}. A detailed analysis of the datasets, including class distribution, linguistic structure of captions, and feature embedding visualizations, is presented in Section~\ref{sec:data_analysis}. Our work addresses the limitations of existing HSI datasets and makes the following contributions:  
\begin{itemize}  
    \item We propose \textbf{HyperCap}, the first large-scale HSI captioning dataset for Remote Sensing, providing fine-grained, pixel-wise textual descriptions for well-known and widely used HSI images.  
    \item Unlike traditional HSI datasets that focus solely on classification, HyperCap combines spectral data with textual annotations. This integration allows models to generate human-readable explanations, thereby enhancing semantic understanding. 
    \item We evaluate the effectiveness of existing methods on HyperCap, establishing a foundation for future research in vision-language learning for HSI imaging.  
\end{itemize}  
\vspace{-1mm}

The rest of the article is organized as follows. Section~\ref{sec:literature} provides a brief overview of the related work. Section~\ref{sec:dataser_prepro} introduces the dataset acquisition and preprocessing approach. Section~\ref{sec:experiments} reports the experimental results and Section~\ref{sec:data_ethics} discusses data ethics considerations. Finally, the conclusions are presented in Section~\ref{sec:con}.

%%%%%%%%%% Related work%%%%%%%%%%%%%%%%%%%%%

\section{Related work}
\label{sec:literature}

HSI datasets have long served as extensive repositories of spectral information, enabling precise pixel-level classification across diverse land cover types, including forests, urban areas, and agricultural fields~\cite{1}~\cite{2}.
While these datasets significantly enhance classification accuracy, they lack interpretability, providing little insight into why specific pixels are assigned particular classes. This gap between computational precision and human understanding is a major challenge, particularly in environmental monitoring, precision agriculture, and disaster management, where explainability is crucial for informed decision-making. To address this, research has shifted toward multimodal approaches that integrate HSI data with textual descriptions, bridging the gap between raw numerical outputs and meaningful semantic interpretation to enhance both reliability and comprehensibility.

\textbf{Evolution of HSI Datasets}: Early HSI research focused on developing datasets for land cover classification using pixel-wise numerical labels. Notable datasets include Indian Pines (1992)~\cite{14}, Pavia University (2001)~\cite{PaviaU}, Salinas Scene (2002)~\cite{15}, and Houston University (2013, 2018)~\cite{16}, each targeting specific applications such as general classification, urban analysis, and agricultural studies. The Chikusei dataset (2016)~\cite{17} captured agricultural landscapes, while the Kennedy Space Center dataset (1996)~\cite{18} provided insights into complex ecosystems. Despite enhancing classification accuracy, these datasets lacked semantic context, offering limited interpretability and leaving analysts without clear explanations for pixel-level class assignments. While early HSI datasets significantly improved classification accuracy, their reliance on pixel-wise numerical labels without contextual information limits their usefulness for semantic understanding and decision-making~\cite{19}. Moreover, alongside standalone HSI archives, early remote sensing also integrated complementary modalities such as MultiSpectral Imaging (MSI), Light Detection and Ranging (LiDAR), and Synthetic Aperture Radar (SAR) to enhance scene understanding~\cite{20}. Sensors like Sentinel-2 \cite{claverie2018harmonized} and WorldView-2~\cite{lake2022deep} provided multispectral views suitable for large-scale monitoring, while LiDAR datasets, including the International Society for Photogrammetry and Remote Sensing (ISPRS) Vaihingen benchmark~\cite{sun2022fair1m} and integrated LiDAR-HSI collections, contributed precise elevation and structural data. SAR, known for its resilience to weather conditions, provided crucial backscatter information, aiding terrain analysis. Despite the richness of these modalities, early datasets primarily relied on numerical labels or sparse metadata, limiting their interpretability and broader applicability beyond classification tasks. MSI, LiDAR, and SAR datasets enhance scene comprehension but lack standardized fusion frameworks and cross-modal interactions, hindering their effectiveness in complex geospatial analysis.

%%%%%%%%%%%%%%%%%%%%%%%%%%%%%%%%%%%%%%%%%%%%%%%%%%%%%%%%%%%%%%%%%%%%%%%%%%%%%%%%%%%%%%%%%%%%%%
\begin{figure*}[t]
    \centering
    \begin{subfigure}{0.49\textwidth}
        \centering
        \includegraphics[height=0.50\textwidth,width=0.9\textwidth, trim=0 0 0 150, clip]{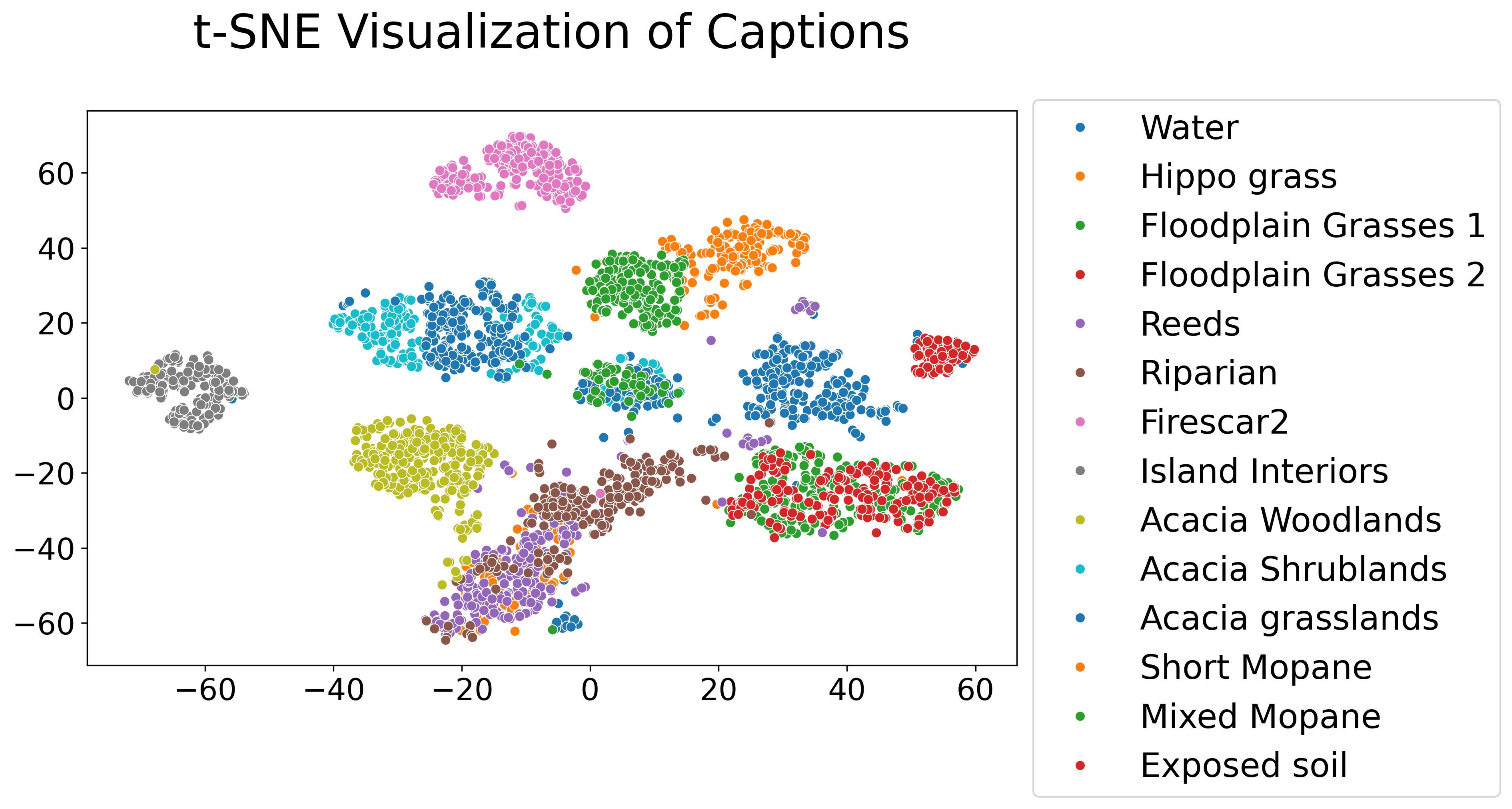}
        \caption{Botswana}
        \label{subfig:tsne_botswana}
    \end{subfigure}
    \hfill
    \begin{subfigure}{0.49\textwidth}
        \centering
        \includegraphics[height=0.50\textwidth,width=0.9\textwidth, trim=0 0 0 150, clip]{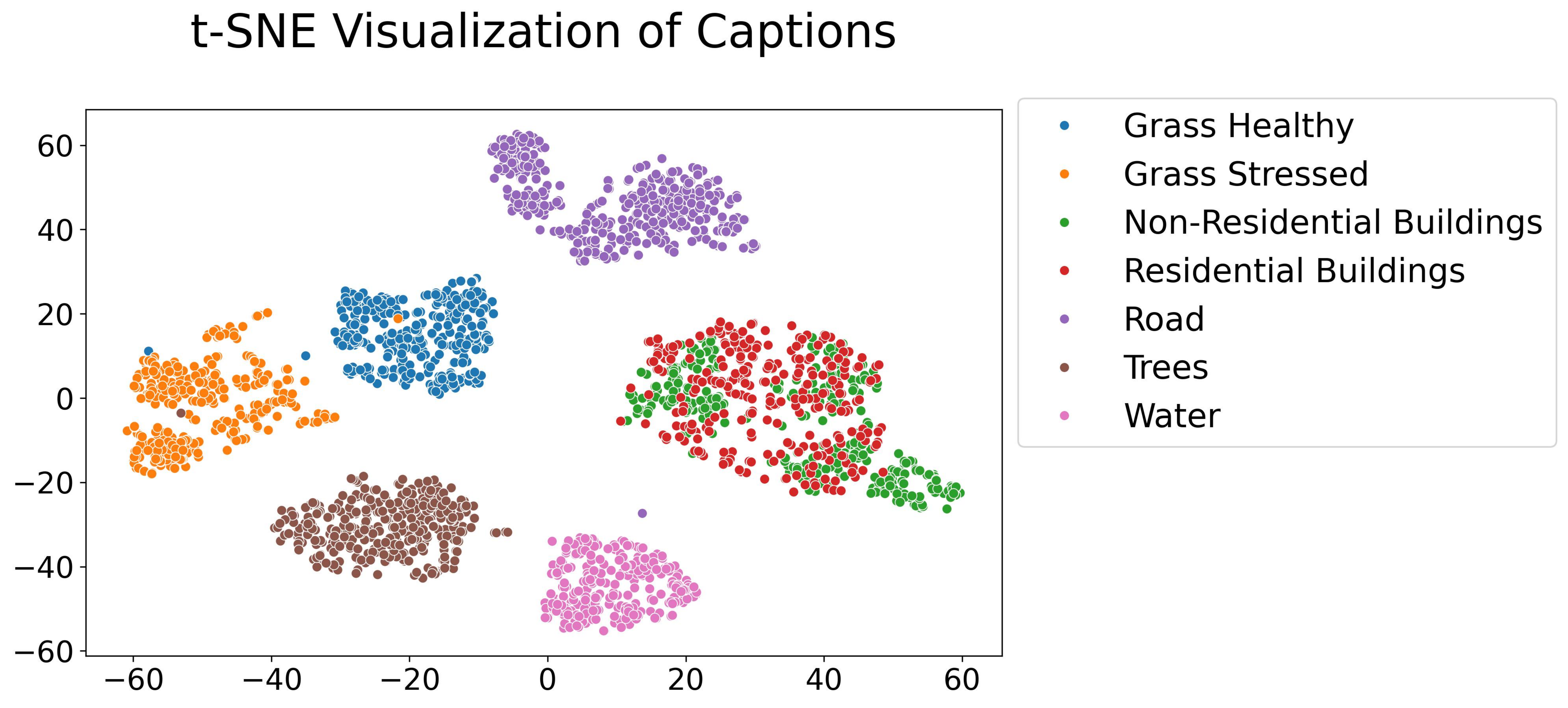}
        \caption{Houston13}
        \label{subfig:tsne_houston13}
    \end{subfigure}
    \begin{subfigure}{0.49\textwidth}
        \centering
        \includegraphics[height=0.50\textwidth,width=0.9\textwidth, trim=0 0 0 150, clip]{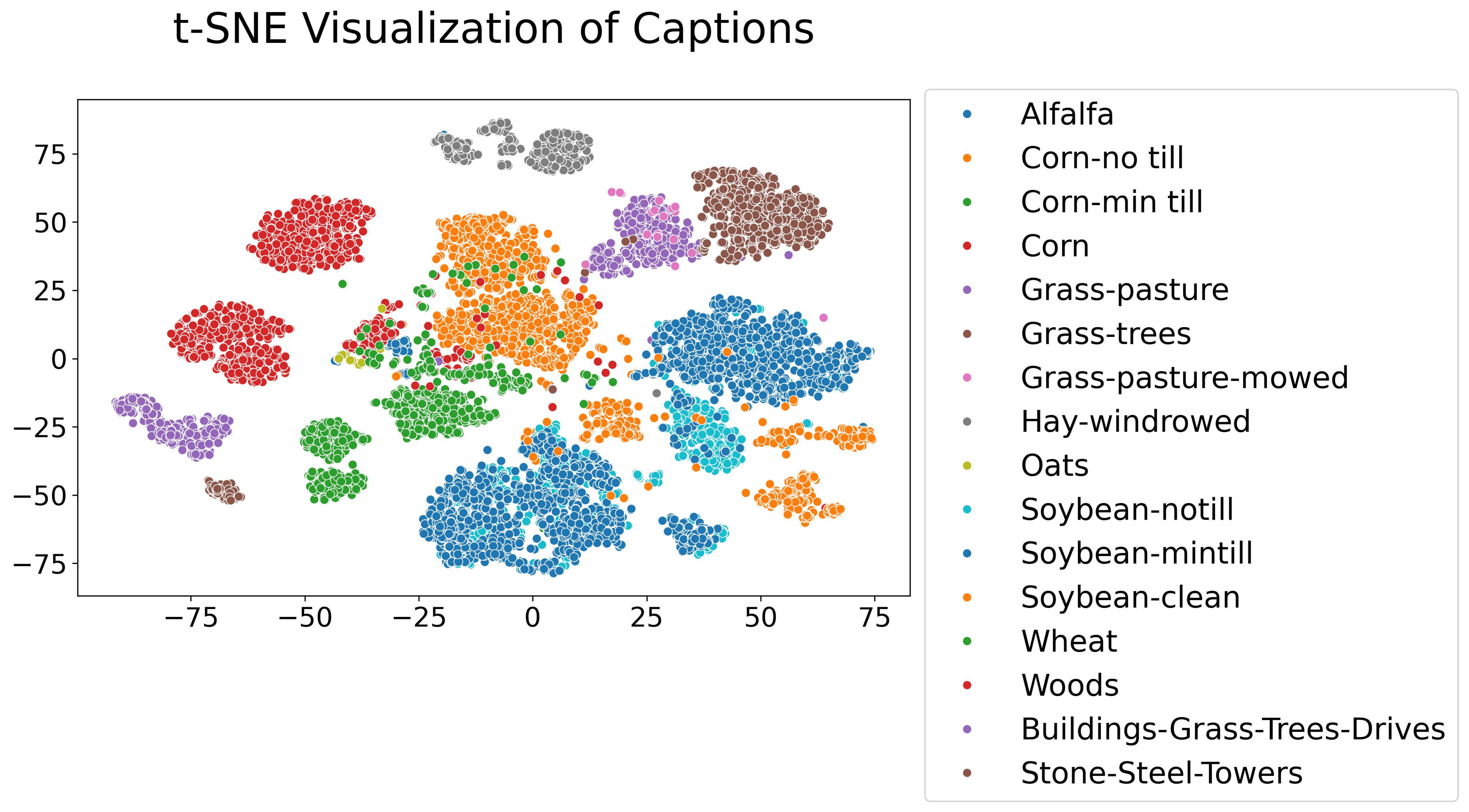}
        \caption{Indian Pines}
        \label{subfig:tsne_indian_pines}
    \end{subfigure}
    \hfill
    \begin{subfigure}{0.49\textwidth}
        \centering
        \includegraphics[height=0.50\textwidth,width=0.9\textwidth, trim=0 0 0 150, clip]{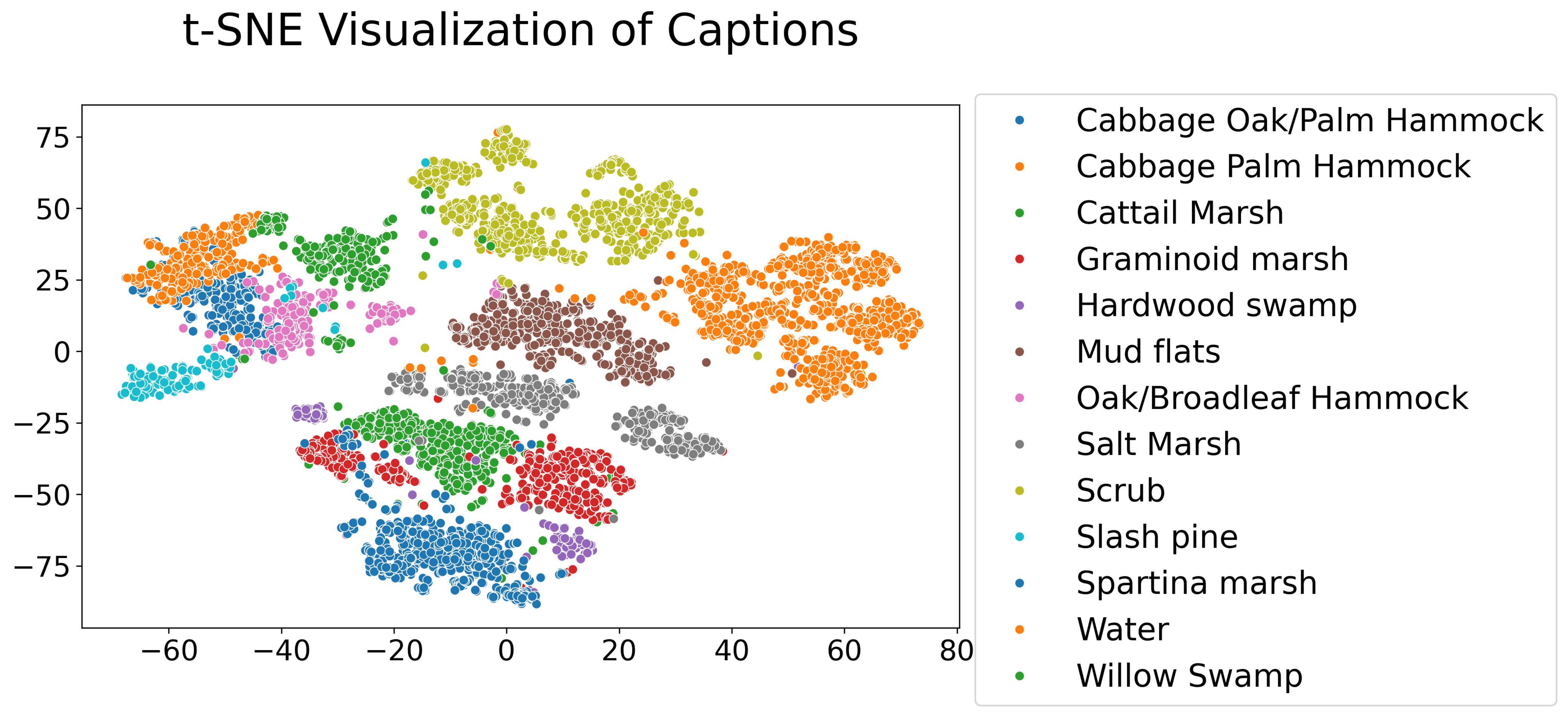}
        \caption{Kennedy Space Center}
        \label{subfig:tsne_ksc}
    \end{subfigure}
    %\vspace{+3mm}
    \caption{t-SNE visualizations of BERT-based caption embeddings across the Botswana, Houston13, Indian Pines and Kennedy Space Center datasets.}
    \vspace{-4mm}
    \label{fig:tsne_visualizations}
\end{figure*}
%%%%%%%%%%%%%%%%%%%%%%%%%%%%%%%%%%%%%%%%%%%%%%%%%%%%%%%%%%%%%%%%%%%%%%%%%%%%%%%%%%%%%%%%%%%%%%

\textbf{Shift Towards Semantic Awareness}: Between 2015 and 2020, researchers recognized the limitations of numerical-only outputs in HSI datasets and began integrating textual descriptions to enhance interpretability. While early efforts in remote sensing image captioning focused on RGB datasets, such as UCM-Captions~\cite{UC} and Sydney-Captions~\cite{Sydney-Captions}, HSI datasets lacked similar advancements. The RSICD~\cite{RSCID} and NWPUCaptions~\cite{NWPUCaptions} datasets expanded scene understanding by providing diverse image-caption pairs, while RSICap incorporated object-level annotations based on the DOTA dataset~\cite{chen2024object}. Despite these improvements, these captioning efforts remained focused on RGB imagery, leaving HSI datasets without detailed semantic labels necessary for a more refined spectral-contextual understanding in classification and decision-making applications.
During this period, researchers explored multimodal integration beyond HSI, incorporating complementary modalities like MSI, LiDAR, and SAR. MSI enhanced spectral range coverage, LiDAR contributed 3D structural details, and SAR improved robustness under diverse conditions~\cite{10574165}. AeroRIT (2019)~\cite{AERORIT} exemplified this trend by integrating HSI with object-level annotations and additional sensor modalities. However, while multimodal datasets advanced classification accuracy, they primarily provided scene-level descriptions rather than pixel-wise annotations. This lack of semantic information limited their effectiveness in fully leveraging the spectral and spatial details of HSI data, highlighting the need for improved annotation techniques to bridge this gap.

A notable advancement in HSI captioning is Language-aware Domain Generalization Network LDGNet (2023), which establishes a benchmark by mapping spectral-spatial features directly to linguistic representations~\cite{LDGNet}. Unlike earlier datasets that lacked semantic annotations, LDGNet provides structured captions, enhancing interpretability in HSI classification. It includes multiple datasets, such as Pavia University, Pavia Centre, Houston13, Houston18, GID-wh, and GID-nc, covering a range of spectral bands and classification tasks. However, despite its pioneering approach, LDGNet relies on template-based captions and offers descriptions at the patch level rather than at the pixel level, limiting the granularity of its semantic information. Furthermore, LDGNet provides only two captions per class, which severely restricts the diversity and contextual depth of the textual descriptions. This constraint underscores the need for datasets with fine-grained, natural-language annotations to further improve explainability and contextual understanding in HSI remote sensing.

\vspace{-3mm}
\section{Dataset Acquisition and Preprocessing} 
\label{sec:dataser_prepro}

This study presents a novel dataset with pixel-level annotations of available hyperspectral images, enabling improved semantic learning. Four well-known and widely used benchmark datasets \textit{Botswana}, \textit{Houston13}, \textit{Indian Pines (IP)}, and \textit{Kennedy Space Center} are employed to ensure diverse spectral and spatial evaluation.

\textbf{Botswana~\cite{bots}:} Acquired by NASA’s EO-1 satellite using the Hyperion sensor, this dataset initially contains 242 spectral bands, and spectral resolution of 10 nm. Following preprocessing, the number of usable bands is reduced to 145, representing 14 distinct land cover classes in the Okavango Delta region. The image size is $1476 \times 256$ pixels.

\textbf{Houston 2013 (Houston13)~\cite{hou}:} Acquired by the Compact Airborne Spectrographic Imager (CASI) and used in the 2013 IEEE Geoscience and Remote Sensing Society (GRSS) Data Fusion Contest, it originally comprised 144 spectral bands but the released version consists of 48 spectral bands, covering wavelengths from 380 to 1050 nm, with a spatial resolution of 2.5 m/pixel. The image dimensions are $349 \times 1905$ pixels, and the dataset includes land cover classes, such as urban areas, vegetation etc.

\textbf{Indian Pines~\cite{IP}:} Acquired by the Airborne Visible Infra-Red Imaging Spectrometer (AVIRIS), the dataset comprises HSI data with a spatial dimension of \(145 \times 145\) pixels and 224 spectral bands spanning wavelengths from 400 to 2500 nm. After removing 24 spectral bands affected by water absorption, 200 bands remain for processing. The ground truth consists of 16 vegetation classes, representing different crop types and forested areas.

\textbf{Kennedy Space Center~\cite{ks}:} Collected using AVIRIS over Kennedy Space Center, Florida, this dataset comprises 13 land cover classes. After removing water absorption and low signal-to-noise ratio (SNR) bands, 176 bands were used for the analysis. The spectral bands span wavelengths from 400 to 2500 nm, and the image dimensions are $512\times614$ pixels.

The datasets have been preprocessed via pixel-wise patching, extracting each pixel’s spectral signature to form localized patches. These are paired with textual descriptions, creating a novel HSI captioning dataset. Preprocessing ensures uniform input dimensions while preserving spectral and spatial integrity, supporting models that link spectral data with semantic meaning.

Let \( \mathbf{D}_{\text{HSI}} \in \mathbb{R}^{B \times H \times W} \) represent the original HSI dataset, where \( B \) is the number of spectral bands, and \( H \) and \( W \) respectively represent the two spatial dimensions. Initially, the dataset is processed at a pixel level, where each spatial coordinate \( (h, w) \in (H, W) \) is set to 1. Each pixel-wise sample is then padded to form patches of size \( (k \times k) \), ensuring a structured input format while preserving spectral integrity. This transformation results in a dataset with dimensions \( (S, B, k, k) \), where \( S = \frac{H \times W}{k^2} \), which is the total number of patched samples. In all experiments reported in this paper, we set \( k = 1 \), so that each pixel is treated as an independent sample with \( S = H \times W \) and no spatial dimension reduction occurs. Each pixel is independently annotated at \( 1 \times 1 \) spatial granularity, distinguishing HyperCap from patch-level captioning approaches where scene descriptions are uniformly upscaled across regions. The generalized formulation below is provided for extensibility to future work. The transformation is mathematically expressed as:
%%%%%%%%%%%%%%%%%%%%
% \vspace{-1.5mm}
\begin{equation}  
\mathbf{D}_{\text{patched}} = {Reshape} \left( \mathbf{D}_{\text{HSI}}, (S, B, k, k) \right)
\vspace{-1.5mm}
\end{equation}  
%%%%%%%%%%%%%%%%%%%%

The ground truth (GT) data, denoted as \( \mathbf{GT} \in \mathbb{R}^{H \times W} \), is processed to maintain alignment with the HSI patches. Since each patch corresponds to a single label, the GT is reshaped accordingly:  
%%%%%%%%%%%%%%%%%%%%
\vspace{-1.5mm}
\begin{equation}
\mathbf{GT}_{\text{patched}} = {Reshape} \left( \mathbf{GT}, (S, 1) \right).
% \vspace{-2mm}
\end{equation} 

\begin{figure}[!t]
\centering
\begin{tcolorbox}[colback=red!5!white,
                  colframe=red!75!black,
                  title=Expert Remote Sensing Data Annotator --- Hyperspectral Captioning Task,
                  boxrule=0.8mm,
                  left=2mm,
                  right=2mm,
                  top=1mm,
                  bottom=2mm]
\textbf{Role:} Expert Remote Sensing Data Annotator tasked with creating a textual description for a specific hyperspectral data point.

\textbf{Objective:} Generate a single, concise, visually grounded descriptive sentence corresponding to a hyperspectral signature.

\textbf{Metadata:} Include acquisition date, sensor type, spatial resolution if available.

\textbf{Provided Information:}
\begin{itemize}
    \item High-Level Context: [CLASS\_LABEL]
    \item Spectral Signature: [SPECTRAL\_SIGNATURE]
    \item Environment: [DATASET\_CONTEXT]
\end{itemize}

\textbf{Instructions:}
\begin{enumerate}
    \item Analyze the context to understand the scene.
    \item Generate one descriptive caption based on visual characteristics.
\end{enumerate}

\textbf{Constraints:}
\begin{itemize}
    \item Do not use words from [CLASS\_LABEL].
    \item Focus on physical attributes only.
    \item Avoid functions, purposes, or invented details.
    \item Output must be a single sentence.
\end{itemize}
\end{tcolorbox}
\caption{Prompt template used for hyperspectral caption generation in the proposed annotation framework.}
\vspace{-3mm}
\label{fig:hyperspectral_caption_prompt}
\end{figure}

%%%%%%%%%%%%%%%%%%%%
\subsection{Dataset Annotation}
\label{sec:annotation}

\begin{figure*}[t]
    \centering
    \begin{subfigure}{0.24\textwidth}
        \centering
        \includegraphics[width=\textwidth]{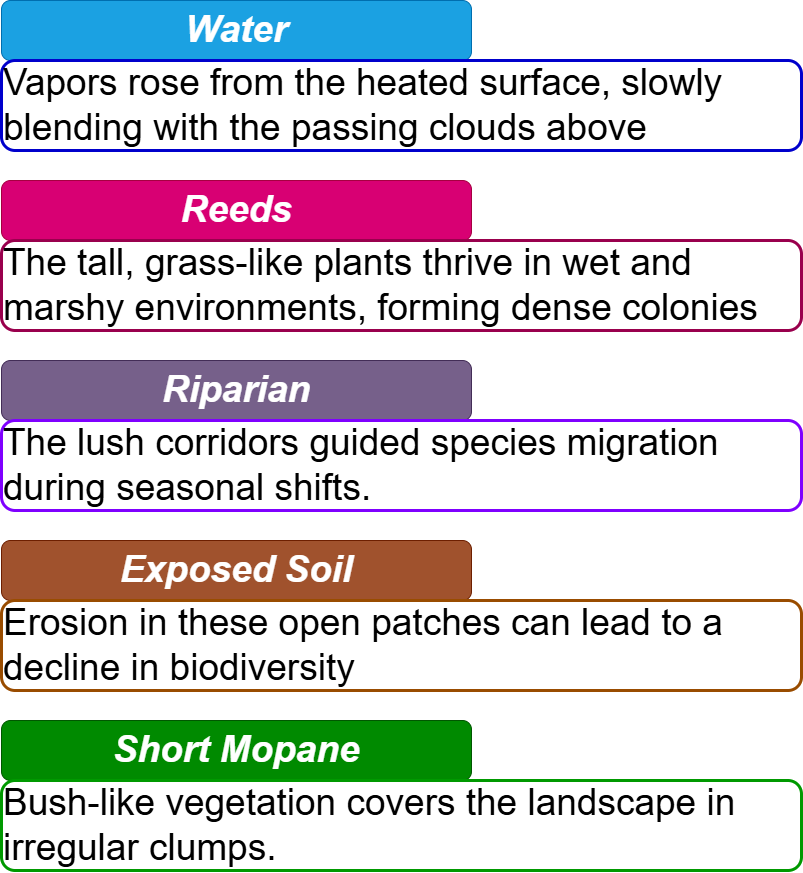}
        \caption{Botswana}
        \label{fig:botswana_caption}
    \end{subfigure}
    \begin{subfigure}{0.24\textwidth}
        \centering
        \includegraphics[width=\textwidth]{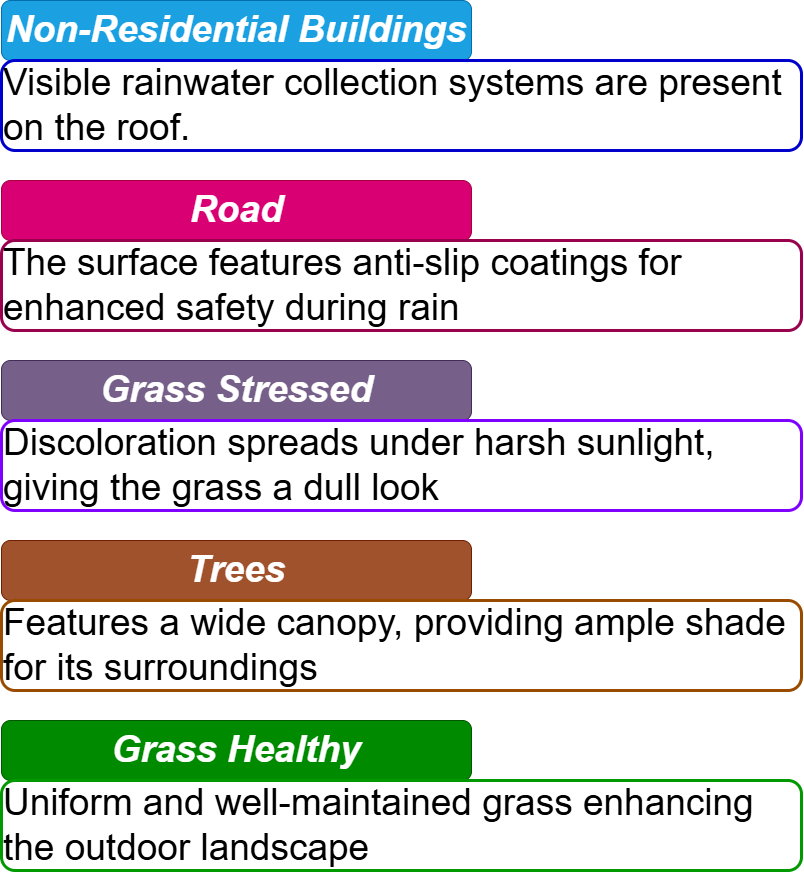}
        \caption{Houston13}
        \label{fig:houston_caption}
    \end{subfigure}
    % \vspace{-4mm}
    \begin{subfigure}{0.24\textwidth}
        \centering
        \includegraphics[width=\textwidth]{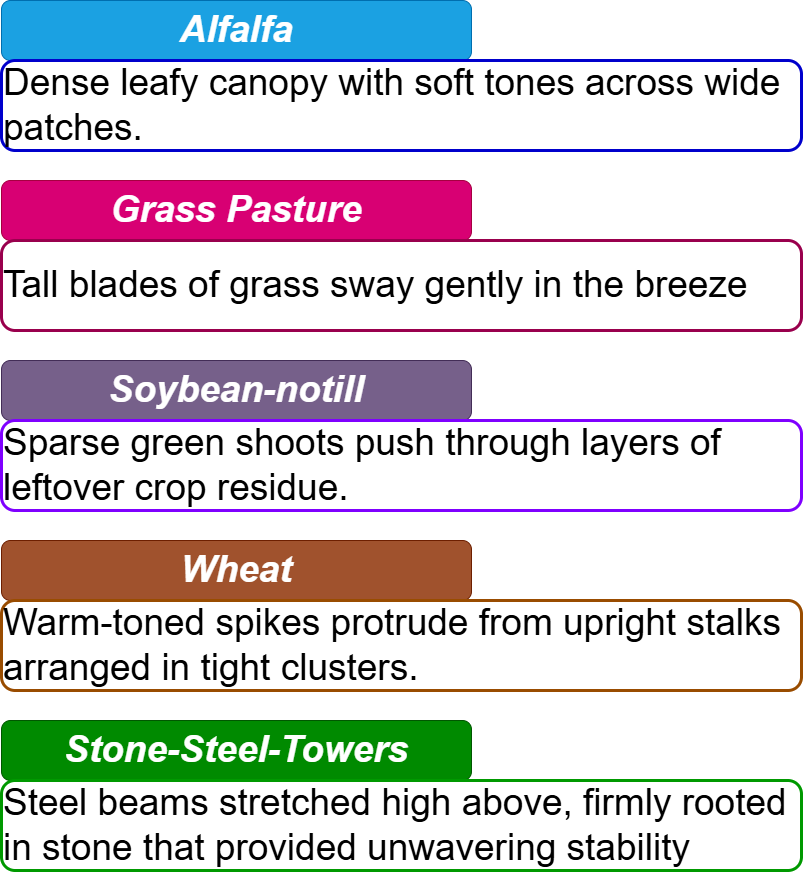}
        \caption{Indian Pines}
        \label{fig:indian_pines_caption}
    \end{subfigure}
    \begin{subfigure}{0.24\textwidth}
        \centering
        \includegraphics[width=\textwidth]{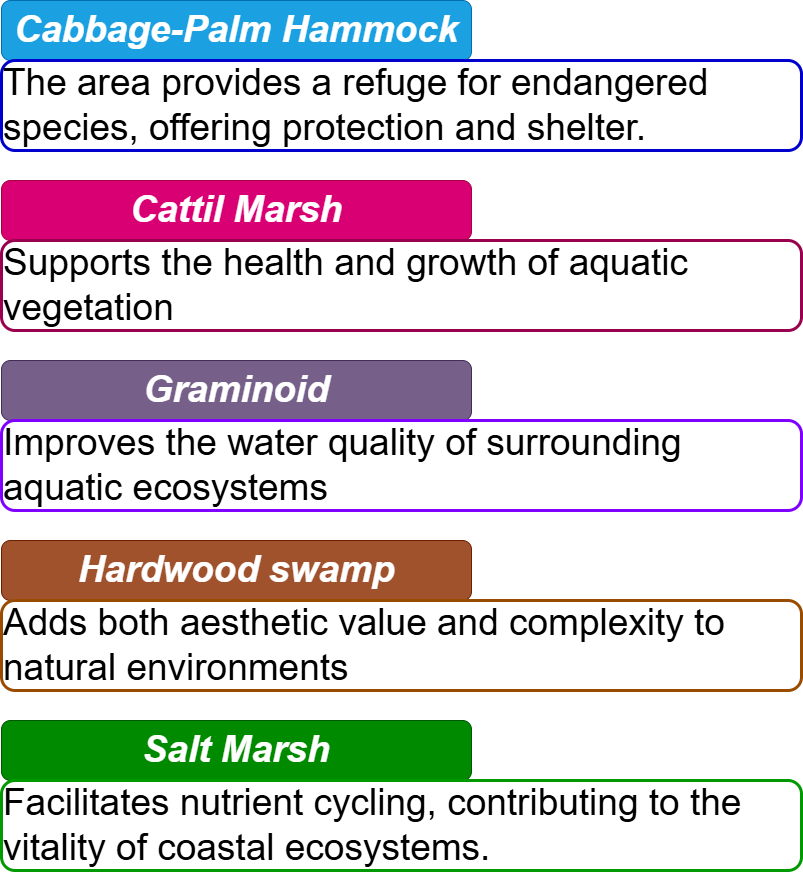}
        \caption{Kennedy Space Center}
        \label{fig:ksc_caption}
    \end{subfigure}
    % \vspace{+2mm}
    \caption{A few sample captions generated for all four datasets used in our study.}
    \label{fig:datasets_captions}
    \vspace{-3mm}
\end{figure*}

%%%%%%%%%%%%%%%%%%%%%%%%%%%%%%%%%%%

\begin{figure*}[!t]
\centering
\includegraphics[width=0.8\textwidth]{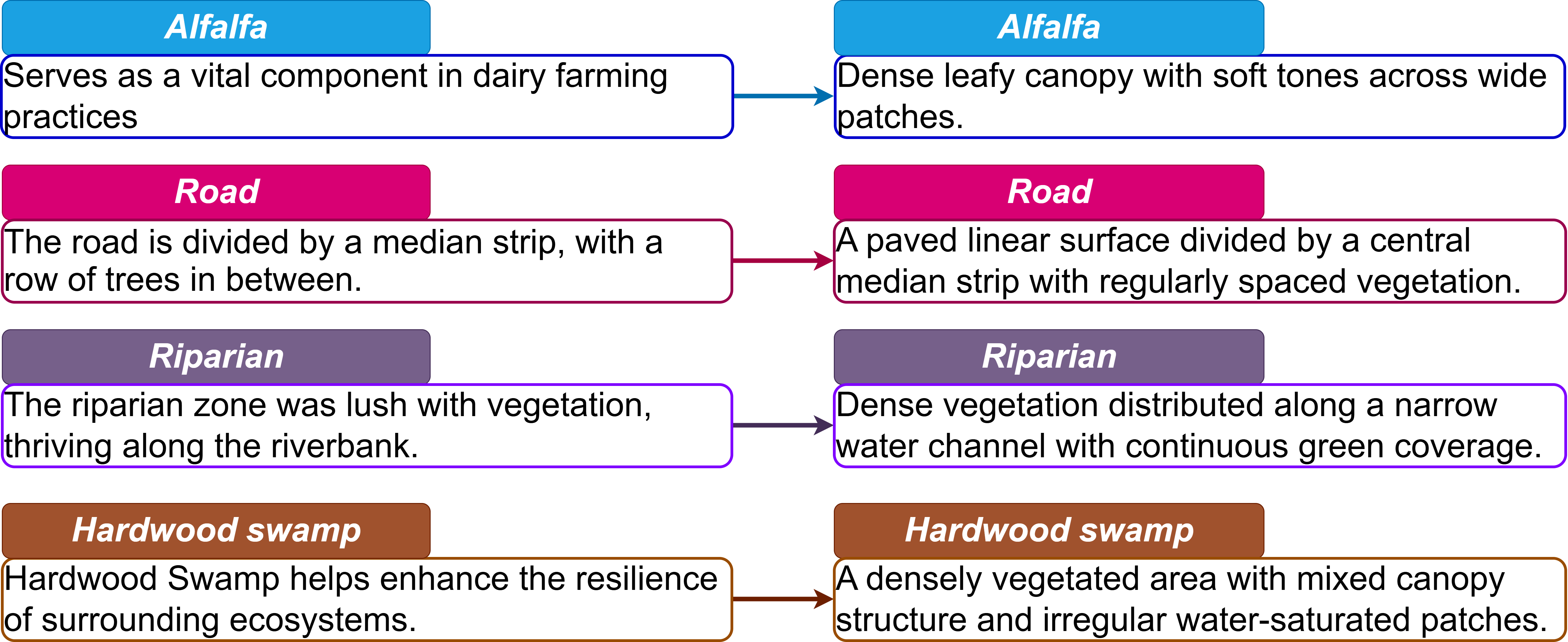}
\caption{Captions before and after manual refinement.}
\vspace{-3mm}
\label{fig:PnP}
\end{figure*}

To construct the HyperCap dataset, the four benchmark HSI datasets were systematically annotated using a hybrid strategy that combined automated caption generation with expert refinement. A structured prompt template (Figure~\ref{fig:hyperspectral_caption_prompt}) was designed to guide Large Language Models (LLMs) in generating visually grounded textual descriptions conditioned on hyperspectral signatures. The template incorporated spectral metadata and environmental context while enforcing constraints to avoid the use of class names, functional descriptions, or speculative details, thereby encouraging physically interpretable outputs. Initially, both ChatGPT-4o\footnote{\url{https://chatgpt.com/}} and Mistral Large\footnote{\url{https://mistral.ai/news/mistral-large}} were applied in parallel to every labeled pixel across all four datasets using the identical structured prompt. The dual-generation strategy was adopted to maximize candidate diversity. The outputs from both models were subsequently evaluated by three expert annotators with backgrounds in remote sensing and hyperspectral image analysis. For each sample, the annotators selected and refined the most spectrally consistent and visually grounded description. The final dataset retains exactly one expert-refined caption per labeled pixel, irrespective of which LLM produced the initial candidate. To ensure annotation consistency and physical plausibility, the expert annotators reviewed and refined the LLM-generated descriptions by comparing them with the corresponding spectral signatures and ground-truth labels. In total, the HyperCap dataset comprises 21,237 pixel-wise captions across the four benchmark datasets. All automatically generated captions were manually reviewed by three expert annotators under a shared refinement protocol. The refinement criteria included: (1) removal of any class names or explicit categorical identifiers; (2) elimination of functional, speculative, or hallucinated content introduced by the LLM; and (3) preservation of physically observable spectral–spatial characteristics while ensuring linguistic diversity.  This verification process enhanced annotation reliability and reduced potential biases introduced by automated generation. Unlike patch-based captioning approaches that assign a single description to a spatial region potentially containing multiple land-cover classes, HyperCap enforces strict one-to-one pixel-to-caption alignment. Each labeled pixel $(h,w)$ is independently indexed and associated with a unique expert-refined caption, without any region-to-pixel projection. This structural design preserves fine spatial granularity and ensures that captions reflect pixel-level spectral--spatial characteristics rather than aggregated regional semantics. Representative caption examples for all four datasets are illustrated in Figure~\ref{fig:datasets_captions}, and Table~\ref{tab:dataset_comparison} summarizes detailed statistics of HyperCap in comparison with LDGNet and the original HSI datasets. Figure~\ref{fig:PnP} presents qualitative examples of captions before and after expert refinement. In Figure~\ref{fig:PnP}, the captions shown in the left column correspond to raw LLM outputs that were \emph{discarded} during expert refinement, whereas only the right-column, expert-edited captions are included in HyperCap. Together, Figure~\ref{fig:PnP} and Table~\ref{tab:error_metrics} provide qualitative and quantitative analyses of caption refinement and annotation quality. The resulting captions encode observable spectral--spatial attributes at the pixel level, providing complementary semantic information rather than explicit target-label representations.

\begin{table}[]
\centering
{\fontsize{8}{8}\selectfont
\setlength{\tabcolsep}{0.2mm}
\caption{Dataset Comparison Based on Various Attributes.}
\begin{tabular}{l |
                >{\centering\arraybackslash}p{1cm} |
                >{\centering\arraybackslash}p{1.2cm} |
                >{\centering\arraybackslash}p{1.1cm} |
                >{\centering\arraybackslash}p{1.3cm} |
                >{\centering\arraybackslash}p{0.8cm}}
    \toprule
    \textbf{Dataset Name} & \textbf{Total Bands} & \textbf{Total Samples} & \textbf{Number of Classes} & \textbf{Captions} & \textbf{Pixel-Level} \\
    \midrule
    Indian Pines~\cite{14}         & 200 & 10,248 & 16 & \textcolor{red}{$\times$} & \textcolor{red}{$\times$} \\
    Kennedy Space Center~\cite{ks} & 176 & 5,211  & 13 & \textcolor{red}{$\times$} & \textcolor{red}{$\times$} \\
    Botswana~\cite{bots}           & 145 & 3,248  & 14 & \textcolor{red}{$\times$} & \textcolor{red}{$\times$} \\
    Houston13~\cite{hou}           & 48  & 2,530  & 7  & \textcolor{red}{$\times$} & \textcolor{red}{$\times$} \\
    \midrule
    \textbf{LDGnet}~\cite{LDGNet} & & & & & \\
    Pavia University & 103 & 39,332 & 7 & 14 & \textcolor{red}{$\times$} \\
    Pavia Centre     & 102 & 39,355 & 7 & 14 & \textcolor{red}{$\times$} \\
    Houston13        & 48  & 2,530  & 7 & 14 & \textcolor{red}{$\times$} \\
    Houston18        & 48  & 53,200 & 7 & 14 & \textcolor{red}{$\times$} \\
    GID-wh           & 4   & 23,339 & 5 & 10 & \textcolor{red}{$\times$} \\
    GID-nc           & 4   & 30,812 & 5 & 10 & \textcolor{red}{$\times$} \\
    \midrule
    \textbf{HyperCAP (Ours)} & & & & & \\
    Indian Pines         & 200 & 10,248 & 16 & 10,248 & \textcolor{blue}{$\checkmark$} \\
    Kennedy Space Center & 176 & 5,211  & 13 & 5,211  & \textcolor{blue}{$\checkmark$} \\
    Botswana             & 145 & 3,248  & 14 & 3,248  & \textcolor{blue}{$\checkmark$} \\
    Houston13            & 48  & 2,530  & 7  & 2,530  & \textcolor{blue}{$\checkmark$} \\
    \bottomrule
\end{tabular}
\label{tab:dataset_comparison}
}
\vspace{-4mm}
\end{table}

\begin{table}[h]
\centering
{\fontsize{8}{8}\selectfont
\setlength{\tabcolsep}{5pt}
\caption{Inter-Annotator Agreement Error Rates: BLEU Error (BE1--BE4), METEOR Error (MTRE), and ROUGE-L Error (R-LE).}
\vspace{2mm}
\begin{tabular}{l | c | c | c | c | c | c}
    \toprule
    \textbf{Dataset} & \textbf{BE1} & \textbf{BE2} & \textbf{BE3} & \textbf{BE4} & \textbf{MTRE} & \textbf{R-LE} \\
    \midrule
    Indian Pines         & 0.73 & 0.87 & 0.92 & 0.95 & 0.83 & 0.74 \\
    Houston13            & 0.84 & 0.91 & 0.95 & 0.97 & 0.89 & 0.84 \\
    Botswana             & 0.81 & 0.90 & 0.95 & 0.97 & 0.87 & 0.81 \\
    Kennedy Space Center & 0.79 & 0.86 & 0.92 & 0.95 & 0.87 & 0.78 \\
    \bottomrule
\end{tabular}
\vspace{-3mm}
\label{tab:error_metrics}
}
\end{table}

\subsection{Dataset Analysis}
\label{sec:data_analysis}

In this work, we conduct a systematic analysis of the HyperCap dataset through four sequential steps: \textbf{Step~1} quantifies class and caption distribution across all four datasets; \textbf{Step~2} examines the linguistic structure of the captions via part-of-speech analysis; \textbf{Step~3} evaluates the spectral separability of pixel-wise HSI samples via t-SNE visualizations; and \textbf{Step~4} measures inter-annotator agreement and caption refinement quality using standard captioning error metrics. Together, these steps provide a thorough understanding of both the statistical properties of the data and the quality of its textual annotations.

\noindent\textbf{Step~1: Class and Caption Distribution.} Figure~\ref{fig:sunburst_visualizations} presents pie charts illustrating the joint class distribution and caption count per class across all four datasets. Since HyperCap provides exactly one caption per pixel-level sample (Table~\ref{tab:dataset_comparison}), the caption distribution directly mirrors the class distribution, and the slice sizes simultaneously encode both quantities. Figure~\ref{subfig:sun_vis_a} shows Botswana's relatively balanced distribution, with Acacia Woodlands (314) and Floodplain Grasses 1 (251) more frequent than Short Mopane (95). Figure~\ref{subfig:sun_vis_b} for Houston13 highlights dominance by Non-Residential Buildings (408) and Road (443), while Water (285) is underrepresented. Figure~\ref{subfig:sun_vis_c} for Indian Pines reveals extreme imbalance, with Soybean-mintill (2,455) dominating over Oats (20). Figure~\ref{subfig:sun_vis_d} for KSC shows Scrub (927) and Water (761) prevailing over Hardwood Swamp (243).

\noindent\textbf{Step~2: Linguistic Structure Analysis.} Figure~\ref{fig:pos_visualizations} presents part-of-speech\footnote{\url{https://www.ibm.com/docs/en/wca/3.5.0?topic=analytics-part-speech-tag-sets}} distributions across all dataset captions. Figures~\ref{subfig:pos_vis_a}--\ref{subfig:pos_vis_d} show that the captions are strongly noun-heavy: nouns (NN) are the single most frequent tag in every dataset (approximately 6{,}000 in Botswana, 5{,}600 in Houston13, 21{,}000 in Indian Pines, and 10{,}500 in Kennedy Space Center). 
Adjectives (JJ) and prepositions (IN) also appear frequently, whereas verbs (VB) are comparatively underrepresented, indicating a predominantly descriptive and object-centric linguistic structure characteristic of physically grounded remote-sensing annotations.

\noindent\textbf{Step~3: Semantic Embedding Analysis.} Figure~\ref{fig:tsne_visualizations} presents t-SNE visualizations of caption embeddings generated using the BERT~\cite{hosseini2023bert} pretrained model. Each point corresponds to one pixel's textual caption encoded through BERT, illustrating the semantic clustering and separability of captions across different land cover classes. Figures~\ref{subfig:tsne_botswana}--\ref{subfig:tsne_ksc} illustrate distinct clustering for major classes, while certain classes exhibit overlap due to semantic similarities in their caption descriptions. Notably, in the Botswana dataset, Acacia types show significant overlap, while in the KSC dataset, Salt Marsh and Spartina Marsh classes blend due to their similar material composition. These visualizations indicate that captions of the same land-cover class form cohesive semantic clusters---much like birds of the same species forming a flock, where each individual is distinct yet they group together because of shared characteristics. This confirms that same-class captions are semantically consistent while maintaining the lexical diversity evidenced in Table~\ref{tab:error_metrics}.

\noindent\textbf{Step~4: Inter-Annotator Agreement and Caption Refinement Quality.} 
Table~\ref{tab:error_metrics} reports inter-annotator agreement error rates derived from BLEU (B1--B4), METEOR (MTR), and ROUGE-L (R-L)~\cite{coco} scores by inverting their respective values, such that higher error rates indicate lower similarity between annotator descriptions. The BLEU metrics (B1--B4) assess n-gram overlap reflecting lexical similarity; METEOR captures semantic alignment through synonymy, stemming, and word order; and ROUGE-L measures structural similarity via the longest common subsequence. The consistently high error rates across all datasets --- for instance, BLEU-4 errors ranging from 0.95 to 0.97 and METEOR errors ranging from 0.83 to 0.89 --- indicate that annotators produced lexically and semantically diverse captions for the same pixel samples, rather than relying on template-like repetitions. Importantly, lexical diversity is not treated as a standalone proxy for annotation quality. Instead, physical grounding and spectral consistency were enforced through the expert refinement protocol (Fig.~\ref{fig:PnP}), which systematically removed captions containing functional interpretations, aesthetic language, class-name references, or non-observable world knowledge. Collectively, these error rates confirm that the captions exhibit genuine linguistic variation across samples and are not template-like reformulations of class labels. Figure~\ref{fig:PnP} demonstrates the constraining of LLM generated captions to visually grounded captions through the HyperCap framework of manual refinement. For instance, in ``Alfalfa,'' the initial caption---``Serves as a vital component in dairy farming practices.''---was discarded since it speaks of functional and agricultural context. The new caption---``Dense leafy canopy with soft tones across wide patches.''---merely lists the visual characteristics without class name or function. Likewise, for ``Road,'' the LLM-generated caption ``The road is divided by a median strip, with a row of trees in between.'' was refined to ``A paved linear surface divided by a central median strip with regularly spaced vegetation.''---removing the class name while retaining only directly observable structural attributes.
In all the examples, captions were revised to exclude the use of class names to avoid class leakage and world knowledge, only what can be seen visually. This favours self-supervised learning and vision-language grounding through all descriptions being based on what one sees.

\vspace{-2mm}

%%%%%%%%%%%%%%%%%%%%%%%%%%%%%%%%%%%%%%%%%%%%%%%%%%%%%%%%%%%%%%%%%%%%%%%%%%%%%%%%%%%%%%%%%%%%%%%%%%%%%%%%%%%%%%%%%%%%%%%%%%%%%%%%
%  Bert~\cite{hosseini2023bert} and T5\cite{t5} 
\begin{table*}[]
    % \vspace{-4mm}
	\large
	\centering
        \caption{Evaluation of Vision Encoders with/without Text Encoders on Botswana \& Houston13 Datasets.}
	\resizebox{1\linewidth}{!}{
		\begin{tabular}{c | l | l | c | c | c | c | c | c | c | c | c | c | c}
			\toprule
			                  & \multirow{3}{*}{\textbf{Vision Model}} & \multirow{3}{*}{\textbf{Metric}} & \textbf{IMG ONLY}  & \multicolumn{10}{c}{\textbf{IMG+TXT}} \\ 
			\cline{4-14}
			\textbf{DATASET} & & &\multirow{2}{*}{\textbf{Vision}}  & \multicolumn{2}{c|}{\textbf{CA}} & \multicolumn{2}{c|}{\textbf{CONCAT}} & \multicolumn{2}{c|}{\textbf{MHA}} & \multicolumn{2}{c|}{\textbf{PWA}} & \multicolumn{2}{c}{\textbf{PWM}} \\ \cline{5-14}
			& & & & \textbf{Bert} & \textbf{T5} & \textbf{Bert} & \textbf{T5} & \textbf{Bert} & \textbf{T5} & \textbf{Bert} & \textbf{T5} & \textbf{Bert} & \textbf{T5} \\ \hline
			
            \multirow{16}{*}{{\rotatebox[origin=c]{90}{\textbf{BOTSWANA}}}}& \multirow{4}{*}{\textbf{3D-RCNet}} & \textbf{OA}        & 86.59 & \textcolor{ForestGreen}{99.64} & \textcolor{BurntOrange}{99.82} & 99.16 & 99.60 & 99.56 & \textcolor{BurntOrange}{99.82} & 98.90 & \textcolor{Blue}{99.86} & 98.68 & \textcolor{ForestGreen}{99.64} \\
			& & \textbf{Precision} & 90.47 & \textcolor{ForestGreen}{99.67} & \textcolor{BurntOrange}{99.72} & 99.27 & 99.62 & 99.58 & \textcolor{ForestGreen}{99.67} & 99.04 & \textcolor{Blue}{99.82} & 98.65 & 99.47 \\
			& & \textbf{Kappa}     & 85.47 & \textcolor{ForestGreen}{99.61} & \textcolor{BurntOrange}{99.80} & 99.09 & 99.57 & 99.52 & \textcolor{BurntOrange}{99.80} & 98.80 & \textcolor{Blue}{99.85} & 98.57 & \textcolor{ForestGreen}{99.61} \\
			& & \textbf{F1-Score}  & 87.78 & 99.67 & \textcolor{BurntOrange}{99.78} & 99.16 & 99.64 & 99.61 & \textcolor{ForestGreen}{99.75} & 99.01 & \textcolor{Blue}{99.85} & 98.64 & 99.53 \\ \cline{2-14}
			& \multirow{4}{*}{\textbf{3D-ConvSST}} & \textbf{OA}        & \textcolor{BurntOrange}{99.95} & \textcolor{BurntOrange}{99.95} & \textcolor{Blue}{100.00} & \textcolor{BurntOrange}{99.95} & \textcolor{Blue}{100.00} & \textcolor{Blue}{100.00} & \textcolor{Blue}{100.00} & \textcolor{BurntOrange}{99.95} & \textcolor{Blue}{100.00} & \textcolor{ForestGreen}{99.86} & \textcolor{BurntOrange}{99.95} \\
			& & \textbf{Precision} & \textcolor{BurntOrange}{99.96} & \textcolor{BurntOrange}{99.96} & \textcolor{Blue}{100.00} & \textcolor{BurntOrange}{99.96} & \textcolor{Blue}{100.00} & \textcolor{Blue}{100.00} & \textcolor{Blue}{100.00} & \textcolor{BurntOrange}{99.96} & \textcolor{Blue}{100.00} & 99.86 & \textcolor{ForestGreen}{99.94} \\
			& & \textbf{Kappa}     & \textcolor{BurntOrange}{99.95} & \textcolor{BurntOrange}{99.95} & \textcolor{Blue}{100.00} & \textcolor{BurntOrange}{99.95} & \textcolor{Blue}{100.00} & \textcolor{Blue}{100.00} & \textcolor{Blue}{100.00} & \textcolor{BurntOrange}{99.95} & \textcolor{Blue}{100.00} & \textcolor{ForestGreen}{99.85} & \textcolor{BurntOrange}{99.95} \\
			& & \textbf{F1-Score}  & \textcolor{BurntOrange}{99.96} & \textcolor{BurntOrange}{99.96} & \textcolor{Blue}{100.00} & \textcolor{BurntOrange}{99.96} & \textcolor{Blue}{100.00} & \textcolor{Blue}{100.00} & \textcolor{Blue}{100.00} & \textcolor{BurntOrange}{99.96} & \textcolor{Blue}{100.00} & 99.87 & \textcolor{ForestGreen}{99.92} \\ \cline{2-14}
			& \multirow{4}{*}{\textbf{DBCTNet}} & \textbf{OA}        & 75.95 & 86.32 & 97.58 & 92.87 & 98.02 & 90.72 & 96.26 & 95.16 & \textcolor{ForestGreen}{98.81} & \textcolor{Blue}{99.56} & \textcolor{BurntOrange}{99.16} \\
			& & \textbf{Precision} & 65.32 & 92.38 & 98.00 & 93.97 & 98.38 & 92.92 & 97.18 & 96.08 & \textcolor{ForestGreen}{98.81} & \textcolor{Blue}{99.43} & \textcolor{BurntOrange}{99.26} \\
			& & \textbf{Kappa}     & 73.80 & 85.14 & 97.37 & 92.27 & 97.85 & 89.92 & 95.94 & 94.75 & \textcolor{ForestGreen}{98.71} & \textcolor{Blue}{99.52} & \textcolor{BurntOrange}{99.09} \\
			& & \textbf{F1-Score}  & 65.64 & 80.42 & 97.25 & 88.89 & 97.80 & 85.12 & 95.25 & 92.11 & \textcolor{ForestGreen}{98.18} & \textcolor{Blue}{99.37} & \textcolor{BurntOrange}{98.75} \\ \cline{2-14}
			& \multirow{4}{*}{\textbf{FAHM}} & \textbf{OA}        & \textcolor{ForestGreen}{99.91} & \textcolor{BurntOrange}{99.95} & \textcolor{Blue}{100.00} & \textcolor{BurntOrange}{99.95} & \textcolor{Blue}{100.00} & \textcolor{Blue}{100.00} & \textcolor{Blue}{100.00} & \textcolor{Blue}{100.00} & \textcolor{Blue}{100.00} & 99.64 & \textcolor{BurntOrange}{99.95} \\
			& & \textbf{Precision} & \textcolor{ForestGreen}{99.92} & \textcolor{BurntOrange}{99.96} & \textcolor{Blue}{100.00} & \textcolor{BurntOrange}{99.96} & \textcolor{Blue}{100.00} & \textcolor{Blue}{100.00} & \textcolor{Blue}{100.00} & \textcolor{Blue}{100.00} & \textcolor{Blue}{100.00}& 99.68 & \textcolor{BurntOrange}{99.96} \\
			& & \textbf{Kappa}     & \textcolor{ForestGreen}{99.90} & \textcolor{BurntOrange}{99.95} & \textcolor{Blue}{100.00} & \textcolor{BurntOrange}{99.95} & \textcolor{Blue}{100.00} & \textcolor{Blue}{100.00} & \textcolor{Blue}{100.00} & \textcolor{Blue}{100.00}& \textcolor{Blue}{100.00} & 99.61 & \textcolor{BurntOrange}{99.95} \\
			& & \textbf{F1-Score}  & 99.92 & \textcolor{BurntOrange}{99.96} & \textcolor{Blue}{100.00} & \textcolor{BurntOrange}{99.96} & \textcolor{Blue}{100.00} & \textcolor{Blue}{100.00} & \textcolor{Blue}{100.00} & \textcolor{Blue}{100.00} & \textcolor{Blue}{100.00} & 99.70 & \textcolor{ForestGreen}{99.93} \\ \hline
			
            \multirow{16}{*}{{\rotatebox[origin=c]{90}{\textbf{HOUSTON13}}}} & \multirow{4}{*}{\textbf{3D-RCNet}} & \textbf{OA}        & 97.45 & \textcolor{BurntOrange}{99.77} & 99.37 & 99.37 & \textcolor{ForestGreen}{99.71} & 99.66 & \textcolor{Blue}{99.94} & 99.32 & 99.71 & 98.98 & 99.37 \\
			& & \textbf{Precision} & 97.64 & \textcolor{BurntOrange}{99.75} & 99.41 & 99.35 & \textcolor{ForestGreen}{99.68} & 99.61 & \textcolor{Blue}{99.93} & 99.36 & 99.68 & 99.01 & 99.41 \\
			& & \textbf{Kappa}     & 97.02 & \textcolor{BurntOrange}{99.73} & 99.27 & 99.27 & \textcolor{ForestGreen}{99.66} & 99.60 & \textcolor{Blue}{99.93} & 99.20 & 99.66 & 98.81 & 99.27 \\
			& & \textbf{F1-score}  & 97.53 & \textcolor{BurntOrange}{99.78} & 99.42 & 99.39 & \textcolor{ForestGreen}{99.72} & 99.66 & \textcolor{Blue}{99.94} & 99.38 & 99.72 & 99.04 & 99.43 \\ \cline{2-14}
			& \multirow{4}{*}{\textbf{3D-ConvSST}} & \textbf{OA}        & 99.43 & 99.43 & \textcolor{BurntOrange}{99.54} & 99.43 & 99.43 & \textcolor{ForestGreen}{99.49} & \textcolor{Blue}{100.00} & 99.43 & 99.37 & 99.43 & 99.37 \\
			& & \textbf{Precision} & \textcolor{ForestGreen}{99.53} & \textcolor{ForestGreen}{99.53} & 99.50 & \textcolor{ForestGreen}{99.53} & \textcolor{ForestGreen}{99.53} & \textcolor{BurntOrange}{99.59} & \textcolor{Blue}{100.00} & \textcolor{ForestGreen}{99.53} & 99.46 & \textcolor{ForestGreen}{99.53} & 99.46 \\
			& & \textbf{Kappa}     & 99.33 & 99.33 & \textcolor{BurntOrange}{99.47} & 99.33 & 99.33 & \textcolor{ForestGreen}{99.40} & \textcolor{Blue}{100.00} & 99.33 & 99.27 & 99.33 & 99.27 \\
			& & \textbf{F1-score}  & 99.50 & 99.50 & 99.54 & 99.50 & 99.50 & \textcolor{BurntOrange}{99.56} & \textcolor{Blue}{100.00} & 99.50 & 99.44 & 99.50 & 99.44 \\ \cline{2-14}
			& \multirow{4}{*}{\textbf{DBCTNet}} & \textbf{OA}        & 94.97 & 96.89 & 99.37 & 98.87 & 99.26 & 97.91 & 99.20 & \textcolor{BurntOrange}{99.71} & \textcolor{ForestGreen}{99.60} & 99.43 & \textcolor{Blue}{99.88} \\
			& & \textbf{Precision} & 95.42 & 97.74 & 99.33 & 98.91 & 99.30 & 97.90 & 99.18 & \textcolor{BurntOrange}{99.69} & \textcolor{ForestGreen}{99.63} & 99.53 & \textcolor{Blue}{99.89} \\
			& & \textbf{Kappa}     & 94.12 & 96.35 & 99.27 & 98.67 & 99.14 & 97.55 & 99.07 & \textcolor{BurntOrange}{99.66} & \textcolor{ForestGreen}{99.53} & 99.33 & \textcolor{Blue}{99.86} \\
			& & \textbf{F1-score}  & 95.16 & 96.90 & 99.40 & 98.84 & 99.33 & 97.94 & 99.14 & \textcolor{BurntOrange}{99.71} & \textcolor{ForestGreen}{99.62} & 99.50 & \textcolor{Blue}{99.89} \\ \cline{2-14}
			& \multirow{4}{*}{\textbf{FAHM}} & \textbf{OA}        & 99.37 & \textcolor{BurntOrange}{99.66} & 99.37 & 99.43 & 99.43 & \textcolor{ForestGreen}{99.60} & \textcolor{Blue}{99.94} & 99.43 & 99.43 & 99.32 & 99.43 \\
			& & \textbf{Precision} & 99.48 & \textcolor{BurntOrange}{99.65} & 99.47 & 99.53 & 99.51 & \textcolor{ForestGreen}{99.61} & \textcolor{Blue}{99.93} & 99.49 & 99.53 & 99.41 & 99.49 \\
			& & \textbf{Kappa}     & 99.27 & \textcolor{BurntOrange}{99.60} & 99.27 & 99.33 & 99.33 & \textcolor{ForestGreen}{99.53} & \textcolor{Blue}{99.93} & 99.33 & 99.33 & 99.20 & 99.33 \\
			& & \textbf{F1-score}  & 99.45 & \textcolor{BurntOrange}{99.67} & 99.44 & 99.50 & 99.49 & \textcolor{ForestGreen}{99.62} & \textcolor{Blue}{99.93} & 99.48 & 99.50 & 99.37 & 99.48 \\ \bottomrule
		\end{tabular}
	}
	\label{tab:table1}
    % \vspace{-4mm}
\end{table*}

%%%%%%%%%%%%%%%%%%%%%%%%%%%%%%%%%%%%%%%%%%%%%%%%%%%%%%%%%%%%%%%%%%%%%%%%%%%%%%%%%%%%%%%%%%%%%%%%%%%%%%%%%%%%%%%%%%%%%%%%%%%%%%%%%%

\section{Experiments}
\label{sec:experiments}
In this section, we evaluate our HyperCap dataset for classification by benchmarking it against state-of-the-art image and text encoders. Specifically, DBCTNet \cite{DBCTNet}, FAHM \cite{FAHM}, 3D-ConvSST \cite{3DConvSST}, and 3D-RCNet \cite{3DRC-Net} are utilized for HSI feature extraction, while BERT-Large-Uncased \cite{hosseini2023bert} and T5 \cite{t5} serve as pretrained text encoders to align spectral data with semantic representations. Experiments on four benchmark datasets—Indian Pines, Houston13, KSC, and Botswana—demonstrate that captions not only enhance classification performance but also help mitigate class imbalance as observed in the dataset analysis in subsection~\ref{sec:data_analysis}. This highlights the potential of captions in improving model robustness and fairness across under-represented classes.

To conduct a rigorous classification assessment of HyperCap, we integrate five fusion techniques—Cross Attention (CA), Concatenation (CONCAT), Multi-Head Attention (MHA), Pixel-Wise Addition (PWA), and Pixel-Wise Multiplication (PWM)—each designed to enhance the fusion of spectral and textual information. A structured pipeline was developed to ensure a fair comparison by training all baseline vision models and recording their scores in the `Vision' column of Tables~\ref{tab:table1}, \ref{tab:table2}. The integration of vision models with text encoders resulted in notable performance improvements across architectures. For performance evaluation, we utilize Overall Accuracy (OA) to measure classification effectiveness across all classes, Precision to assess the reliability of positive predictions, F1-Score to balance precision and recall, and Kappa Score to quantify classification agreement beyond chance. Our experiments are conducted on the proposed HyperCap benchmark dataset to ensure robust generalization and semantic understanding. 

We also evaluate captioning with five models; results are in Tables~\ref{tab:table5}-\ref{tab:table6}, using BLUE, METEOR, and ROUGE-L. Since the original image encoders in these models were not designed to process HSI data, we adapted them by replacing their image encoders with FAHM~\cite{FAHM} to ensure compatibility and optimal performance.

\subsection{Experimental Setup}
The experiments were conducted on a system equipped with an Intel(R) Xeon(R) Silver 4214R CPU @ 2.40\,GHz and three NVIDIA RTX A30 GPUs, each with \textbf{24\,GB} of VRAM. All experiments follow a standard supervised learning protocol with random train/validation/test splits performed at the pixel-index level. Each pixel and its corresponding caption are treated as an independent sample, and splits are generated via random shuffling at the pixel level without regard to which specific pixels belong to which class. This ensures that no test pixel's caption appears in the training set, even when train and test pixels share the same class label, requiring the classifier to generalize to wholly unseen captions at test time. For the classification task, the dataset was split into \textbf{10\%} for training, \textbf{10\%} for validation, and \textbf{80\%} for testing, following standard practices in HSI classification where models are typically trained on limited data due to the scarcity of labelled spectral samples across numerous bands. Models were trained for \textbf{50} epochs, and the best checkpoint based on validation performance was selected for final evaluation. Optimization was performed using the Adam optimizer with a learning rate of \textit{1E-4}, ensuring stable training across all encoders. The Cross-Entropy loss function was employed to minimize classification error and promote model convergence. For the captioning task, the dataset was divided into \textbf{70\%} for training, \textbf{10\%} for validation, and \textbf{20\%} for testing, in accordance with general practice for vision-language captioning benchmarks. The same training methodology was adopted, utilizing the official codebase released by the original authors to maintain consistency in experimental settings. Models were trained for up to \textbf{50} epochs, with early stopping applied to prevent overfitting. In the result tables, the highest scores are highlighted in \textcolor{Blue}{\textbf{Blue}}, the second-highest in \textcolor{BurntOrange}{\textbf{Orange}}, and the third-highest in \textcolor{ForestGreen}{\textbf{Green}}.

\subsection{Results and Discussion}\label{sec:result}
%%%%%%%%%%%%%%%%%%%%%%%%%%%%%%%%%%%%%
{In Table~\ref{tab:table1}, the classification results for the Botswana dataset exhibit significant gains when text encoders are used. 3D-RCNet with PWA-T5 achieves 99.86\% OA (\(+13.27\%\)) and 99.82\% Precision (\(+9.35\%\)), confirming PWA's spatial modeling strength. 3D-ConvSST with CONCAT-T5, MHA-BERT, MHA-T5, and PWA-T5, as well as FAHM with various types of fusions reach 100\% across all metrics (from 99.95\% baseline). DBCTNet with PWM-BERT improves from 75.95\% to 99.56\% OA and from 65.32\% to 99.43\% Precision. For the Houston13 dataset, 3D-RCNet with CA-BERT lifts OA from 97.45\% to 99.77\% (\(+2.32\%\)), Precision from 97.64\% to 99.75\%, and F1 from 97.53\% to 99.78\%. MHA-T5 records the best F1 (99.94\%). 3D-ConvSST with MHA-T5 achieves 100\% from a 99.43\% OA baseline. DBCTNet with PWM-T5 improves OA from 94.97\% to 99.88\%, and FAHM with MHA-T5 achieves 99.94\% OA and 99.93\% F1, reflecting strong caption-text fusion benefits.}

\begin{figure*}[t]
    \centering
        \begin{subfigure}[b]{0.3\textwidth}
            \centering
          \includegraphics[width=0.18\linewidth, trim=700 50 250 60, clip, angle=90]{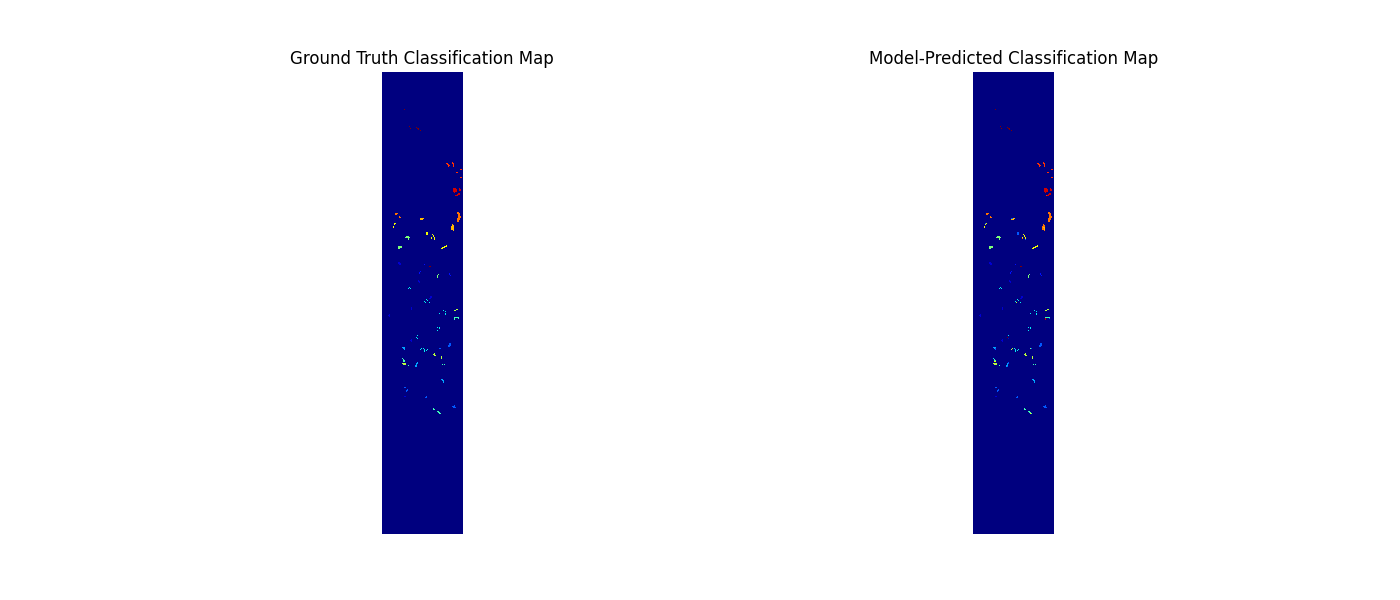}
        \caption{3D-RCNet}
    \end{subfigure}
        \begin{subfigure}[b]{0.3\textwidth}
            \centering
         \includegraphics[width=0.18\linewidth, trim=700 50 250 60, clip, angle=90]{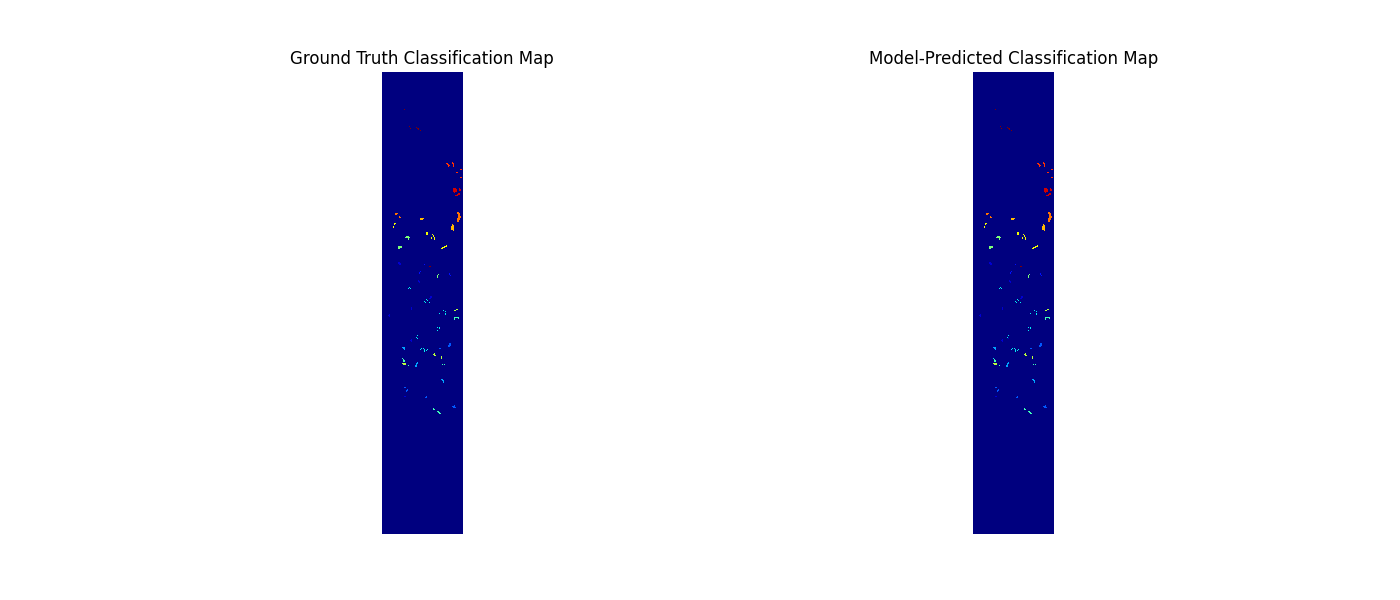}
        \caption{3D-ConvSST}
    \end{subfigure}
        \begin{subfigure}[b]{0.3\textwidth}
            \centering
         \includegraphics[width=0.18\linewidth, trim=700 50 250 60, clip, angle=90]{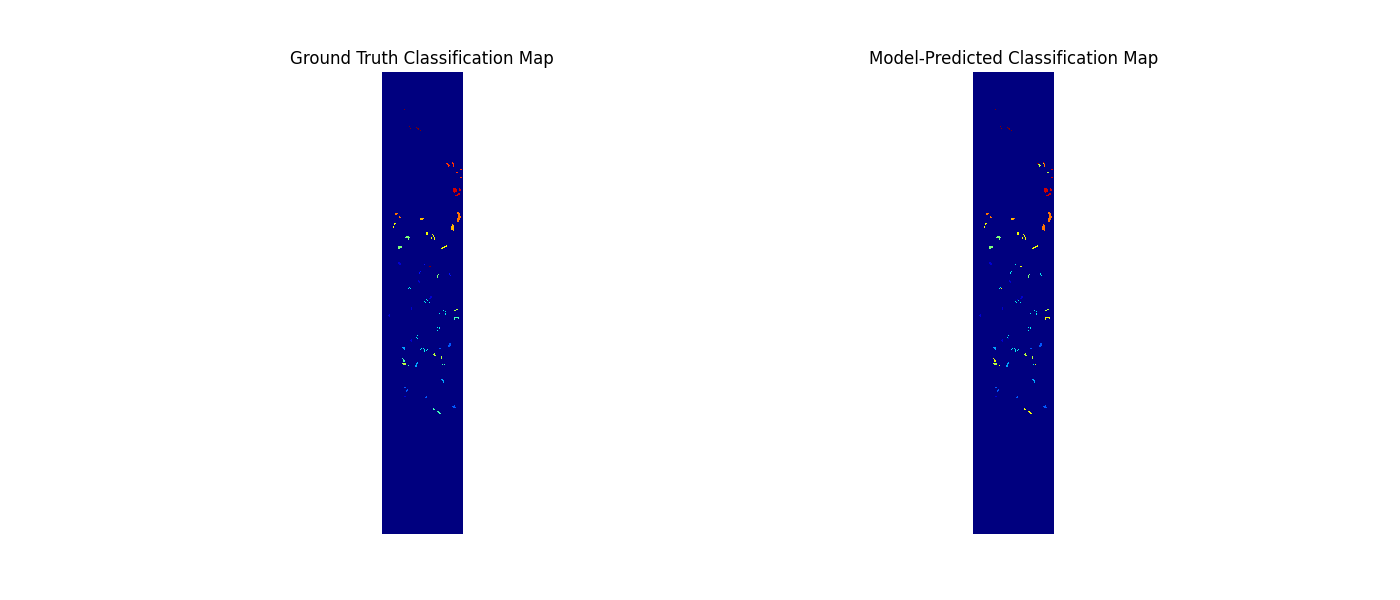}
        \caption{DBCTNet}
    \end{subfigure}
        \begin{subfigure}[b]{0.3\textwidth}
            \centering
         \includegraphics[width=0.18\linewidth, trim=700 50 250 60, clip, angle=90]{Output_Map/Botswana/FAHM-None/vision_prediction_map_run1.png}
        \caption{FAHM}
    \end{subfigure}
        \begin{subfigure}[b]{0.3\textwidth}
            \centering
        \includegraphics[width=0.18\linewidth, trim=275 50 675 60, clip, angle=90]{Output_Map/Botswana/FAHM-None/vision_prediction_map_run1.png}
        \caption{GT}
    \end{subfigure}
    \caption{Comparison of classification maps for the Botswana dataset, showing different maps: 3D-RCNet, 3D-ConvSST, DBCTNet, FAHM and Ground Truth (GT).}
    \vspace{-3mm}
    \label{VO-Botswana}
\end{figure*}

\begin{figure*}[t]
    \centering
    % \vspace{-4mm}
        \begin{subfigure}[b]{0.3\textwidth}
            \centering
          \includegraphics[width=0.18\linewidth, trim=700 50 250 60, clip, angle=90]{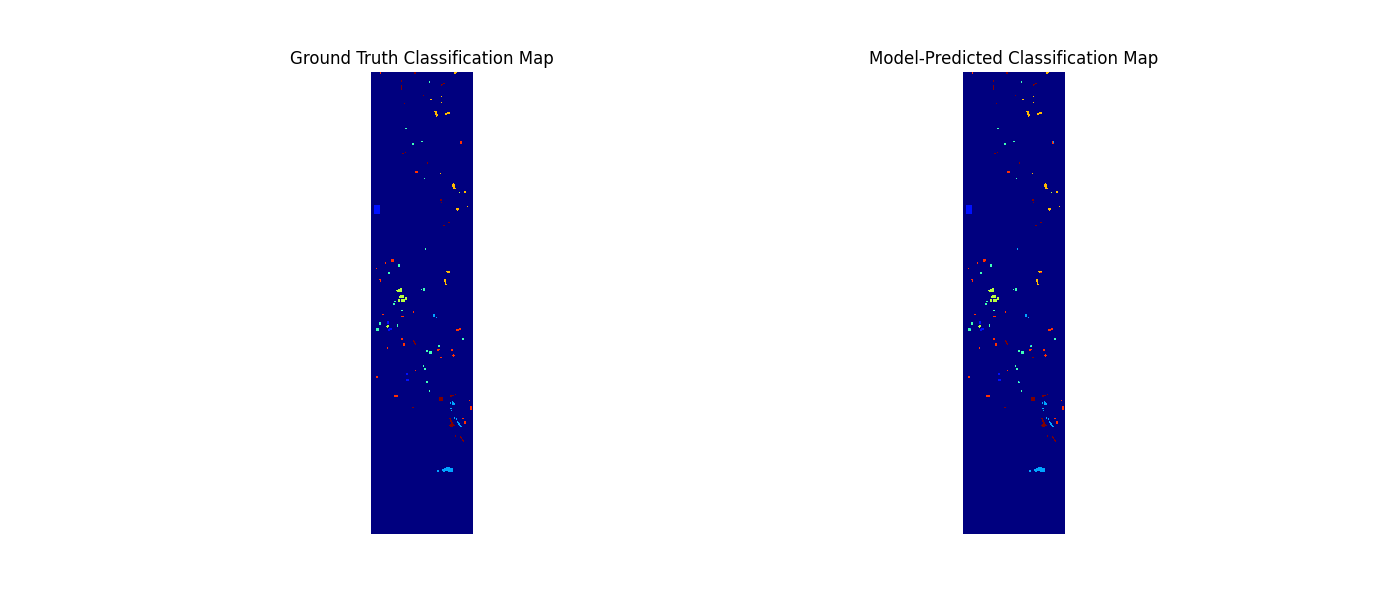}
        \caption{3D-RCNet}
    \end{subfigure}
        \begin{subfigure}[b]{0.3\textwidth}
            \centering
         \includegraphics[width=0.18\linewidth, trim=700 50 250 60, clip, angle=90]{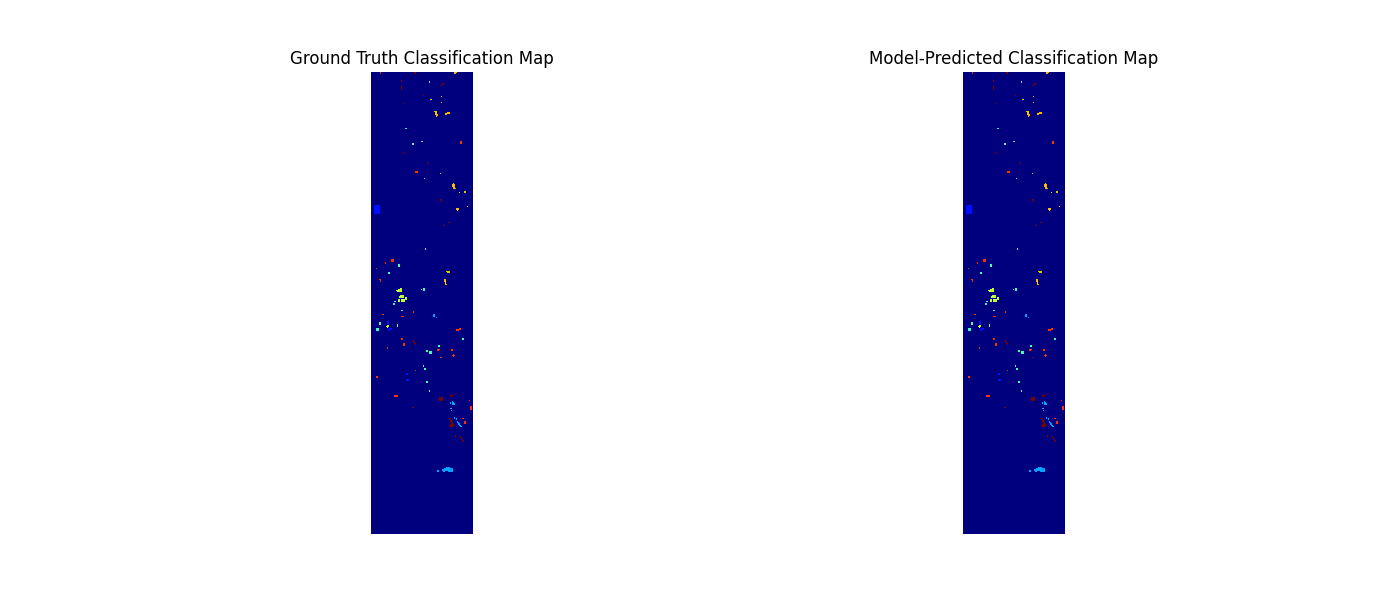}
        \caption{3D-ConvSST}
    \end{subfigure}
        \begin{subfigure}[b]{0.3\textwidth}
            \centering
         \includegraphics[width=0.18\linewidth, trim=700 50 250 60, clip, angle=90]{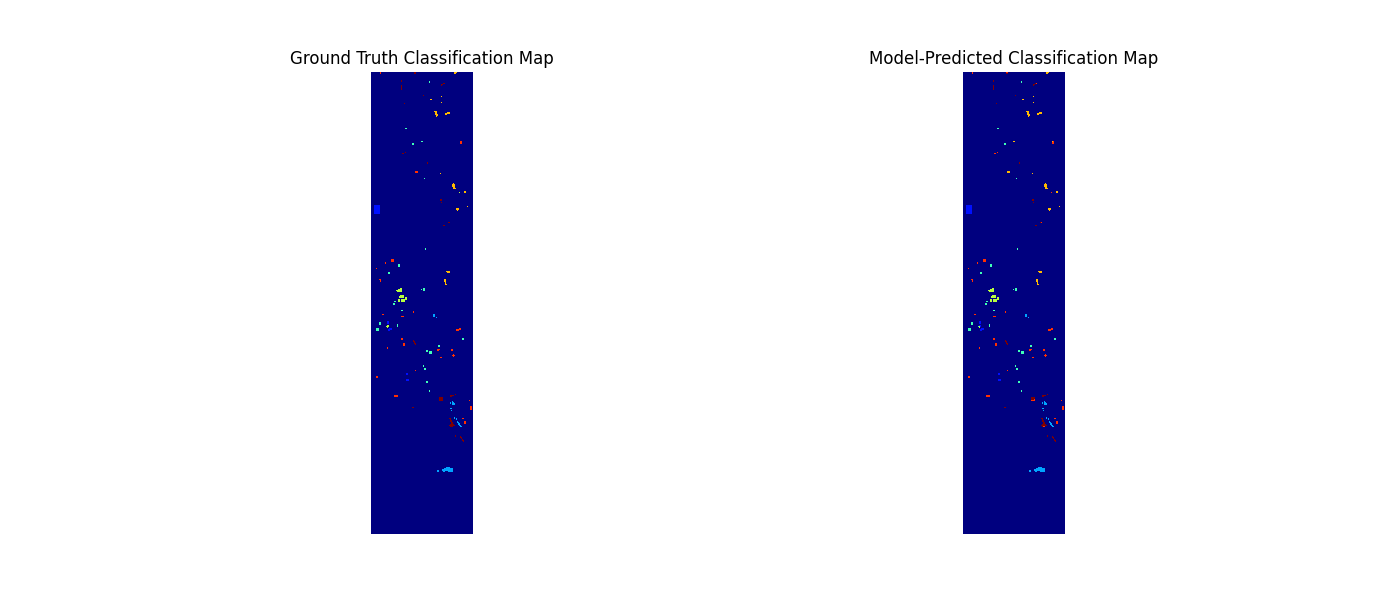}
        \caption{DBCTNet}
    \end{subfigure}
        \begin{subfigure}[b]{0.3\textwidth}
            \centering
         \includegraphics[width=0.18\linewidth, trim=700 50 250 60, clip, angle=90]{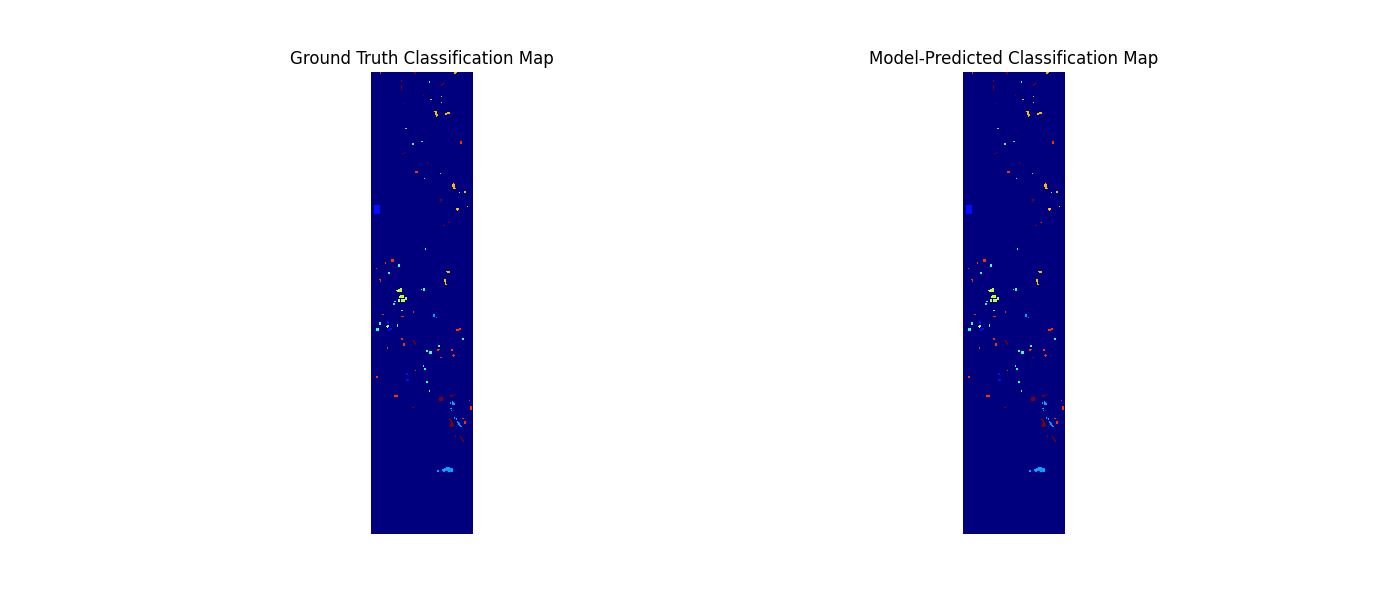}
        \caption{FAHM}
    \end{subfigure}
        \begin{subfigure}[b]{0.3\textwidth}
            \centering
        \includegraphics[width=0.18\linewidth, trim=275 50 675 60, clip, angle=90]{Output_Map/Houston13/FAHM-None/vision_prediction_map_run1.png}
        \caption{GT}
    \end{subfigure}
    \caption{Comparison of classification maps for the Houston13 dataset, showing different maps: 3D-RCNet, 3D-ConvSST, DBCTNet, FAHM and Ground Truth (GT).}
    \vspace{-3mm}
    \label{VO-Houston13}
\end{figure*}

In Table~\ref{tab:table2}, the Indian Pines dataset benefits from substantial gains in classification accuracy across models. 3D-RCNet with MHA-BERT achieves 99.83\% OA (\(+17.74\%\)) and 99.69\% Precision (\(+7.52\%\)), highlighting MHA’s effectiveness. DBCTNet with PWM-T5 shows the highest jump: OA from 76.01\% to 99.37\% (\(+23.36\%\)), Precision from 43.68\% to 99.41\% (\(+55.73\%\)), and F1 from 43.21\% to 97.92\% (\(+54.71\%\)). FAHM reaches 99.91\% OA with PWM-T5, and 3D-ConvSST with MHA-T5 improves OA from 98.80\% to 99.90\%. CONCAT-BERT also performs well (99.88\% OA, 99.90\% Precision). On the KSC dataset, CONCAT-BERT with 3D-RCNet boosts OA from 77.05\% to 96.57\% (\(+19.52\%\)), Precision from 73.95\% to 95.25\% (\(+21.3\%\)), and F1 from 59.01\% to 94.04\% (\(+35.03\%\)). 3D-ConvSST with MHA-T5 improves OA from 71.87\% to 88.95\% (\(+17.08\%\)), Precision from 46.78\% to 94.90\% (\(+48.12\%\)). DBCTNet with PWA-T5 lifts OA from 70.50\% to 97.58\% (\(+27.08\%\)), and Precision from 55.13\% to 97.04\% (\(+41.91\%\)). FAHM, starting at 99.78\% OA, achieves 100\% with MHA-T5, showing even top models benefit from textual fusion.

%%%%%%%%%%%%%%%%%%%%%%%%%%%%%%%%%%%%%%%%%%%%%%%%%%%%%%%%%%%%%%%%%%%%%%%%%%%%%%%%%%%%%%%%%%%%%%%%%%%%%%%%%%%%%%%%%%%%%%%%%%%%%%%%%%%%%%%%%%%%%%%%%%%%%%%%%%%%%%%%%%%%%%%%%%%

\begin{table*}[!t]
    % \vspace{-2mm}
	\large
	\centering
    	\caption{Evaluating Vision Encoders with/without Text Encoders on Indian Pines \& KSC Datasets.}
	\resizebox{1\linewidth}{!}{%
		\begin{tabular}{c | l | l | c | c | c | c | c | c | c | c | c | c | c}
			\toprule
			                 & \multirow{3}{*}{\textbf{Vision Model}} & \multirow{3}{*}{\textbf{Metric}} & \textbf{IMG ONLY}  & \multicolumn{10}{c}{\textbf{IMG+TXT}} \\
			\cline{4-14}
			\textbf{DATASET} & & &\multirow{2}{*}{\textbf{Vision}}  & \multicolumn{2}{c|}{\textbf{CA}} & \multicolumn{2}{c|}{\textbf{CONCAT}} & \multicolumn{2}{c|}{\textbf{MHA}} & \multicolumn{2}{c|}{\textbf{PWA}} & \multicolumn{2}{c}{\textbf{PWM}} \\ \cline{5-14}
			& & & & \textbf{Bert} & \textbf{T5} & \textbf{Bert} & \textbf{T5} & \textbf{Bert} & \textbf{T5} & \textbf{Bert} & \textbf{T5} & \textbf{Bert} & \textbf{T5} \\ \hline
            
			\multirow{16}{*}{{\rotatebox[origin=c]{90}{\textbf{INDIAN PINES}}}} & \multirow{4}{*}{\textbf{3D-RCNet}} & \textbf{OA}        & 82.09 & 98.39 & 99.05 & \textcolor{BurntOrange}{99.48} & \textcolor{ForestGreen}{99.23} & \textcolor{Blue}{99.83} & 99.05 & 98.28 & 98.92 & 98.04 & 99.12 \\
			& & \textbf{Precision} & 92.17 & 98.10 & \textcolor{ForestGreen}{99.36} & \textcolor{BurntOrange}{99.47} & 98.70 & \textcolor{Blue}{99.69} & 99.13 & 98.64 & 98.54 & 95.35 & 99.10 \\
			& & \textbf{Kappa}     & 79.17 & 98.17 & 98.91 & \textcolor{BurntOrange}{99.41} & \textcolor{ForestGreen}{99.12} & \textcolor{Blue}{99.80} & 98.92 & 98.04 & 98.77 & 97.77 & 98.99 \\
			& & \textbf{F1-score}  & 86.68 & 97.19 & 99.05 & \textcolor{BurntOrange}{99.15} & 98.88 & \textcolor{Blue}{99.70} & \textcolor{ForestGreen}{99.09} & 98.51 & 98.75 & 96.42 & 98.98 \\ \cline{2-14}
			& \multirow{4}{*}{\textbf{3D-ConvSST}} & \textbf{OA}        & 98.80 & 99.72 & 99.69 & \textcolor{BurntOrange}{99.88} & 99.77 & 99.80 & \textcolor{Blue}{99.90} & \textcolor{ForestGreen}{99.81} & 99.76 & 99.10 & 99.70 \\
			& & \textbf{Precision} & 99.22 & 99.70 & 99.55 & \textcolor{BurntOrange}{99.90} & 99.76 & 99.76 & \textcolor{Blue}{99.95} & \textcolor{ForestGreen}{99.84} & 99.72 & 99.31 & 99.68 \\
			& & \textbf{Kappa}     & 98.63 & 99.68 & 99.65 & \textcolor{BurntOrange}{99.87} & 99.74 & 99.77 & \textcolor{Blue}{99.88} & \textcolor{ForestGreen}{99.79} & 99.72 & 98.98 & 99.66 \\
			& & \textbf{F1-score}  & 97.06 & 99.41 & 99.50 & \textcolor{Blue}{99.87} & 99.72 & 99.60 & \textcolor{BurntOrange}{99.85} & \textcolor{ForestGreen}{99.74} & 99.65 & 98.55 & 99.44 \\ \cline{2-14}
			& \multirow{4}{*}{\textbf{DBCTNet}} & \textbf{OA}        & 76.01 & 96.27 & 97.15 & 98.66 & 98.34 & 96.45 & 97.82 & \textcolor{BurntOrange}{99.03} & 98.64 & \textcolor{ForestGreen}{98.82} & \textcolor{Blue}{99.37} \\
			& & \textbf{Precision} & 43.68 & 78.69 & 78.79 & 86.00 & 92.25 & 77.86 & \textcolor{ForestGreen}{97.70} & 86.26 & 92.35 & \textcolor{BurntOrange}{98.72} & \textcolor{Blue}{99.41} \\
			& & \textbf{Kappa}     & 71.85 & 95.74 & 96.75 & 98.47 & 98.10 & 95.95 & 97.51 & \textcolor{BurntOrange}{98.90} & 98.45 & \textcolor{ForestGreen}{98.66} & \textcolor{Blue}{99.28} \\
			& & \textbf{F1-score}  & 43.21 & 77.11 & 78.79 & 82.16 & 89.23 & 74.99 & \textcolor{ForestGreen}{91.19} & 85.09 & 89.29 & \textcolor{BurntOrange}{96.69} & \textcolor{Blue}{97.92} \\ \cline{2-14}
			& \multirow{4}{*}{\textbf{FAHM}} & \textbf{OA}        & 98.45 & 99.67 & 99.77 & 98.64 & 99.76 & 99.74 & \textcolor{BurntOrange}{99.86} & 99.56 & 99.79 & \textcolor{ForestGreen}{99.84} & \textcolor{Blue}{99.91} \\
			& & \textbf{Precision} & 98.12 & 99.76 & 99.83 & 99.18 & 99.81 & 99.62 & \textcolor{Blue}{99.90} & 99.69 & 99.36 & \textcolor{BurntOrange}{99.85} & \textcolor{ForestGreen}{99.89} \\
			& & \textbf{Kappa}     & 98.23 & 99.63 & 99.74 & 98.45 & 99.72 & 99.71 & \textcolor{BurntOrange}{99.84} & 99.50 & 99.76 & \textcolor{ForestGreen}{99.82} & \textcolor{Blue}{99.90} \\
			& & \textbf{F1-score}  & 97.72 & 99.01 & 99.35 & 98.93 & \textcolor{Blue}{99.74} & 99.25 & 98.87 & \textcolor{ForestGreen}{99.45} & 99.40 & \textcolor{BurntOrange}{99.61} & 99.35 \\ \hline
            
			\multirow{16}{*}{{\rotatebox[origin=c]{90}{\textbf{KENNEDY SPACE CENTER}}}} & \multirow{4}{*}{\textbf{3D-RCNet}} & \textbf{OA}        & 77.05 & \textcolor{ForestGreen}{94.51} & 94.27 & \textcolor{Blue}{96.57} & 91.74 & 90.78 & \textcolor{BurntOrange}{95.55} & 91.69 & 87.71 & 83.08 & 92.90 \\
			& & \textbf{Precision} & 73.95 & \textcolor{ForestGreen}{92.65} & 92.35 & \textcolor{Blue}{95.25} & 90.46 & 92.29 & \textcolor{BurntOrange}{94.61} & 90.82 & 89.47 & 85.25 & 91.97 \\
			& & \textbf{Kappa}     & 74.36 & \textcolor{ForestGreen}{93.89} & 93.62 & \textcolor{Blue}{96.18} & 90.81 & 89.73 & \textcolor{BurntOrange}{95.05} & 90.75 & 86.18 & 81.13 & 92.06 \\
			& & \textbf{F1-score}  & 59.01 & \textcolor{ForestGreen}{90.47} & 90.39 & \textcolor{Blue}{94.04} & 89.03 & 87.19 & \textcolor{BurntOrange}{92.20} & 85.04 & 80.58 & 66.90 & 86.17 \\ \cline{2-14}
			& \multirow{4}{*}{\textbf{3D-ConvSST}} & \textbf{OA}        & 71.87 & 69.49 & \textcolor{ForestGreen}{75.60} & 72.75 & \textcolor{BurntOrange}{80.23} & 67.90 & \textcolor{Blue}{88.95} & 71.65 & 70.99 & 64.19 & 57.83 \\
			& & \textbf{Precision} & 46.78 & 57.04 & 66.37 & 65.32 & \textcolor{BurntOrange}{79.90} & 50.61 & \textcolor{Blue}{94.90} & 64.63 & \textcolor{ForestGreen}{76.41} & 36.55 & 46.57 \\
			& & \textbf{Kappa}     & 68.44 & 65.30 & \textcolor{ForestGreen}{72.34} & 69.17 & \textcolor{BurntOrange}{77.49} & 63.43 & \textcolor{Blue}{87.55} & 67.80 & 66.91 & 59.01 & 51.27 \\
			& & \textbf{F1-score}  & 47.61 & 48.31 & \textcolor{ForestGreen}{56.92} & 54.98 & \textcolor{BurntOrange}{62.48} & 45.04 & \textcolor{Blue}{83.79} & 52.78 & 52.29 & 38.86 & 35.00 \\ \cline{2-14}
			& \multirow{4}{*}{\textbf{DBCTNet}} & \textbf{OA}        & 70.50 & 82.62 & 93.33 & \textcolor{ForestGreen}{96.79} & 95.06 & 89.47 & \textcolor{BurntOrange}{97.31} & 96.46 & \textcolor{Blue}{97.58} & 75.52 & 93.61 \\
			& & \textbf{Precision} & 55.13 & 76.23 & 91.92 & 95.82 & \textcolor{ForestGreen}{96.02} & 89.49 & \textcolor{BurntOrange}{96.72} & 95.95 & \textcolor{Blue}{97.04} & 70.71 & 92.73 \\
			& & \textbf{Kappa}     & 66.35 & 80.44 & 92.56 & \textcolor{ForestGreen}{96.42} & 94.48 & 88.25 & \textcolor{BurntOrange}{97.00} & 96.05 & \textcolor{Blue}{97.31} & 72.33 & 92.88 \\
			& & \textbf{F1-score}  & 47.01 & 65.73 & 88.48 & \textcolor{ForestGreen}{95.11} & 93.01 & 78.77 & \textcolor{BurntOrange}{95.71} & 95.02 & \textcolor{Blue}{96.07} & 60.87 & 88.66 \\ \cline{2-14}
			& \multirow{4}{*}{\textbf{FAHM}} & \textbf{OA}        & 99.78 & 99.61 & 99.83 & \textcolor{BurntOrange}{99.97} & 99.83 & 99.83 & \textcolor{Blue}{100.00} & \textcolor{ForestGreen}{99.89} & 99.69 & 99.83 & \textcolor{ForestGreen}{99.89} \\
			& & \textbf{Precision} & 99.68 & 99.52 & 99.74 & \textcolor{BurntOrange}{99.95} & 99.74 & 99.72 & \textcolor{Blue}{100.00} & \textcolor{ForestGreen}{99.83} & 99.63 & 99.79 & 99.81 \\
			& & \textbf{Kappa}     & 99.75 & 99.57 & 99.81 & \textcolor{BurntOrange}{99.96} & 99.81 & 99.81 & \textcolor{Blue}{100.00} & \textcolor{ForestGreen}{99.87} & 99.66 & 99.81 & \textcolor{ForestGreen}{99.87} \\
			& & \textbf{F1-score}  & 99.59 & 99.27 & 99.66 & \textcolor{BurntOrange}{99.95} & 99.66 & 99.72 & \textcolor{Blue}{100.00} & 99.77 & 99.51 & 99.72 & \textcolor{ForestGreen}{99.82} \\ \bottomrule
		\end{tabular}%
	}
	\label{tab:table2}
    \vspace{-3mm}
\end{table*}
 % Tables

\begin{figure*}[!h]
    \centering
    % \vspace{-5mm}
       \begin{subfigure}[b]{0.16\textwidth}
            \centering
          \includegraphics[width=1\linewidth, trim=565 50 120 60, clip, angle=90, clip, angle=90]{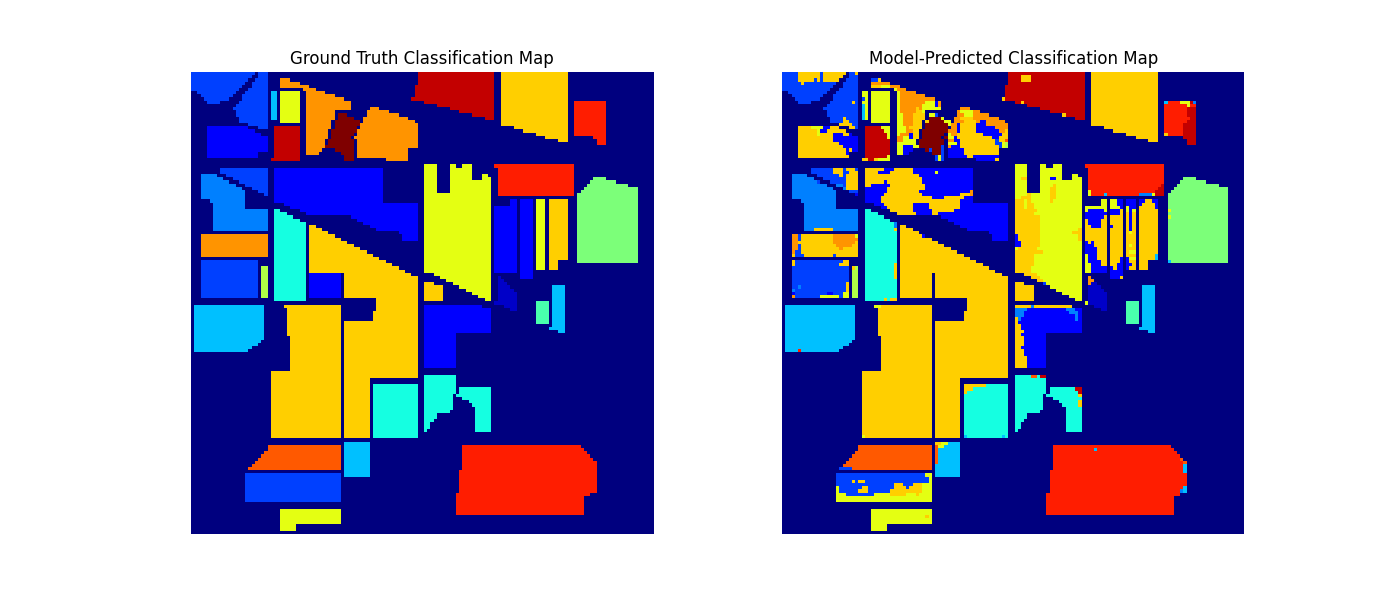}
        \caption{3D-RCNet}
    \end{subfigure}
       \begin{subfigure}[b]{0.16\textwidth}
            \centering
          \includegraphics[width=1\linewidth, trim=565 50 120 60, clip, angle=90]{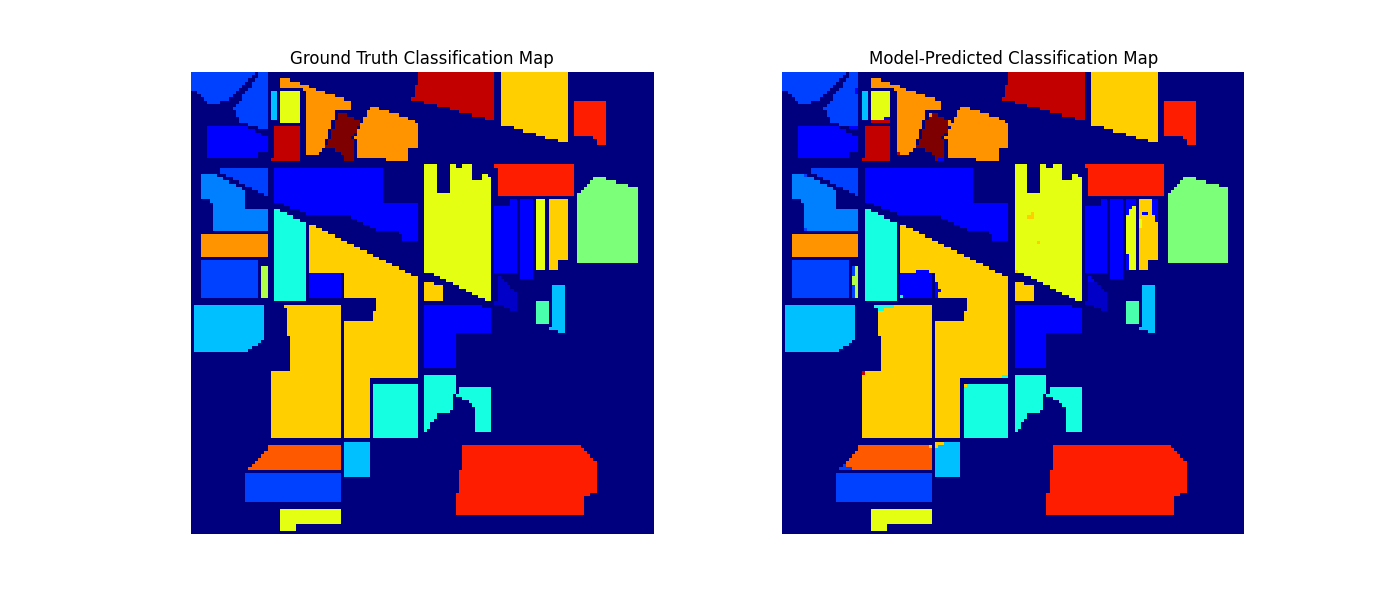}
        \caption{3D-ConvSST}
    \end{subfigure}
       \begin{subfigure}[b]{0.16\textwidth}
            \centering
          \includegraphics[width=1\linewidth, trim=565 50 120 60, clip, angle=90]{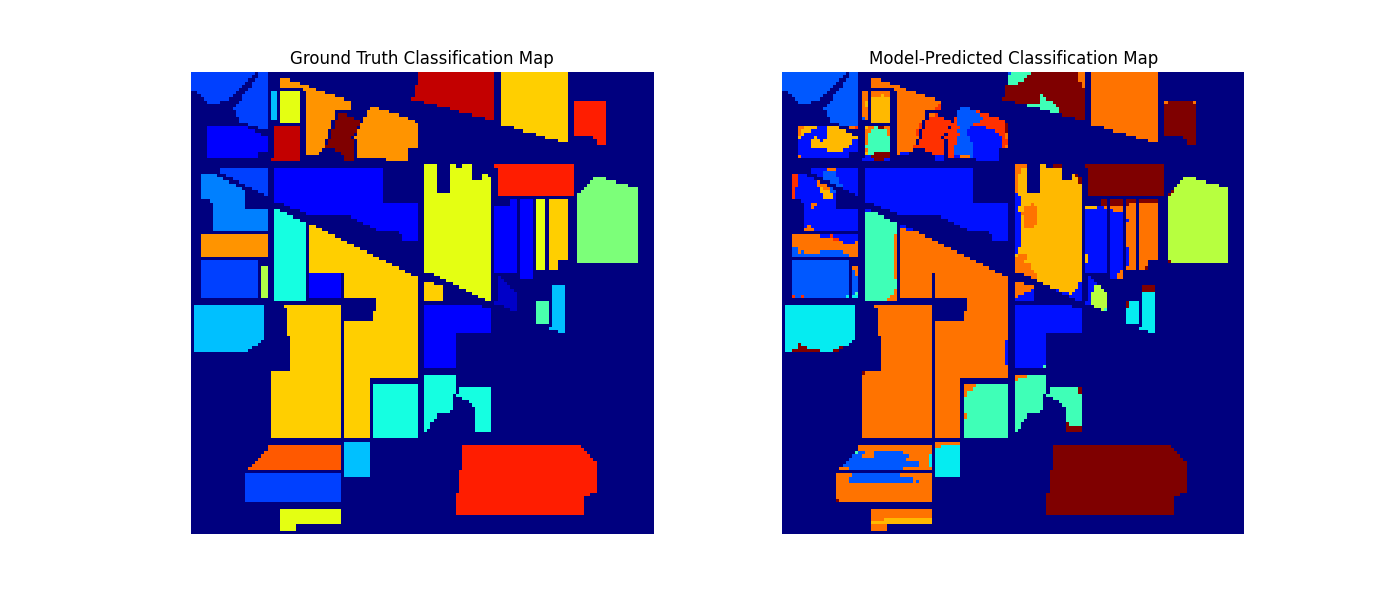}
        \caption{DBCTNet}
    \end{subfigure}
       \begin{subfigure}[b]{0.16\textwidth}
            \centering
          \includegraphics[width=1\linewidth, trim=565 50 120 60, clip, angle=90]{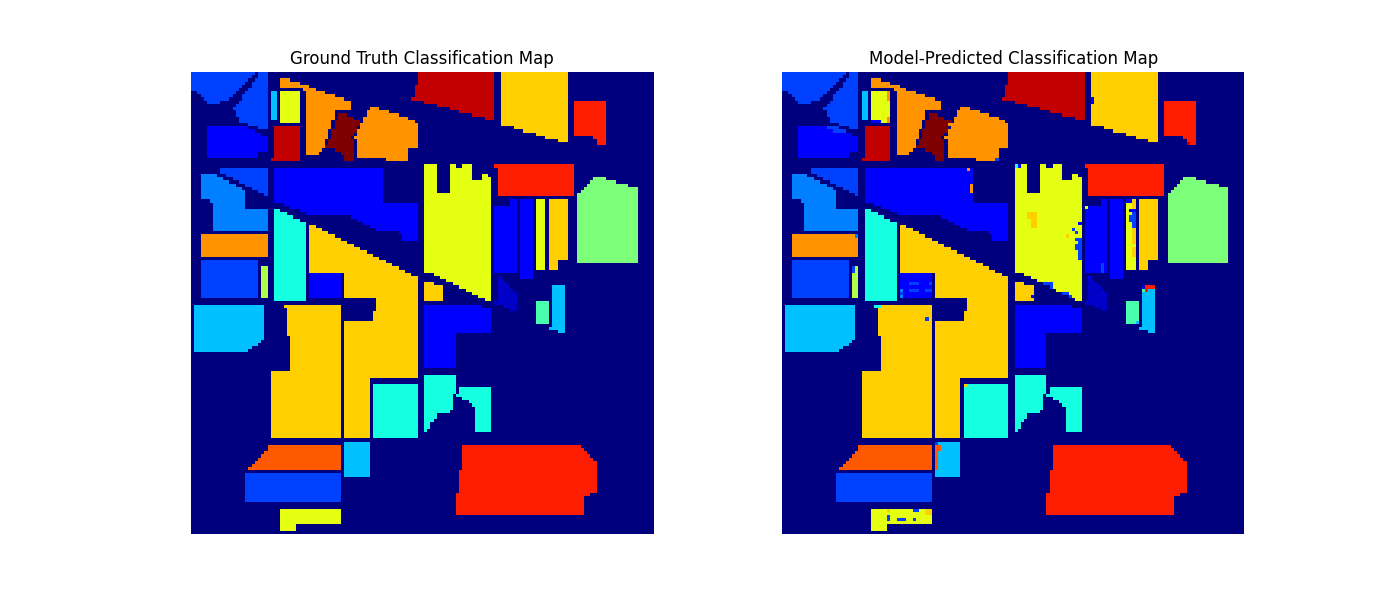}
        \caption{FAHM}
    \end{subfigure}
       \begin{subfigure}[b]{0.16\textwidth}
            \centering
          \includegraphics[width=1\linewidth, trim=130 70 540 70, clip, angle=90]{Output_Map/Indian_Pines/FAHM-None/vision_prediction_map_run1.png}
        \caption{GT}
    \end{subfigure}
    \caption{Comparison of classification maps for the Indian Pines dataset, showing different maps: 3D-RCNet, 3D-ConvSST, DBCTNet, FAHM and Ground Truth (GT).}
    % \vspace{-3mm}
    \label{VO-Indian Pines}
\end{figure*}

\begin{figure*}[!t]
    \centering
       \begin{subfigure}[b]{0.16\textwidth}
            \centering
          \includegraphics[width=1\linewidth, trim=575 70 120 70, clip, angle=90, clip, angle=90]{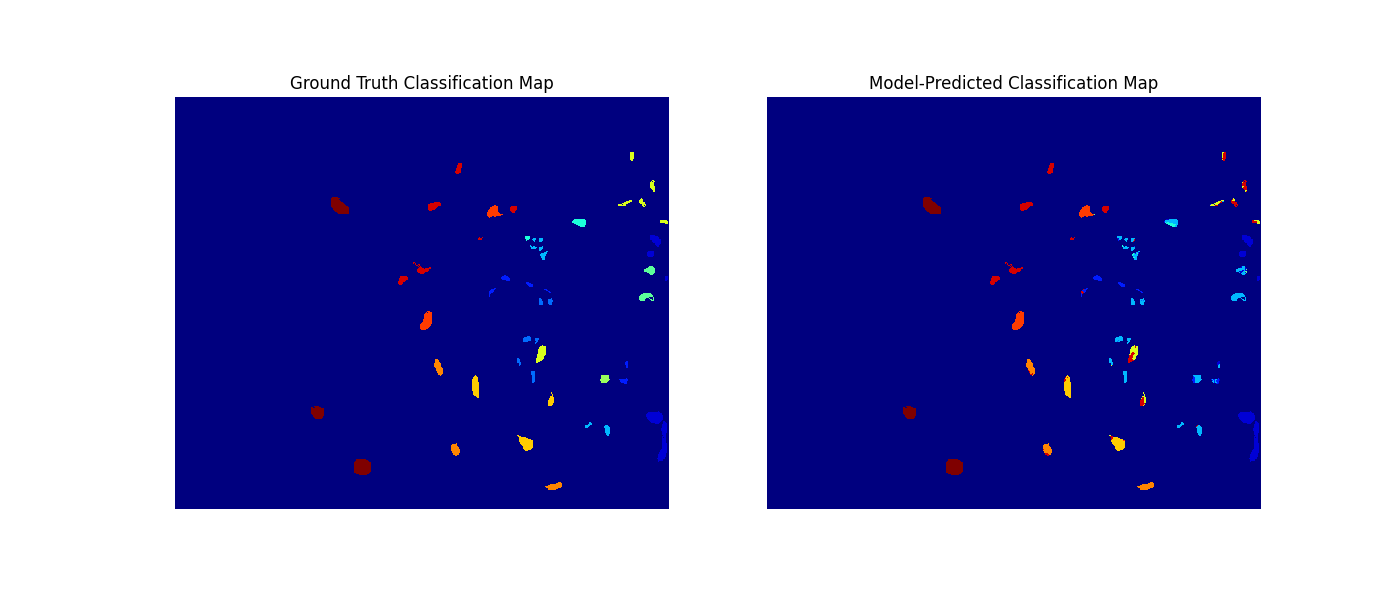}
        \caption{3D-RCNet}
    \end{subfigure}
       \begin{subfigure}[b]{0.16\textwidth}
            \centering
          \includegraphics[width=1\linewidth, trim=575 70 120 70, clip, angle=90, clip, angle=90]{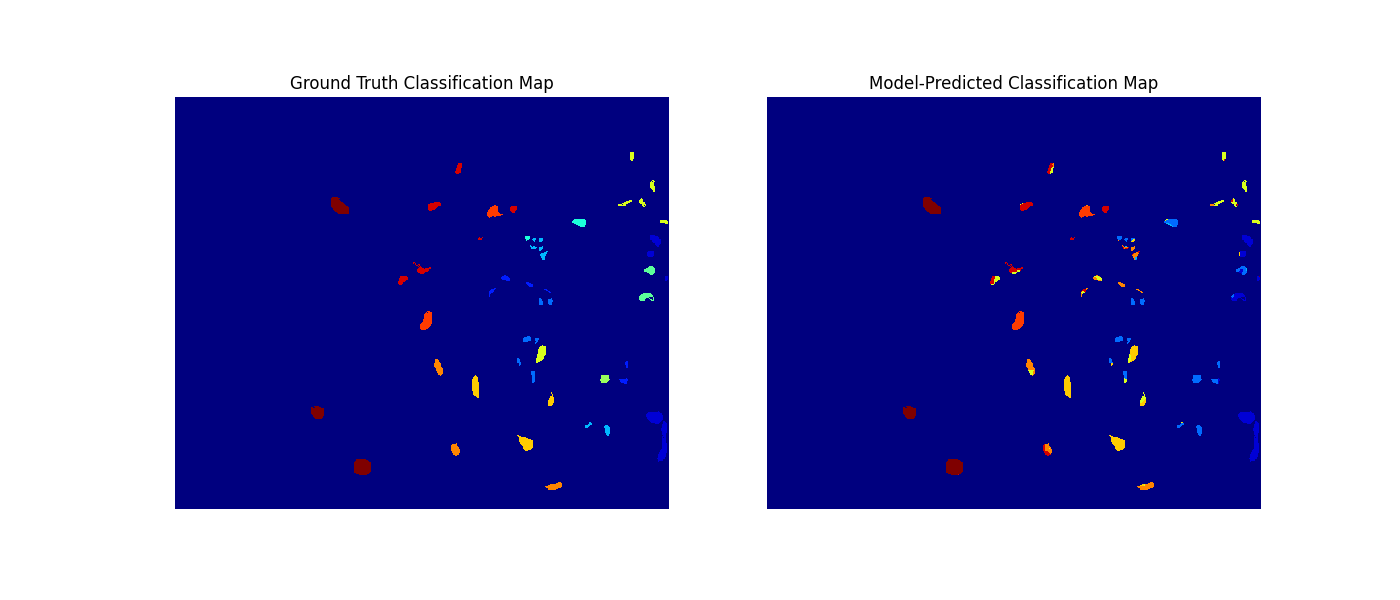}
        \caption{3D-ConvSST}
    \end{subfigure}
       \begin{subfigure}[b]{0.16\textwidth}
            \centering
          \includegraphics[width=1\linewidth, trim=575 70 120 70, clip, angle=90, clip, angle=90]{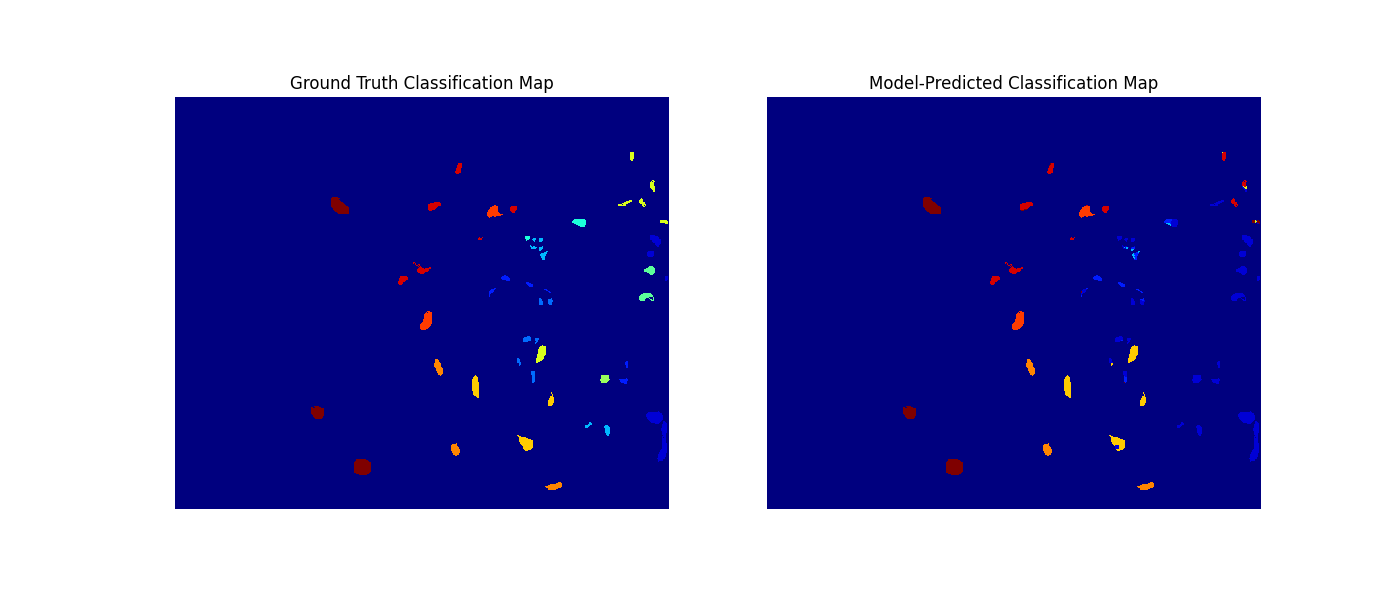}
        \caption{DBCTNet}
    \end{subfigure}
       \begin{subfigure}[b]{0.16\textwidth}
            \centering
          \includegraphics[width=1\linewidth, trim=575 70 120 70, clip, angle=90, clip, angle=90]{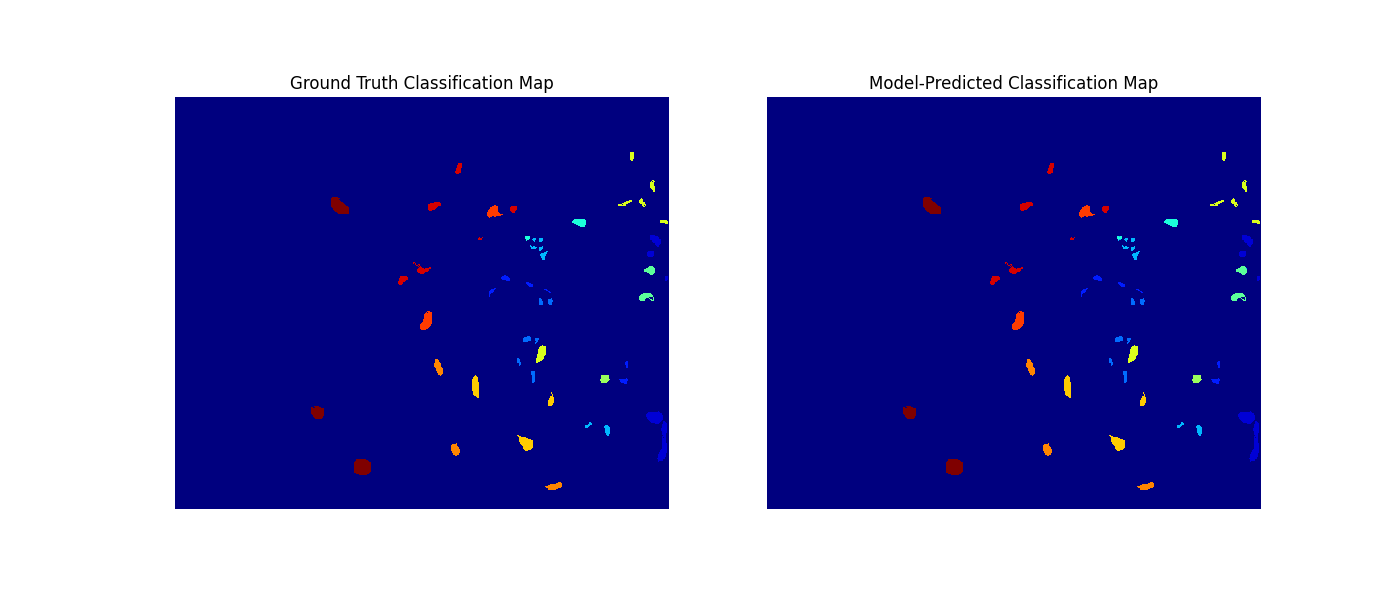}
        \caption{FAHM}
    \end{subfigure}
       \begin{subfigure}[b]{0.16\textwidth}
            \centering
          \includegraphics[height=0.94\textwidth, width=1\linewidth, trim=130 70 540 70, clip, angle=180]{Output_Map/KSC/FAHM-None/vision_prediction_map_run1.png}
        \caption{GT}
    \end{subfigure}
    \caption{Comparison of classification maps for the KSC dataset, showing different maps: 3D-RCNet, 3D-ConvSST, DBCTNet, FAHM and Ground Truth (GT).}
    \vspace{-3mm}
    \label{VO-KSC}
\end{figure*}

From Table~\ref{tab:table5}, GIT and mPLUG exhibit top performance on the Botswana dataset, with GIT achieving the highest BLEU-1 (0.4331) and mPLUG closely following (0.4291), reflecting strong unigram precision. mPLUG leads in METEOR (0.1390) and ROUGE-L (0.4225), indicating superior semantic alignment and fluency. On the Houston13 dataset, GIT attains the highest BLEU-1 (0.3637), BLEU-4 (0.2642), and METEOR (0.2096), along with a strong ROUGE-L (0.3570), showcasing its semantic richness and structural alignment. Table~\ref{tab:table6} confirms GIT’s dominance across BLEU-1 to BLEU-4, METEOR, and ROUGE-L on Indian Pines and Kennedy Space Center. Specifically, on Indian Pines, GIT (0.3816 BLEU-1) surpasses BLIP (0.3588) by \(\sim3\%\) and slightly outperforms mPLUG (0.3794), proving its robustness in captioning complex remote sensing scenes.

The benchmark results above demonstrate that integrating textual information through transformer-based text encoders significantly enhances classification accuracy, particularly in addressing data imbalance for minority classes, as discussed in subsection~\ref{sec:data_analysis}. 
Importantly, the modality ablation conducted under identical data splits (Image-only, Text-only, and Image+Text) further clarifies that the textual modality does not function as a direct encoding of class labels. As observed in Table~\ref{tab:ablation_study}, Text-only performance remains substantially lower than the corresponding multimodal configurations. If the captions were merely alternative representations of the target labels, Text-only models would be expected to approach multimodal performance. The consistent performance gap therefore indicates that captions provide complementary semantic information rather than introducing target leakage.
 The textual features further refine spectral representations, improving discrimination for minority classes and promoting more balanced predictions. Furthermore, certain vision encoders exhibit reduced agreement between class assignments, suggesting inconsistencies in feature extraction. This limitation is mitigated through the incorporation of textual information, which provides complementary semantic context and improves alignment with ground-truth labels across diverse datasets. The captioning benchmark additionally provides a comprehensive evaluation across datasets, where GIT and mPLUG emerge as the strongest performers, demonstrating their effectiveness in generating accurate and semantically grounded descriptions.

Across all four datasets, MHA and PWM fusion strategies consistently deliver the most robust classification performance regardless of the vision encoder employed. Moreover, T5-based text encoding generally outperforms BERT-Large on imbalanced datasets such as Indian Pines and KSC, suggesting that attention-based fusion with transformer text encoders represents the most practically reliable configuration for HSI vision--language classification.

Tables~\ref{tab:classfication_models} and~\ref{tab:Captioning_models} present a comparative analysis of model parameters and computational cost in floating point operations (FLOPs) for classification and captioning architectures, respectively. Table~\ref{tab:classfication_models} includes both unimodal and multimodal classification models, while Table~\ref{tab:Captioning_models} focuses on representative vision-language captioning frameworks.

%%%%%%%%%%%%%%%%%%%%%%%%%%%%%%%%%%

\begin{table*}[h]
\centering
% \vspace{-4mm}
\caption{Image Retrieval (IR) and Text Retrieval (TR) Results (\%) for Several Popular Vision-Language models on HyperCap Across All Considered Datasets.}
\label{tab:retrieval_results}
\resizebox{\textwidth}{!}{
\begin{tabular}{l|c|c|c|c|c|c|c|c}
\hline
\textbf{Model} & \textbf{Botswana IR} & \textbf{Botswana TR} & \textbf{Houston13 IR} & \textbf{Houston13 TR} & \textbf{IP IR} & \textbf{IP TR} & \textbf{KSC IR} & \textbf{KSC TR} \\
\hline
\textbf{BLIP~\cite{li2022blip}}       & 61.58 & 68.67 & 60.01 & 67.69 & 61.50 & 70.37 & 61.12 & 69.77 \\
\textbf{GIT~\cite{wang2022git}}        & 62.66 & 70.26 & 61.08 & 69.02 & 62.35 & 71.15 & 61.75 & 71.05 \\
\textbf{mPLUG~\cite{li2022mplug}}      & 62.96 & 69.92 & 60.60 & 68.31 & 62.45 & 71.25 & 61.46 & 70.96 \\
\textbf{VinVL~\cite{zhang2021vinvl}}      & 62.49 & 68.69 & 60.05 & 67.42 & 61.89 & 69.57 & 60.40 & 69.74 \\
\textbf{VisualBERT~\cite{li2019visualbertsimpleperformantbaseline}} & 60.62 & 67.48 & 59.05 & 66.50 & 60.38 & 69.13 & 59.34 & 69.07 \\
\hline
\end{tabular}
}
\vspace{-3mm}
\end{table*}
%%%%%%%%%%%%%%%%%%%%%%%%%%%%%%%%%%%%%%%%%%%%%%%%%%%%%%%%%%%%%%%%%%%%%%%%%%%%%%%%%%%%%%%%%%%%%% 

\begin{table*}[]
\centering
% \vspace{-4mm}
\caption{Performance of Captioning Models on Botswana and Houston13 Datasets: BLEU (B1–B4), METEOR (MTR), and ROUGE-L (R-L).}
% \vspace{2mm}
\fontsize{8}{8}\selectfont
\renewcommand{\arraystretch}{1}
\setlength{\tabcolsep}{3pt}
\begin{tabular}{l|cccccc|cccccc}
\hline
\multirow{2}{*}{\textbf{Model}} 
& \multicolumn{6}{c|}{\textbf{BOTSWANA}} 
& \multicolumn{6}{c}{\textbf{HOUSTON13}} \\ \cline{2-13}
& B1 & B2 & B3 & B4 & MET & R-L 
& B1 & B2 & B3 & B4 & MET & R-L \\ \hline

\textbf{BLIP~\cite{li2022blip}}       
  & \textcolor{ForestGreen}{0.4036} 
  & \textcolor{ForestGreen}{0.3747} 
  & \textcolor{ForestGreen}{0.3603} 
  & \textcolor{ForestGreen}{0.3536} 
  & \textcolor{ForestGreen}{0.1197} 
  & \textcolor{ForestGreen}{0.3945}
  & \textcolor{ForestGreen}{0.3385}
  & 0.2853
  & 0.2538
  & {0.2368}
  & 0.1861
  & 0.3321 \\

\textbf{GIT~\cite{wang2022git}}       
  & \textcolor{Blue}{0.4331} 
  & \textcolor{BurntOrange}{0.3980} 
  & \textcolor{Blue}{0.3899} 
  & \textcolor{Blue}{0.3828} 
  & \textcolor{Blue}{0.1423} 
  & \textcolor{BurntOrange}{0.4167}
  & \textcolor{Blue}{0.3637}
  & \textcolor{BurntOrange}{0.3077}
  & \textcolor{BurntOrange}{0.2755}
  & \textcolor{Blue}{0.2642}
  & \textcolor{Blue}{0.2106}
  & \textcolor{Blue}{0.3570} \\

\textbf{mPlug~\cite{li2022mplug}}       
  & \textcolor{BurntOrange}{0.4291} 
  & \textcolor{Blue}{0.4024} 
  & \textcolor{BurntOrange}{0.3867} 
  & \textcolor{BurntOrange}{0.3762} 
  & \textcolor{BurntOrange}{0.1390} 
  & \textcolor{Blue}{0.4225}
  & \textcolor{BurntOrange}{0.3546}
  & \textcolor{ForestGreen}{0.3012}
  & \textcolor{ForestGreen}{0.2737}
  & \textcolor{BurntOrange}{0.2615}
  & \textcolor{BurntOrange}{0.2096}
  & \textcolor{ForestGreen}{0.3485} \\

\textbf{VinVL~\cite{zhang2021vinvl}}   
  & 0.4004 
  & 0.3736 
  & 0.3542 
  & 0.3449 
  & 0.1154 
  & 0.3913
  & 0.3305
  & \textcolor{Blue}{0.3085}
  & \textcolor{Blue}{0.2769}
  & \textcolor{ForestGreen}{0.2592}
  & \textcolor{ForestGreen}{0.2044}
  & \textcolor{BurntOrange}{0.3565} \\

\textbf{VisualBERT}~\cite{li2019visualbertsimpleperformantbaseline}
  & 0.3948 
  & 0.3645 
  & 0.3519 
  & 0.3416 
  & 0.1058 
  & 0.3877
  & 0.3244
  & 0.2795
  & 0.2473
  & 0.2253
  & 0.1732
  & 0.3172 \\ \hline
\end{tabular}
% \vspace{-4mm}
\label{tab:table5}
\end{table*}
%%%%%%%%%%%%%%%%%%%%%%%%%%%%%%%%%%%%%%%%%%%%%%%%%%%%%%%%%%%%%%%%%%%%%%%%%%%%%%%%%%
\begin{table*}[]
\centering
% \vspace{-5mm}
\caption{Performance of Captioning Models on Indian Pines and KSC Datasets: BLEU (B1–B4), METEOR (MTR), and ROUGE-L (R-L).}
% \vspace{2mm}
\fontsize{8}{8}\selectfont
\renewcommand{\arraystretch}{1}
\setlength{\tabcolsep}{3pt}
\begin{tabular}{l|cccccc|cccccc}
\hline
\multirow{2}{*}{\textbf{Model}} 
& \multicolumn{6}{c|}{\textbf{INDIAN PINES}} 
& \multicolumn{6}{c}{\textbf{KENNEDY SPACE CENTER}} \\ \cline{2-13}
& B1 & B2 & B3 & B4 & MET & R-L 
& B1 & B2 & B3 & B4 & MET & R-L \\ \hline
\textbf{BLIP~\cite{li2022blip}}        
  & \textcolor{ForestGreen}{0.3588} 
  & \textcolor{ForestGreen}{0.2571} 
  & \textcolor{ForestGreen}{0.1966} 
  & \textcolor{ForestGreen}{0.1456} 
  & \textcolor{ForestGreen}{0.1777} 
  & \textcolor{ForestGreen}{0.3405}
  & \textcolor{ForestGreen}{0.3749}
  & \textcolor{ForestGreen}{0.3122}
  & \textcolor{ForestGreen}{0.2748}
  & \textcolor{ForestGreen}{0.2528}
  & \textcolor{ForestGreen}{0.2037}
  & \textcolor{ForestGreen}{0.3654} \\

\textbf{GIT~\cite{wang2022git}}        
  & \textcolor{Blue}{0.3816} 
  & \textcolor{Blue}{0.2786} 
  & \textcolor{Blue}{0.2265} 
  & \textcolor{Blue}{0.1746} 
  & \textcolor{Blue}{0.2022} 
  & \textcolor{Blue}{0.3629}
  & \textcolor{Blue}{0.4118}
  & \textcolor{Blue}{0.3485}
  & \textcolor{Blue}{0.3134}
  & \textcolor{Blue}{0.2847}
  & \textcolor{Blue}{0.2398}
  & \textcolor{Blue}{0.3976} \\

\textbf{mPlug~\cite{li2022mplug}}       
  & \textcolor{BurntOrange}{0.3794} 
  & \textcolor{BurntOrange}{0.2775} 
  & \textcolor{BurntOrange}{0.2188} 
  & \textcolor{BurntOrange}{0.1711} 
  & \textcolor{BurntOrange}{0.2003} 
  & \textcolor{BurntOrange}{0.3589}
  & \textcolor{BurntOrange}{0.3977}
  & \textcolor{BurntOrange}{0.3307}
  & \textcolor{BurntOrange}{0.2976}
  & \textcolor{BurntOrange}{0.2710}
  & \textcolor{BurntOrange}{0.2271}
  & \textcolor{BurntOrange}{0.3863} \\

\textbf{VinVL~\cite{zhang2021vinvl}}      
  & 0.3478 & 0.2412 & 0.1845 & 0.1332 & 0.1592 & 0.3264
  & 0.3653 & 0.3045 & 0.2611 & 0.2439 & 0.1963 & 0.3522 \\

\textbf{VisualBERT~\cite{li2019visualbertsimpleperformantbaseline}}
  & 0.3322 & 0.2368 & 0.1775 & 0.1246 & 0.1593 & 0.3208
  & 0.3618 & 0.2978 & 0.2690 & 0.2415 & 0.1967 & 0.3571 \\ \hline
\end{tabular}
\label{tab:table6}
% \vspace{-3mm}
\end{table*}

% \begin{table*}[t]
% \centering
% \caption{Performance of Captioning Models on Indian Pines and KSC Datasets: BLEU (B1-B4), METEOR (MTR), and ROUGE-L (R-L).}
% \fontsize{8}{8}\selectfont
% \renewcommand{\arraystretch}{1}
% \setlength{\tabcolsep}{3pt}
% \begin{tabular}{l|cccccc|cccccc}
% \hline
% \multirow{2}{*}{\textbf{Model}} 
% & \multicolumn{6}{c|}{\textbf{INDIAN PINES}} 
% & \multicolumn{6}{c}{\textbf{KSC}} \\ \cline{2-13}
% & B1 & B2 & B3 & B4 & MET & R-L 
% & B1 & B2 & B3 & B4 & MET & R-L \\ \hline

% \textbf{BLIP}        & 0.3588 & 0.2571 & 0.1966 & 0.1456 & 0.1777 & 0.3405 
%                     & 0.3749 & 0.3122 & 0.2748 & 0.2528 & 0.2037 & 0.3654 \\
% \textbf{GIT}         & 0.3816 & 0.2786 & 0.2265 & 0.1746 & 0.2022 & 0.3629 
%                     & 0.4118 & 0.3485 & 0.3134 & 0.2847 & 0.2398 & 0.3976 \\
% \textbf{mPlug}       & 0.3794 & 0.2775 & 0.2188 & 0.1711 & 0.2003 & 0.3589 
%                     & 0.3977 & 0.3307 & 0.2976 & 0.2710 & 0.2271 & 0.3863 \\
% \textbf{VinVL}       & 0.3478 & 0.2412 & 0.1845 & 0.1332 & 0.1592 & 0.3264 
%                     & 0.3653 & 0.3045 & 0.2611 & 0.2439 & 0.1963 & 0.3522 \\
% \textbf{VisualBERT}  & 0.3322 & 0.2368 & 0.1775 & 0.1246 & 0.1593 & 0.3208 
%                     & 0.3618 & 0.2978 & 0.2690 & 0.2415 & 0.1967 & 0.3571 \\ \hline
% \end{tabular}
% \vspace{-2mm}
% \end{table*}
%%%%%%%%%%%%%%%%
Figures~\ref{VO-Botswana}, \ref{VO-Houston13}, \ref{VO-Indian Pines} and \ref{VO-KSC}  present the vision-only classification maps for all four datasets, showing the baseline performance of 3D-RCNet, 3D-ConvSST, DBCTNet, and FAHM without textual information and providing a clear reference point for assessing the benefits of multimodal fusion. The corresponding multimodal classification maps for the different vision–text combinations (3D-RCNet-BERT/T5, DBCTNet-BERT/T5, FAHM-BERT, and 3D-ConvSST-BERT/T5) are now provided in the Supplementary Material to maintain readability of the main manuscript while still offering detailed qualitative comparisons for interested readers.

\textbf{Limitations}: Large Language Models (LLMs) are not inherently equipped to interpret HSI data and, as such, cannot generate captions directly from it. HyperCap addresses this limitation by leveraging LLMs to produce detailed, human-readable captions aligned with HSI pixels. Although the initial captions generated by LLMs have been carefully refined, they still tend to follow a template-like structure. {Details on both the Pre-edited and Post-edited captions are provided in Figure~\ref{fig:PnP}}. In future work, we plan to scale HyperCap with larger datasets and incorporate more diverse, context-rich annotations to enhance caption quality further and improve generalizability.
% \vspace{-mm}

%%%%%%%%%%%%%%%%%%%%%%%%%%%%%
%%%%%%%%%%%%%%%%% Table on label leakage 
\begin{table*}[!t]
% \vspace{-5mm}
\large
\centering
\caption{Performance Metrics for Botswana Dataset Using Different Models, Input Modalities, and Merging Methods.}
% \vspace{2mm}
\resizebox{0.7\textwidth}{!}{ % Adjust the width as needed
\begin{tabular}{l|c|c|c|ccccc}
\hline
 &
   &
   &
  \multicolumn{6}{c}{\textbf{Merging   Method}} \\ \cline{4-9}
 &
   &
   &
  \multicolumn{3}{c|}{\textbf{PWA}} &
  \multicolumn{3}{c}{\textbf{PWM}} \\ \cline{4-9}
\multirow{-3}{*}{\textbf{DATASET}} &
  \multirow{-3}{*}{\textbf{Vision-Text}} &
  \multirow{-3}{*}{\textbf{Metric}} &
  \multicolumn{1}{c|}{\textbf{IMG+TXT}} &
  \multicolumn{1}{c|}{\textbf{IMG}} &
  \multicolumn{1}{c|}{\textbf{TXT}} &
  \multicolumn{1}{c|}{\textbf{IMG+TXT}} &
  \multicolumn{1}{c|}{\textbf{IMG}} &
  {\textbf{TXT}} \\ \hline
 &
   &
  \textbf{OA} &
  \multicolumn{1}{c|}{\textcolor{Blue}{95.16}} &
  \multicolumn{1}{c|}{\textcolor{ForestGreen}{72.66}} &
  \multicolumn{1}{c|}{\textcolor{BurntOrange}{75.43}} &
  \multicolumn{1}{c|}{\textcolor{Blue}{99.16}} &
  \multicolumn{1}{c|}{\textcolor{ForestGreen}{78.14}} &
  \textcolor{BurntOrange}{81.03} \\
 &
   &
  \textbf{Precision} &
  \multicolumn{1}{c|}{\textcolor{Blue}{94.75}} &
  \multicolumn{1}{c|}{\textcolor{BurntOrange}{75.30}} &
  \multicolumn{1}{c|}{\textcolor{ForestGreen}{69.26}} &
  \multicolumn{1}{c|}{\textcolor{Blue}{99.09}} &
  \multicolumn{1}{c|}{\textcolor{ForestGreen}{69.81}} &
  \textcolor{BurntOrange}{76.55} \\
 &
   &
  \textbf{Kappa} &
  \multicolumn{1}{c|}{\textcolor{Blue}{96.08}} &
  \multicolumn{1}{c|}{\textcolor{BurntOrange}{76.44}} &
  \multicolumn{1}{c|}{\textcolor{ForestGreen}{69.25}} &
  \multicolumn{1}{c|}{\textcolor{Blue}{99.26}} &
  \multicolumn{1}{c|}{\textcolor{ForestGreen}{78.90}} &
  \textcolor{BurntOrange}{78.93} \\
 &
  \multirow{-4}{*}{\textbf{DBCTNet-Bert}} &
  \textbf{F1-score} &
  \multicolumn{1}{c|}{\textcolor{Blue}{92.11}} &
  \multicolumn{1}{c|}{\textcolor{BurntOrange}{76.43}} &
  \multicolumn{1}{c|}{\textcolor{ForestGreen}{71.43}} &
  \multicolumn{1}{c|}{\textcolor{Blue}{98.75}} &
  \multicolumn{1}{c|}{\textcolor{ForestGreen}{78.46}} &
  \textcolor{BurntOrange}{83.59} \\ \cline{2-9}
 &
   &
  \textbf{OA} &
  \multicolumn{1}{c|}{\textcolor{Blue}{98.81}} &
  \multicolumn{1}{c|}{\textcolor{ForestGreen}{74.88}} &
  \multicolumn{1}{c|}{\textcolor{BurntOrange}{75.55}} &
  \multicolumn{1}{c|}{\textcolor{Blue}{99.56}} &
  \multicolumn{1}{c|}{\textcolor{BurntOrange}{76.35}} &
  \textcolor{ForestGreen}{72.58} \\
 &
   &
  \textbf{Precision} &
  \multicolumn{1}{c|}{\textcolor{Blue}{98.71}} &
  \multicolumn{1}{c|}{\textcolor{ForestGreen}{76.71}} &
  \multicolumn{1}{c|}{\textcolor{BurntOrange}{80.96}} &
  \multicolumn{1}{c|}{\textcolor{Blue}{99.52}} &
  \multicolumn{1}{c|}{\textcolor{ForestGreen}{69.2}} &
  \textcolor{BurntOrange}{77.77} \\
 &
   &
  \textbf{Kappa} &
  \multicolumn{1}{c|}{\textcolor{Blue}{98.81}} &
  \multicolumn{1}{c|}{\textcolor{ForestGreen}{73.27}} &
  \multicolumn{1}{c|}{\textcolor{BurntOrange}{76.71}} &
  \multicolumn{1}{c|}{\textcolor{Blue}{99.43}} &
  \multicolumn{1}{c|}{\textcolor{ForestGreen}{78.20}} &
  \textcolor{BurntOrange}{78.82} \\
\multirow{-8}{*}{\textbf{BOTSWANA}} &
  \multirow{-4}{*}{\textbf{DBCTNet-T5}} &
  \textbf{F1-score} &
  \multicolumn{1}{c|}{\textcolor{Blue}{98.18}} &
  \multicolumn{1}{c|}{\textcolor{ForestGreen}{70.47}} &
  \multicolumn{1}{c|}{\textcolor{BurntOrange}{79.47}} &
  \multicolumn{1}{c|}{\textcolor{Blue}{99.37}} &
  \multicolumn{1}{c|}{\textcolor{ForestGreen}{73.34}} &
  \textcolor{BurntOrange}{76.49} \\ \hline
\end{tabular}
}
\vspace{-4mm}
\label{tab:ablation_study}
\end{table*}

%%%%%%%%%%%%%%%%%% Table on 3% dataset

%%%%%%%%%%%%%%%%%%%%%%%%%%%%%
\subsection{Image-text Retrieval}
A major innovation of the HyperCap dataset is its applicability to multimodal retrieval across multiple hyperspectral benchmarks. 
To evaluate retrieval performance, we benchmark five vision--language models---BLIP, GIT, mPLUG, VinVL, and VisualBERT---on Botswana, Houston13, Indian Pines, and Kennedy Space Center, using the same train/val/test splits adopted for the captioning experiments.We evaluate two retrieval directions: (i) \emph{text-to-image} retrieval (denoted as image retrieval, IR), where a caption query retrieves its paired HSI sample, and (ii) \emph{image-to-text} retrieval (denoted as text retrieval, TR), where an HSI query retrieves its paired caption. For each query, we compute a similarity score between the query embedding and all candidate embeddings in the test set (using the model's image/text embedding space), rank candidates in descending order of similarity, and report Top-1 retrieval accuracy (Recall@1, R@1). A query is counted as correct if its paired ground-truth item is ranked at position 1; the final R@1 is averaged over all test queries. As reported in Table~\ref{tab:retrieval_results}, retrieval performance is consistent across datasets but remains below saturation: IR scores are typically in the 59\%--63\% range, while TR ranges from 66\% to 71\%. For instance, on Botswana, GIT achieves 62.66\% IR and 70.26\% TR, and similar trends are observed across models and sites with minor fluctuations. Overall, these results indicate that semantic alignment between pixel-level hyperspectral samples and natural-language descriptions is non-trivial for off-the-shelf RGB vision--language models, motivating future work on domain-adapted architectures for hyperspectral vision--language retrieval.
% \vspace{-4mm}

\subsection{Ablation Study}\label{sec:ablation}
We conduct two complementary ablations to both validate the integrity of our captions and quantify their practical utility. First, we test for potential label leakage by comparing multimodal (IMG+TXT) performance against unimodal Image-only and Text-only variants on Botswana using DBCTNet-BERT and DBCTNet-T5 under the PWM and PWA merging strategies. Second, we stress-test data efficiency by reducing the training set to 3\% of labeled pixels and benchmarking vision-only models alongside our multimodal fusions. Together, these studies assess whether captions inadvertently encode class labels and whether they meaningfully improve performance under label scarcity—two properties that are critical for trustworthy and deployable multimodal HSI systems.

\begin{table*}[t]
% \vspace{-5mm}
\centering
\caption{Parameters (M) and FLOPs (M) of Different Classification Models.}
% \vspace{2mm}
\small
\label{tab:classfication_models}
\renewcommand{\arraystretch}{1.1} % Increase row height
\resizebox{1.8\columnwidth}{!}{%
\begin{tabular}{c|c|c|c|c|c|c|c|c|c|c}
\hline
\multicolumn{3}{c|}{\textbf{Model}} &
  \multicolumn{2}{c|}{\textbf{DBCTNet}} &
  \multicolumn{2}{c|}{\textbf{3D   RCNet}} &
  \multicolumn{2}{c|}{\textbf{FAHM}} &
  \multicolumn{2}{c}{\textbf{3D   ConvSST}} \\ \hline
\multicolumn{3}{c|}{\textbf{Method}} &
  \multicolumn{1}{c|}{\textbf{Params}} &
  \textbf{FLOPs} &
  \multicolumn{1}{c|}{\textbf{Params}} &
  \textbf{FLOPs} &
  \multicolumn{1}{c|}{\textbf{Params}} &
  \textbf{FLOPs} &
  \multicolumn{1}{c|}{\textbf{Params}} &
  \textbf{FLOPs} \\ \hline
\multicolumn{3}{c|}{\textbf{Vision}} &
  \multicolumn{1}{c|}{1.63E-02} &
  1.90E+01 &
  \multicolumn{1}{c|}{3.43E+00} &
  8.99E+02 &
  \multicolumn{1}{c|}{8.88E-01} &
  5.04E+01 &
  \multicolumn{1}{c|}{2.86E-01} &
  1.01E+02 \\ \hline
\multicolumn{1}{c|}{\multirow{10}{*}{\rotatebox{90}{\textbf{IMG+TXT}}}}
 &
  \multicolumn{1}{c|}{\multirow{2}{*}{\textbf{CA}}} &
  \textbf{Bert} &
  \multicolumn{1}{|c|}{5.04E-02} &
  2.10E-05 &
  \multicolumn{1}{|c|}{3.96E+00} &
  9.89E-04 &
  \multicolumn{1}{|c|}{1.02E+00} &
  5.55E-05 &
  \multicolumn{1}{|c|}{4.19E-01} &
  1.11E-04 \\ \cline{3-11}
\multicolumn{1}{c|}{} &
  \multicolumn{1}{|c|}{} &
  \textbf{T5} &
  \multicolumn{1}{|c|}{5.04E-02} &
  2.10E-05 &
  \multicolumn{1}{|c|}{3.96E+00} &
  9.89E-04 &
  \multicolumn{1}{|c|}{1.02E+00} &
  5.55E-05 &
  \multicolumn{1}{|c|}{3.54E-01} &
  1.11E-04 \\ \cline{2-11}
 &
  \multicolumn{1}{|c|}{\multirow{2}{*}{\textbf{CONCAT}}} &
  \textbf{Bert} &
  \multicolumn{1}{|c|}{3.32E-02} &
  2.10E-05 &
  \multicolumn{1}{|c|}{3.70E+00} &
  9.89E-04 &
  \multicolumn{1}{|c|}{9.55E-01} &
  5.55E-05 &
  \multicolumn{1}{|c|}{3.69E-01} &
  1.11E-04 \\ \cline{3-11}
 &
  \multicolumn{1}{|c|}{} &
  \textbf{T5} &
  \multicolumn{1}{|c|}{3.32E-02} &
  2.10E-05 &
  \multicolumn{1}{|c|}{3.70E+00} &
  9.89E-04 &
  \multicolumn{1}{|c|}{9.55E-01} &
  5.55E-05 &
  \multicolumn{1}{|c|}{4.19E-01} &
  1.11E-04 \\ \cline{2-11}
 &
  \multicolumn{1}{|c|}{\multirow{2}{*}{\textbf{MHA}}} &
  \textbf{Bert} &
  \multicolumn{1}{|c|}{3.40E-02} &
  2.10E-05 &
  \multicolumn{1}{|c|}{3.96E+00} &
  9.89E-04 &
  \multicolumn{1}{|c|}{9.71E-01} &
  5.55E-05 &
  \multicolumn{1}{|c|}{3.54E-01} &
  1.11E-04 \\ \cline{3-11}
\multicolumn{1}{c|}{} &
  \multicolumn{1}{|c|}{} &
  \textbf{T5} &
  \multicolumn{1}{|c|}{3.40E-02} &
  2.10E-05 &
  \multicolumn{1}{|c|}{3.96E+00} &
  9.89E-04 &
  \multicolumn{1}{|c|}{9.71E-01} &
  5.55E-05 &
  \multicolumn{1}{|c|}{3.69E-01} &
  1.11E-04 \\ \cline{2-11}
&
  \multicolumn{1}{|c|}{\multirow{2}{*}{\textbf{PWA}}} &
  \textbf{Bert} &
  \multicolumn{1}{|c|}{3.30E-02} &
  2.10E-05 &
  \multicolumn{1}{|c|}{3.70E+00} &
  9.89E-04 &
  \multicolumn{1}{|c|}{9.54E-01} &
  5.55E-05 &
  \multicolumn{1}{|c|}{3.53E-01} &
  1.11E-04 \\ \cline{3-11}
 &
  &
  \textbf{T5} &
  \multicolumn{1}{|c|}{3.30E-02} &
  2.10E-05 &
  \multicolumn{1}{|c|}{3.70E+00} &
  9.89E-04 &
  \multicolumn{1}{|c|}{9.54E-01} &
  5.55E-05 &
  \multicolumn{1}{|c|}{3.53E-01} &
  1.11E-04 \\ \cline{2-11}
&
  \multicolumn{1}{|c|}{\multirow{2}{*}{\textbf{PWM}}} &
  \textbf{Bert} &
  \multicolumn{1}{|c|}{3.30E-02} &
  2.10E-05 &
  \multicolumn{1}{|c|}{3.70E+00} &
  9.89E-04 &
  \multicolumn{1}{|c|}{9.54E-01} &
  5.55E-05 &
  \multicolumn{1}{|c|}{3.53E-01} &
  1.11E-04 \\ \cline{3-11}
&
 &
  \textbf{T5} &
  \multicolumn{1}{|c|}{3.30E-02} &
  2.10E-05 &
  \multicolumn{1}{|c|}{3.70E+00} &
  9.89E-04 &
  \multicolumn{1}{|c|}{9.54E-01} &
  5.55E-05 &
  \multicolumn{1}{|c|}{3.53E-01} &
  1.11E-04 \\ \hline
\end{tabular}%
}
% \vspace{-2mm}
\end{table*}

%%%%%%%%% end %%%%%%%%%%%%%%%%%%%%%%%%
\begin{table*}[]
% \vspace{-4mm}
\caption{Parameters (M) and FLOPs (M) of Captioning Vision-Language Models.}
% \vspace{2mm}
\centering
\begin{tabular}{l|c|c|c|c|c}
\hline
\textbf{Metric} & \textbf{BLIP} & \textbf{GIT} & \textbf{mPLUG} & \textbf{VinVL} & \textbf{Visual BERT} \\
\hline
{Params(M)} & 87.38     & 86.34     & 207.80   & 112.82    & 110.37  \\
\hline
{FLOPs(M)}      & 55590    & 149.14    & 4180.58  & 22560     & 3930  \\
\hline
\end{tabular}
\label{tab:Captioning_models}
\vspace{-4mm}
\end{table*}

\begin{table*}[]
\centering
% \vspace{-3.5mm}
\caption{Comprehensive Results on Botswana Dataset with 3\% Training data: Overall Accuracy (OA), Precision, Kappa Accuracy (KA), and F1-Score (\%) for Vision-Only and Various Multimodal Fusion Models.}
\label{tab:3percent_full}
\resizebox{\textwidth}{!}{
\begin{tabular}{l|l|c|c|c|c|c|c|c|c|c|c|c}
\hline
\textbf{Model} & \textbf{Metric} & \textbf{Vision Only} & \textbf{CA-BERT} & \textbf{CA-T5} & \textbf{CONCAT-BERT} & \textbf{CONCAT-T5} &\textbf{MHA-BERT} & \textbf{MHA-T5} & \textbf{PWA-BERT} & \textbf{PWA-T5} & \textbf{PWM-BERT} & \textbf{PWM-T5} \\
\hline
\textbf{\multirow{4}{*}{FAHM}}
& \textbf{OA}         & 94.02 & 98.61 & 98.53 & 98.96 & 98.82 & 98.46 & 99.08 & 98.67 & 98.75 & 99.21 & 99.21 \\
& \textbf{Precision}  & 94.92 & 99.17 & 99.22 & 98.53 & 99.20 & 99.20 & 99.18 & 98.57 & 98.86 & 99.40 & 98.89 \\
& \textbf{KA}         & 92.99 & 98.03 & 97.38 & 98.33 & 97.57 & 97.56 & 97.63 & 97.56 & 98.08 & 98.19 & 97.39 \\
& \textbf{F1-Score}   & 94.07 & 98.92 & 98.08 & 98.76 & 98.77 & 98.11 & 98.59 & 98.52 & 98.33 & 98.52 & 98.61 \\
\hline
\textbf{\multirow{4}{*}{3D-ConvSST}}
& \textbf{OA}         & 95.11 & 99.25 & 99.67 & 99.31 & 98.91 & 99.35 & 99.07 & 99.90 & 99.51 & 99.68 & 99.39 \\
& \textbf{Precision}  & 95.98 & 99.76 & 99.64 & 99.57 & 99.38 & 99.91 & 99.85 & 99.15 & 99.24 & 99.18 & 99.40 \\
& \textbf{KA}         & 94.26 & 98.49 & 98.82 & 98.43 & 98.67 & 98.48 & 99.02 & 98.23 & 98.35 & 98.79 & 98.45 \\
& \textbf{F1-Score}   & 95.05 & 98.11 & 98.13 & 98.38 & 98.31 & 98.83 & 98.75 & 98.06 & 98.26 & 98.78 & 98.10 \\
\hline
\end{tabular}
}
\end{table*}

To address concerns about potential label leakage through the captions, we conducted a systematic ablation study on the Botswana dataset using DBCTNet-BERT and DBCTNet-T5 under three input settings: Image Only, Text Only, and Image + Text. As shown in Table~\ref{tab:ablation_study}, the values are color-coded to indicate the top three modalities within each fusion strategy: \textcolor{Blue}{Blue} for the highest-performing modality, \textcolor{BurntOrange}{Orange} for the second, and \textcolor{ForestGreen}{Green} for the third. The results definitively rule out label leakage. For DBCTNet-BERT (PWM), the F1-score dropped from 98.75\% to 83.59\% (Text~$\downarrow$15.2\%) and 78.46\% (Image~$\downarrow$20.3\%). DBCTNet-T5 (PWM) exhibit a decline from 99.37\% to 76.49\% (Text~$\downarrow$22.9\%) and 73.34\% (Image~$\downarrow$26.0\%). Under the PWA merging strategy, BERT's F1-score fell from 92.11\% to 71.43\% (Text~$\downarrow$22.4\%) and 76.43\% (Image~$\downarrow$17.0\%), while T5 dropped from 98.18\% to 79.47\% (Text~$\downarrow$18.7\%) and 70.47\% (Image~$\downarrow$28.0\%). These consistent and substantial performance drops for unimodal inputs robustly confirm no label leakage, as neither images nor text alone retain full predictive strength. The fusion strategies CONCAT, CA, and MHA output embeddings that are twice as large, making single-modality input incompatible without duplication. Therefore, PWM and PWA merging methods were selected to maintain fair comparisons.
Having established that our captions provide genuine semantic information without label leakage, we next investigate how this complementary textual knowledge performs under challenging real-world conditions. While the previous ablation confirmed the independence of visual and textual modalities, this experiment explores their synergistic benefits when training data is severely limited—a common scenario in remote sensing applications where ground truth annotation is expensive and time-consuming.

Despite the high accuracy achieved by leading hyperspectral vision models such as FAHM and 3D-ConvSST (often exceeding 99\% OA), HyperCap delivers crucial benefits that extend well beyond incremental improvements on already saturated models. Its true value emerges in two principal areas: the dramatic performance uplift for weaker architectures and new opportunities for multimodal task development in HSI.

First, HyperCap’s rich semantic captions significantly elevate models with initially modest vision-only performance. In Table~\ref{tab:3percent_full}, on the Indian Pines dataset, the OA of DBCTNet surges from 76.01\% (vision only) to 99.37\% with text fusion—a gain of over 23\%. Similarly, on Kennedy Space Center, the OA improves from 70.50\% to 97.58\%, a leap of 27\%. These results demonstrate that HyperCap does not merely polish top-performing methods, but democratizes state-of-the-art performance for a wider spectrum of architectures.

Second, real-world HSI applications frequently face low-annotation scenarios. To rigorously probe model robustness in such settings, we ablate the training set size to just 3\% of labeled pixels. Vision-only baselines, though strong with ample data, degrade rapidly under this regime: FAHM’s OA falls to 94.02\% and 3D-ConvSST to 95.11\%, marking a clear 4–5\% drop. By contrast, all multimodal fusion models leveraging BERT and T5 retain robust performance—regularly exceeding 98\% OA and often reaching above 99\%. This evidences the unique advantage provided by caption-based supervision: text-guided models remain resilient even under severe data scarcity, while vision-only methods are far more susceptible to label depletion.
\vspace{-3mm}

%%%%%%%%%%%%%%%%%%
\section{Data Ethics}
\label{sec:data_ethics}
We only used publicly available hyperspectral datasets in this study, ensuring that no private or sensitive information is involved. All data were handled responsibly, with care to preserve integrity and fairness. The dataset annotations were created and verified ethically, avoiding misuse or leakage of class labels. We share our methods openly to support transparency and responsible research.
%%%%%%%%%%%% conclusion%%%%%%%%%%%%%%%%%%%
\vspace{-3mm}
\section{Conclusion}
\label{sec:con}
In this paper, we propose the HyperCap dataset, which is (to the best of our knowledge) the first pixel-level vision-language benchmark for HSI, thereby enabling fine-grained multimodal learning. Its fine-grained captions bridge the gap between spectral data and semantic understanding, effectively addressing limitations found in existing HSI datasets. Experimental results show that integrating textual descriptions leads to substantial improvements in classification performance across various architectures, underscoring the potential of multimodal approaches in HSI analysis. This work lays a strong foundation for future research in vision-language learning for HSI, paving the way for broader multimodal tasks in remote sensing. We demonstrate the dataset's applicability in tasks such as multimodal classification and caption generation. Potential future directions include image-text retrieval and the development of foundational captioning models specifically tailored for HSI data. These contributions position HyperCap as a benchmark for advancing cross-modal representation learning in the HSI domain. In future work, we are working on annotating additional hyperspectral benchmarks to construct a larger, more diverse, and truly large-scale corpus, thereby further improving generalization and applicability.
% \input{dataset}
% \input{results}
% \vspace{-5mm}
% \input{discussion}

% \vspace{-5mm}
%\section{Why GRSM?}
% \vspace{-5mm}
%Considering the aims and scope of IEEE GRSM, HyperCap is exceptionally well suited for the magazine’s broad audience. As the first large-scale, pixel-wise hyperspectral captioning dataset, HyperCap bridges a crucial gap between rich spectral features and semantic, human-understandable information. This resource enables new vision-language research, advancing transparency and semantic understanding in HSI. The open dataset and comprehensive benchmarks target both students and professionals, supporting fundamental advances in land cover mapping, environmental monitoring, and cross-domain remote sensing. Publishing in GRSM ensures maximum visibility and makes this paper a foundational tutorial reference for the remote sensing community.

\bibliographystyle{ieeetr}
\bibliography{sec_V2/refs}

% \documentclass[twocolumn,10pt]{article}   % Or report/book depending on your thesis/paper format

% % ---------- Essential Packages ----------
% \usepackage{graphicx}     % For including images
% \usepackage{caption}      % Better caption control
% \usepackage{subcaption}   % For subfigure captions
% \usepackage{float}        % Improved figure/table placement
% \usepackage{amsmath}      % Math formatting
% \usepackage{amssymb}      % Extra math symbols
% \usepackage{multirow}     % Multi-row tables
% \usepackage{booktabs}     % Professional tables
% \usepackage{xcolor}       % For text colors (if needed)
% \usepackage{hyperref}     % Clickable references
% \usepackage{geometry}     % Adjust margins
% \usepackage{times}        % Times New Roman font (optional, IEEE style)
% \usepackage{enumitem}     % Better control for lists
% \usepackage{titlesec}     % Control of section formatting

% % ---------- Margin and layout adjustments ----------
% \geometry{
%     left=1.5cm,
%     right=1.5cm,
%     top=1.5cm,
%     bottom=1.5cm
% }

% % ---------- Hyperlink Setup ----------
% \hypersetup{
%     colorlinks=true,
%     linkcolor=blue,
%     citecolor=blue,
%     urlcolor=blue
% }

% % ---------- Section Title formatting (optional, adjust as needed) ----------
% \titlespacing*{\section}{0pt}{2ex plus 1ex minus .2ex}{1ex}
% \titlespacing*{\subsection}{0pt}{1.5ex plus 1ex minus .2ex}{0.8ex}

% % ---------- Begin Document ----------
% \begin{document}

\twocolumn[
\begin{center}
\Huge\textbf{Supplementary Material}
\end{center}
]

Across all datasets, the visual maps align tightly with the numbers: adding text encoders (IMG+TXT) dramatically cleans up the predictions compared to vision-only baselines, especially for models that start lower (e.g., 3D RCNet and DBCTNet). The best-performing fusions yield maps that are nearly indistinguishable from ground truth—large homogeneous regions become uniform, boundaries sharpen, and small classes are better preserved. Differences between BERT and T5 are generally modest relative to the impact of the fusion choice; both encoders enable high-quality maps, with small, backbone- and dataset-specific swings in who wins by a few hundredths.

% ============================================================================
% BOTSWANA DATASET
% ============================================================================

%%%%%%%%%%%%%%%%%%%%%%%%%%%%%%%% Botswana 3DRCNet-T5Encoder %%%%%%%%%%%%%%%%%%%%%%%%%%%%%%%%%%%%%%%%%%%%%%%%
\begin{figure*}[t]
    \centering

    \newcommand{\mapimg}[3][]{%
        \begin{subfigure}[b]{0.3\textwidth}
            \ifx&#1& % optional trim empty
                \includegraphics[width=0.18\linewidth, trim=700 50 250 60, clip, angle=90]{Output_Map/Botswana/3DRCNet-T5Encoder_Large/#2_prediction_map_run1.png}%
            \else
                \includegraphics[width=0.18\linewidth, trim=#1, clip, angle=90]{Output_Map/Botswana/3DRCNet-T5Encoder_Large/#2_prediction_map_run1.png}%
            \fi
            \caption{#2}
        \end{subfigure}
    }
    % ---------- Images ----------
    \mapimg{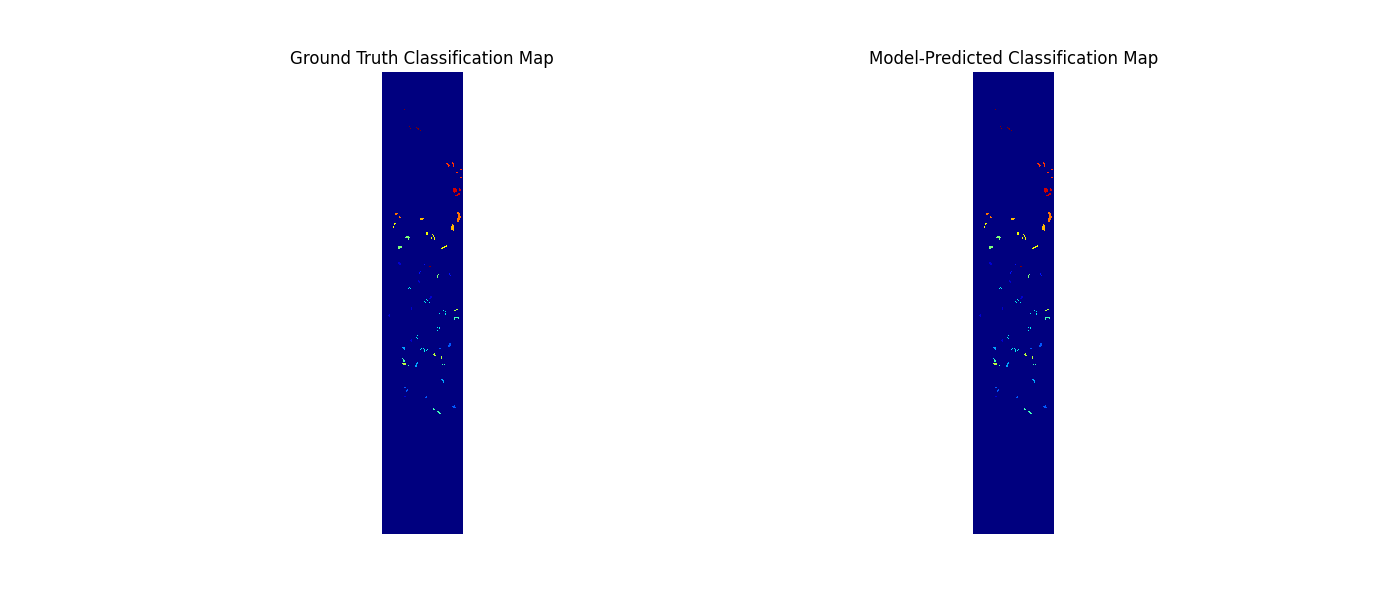}{CA}
    \mapimg{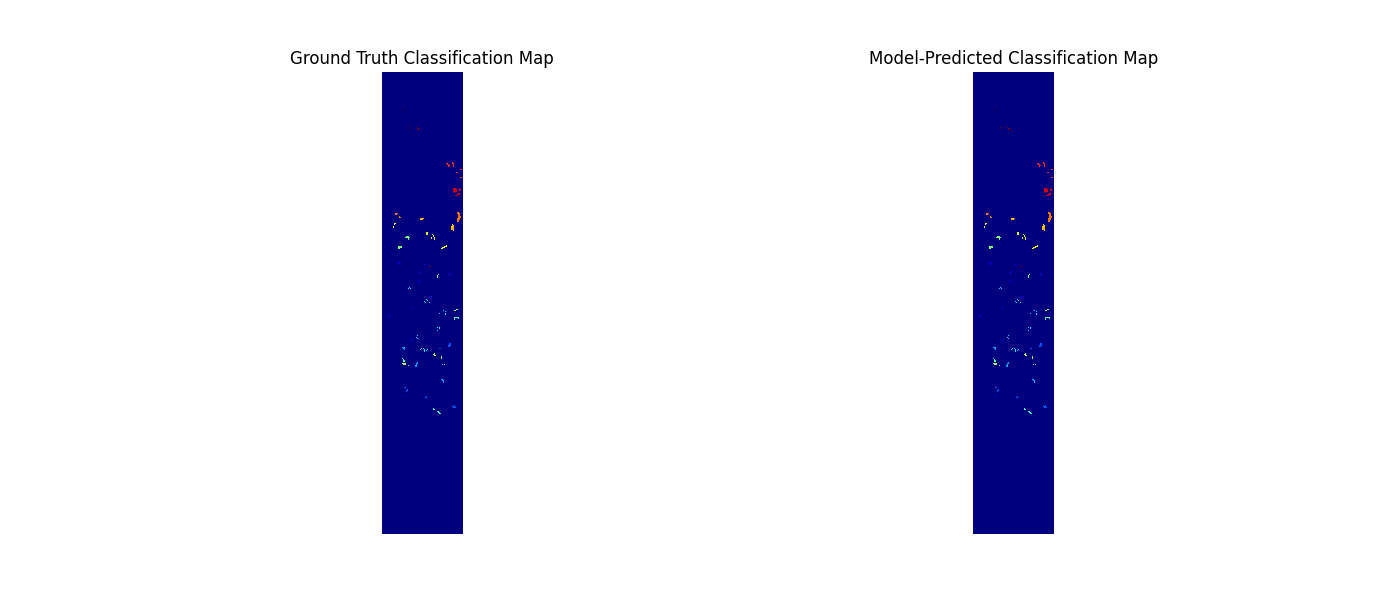}{CONCAT}
    \mapimg{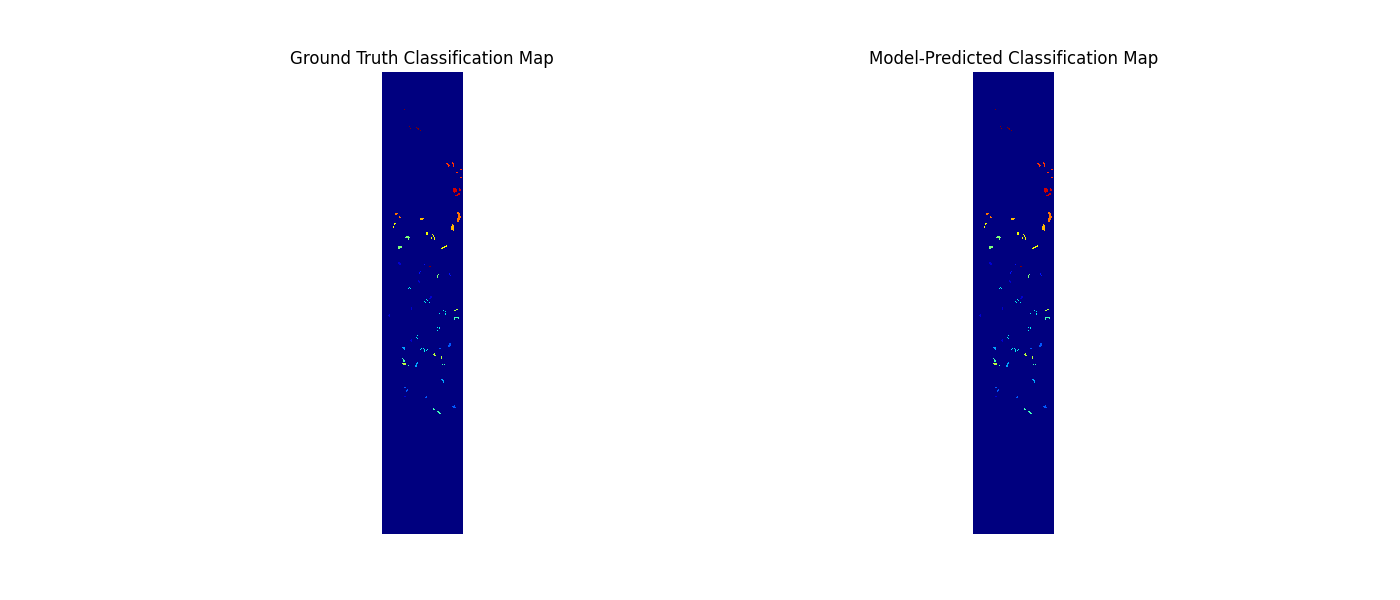}{MHA}
    \mapimg{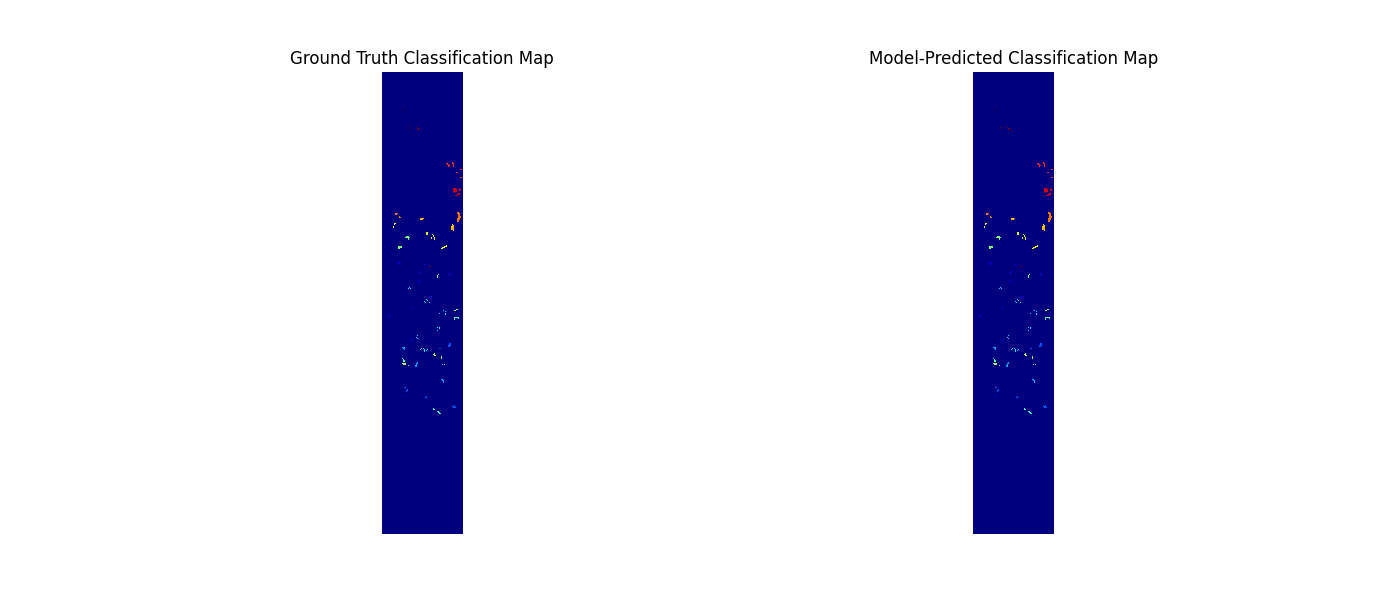}{PWA}
    \mapimg{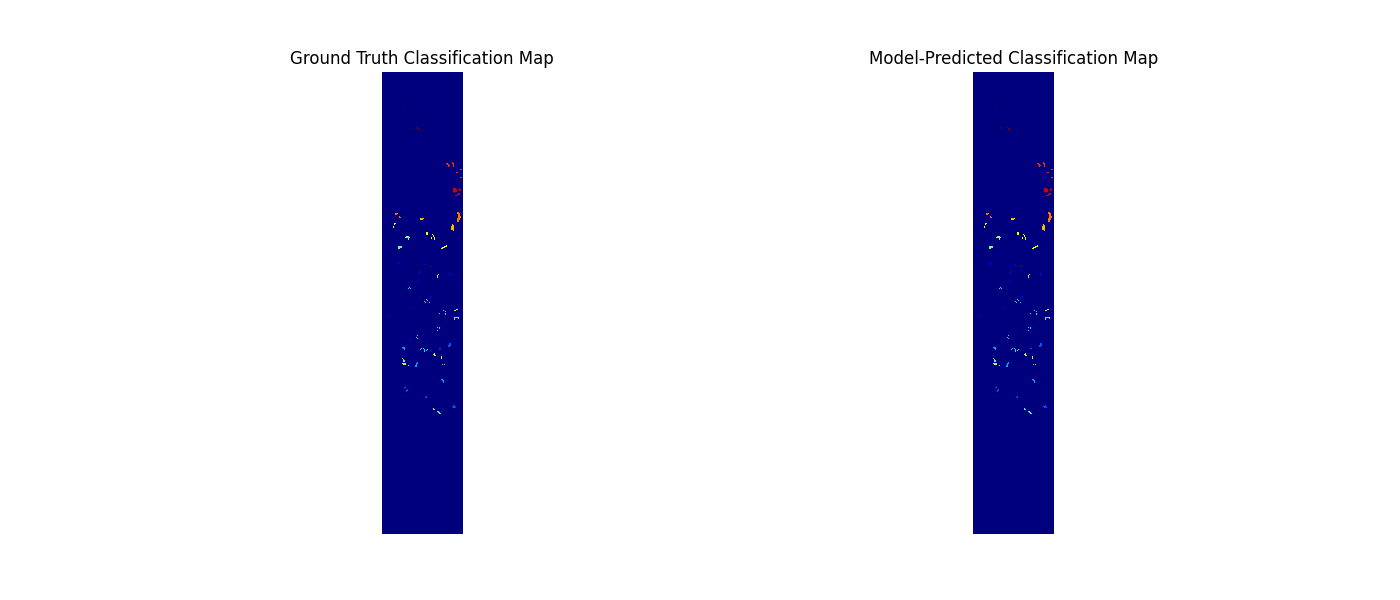}{PWM}
    \mapimg[275 50 675 60]{PWM}{GT} % only GT changes trim

    \caption{Comparison of classification maps for the 3D-RCNet-T5 model on the Botswana dataset, showing different fusion methods: Cross Attention (CA), Concatenation (CONCAT), Multi-Head Attention (MHA), Pixel-Wise Addition (PWA), Pixel-Wise Multiplication (PWM), and Ground Truth (GT).}
    \vspace{-3mm}
    \label{fig:3D-RCNet-T5-Botswana}
\end{figure*}

%%%%%%%%%%%%%%%%%%%%%%%%%%%%%%%%%%%%%%%%%%%%%%%%  Botswana  3DRCNet-BertEncoder   %%%%%%%%%%%%%%%%%%%%%%%%%%%%%%%%%%%%%%%%%%%%%%%%%%%%%%%%%%%%%%%%%%%%%
\begin{figure*}[!t]
    \centering
    \newcommand{\mapimg}[3][]{%
        \begin{subfigure}[b]{0.3\textwidth}
            \ifx&#1&
                \includegraphics[width=0.18\linewidth, trim=700 50 250 60, clip, angle=90]{Output_Map/Botswana/3DRCNet-BertEnocder_Large/#2_prediction_map_run1.png}%
            \else
                \includegraphics[width=0.18\linewidth, trim=#1, clip, angle=90]{Output_Map/Botswana/3DRCNet-BertEnocder_Large/#2_prediction_map_run1.png}%
            \fi
            \caption{#2}
        \end{subfigure}
    }

    \mapimg{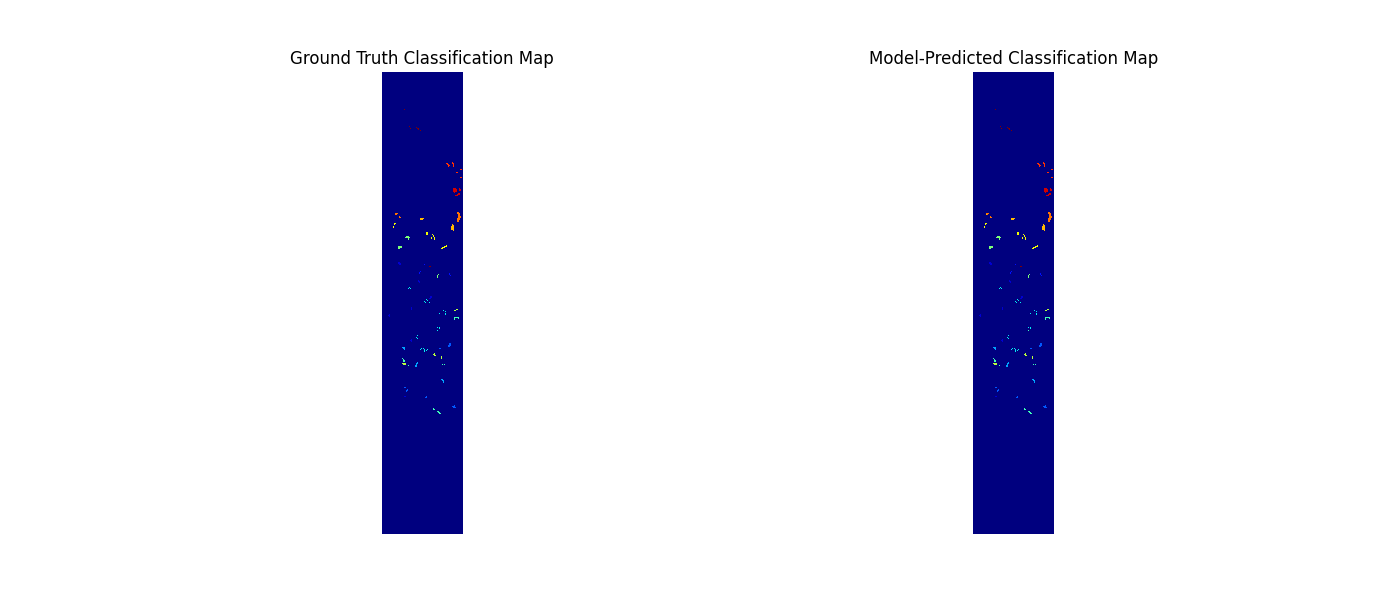}{CA}
    \mapimg{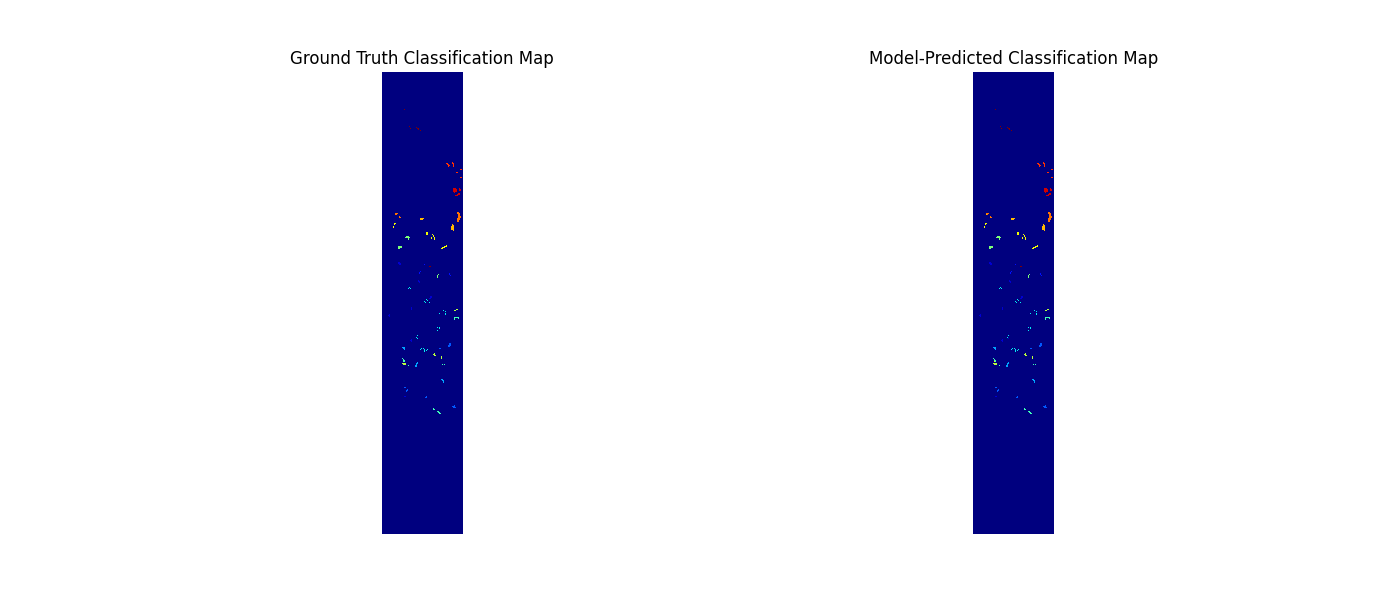}{CONCAT}
    \mapimg{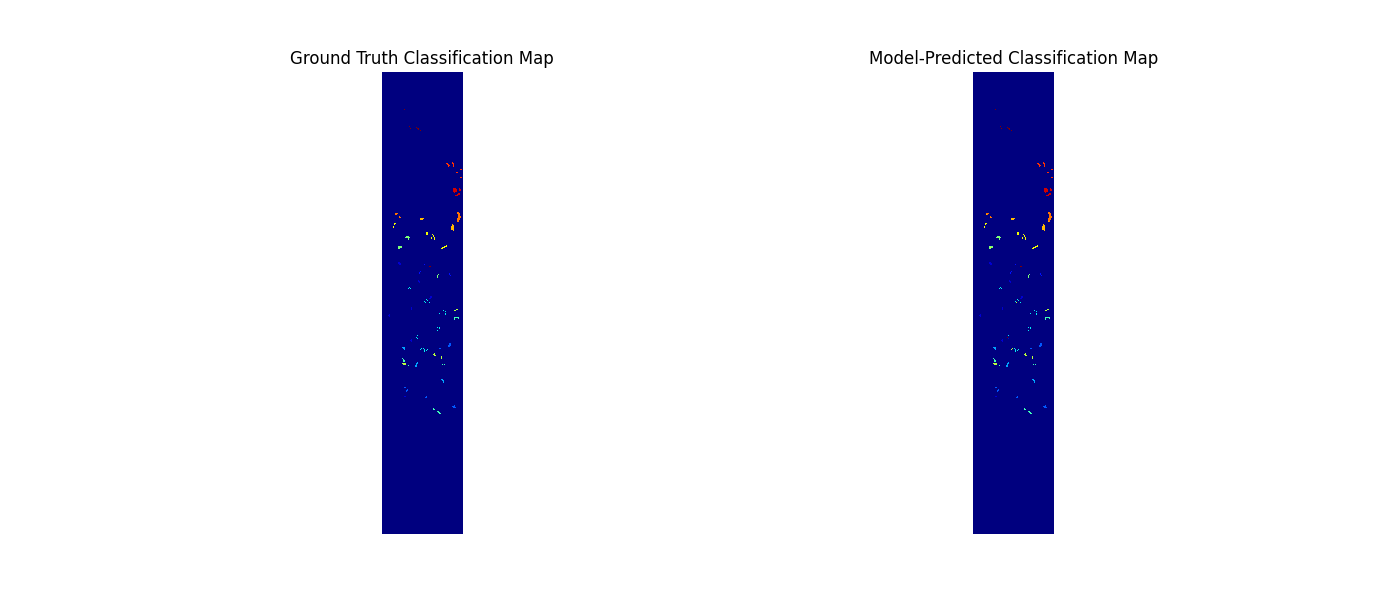}{MHA}
    \mapimg{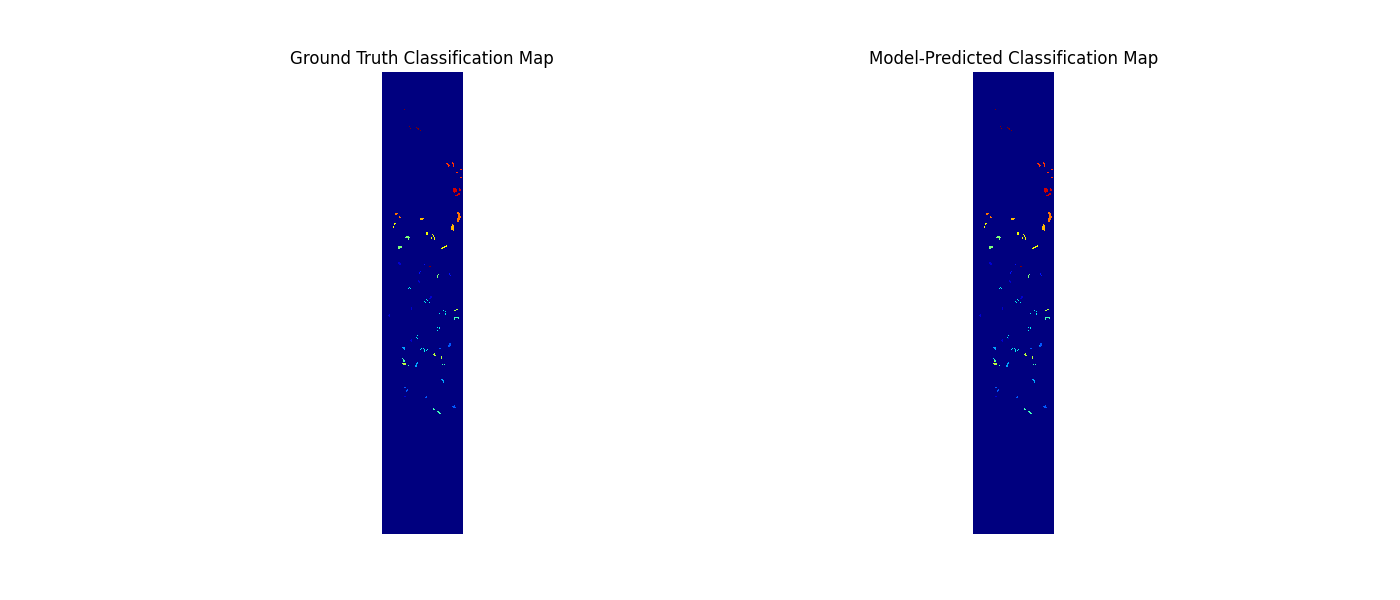}{PWA}
    \mapimg{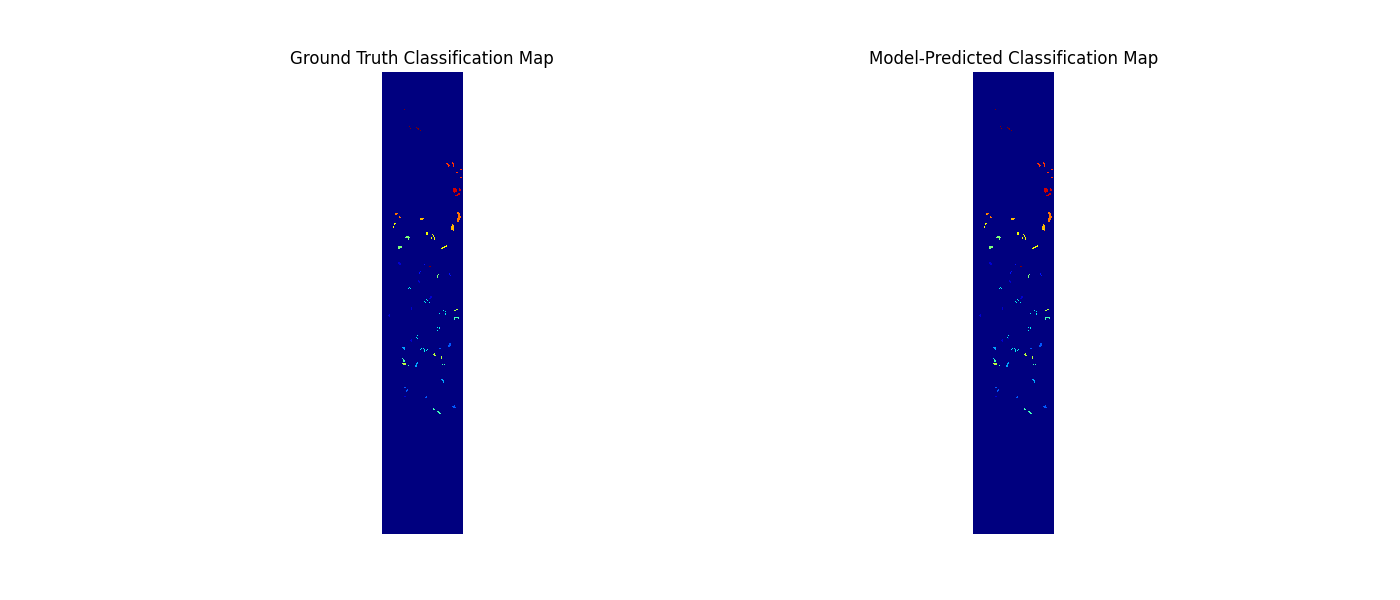}{PWM}
    \mapimg[275 50 675 60]{PWM}{GT}

    \caption{Comparison of classification maps for the 3D-RCNet-BERT model on the Botswana dataset, showing different fusion methods: Cross Attention (CA), Concatenation (CONCAT), Multi-Head Attention (MHA), Pixel-Wise Addition (PWA), Pixel-Wise Multiplication (PWM), and Ground Truth (GT).}
    \vspace{-3mm}
    \label{fig:3D-RCNet-Bert-Botswana}
\end{figure*}

\begin{figure*}[]
    \centering
    \begin{subfigure}[b]{0.3\textwidth}
         \includegraphics[width=0.18\linewidth, trim=700 50 250 60, clip, angle=90]{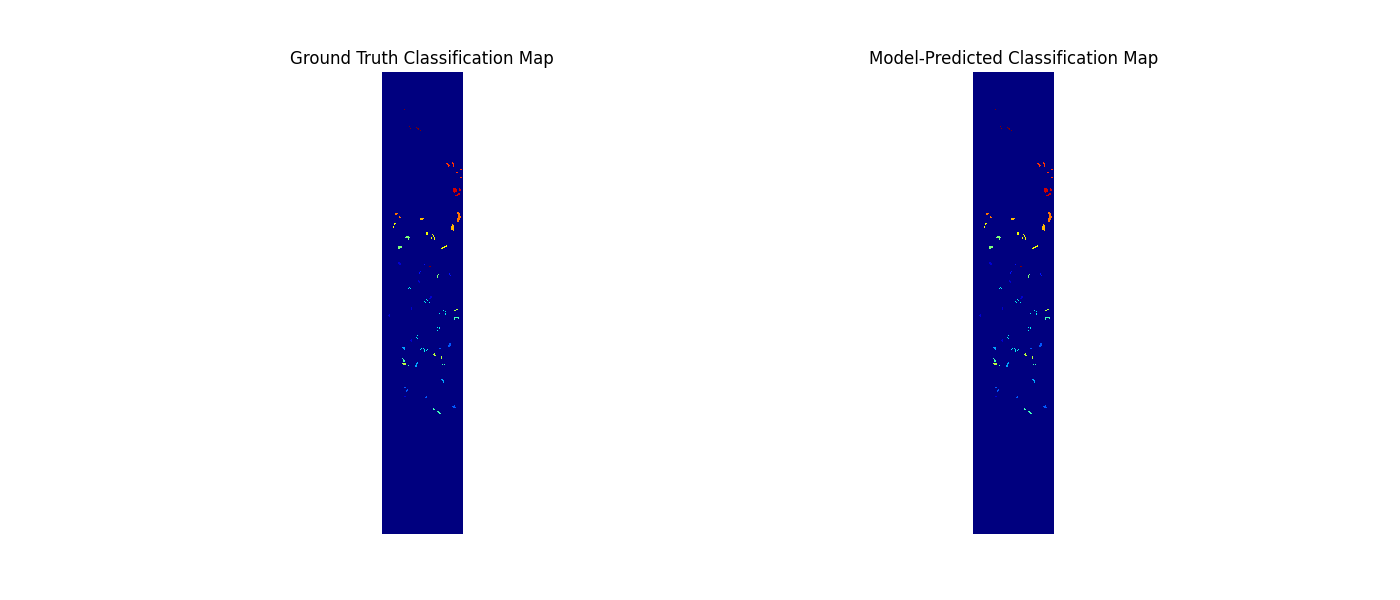}
        \caption{CA}
    \end{subfigure}
    \begin{subfigure}[b]{0.3\textwidth}
        \includegraphics[width=0.18\linewidth, trim=700 50 250 60, clip, angle=90]{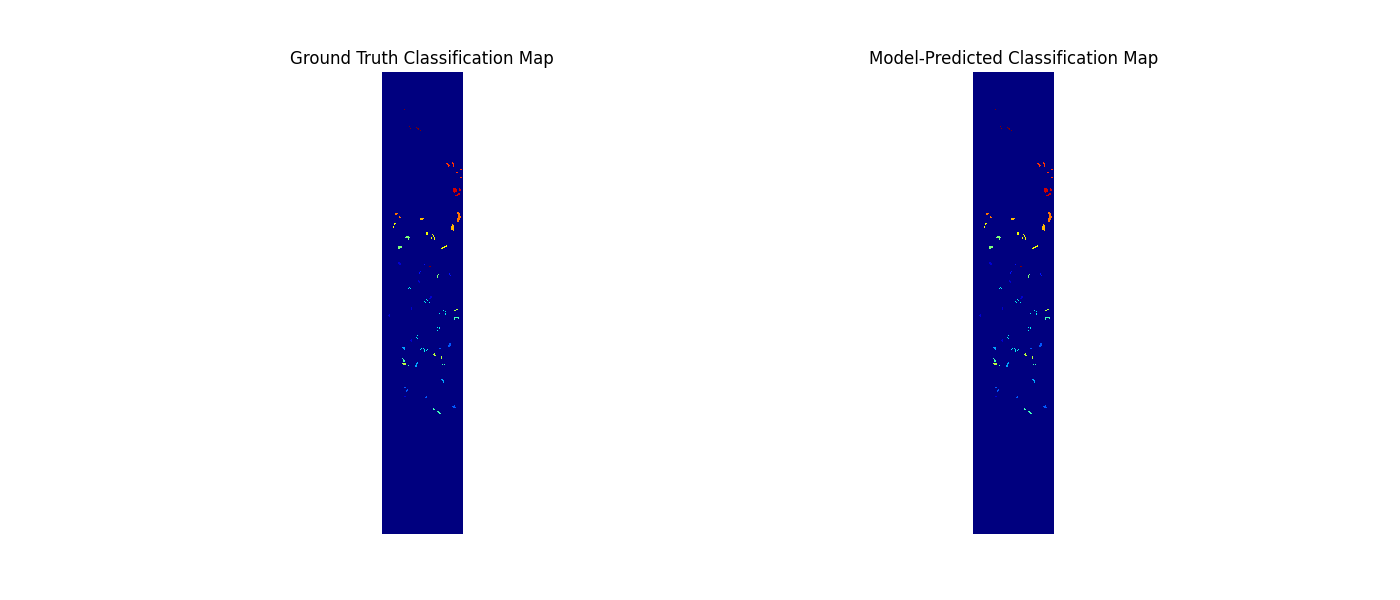}
        \caption{CONCAT}
    \end{subfigure}
    \begin{subfigure}[b]{0.3\textwidth}
        \includegraphics[width=0.18\linewidth, trim=700 50 250 60, clip, angle=90]{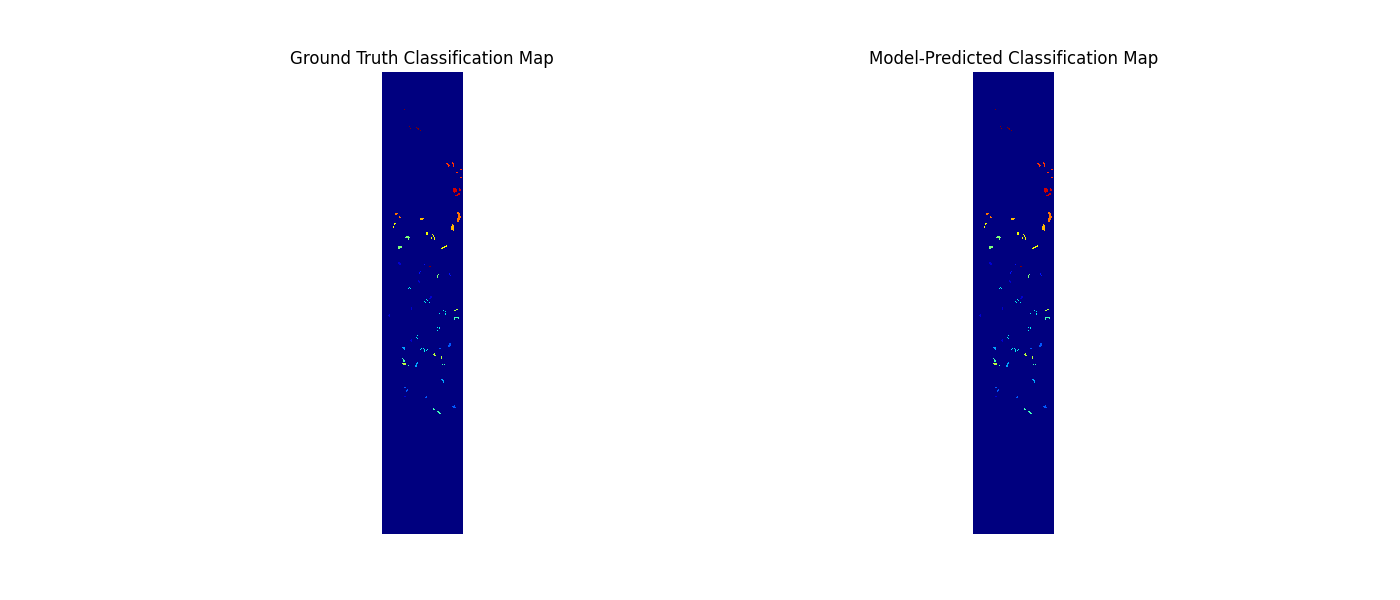}
        \caption{MHA}
    \end{subfigure}
    \begin{subfigure}[b]{0.3\textwidth}
        \includegraphics[width=0.18\linewidth, trim=700 50 250 60, clip, angle=90]{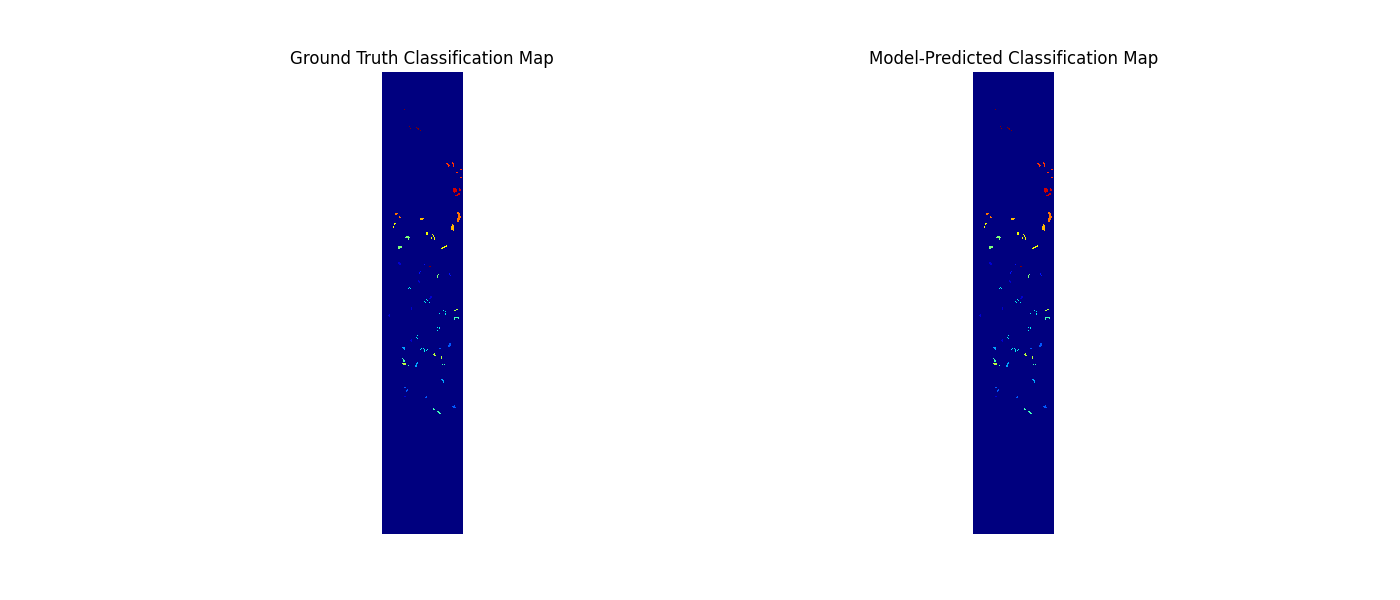}
        \caption{PWA}
    \end{subfigure}
    \begin{subfigure}[b]{0.3\textwidth}
        \includegraphics[width=0.18\linewidth, trim=700 50 250 60, clip, angle=90]{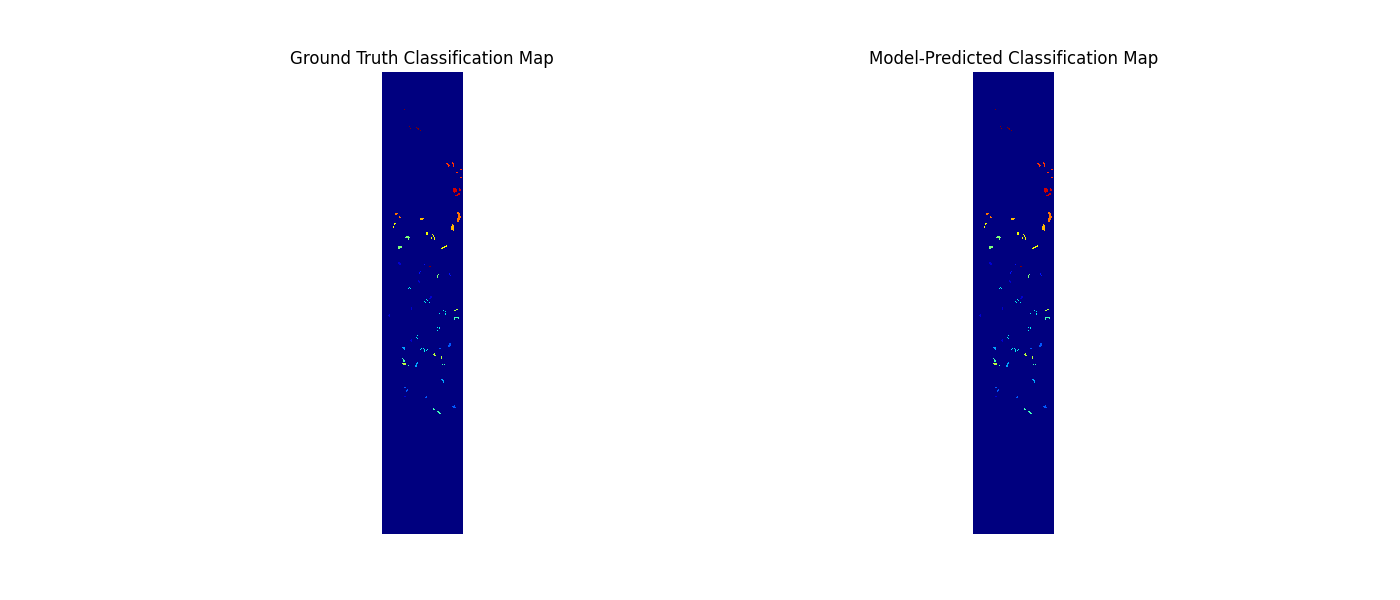}
        \caption{PWM}
    \end{subfigure}
        \begin{subfigure}[b]{0.3\textwidth}
        \includegraphics[width=0.18\linewidth, trim=275 50 675 60, clip, angle=90]{Output_Map/Botswana/3D_ConvSST-T5Encoder_Large/PWM_prediction_map_run1.png}
        \caption{GT}
    \end{subfigure}
    \caption{Comparison of classification maps for the 3D-ConvSST-T5 model on the Botswana dataset, showing different fusion methods: Cross Attention (CA), Concatenation (CONCAT), Multi-Head Attention (MHA), Pixel-Wise Addition (PWA), Pixel-Wise Multiplication (PWM), and Ground Truth (GT).}
    \vspace{-3mm}
    \label{fig:3D-ConvSST-T5-Botswana}
\end{figure*}

%%%%%%%%%%%%%%%%%%%%%%%%%%%%%%%%%%%%%%%%%%%%% Botswana     3D_ConvSST-BertEncoder%%%%%%%%%%%%%%%%%%%%%%%%%%%%%%%%%%%%%%%%%%%%%%%%
\begin{figure*}[]
    \centering
    \newcommand{\mapimg}[3][]{%
        \begin{subfigure}[b]{0.3\textwidth}
            \ifx&#1&
                \includegraphics[width=0.18\linewidth, trim=700 50 250 60, clip, angle=90]{Output_Map/Botswana/3D_ConvSST-BertEncoder_Large/#2_prediction_map_run1.png}%
            \else
                \includegraphics[width=0.18\linewidth, trim=#1, clip, angle=90]{Output_Map/Botswana/3D_ConvSST-BertEncoder_Large/#2_prediction_map_run1.png}%
            \fi
            \caption{#2}
        \end{subfigure}
    }

    \mapimg{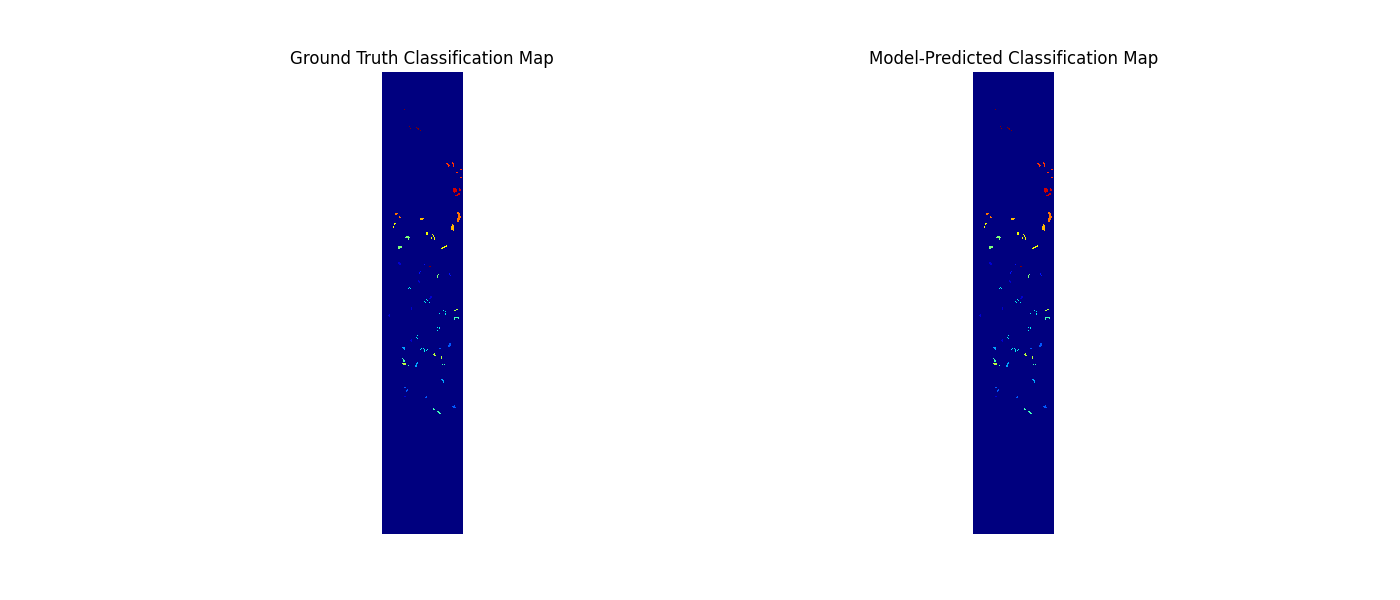}{CA}
    \mapimg{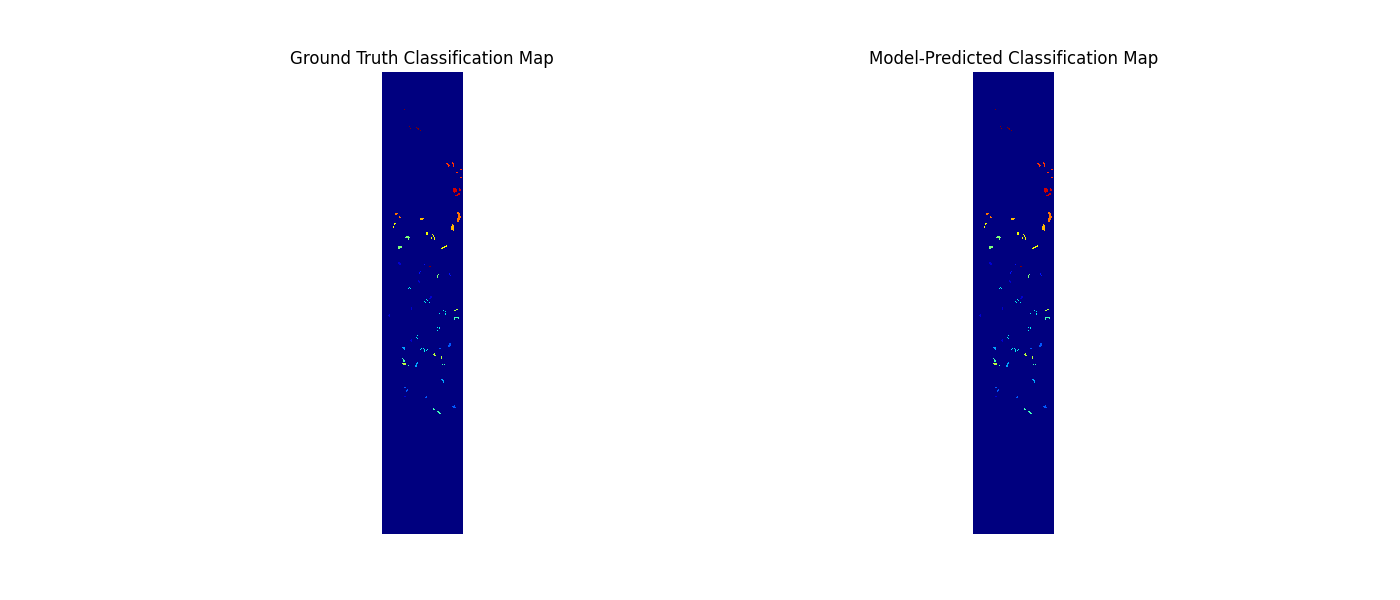}{CONCAT}
    \mapimg{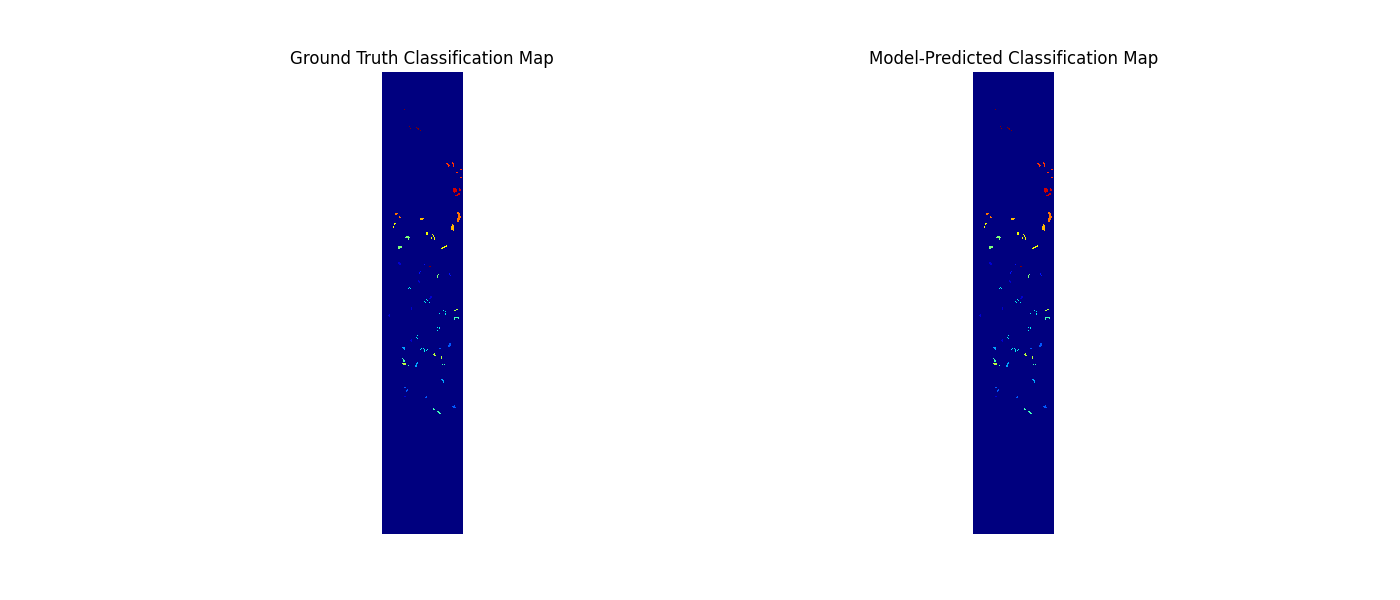}{MHA}
    \mapimg{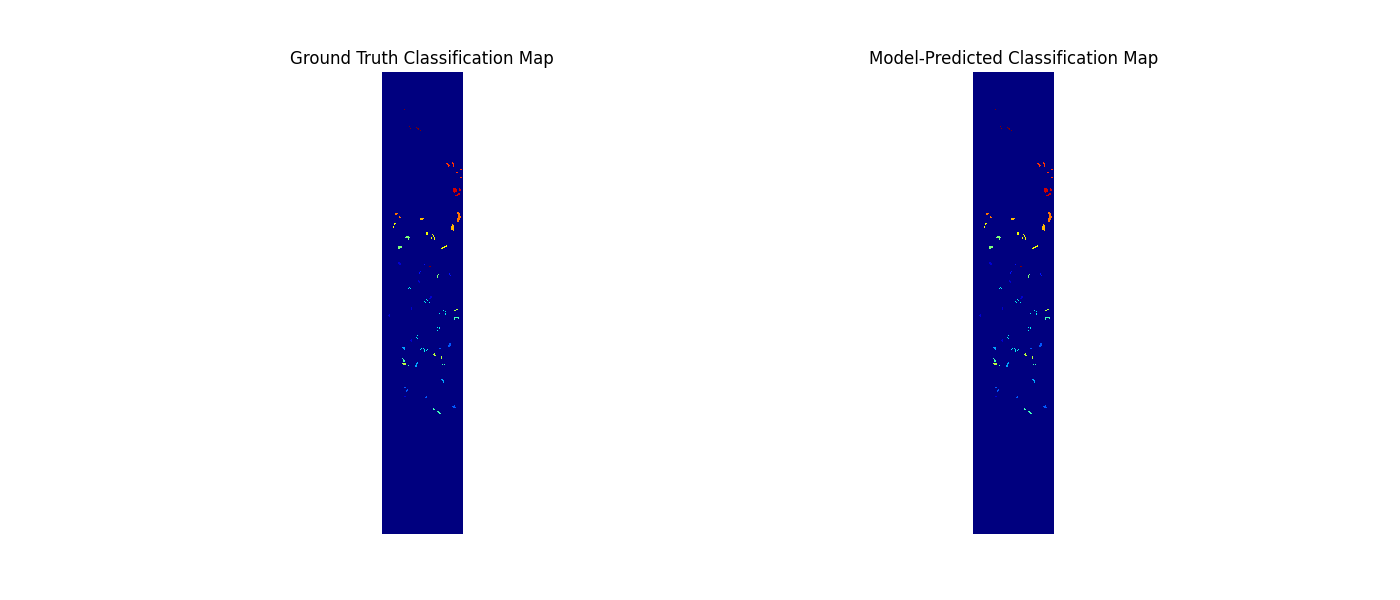}{PWA}
    \mapimg{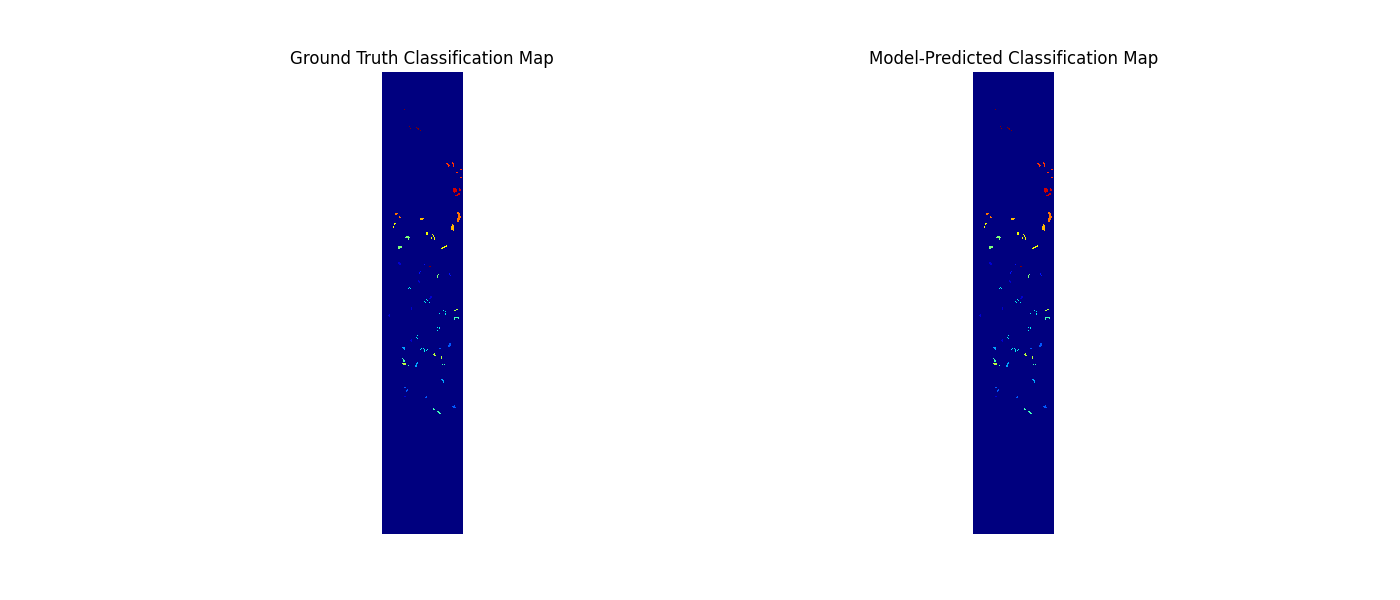}{PWM}
    \mapimg[275 50 675 60]{PWM}{GT}

    \caption{Comparison of classification maps for the 3D-ConvSST-BERT model on the Botswana dataset, showing different fusion methods: Cross Attention (CA), Concatenation (CONCAT), Multi-Head Attention (MHA), Pixel-Wise Addition (PWA), Pixel-Wise Multiplication (PWM), and Ground Truth (GT).}
    \vspace{-3mm}
    \label{fig:3D-ConvSST-Bert-Botswana}
\end{figure*}

%%%%%%%%%%%%%%%%%%%%%%%%%%%%%%%%%%%%%%%%% Botswana     DBCTNet-T5Encoder%%%%%%%%%%%%%%%%%%%%%%%%%%%%%%%%%%%%%%%%%%%%%%%%
\begin{figure*}[]
    \centering
    \newcommand{\mapimg}[3][]{%
        \begin{subfigure}[b]{0.3\textwidth}
            \ifx&#1&
                \includegraphics[width=0.18\linewidth, trim=700 50 250 60, clip, angle=90]{Output_Map/Botswana/DBCTNet-T5Encoder_Large/#2_prediction_map_run1.png}%
            \else
                \includegraphics[width=0.18\linewidth, trim=#1, clip, angle=90]{Output_Map/Botswana/DBCTNet-T5Encoder_Large/#2_prediction_map_run1.png}%
            \fi
            \caption{#2}
        \end{subfigure}
    }

    \mapimg{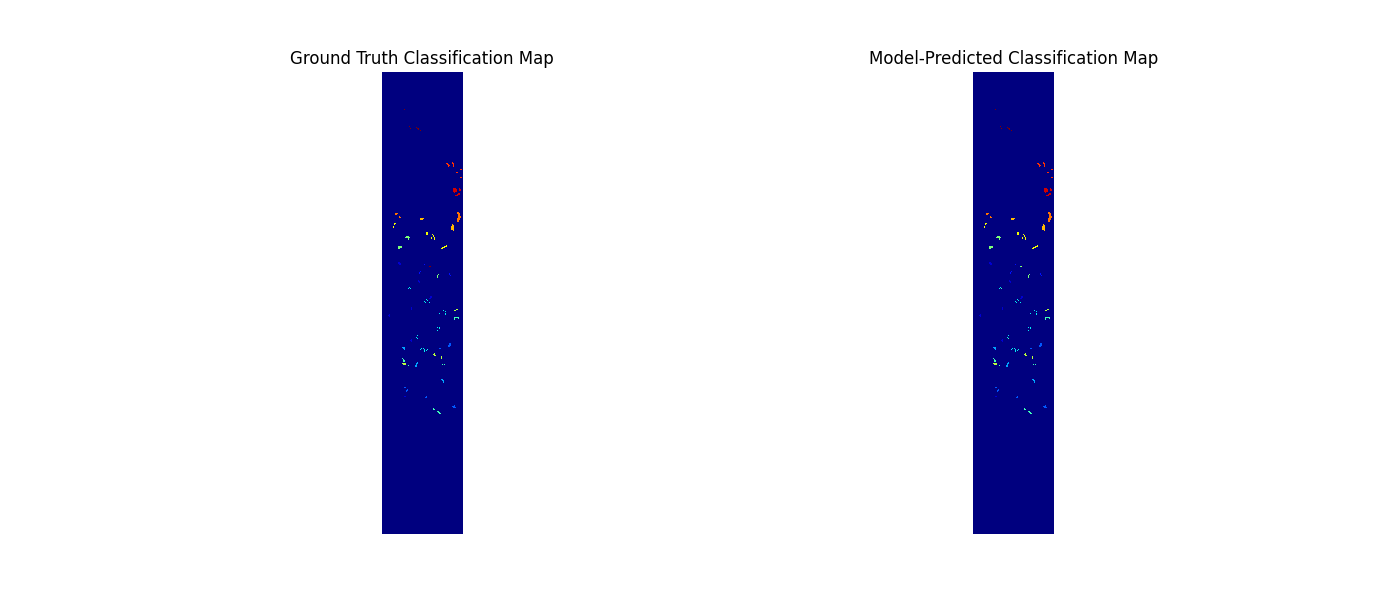}{CA}
    \mapimg{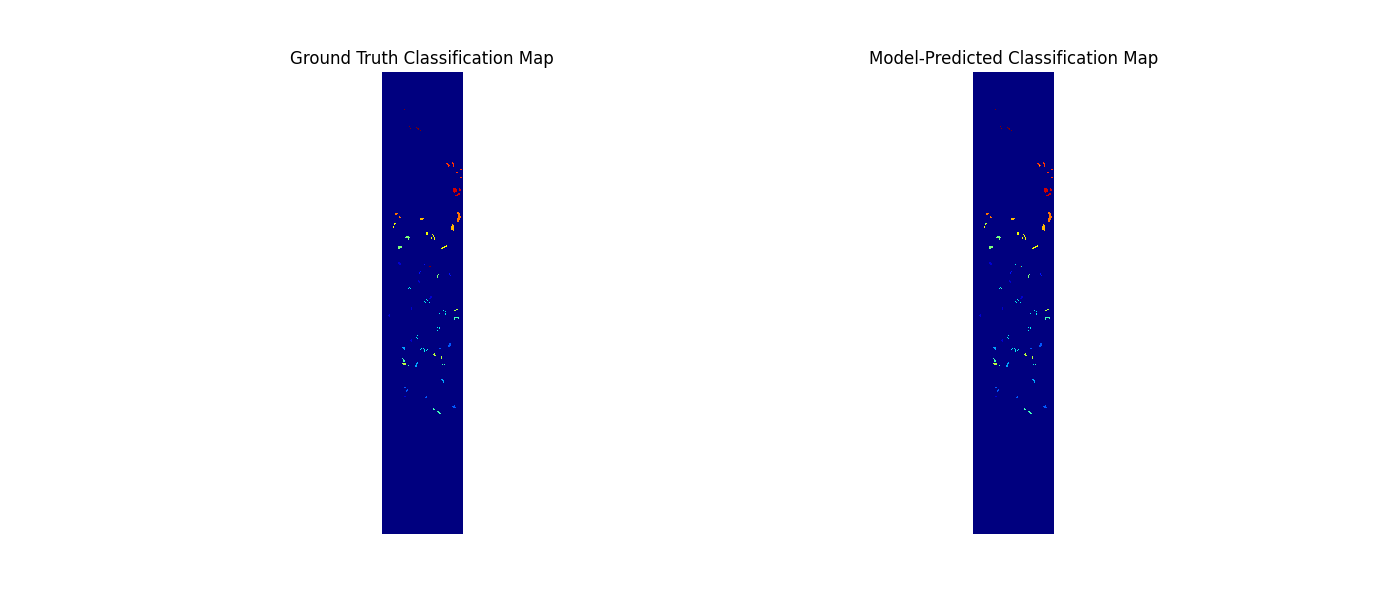}{CONCAT}
    \mapimg{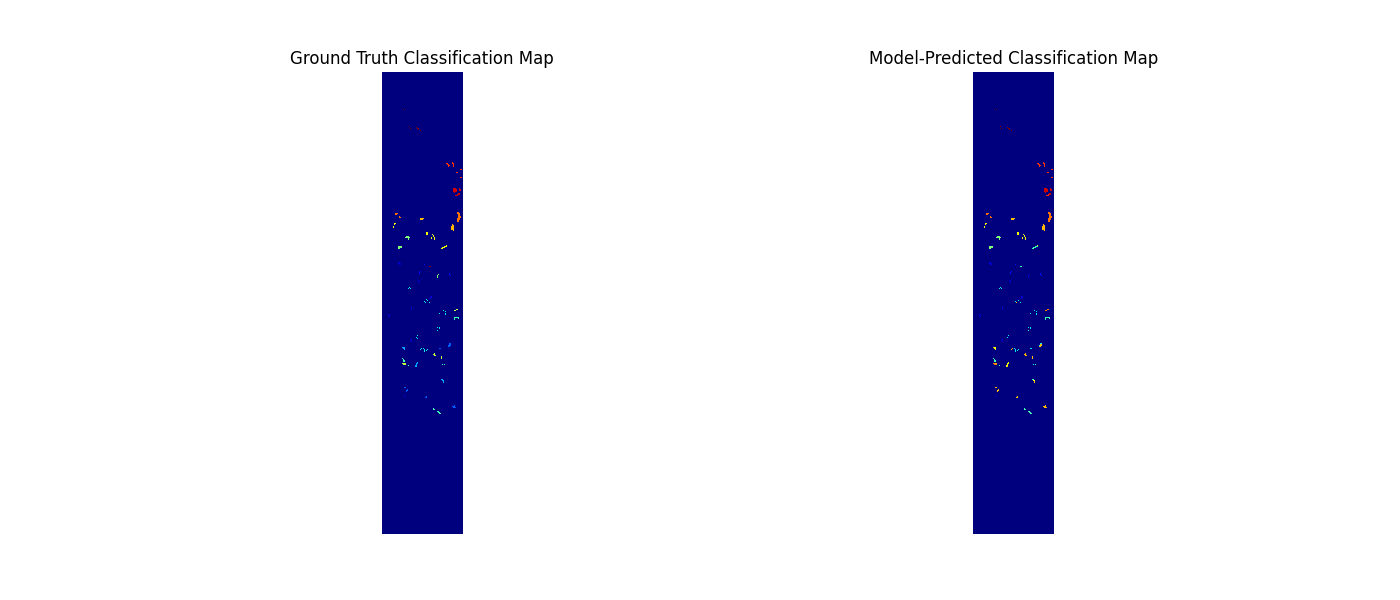}{MHA}
    \mapimg{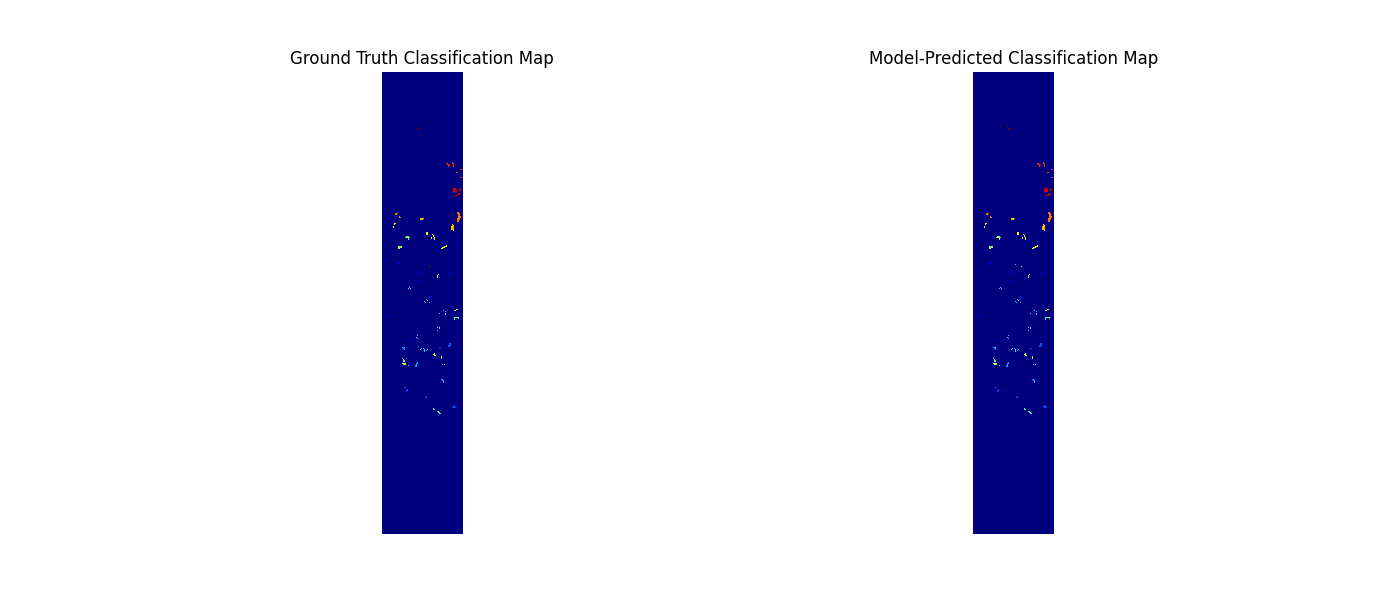}{PWA}
    \mapimg{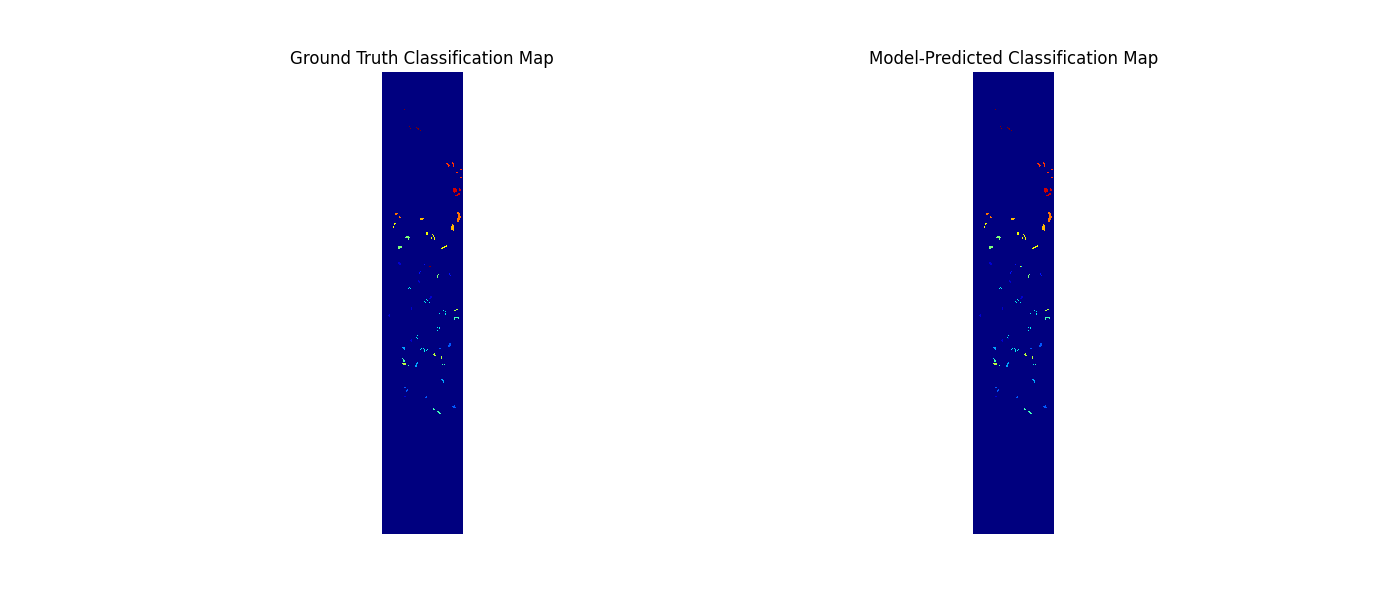}{PWM}
    \mapimg[275 50 675 60]{PWM}{GT}

    \caption{Comparison of classification maps for the DBCTNet-T5 model on the Botswana dataset, showing different fusion methods: Cross Attention (CA), Concatenation (CONCAT), Multi-Head Attention (MHA), Pixel-Wise Addition (PWA), Pixel-Wise Multiplication (PWM), and Ground Truth (GT).}
    \vspace{-3mm}
    \label{fig:DBCTNet-T5-Botswana}
\end{figure*}

%%%%%%%%%%%%%%%%%%%%%%%%%%%%%%%%%%%%%%%%%%%%%%% Botswana     DBCTNet-BertEncoder%%%%%%%%%%%%%%%%%%%%%%%%%%%%%%%%%%%%%%%%%%%%%%%%
\begin{figure*}[]
    \centering
    \newcommand{\mapimg}[3][]{%
        \begin{subfigure}[b]{0.3\textwidth}
            \ifx&#1&
                \includegraphics[width=0.18\linewidth, trim=700 50 250 60, clip, angle=90]{Output_Map/Botswana/DBCTNet-BertEncoder_Large/#2_prediction_map_run1.png}%
            \else
                \includegraphics[width=0.18\linewidth, trim=#1, clip, angle=90]{Output_Map/Botswana/DBCTNet-BertEncoder_Large/#2_prediction_map_run1.png}%
            \fi
            \caption{#2}
        \end{subfigure}
    }

    \mapimg{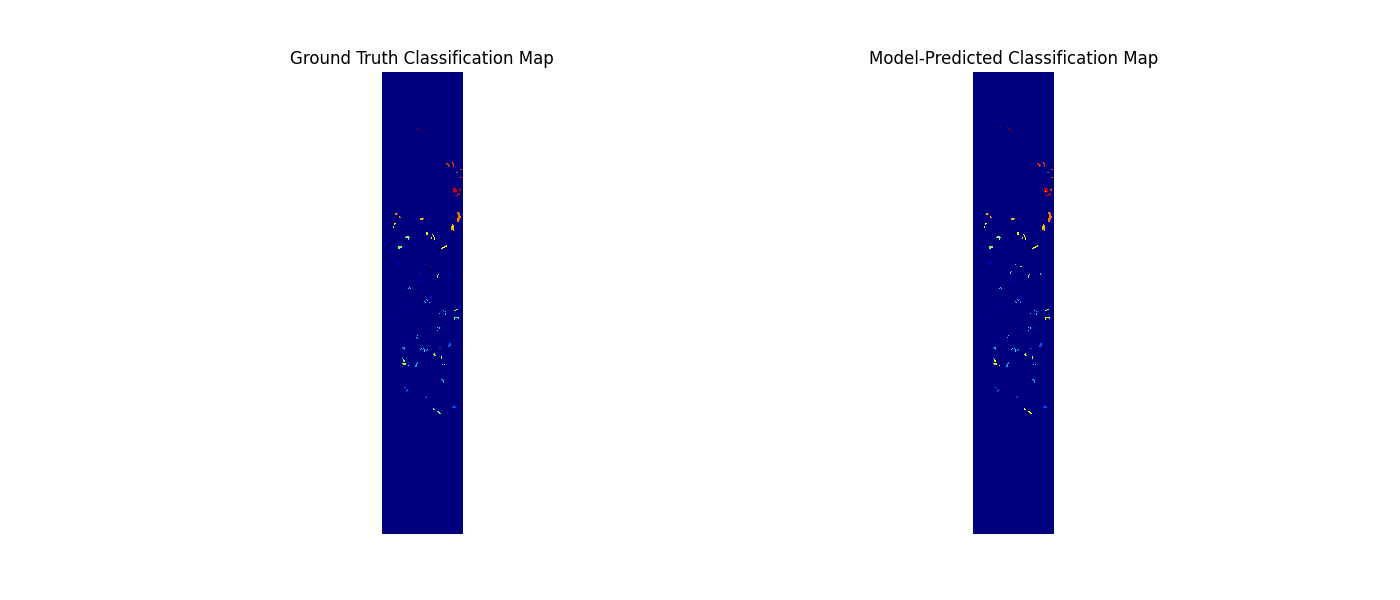}{CA}
    \mapimg{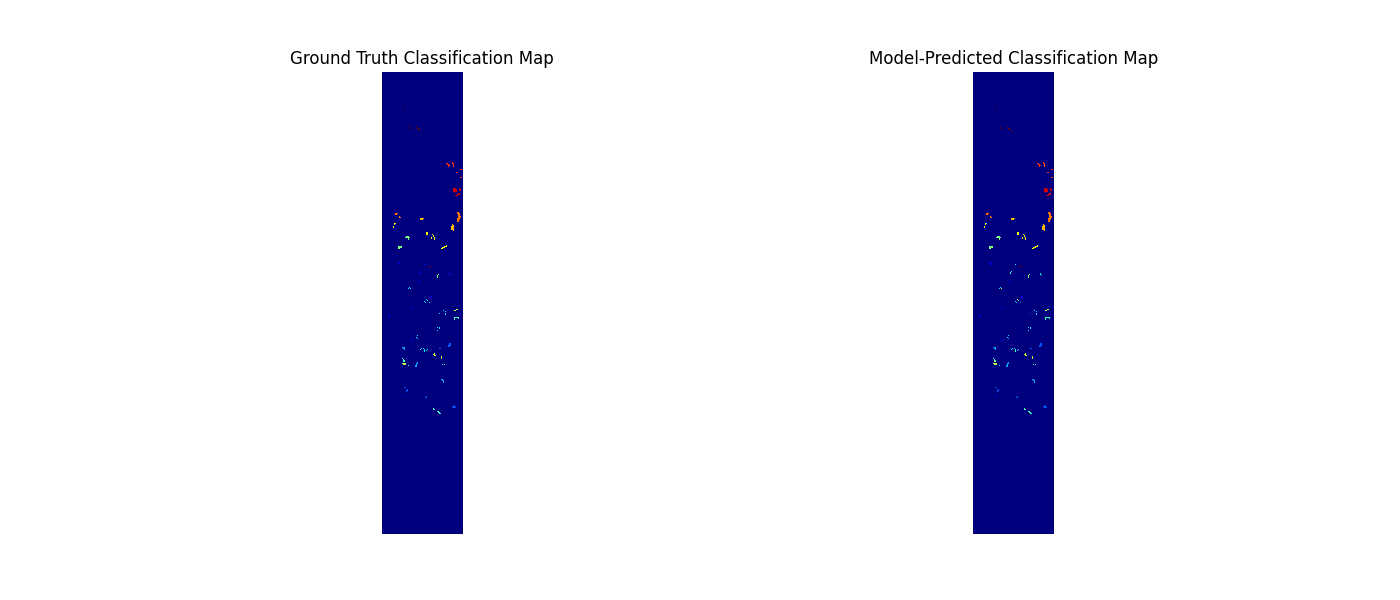}{CONCAT}
    \mapimg{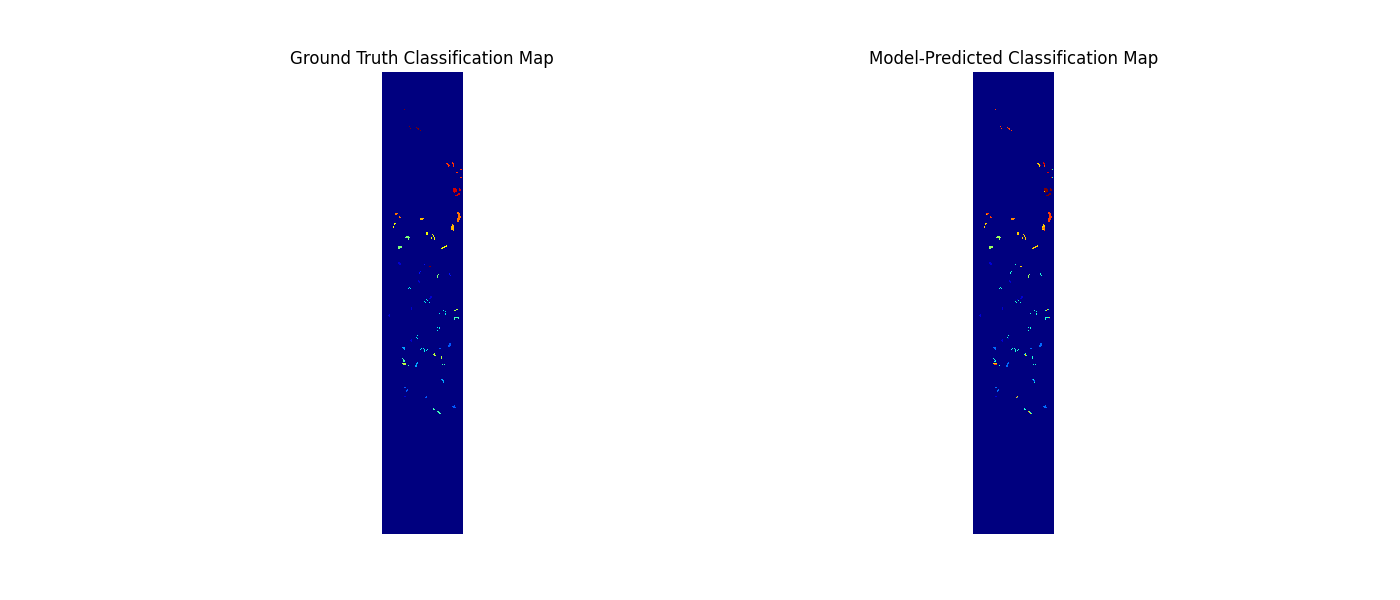}{MHA}
    \mapimg{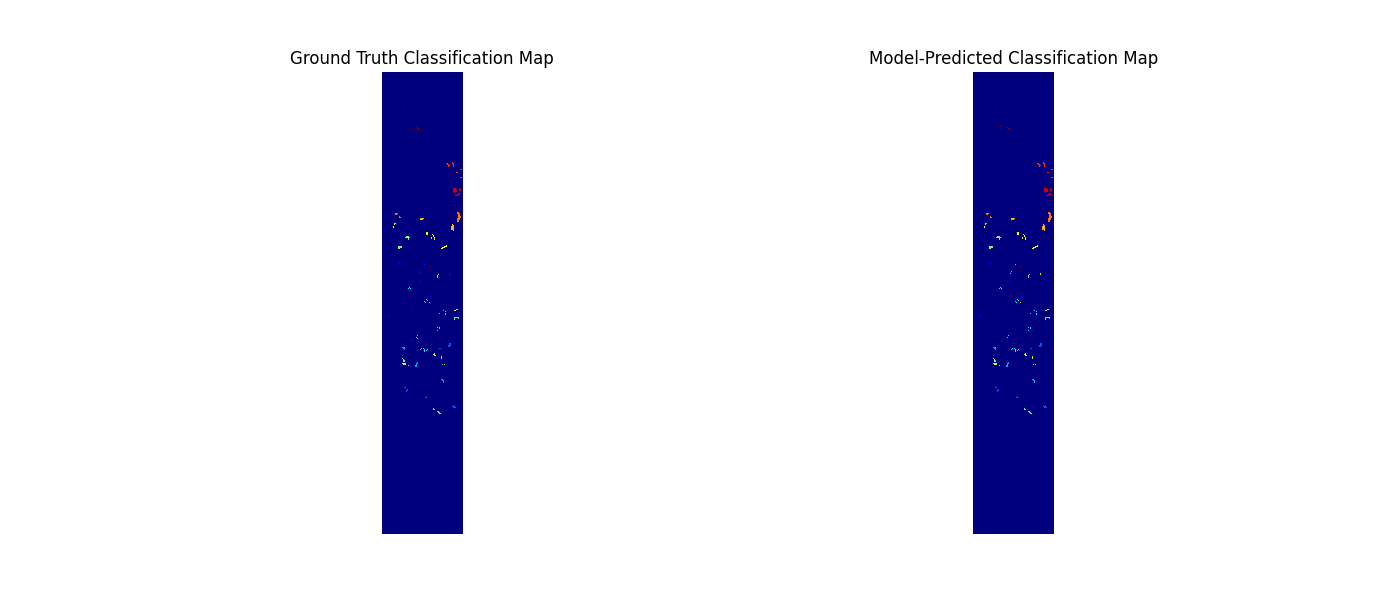}{PWA}
    \mapimg{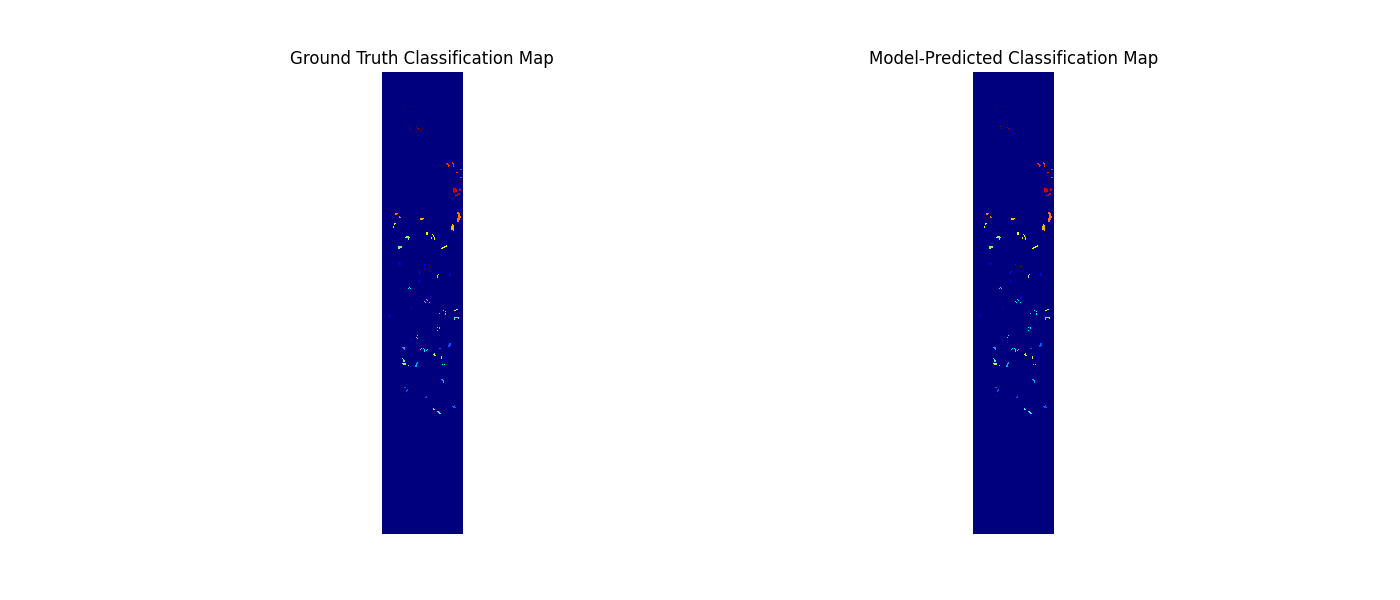}{PWM}
    \mapimg[275 50 675 60]{PWM}{GT}

    \caption{Comparison of classification maps for the DBCTNet-BERT model on the Botswana dataset, showing different fusion methods: Cross Attention (CA), Concatenation (CONCAT), Multi-Head Attention (MHA), Pixel-Wise Addition (PWA), Pixel-Wise Multiplication (PWM), and Ground Truth (GT).}
    \vspace{-3mm}
    \label{fig:DBCTNet-Bert-Botswana}
\end{figure*}

%%%%% Botswana
\begin{figure*}[]
    \centering
    % \vspace{-3mm}
    \begin{subfigure}[b]{0.3\textwidth}
         \includegraphics[width=0.18\linewidth, trim=700 50 250 60, clip, angle=90]{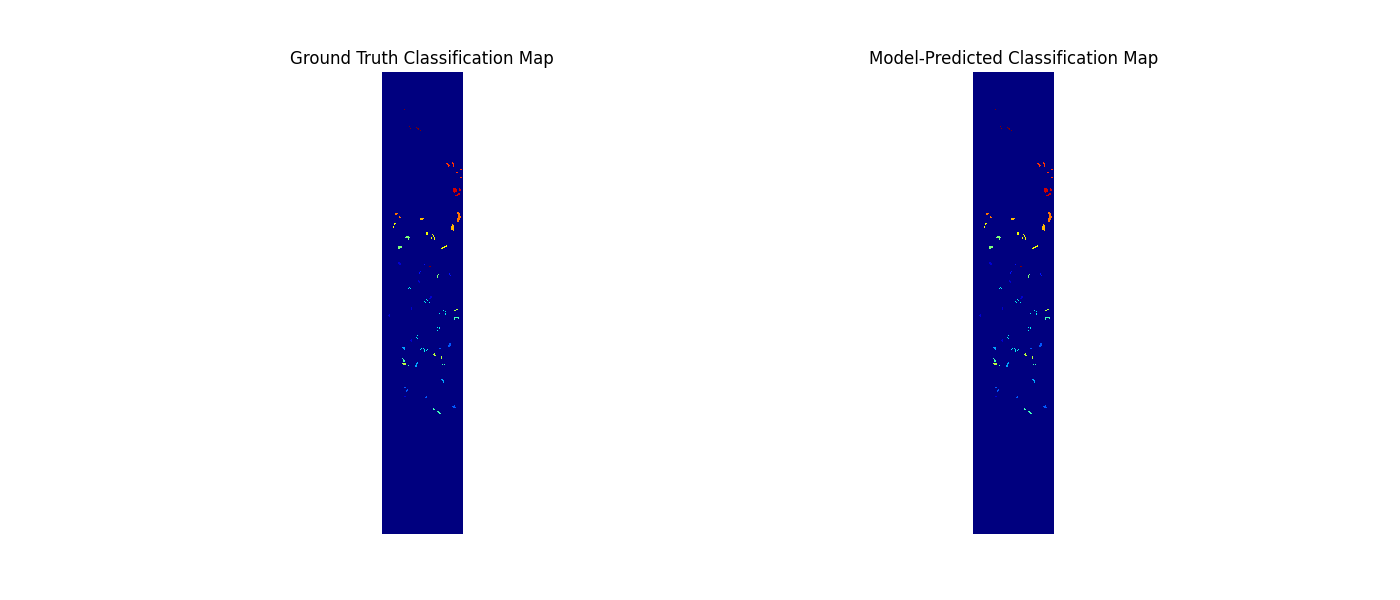}
        \caption{CA}
    \end{subfigure}
    \begin{subfigure}[b]{0.3\textwidth}
        \includegraphics[width=0.18\linewidth, trim=700 50 250 60, clip, angle=90]{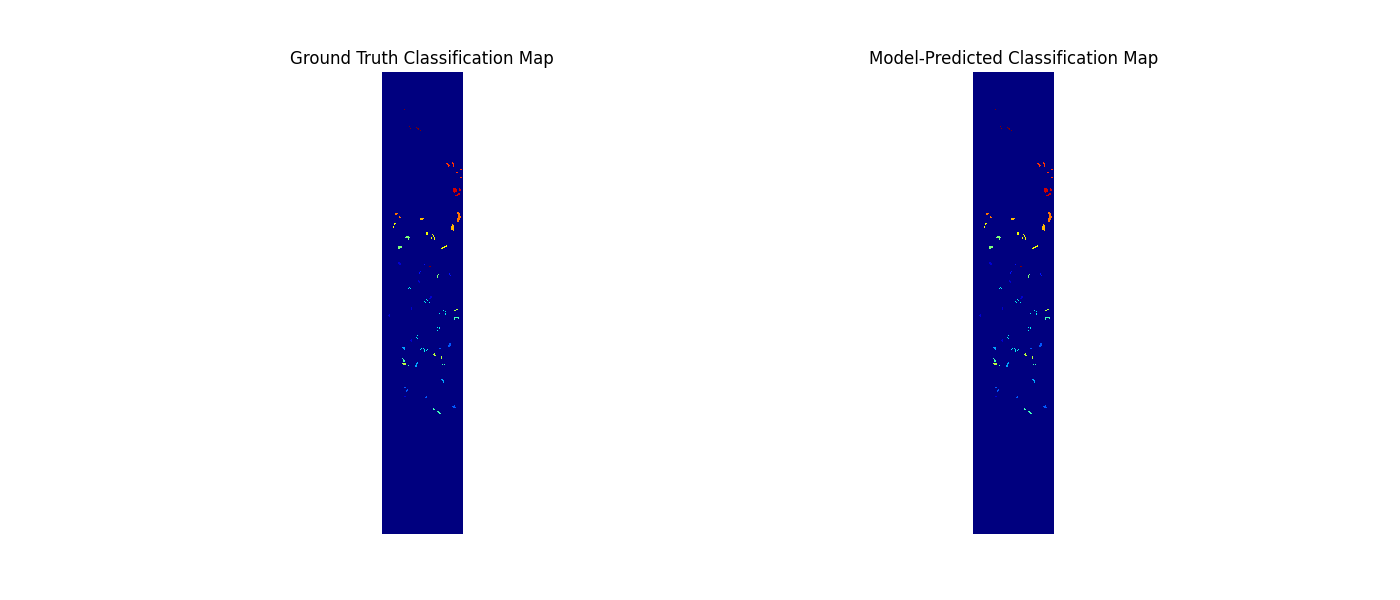}
        \caption{CONCAT}
    \end{subfigure}
    \begin{subfigure}[b]{0.3\textwidth}
        \includegraphics[width=0.18\linewidth, trim=700 50 250 60, clip, angle=90]{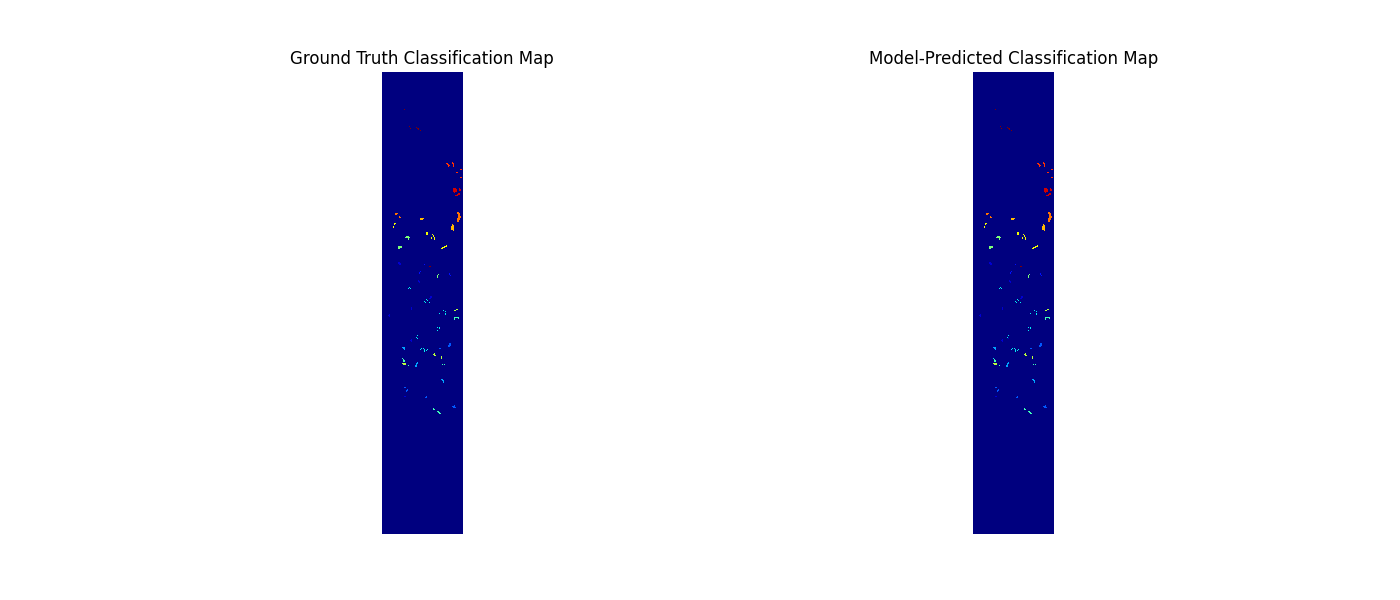}
        \caption{MHA}
    \end{subfigure}
    \begin{subfigure}[b]{0.3\textwidth}
        \includegraphics[width=0.18\linewidth, trim=700 50 250 60, clip, angle=90]{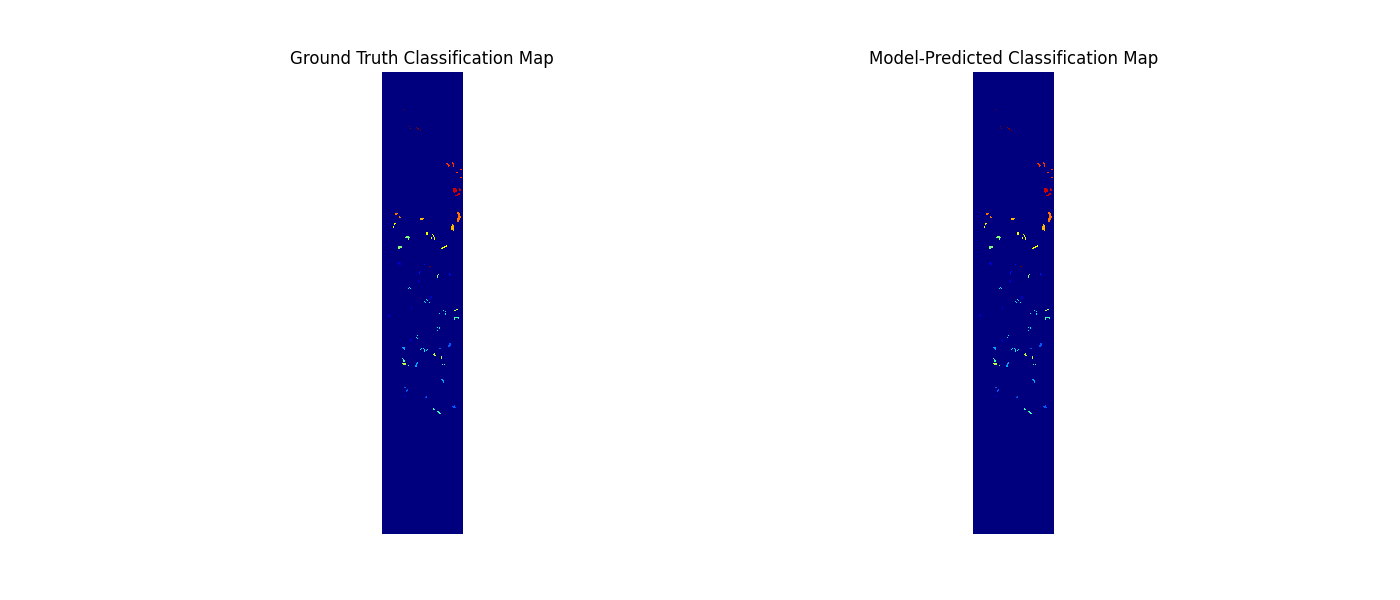}
        \caption{PWA}
    \end{subfigure}
    \begin{subfigure}[b]{0.3\textwidth}
        \includegraphics[width=0.18\linewidth, trim=700 50 250 60, clip, angle=90]{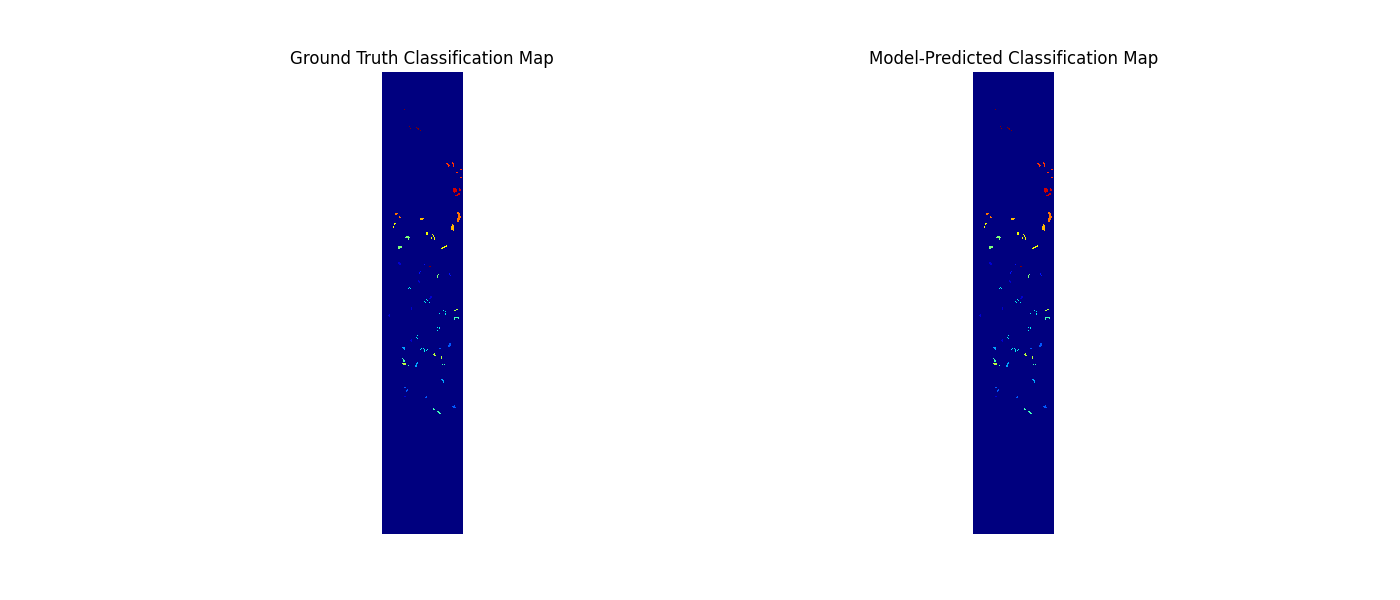}
        \caption{PWM}
    \end{subfigure}
        \begin{subfigure}[b]{0.3\textwidth}
        \includegraphics[width=0.18\linewidth, trim=275 50 675 60, clip, angle=90]{Output_Map/Botswana/FAHM-T5Encoder_Large/PWM_prediction_map_run1.png}
        \caption{GT}
    \end{subfigure}
    \caption{Comparison of classification maps for the FAHM-T5 model on the Botswana dataset, showing different fusion methods: Cross Attention (CA), Concatenation (CONCAT), Multi-Head Attention (MHA), Pixel-Wise Addition (PWA), Pixel-Wise Multiplication (PWM), and Ground Truth (GT).}
    \label{fig:FAHM-T5-Botswana}
    % \vspace{-6mm}
\end{figure*}
%%%%%%%%%%%%%%%%%%%%%%%%%%%%%%%%%%% Botswana     FAHM-BertEncoder%%%%%%%%%%%%%%%%%%%%%%%%%%%%%%%%%%%%%%%%%%%%%%%%
\begin{figure*}[]
    \centering
    \newcommand{\mapimg}[3][]{%
        \begin{subfigure}[b]{0.3\textwidth}
            \ifx&#1&
                \includegraphics[width=0.18\linewidth, trim=700 50 250 60, clip, angle=90]{Output_Map/Botswana/FAHM-BertEncoder_Large/#2_prediction_map_run1.png}%
            \else
                \includegraphics[width=0.18\linewidth, trim=#1, clip, angle=90]{Output_Map/Botswana/FAHM-BertEncoder_Large/#2_prediction_map_run1.png}%
            \fi
            \caption{#2}
        \end{subfigure}
    }

    \mapimg{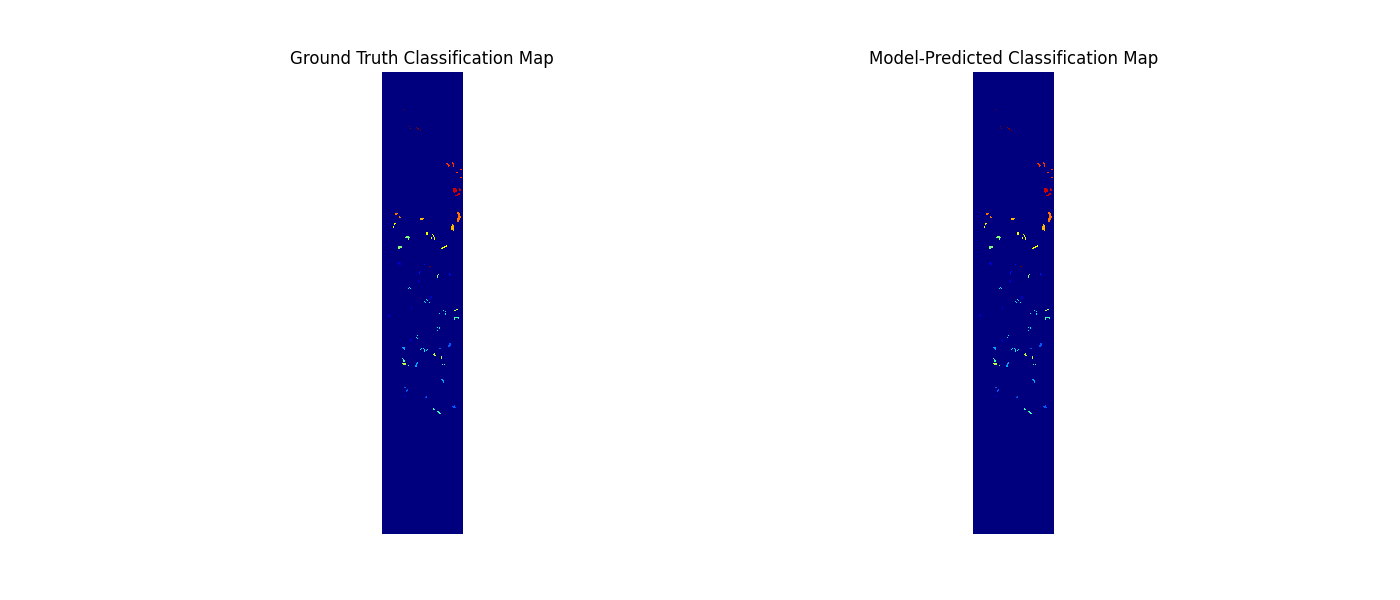}{CA}
    \mapimg{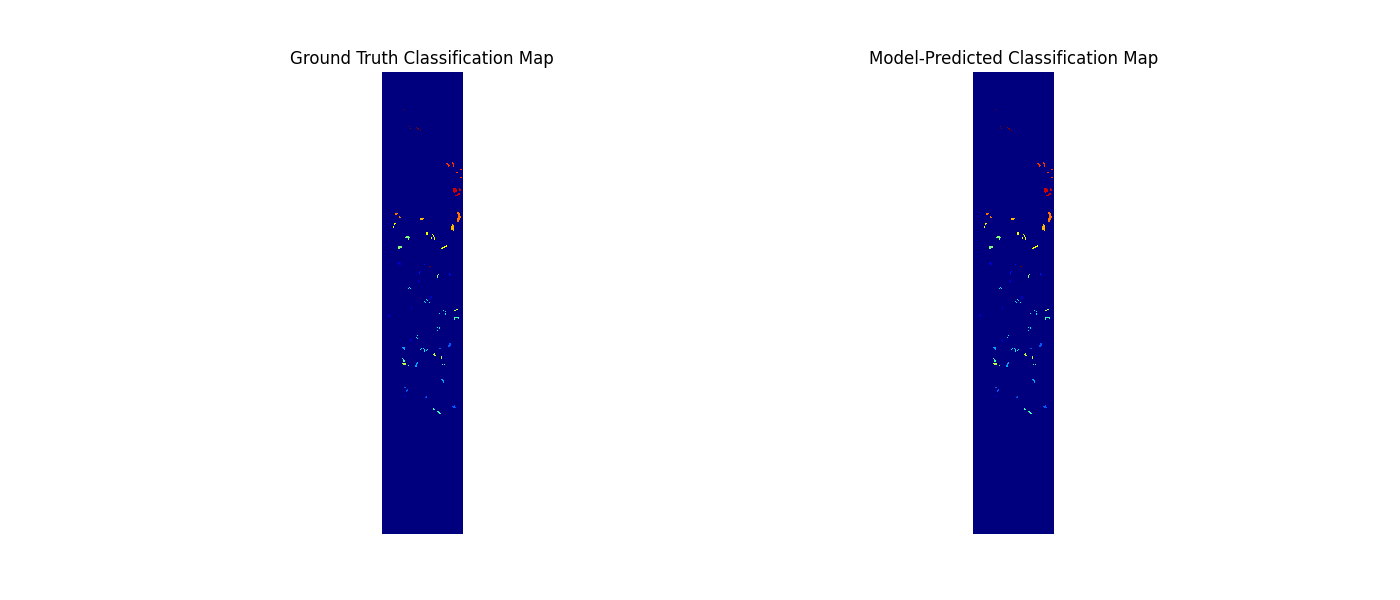}{CONCAT}
    \mapimg{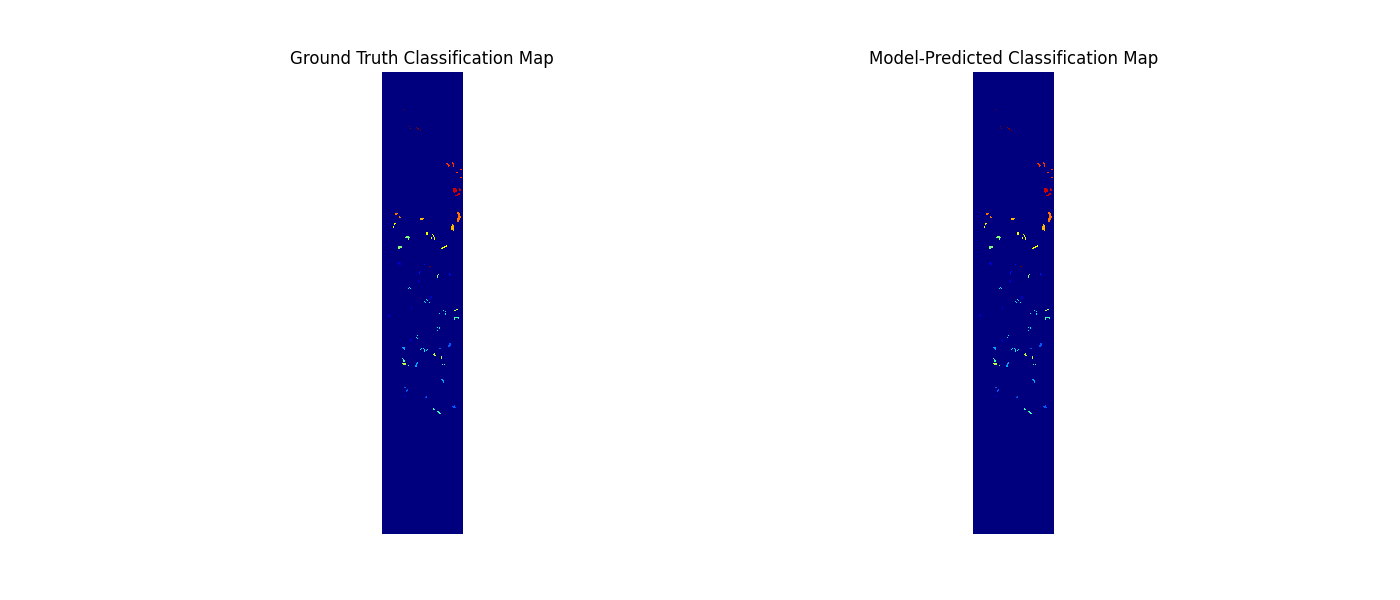}{MHA}
    \mapimg{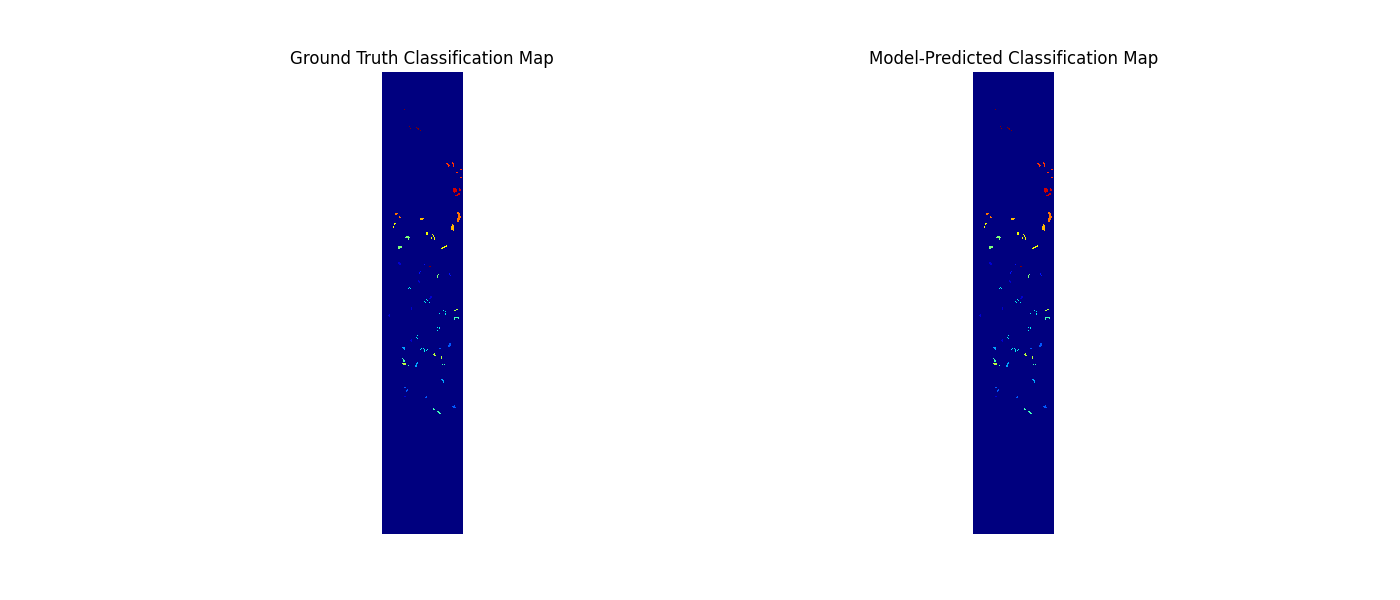}{PWA}
    \mapimg{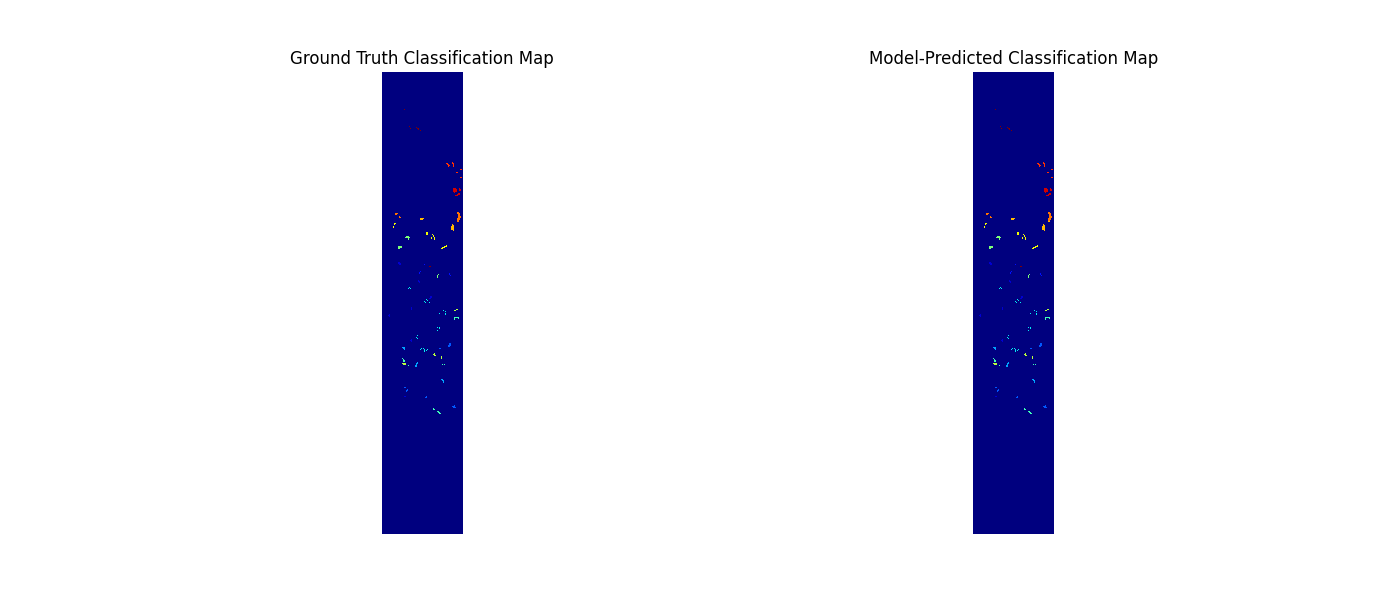}{PWM}
    \mapimg[275 50 675 60]{PWM}{GT}

    \caption{Comparison of classification maps for the FAHM-BERT model on the Botswana dataset, showing different fusion methods: Cross Attention (CA), Concatenation (CONCAT), Multi-Head Attention (MHA), Pixel-Wise Addition (PWA), Pixel-Wise Multiplication (PWM), and Ground Truth (GT).}
    \vspace{-3mm}
    \label{fig:FAHM-Bert-Botswana}
\end{figure*}

The Botswana dataset highlights a clear transformation from fragmented, noisy land cover predictions (in vision-only models) to highly uniform and crisp maps when text encoders (BERT, T5) are fused using cross-attention, concatenation, or multi-head attention strategies. 3D-RCNet and DBCTNet start with modest performance but, with any strong fusion strategy, the output fields become smooth and borders sharply defined—almost matching the visual quality of ground truth. Powerful backbones like 3D-ConvSST and FAHM, especially with attention-based fusions, saturate visually to ground-truth quality, eradicating most mapping errors and noise. The figures (\ref{fig:3D-RCNet-T5-Botswana}, \ref{fig:3D-RCNet-Bert-Botswana}, \ref{fig:3D-ConvSST-T5-Botswana}, \ref{fig:3D-ConvSST-Bert-Botswana}, \ref{fig:DBCTNet-T5-Botswana}, \ref{fig:DBCTNet-Bert-Botswana}, \ref{fig:FAHM-T5-Botswana}, \ref{fig:FAHM-Bert-Botswana}) show that both BERT and T5 consistently improve visual segmentation detail, though the choice of specific fusion (MHA, CONCAT) often has more impact than the specific language model, leading to robust map quality under various backbone configurations.

% ============================================================================
% HOUSTON13 DATASET
% ============================================================================

%%%%%%%%%%%%%%%%%%%%%%%%%%%%%%%% Houston13 3DRCNet-T5Encoder %%%%%%%%%%%%%%%%%%%%%%%%%%%%%%%%%%%%%%%%%%%%%%%%
\begin{figure*}[]
    \centering
    \begin{subfigure}[b]{0.3\textwidth}
         \includegraphics[width=0.18\linewidth, trim=700 50 250 60, clip, angle=90]{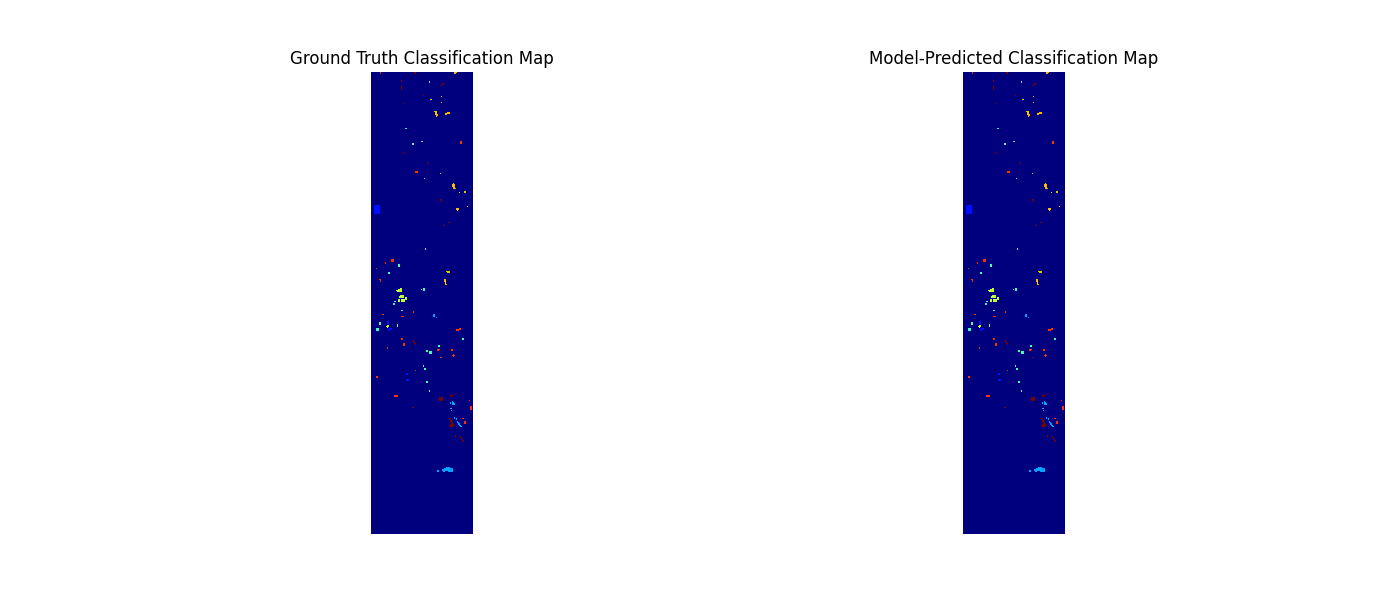}
        \caption{CA}
    \end{subfigure}
    \begin{subfigure}[b]{0.3\textwidth}
        \includegraphics[width=0.18\linewidth, trim=700 50 250 60, clip, angle=90]{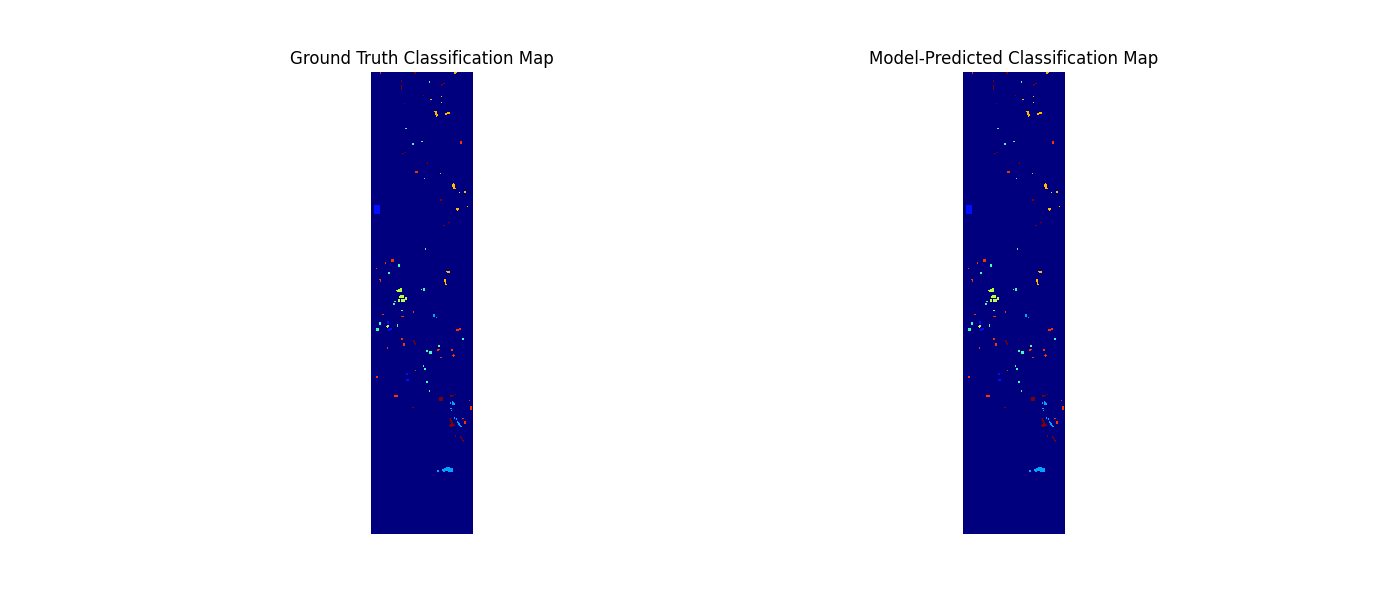}
        \caption{CONCAT}
    \end{subfigure}
    \begin{subfigure}[b]{0.3\textwidth}
        \includegraphics[width=0.18\linewidth, trim=700 50 250 60, clip, angle=90]{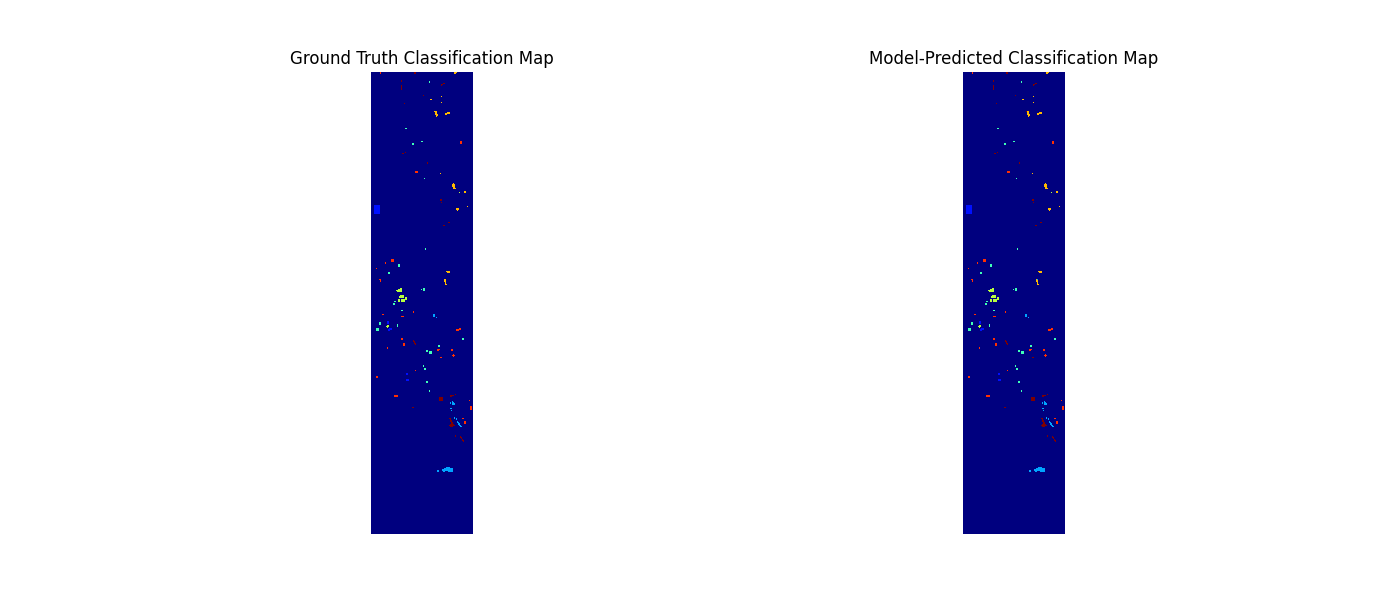}
        \caption{MHA}
    \end{subfigure}
    \begin{subfigure}[b]{0.3\textwidth}
        \includegraphics[width=0.18\linewidth, trim=700 50 250 60, clip, angle=90]{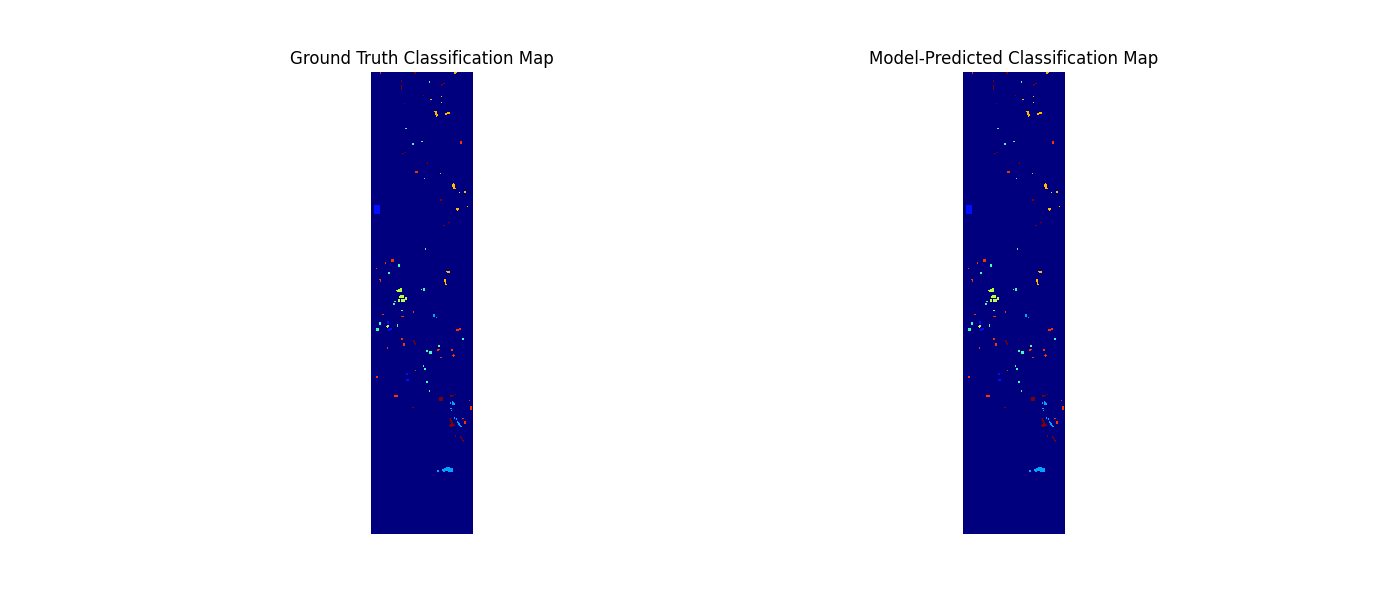}
        \caption{PWA}
    \end{subfigure}
    \begin{subfigure}[b]{0.3\textwidth}
        \includegraphics[width=0.18\linewidth, trim=700 50 250 60, clip, angle=90]{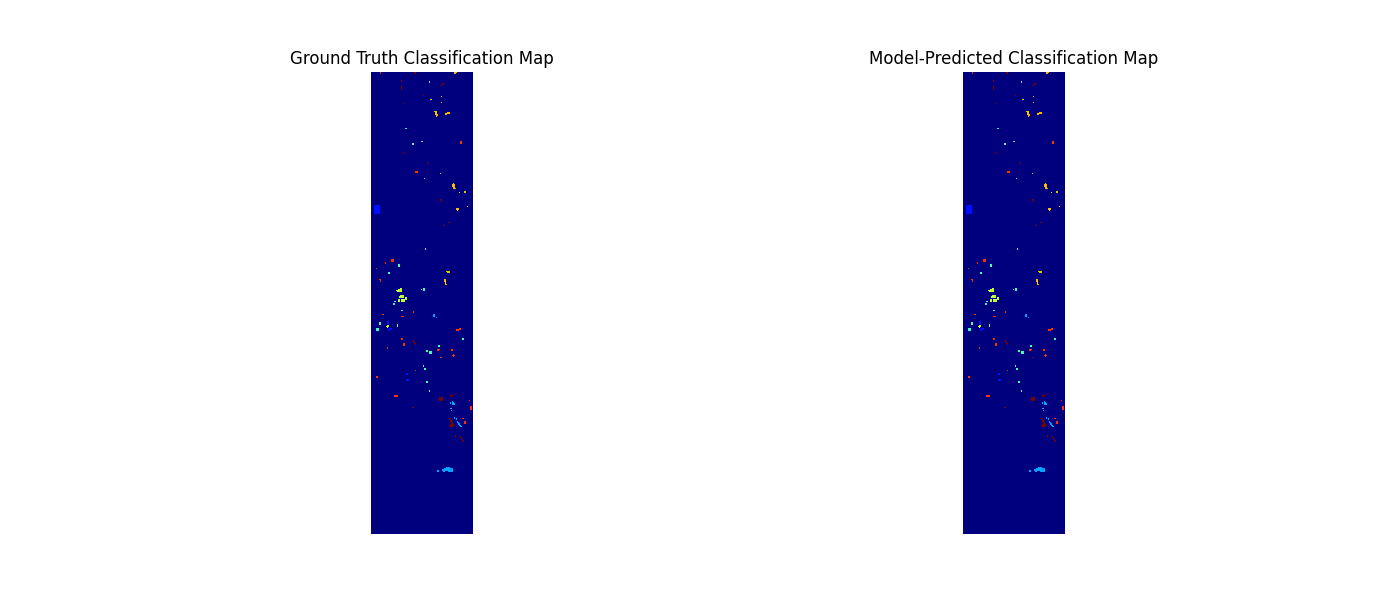}
        \caption{PWM}
    \end{subfigure}
        \begin{subfigure}[b]{0.3\textwidth}
         \includegraphics[width=0.18\linewidth, trim=275 50 675 60, clip, angle=90]{Output_Map/Houston13/3DRCNet-T5Encoder_Large/PWM_prediction_map_run1.png}
        \caption{GT}
    \end{subfigure}
    \caption{Comparison of classification maps for the 3D-RCNet-T5 model on the Houston13 dataset, showing different fusion methods: Cross Attention (CA), Concatenation (CONCAT), Multi-Head Attention (MHA), Pixel-Wise Addition (PWA), Pixel-Wise Multiplication (PWM), and Ground Truth (GT).}
    \vspace{-3mm}
    \label{fig:3D-RCNet-T5-Houston13}
\end{figure*}

%%%%%%%%%%%%%%%%%%%%%%%%%%%%%%%%%%%%%%%%%%%%%%%%  Houston13  3DRCNet-BertEncoder   %%%%%%%%%%%%%%%%%%%%%%%%%%%%%%%%%%%%%%%%%%%%%%%%%%%%%%%%%%%%%%%%%%%%%
\begin{figure*}[]
    \centering
    \begin{subfigure}[b]{0.3\textwidth}
         \includegraphics[width=0.18\linewidth, trim=700 50 250 60, clip, angle=90]{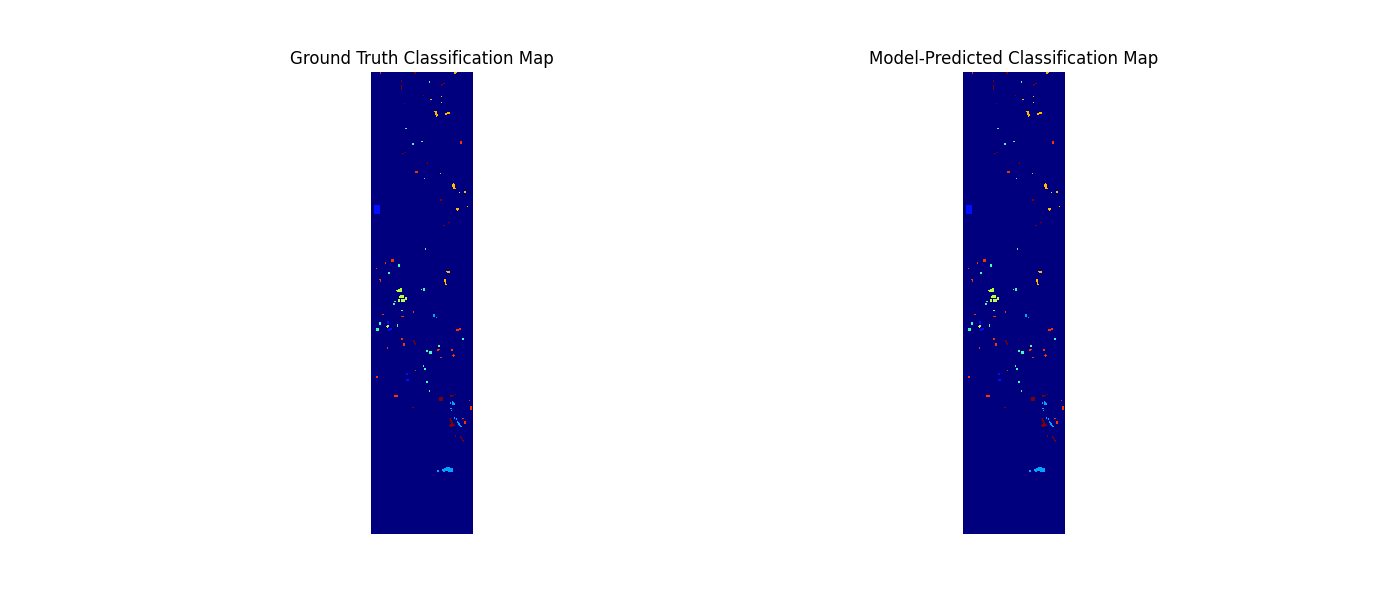}
        \caption{CA}
    \end{subfigure}
    \begin{subfigure}[b]{0.3\textwidth}
        \includegraphics[width=0.18\linewidth, trim=700 50 250 60, clip, angle=90]{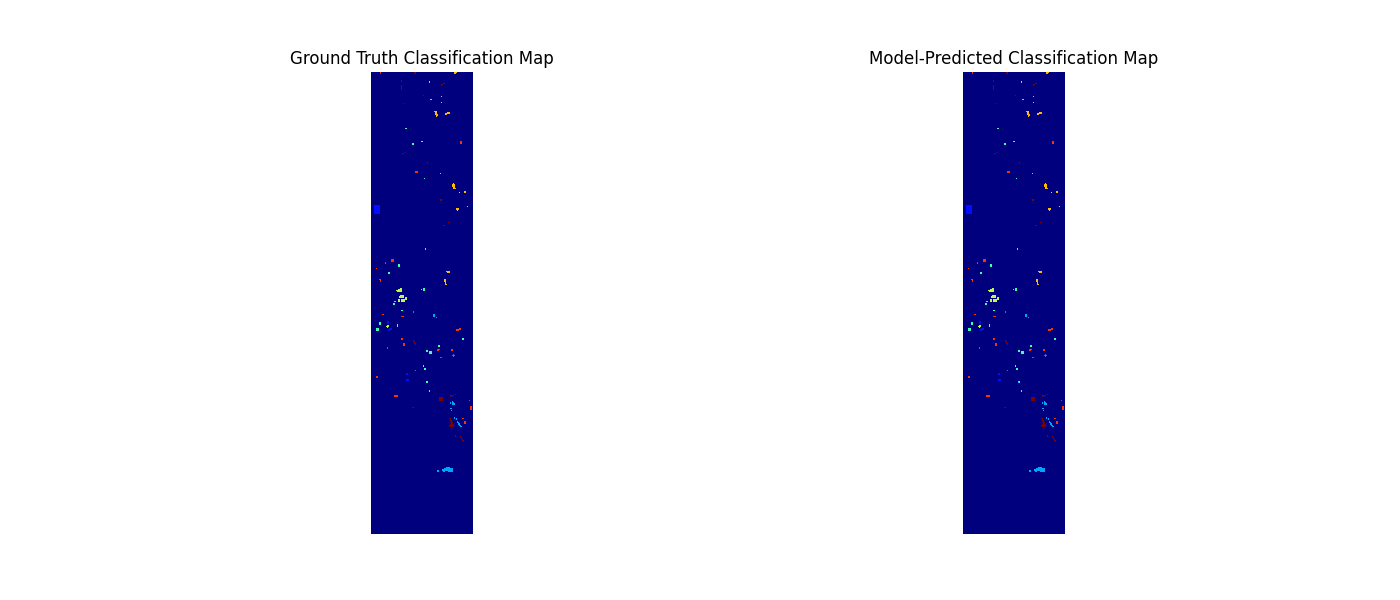}
        \caption{CONCAT}
    \end{subfigure}
    \begin{subfigure}[b]{0.3\textwidth}
        \includegraphics[width=0.18\linewidth, trim=700 50 250 60, clip, angle=90]{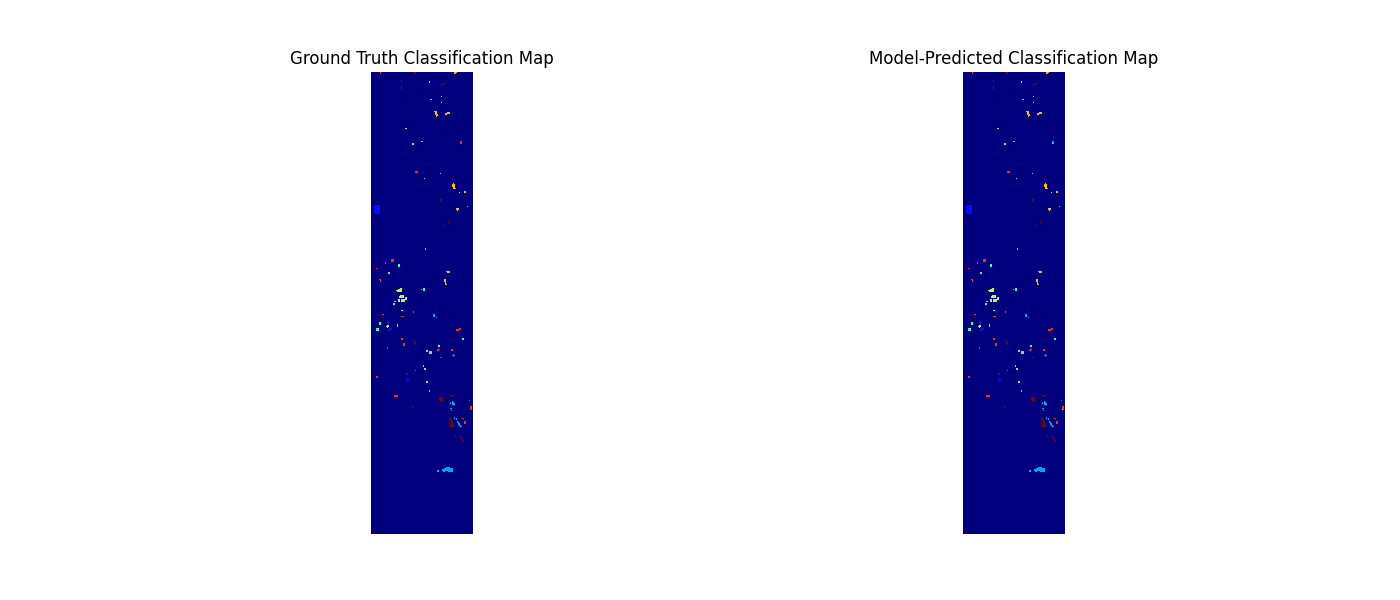}
        \caption{MHA}
    \end{subfigure}
    \begin{subfigure}[b]{0.3\textwidth}
        \includegraphics[width=0.18\linewidth, trim=700 50 250 60, clip, angle=90]{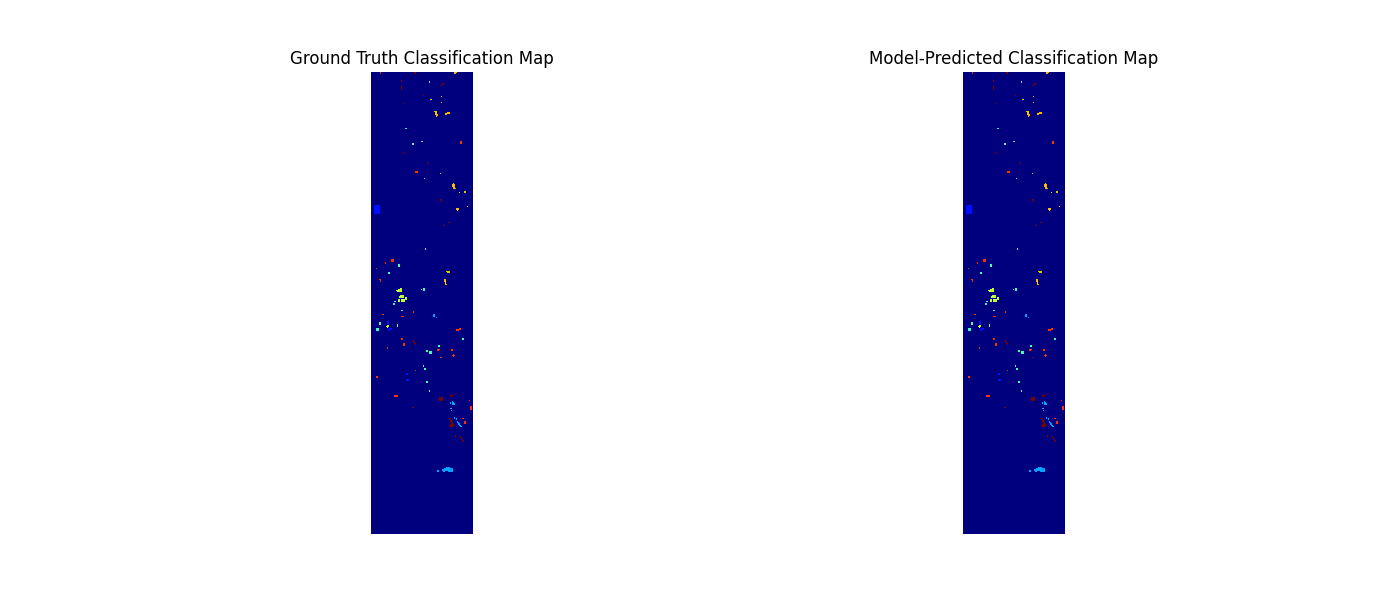}
        \caption{PWA}
    \end{subfigure}
    \begin{subfigure}[b]{0.3\textwidth}
        \includegraphics[width=0.18\linewidth, trim=700 50 250 60, clip, angle=90]{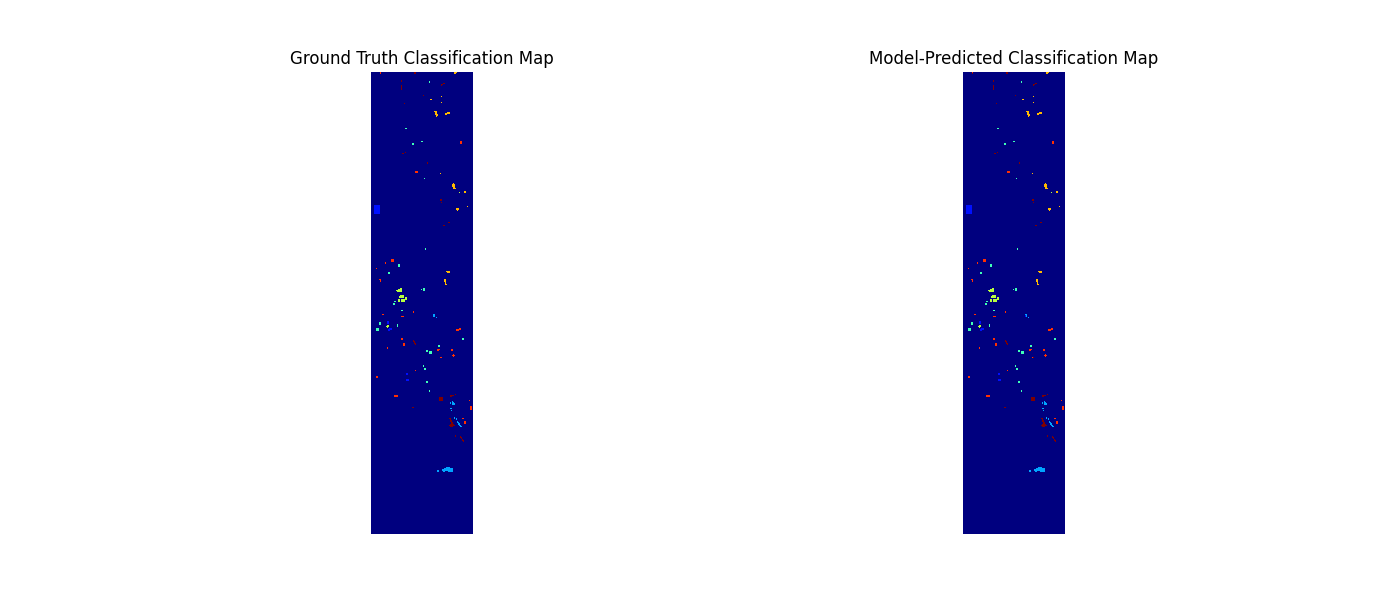}
        \caption{PWM}
    \end{subfigure}
        \begin{subfigure}[b]{0.3\textwidth}
         \includegraphics[width=0.18\linewidth, trim=275 50 675 60, clip, angle=90]{Output_Map/Houston13/3DRCNet-BertEncoder_Large/PWM_prediction_map_run1.png}
        \caption{GT}
    \end{subfigure}
    \caption{Comparison of classification maps for the 3D-RCNet-Bert model on the Houston13 dataset, showing different fusion methods: Cross Attention (CA), Concatenation (CONCAT), Multi-Head Attention (MHA), Pixel-Wise Addition (PWA), Pixel-Wise Multiplication (PWM), and Ground Truth (GT).}
    \vspace{-3mm}
    \label{fig:3D-RCNet-Bert-Houston13}
\end{figure*}

\begin{figure*}[]
    \centering
    \begin{subfigure}[b]{0.3\textwidth}
         \includegraphics[width=0.18\linewidth, trim=700 50 250 60, clip, angle=90]{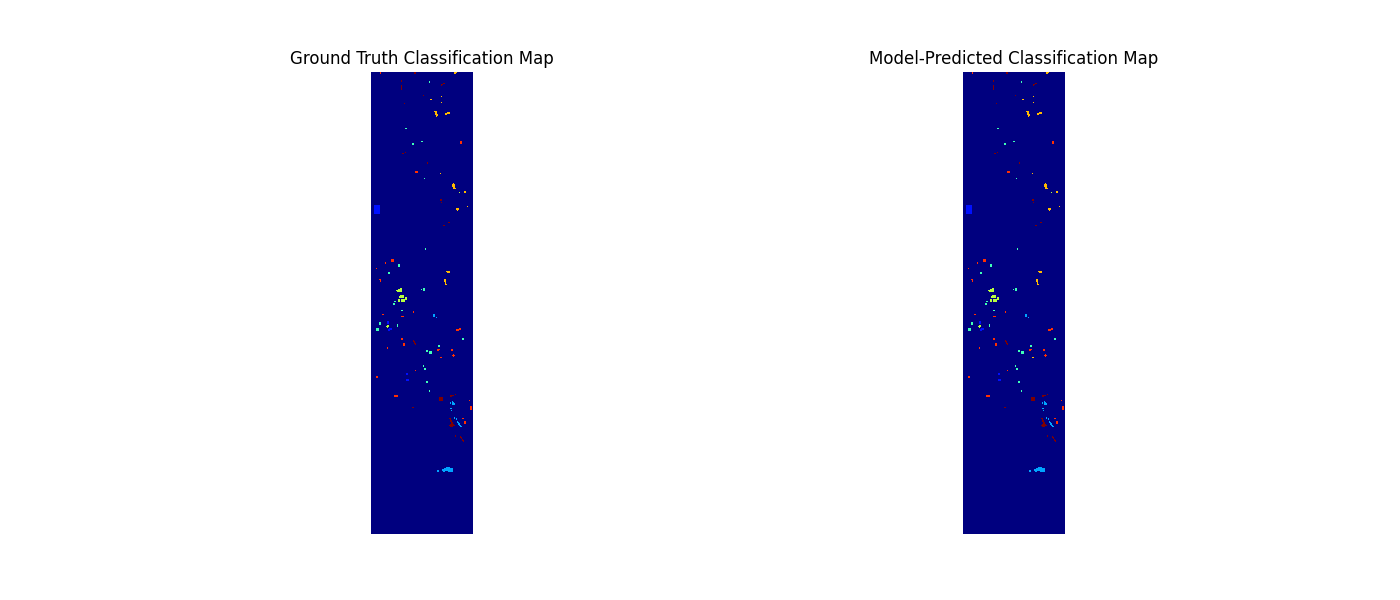}
        \caption{CA}
    \end{subfigure}
    \begin{subfigure}[b]{0.3\textwidth}
        \includegraphics[width=0.18\linewidth, trim=700 50 250 60, clip, angle=90]{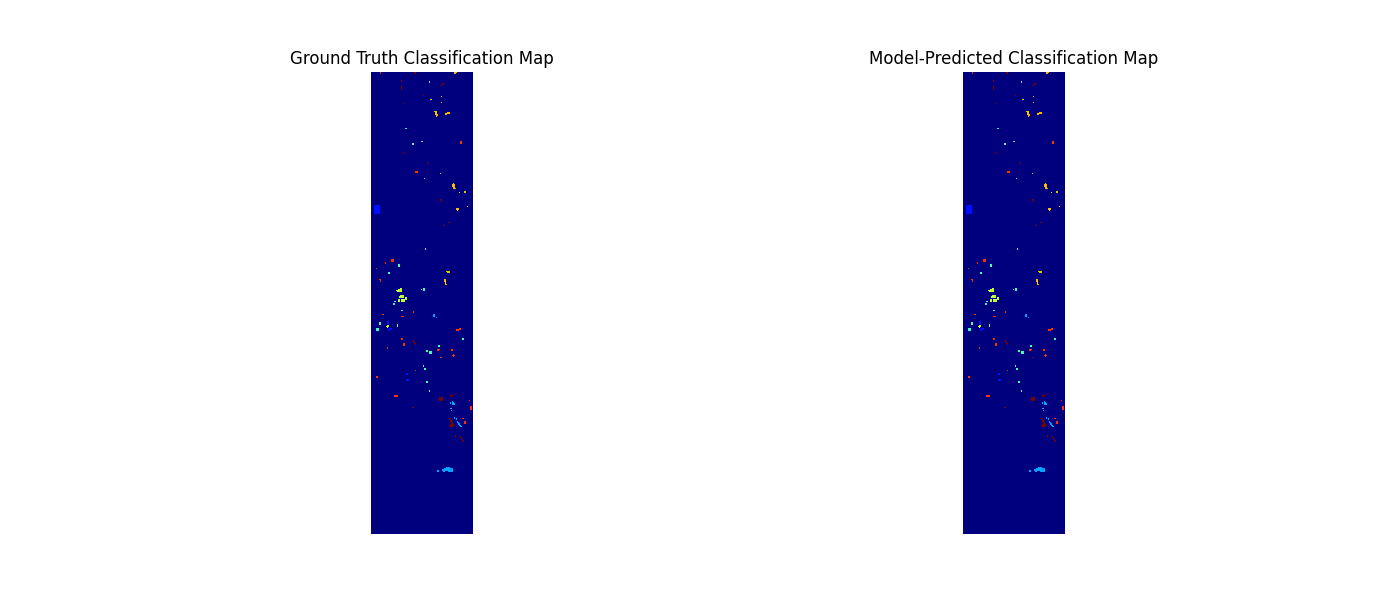}
        \caption{CONCAT}
    \end{subfigure}
    \begin{subfigure}[b]{0.3\textwidth}
        \includegraphics[width=0.18\linewidth, trim=700 50 250 60, clip, angle=90]{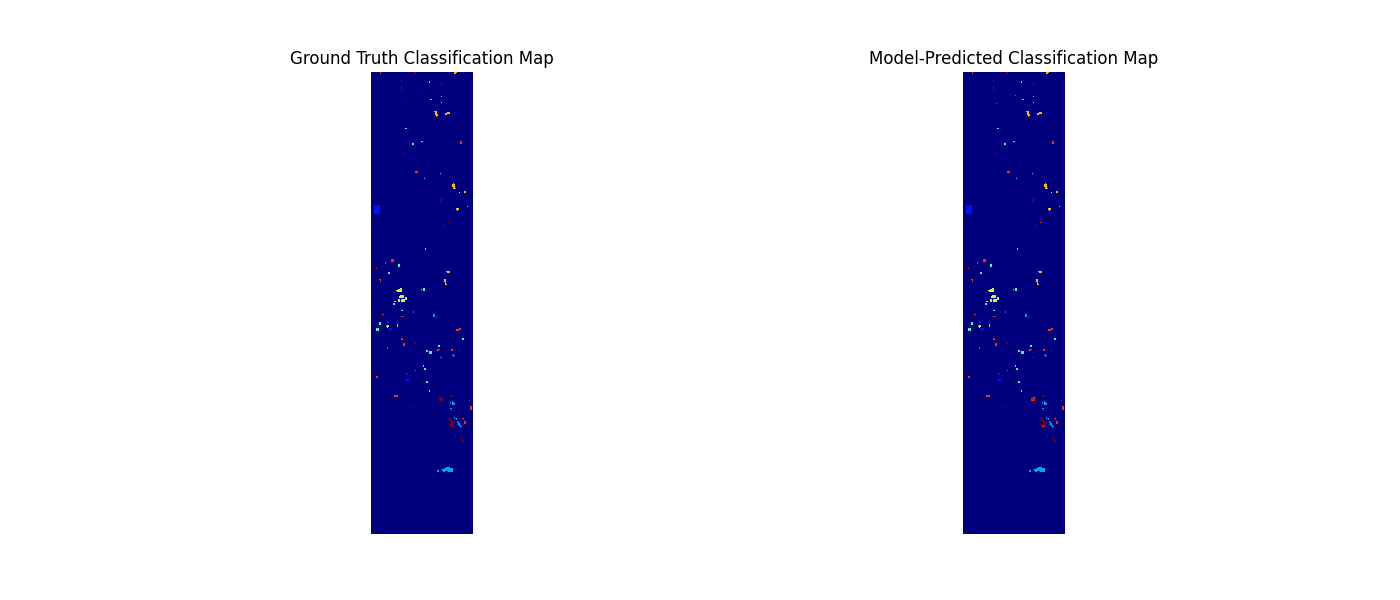}
        \caption{MHA}
    \end{subfigure}
    \begin{subfigure}[b]{0.3\textwidth}
        \includegraphics[width=0.18\linewidth, trim=700 50 250 60, clip, angle=90]{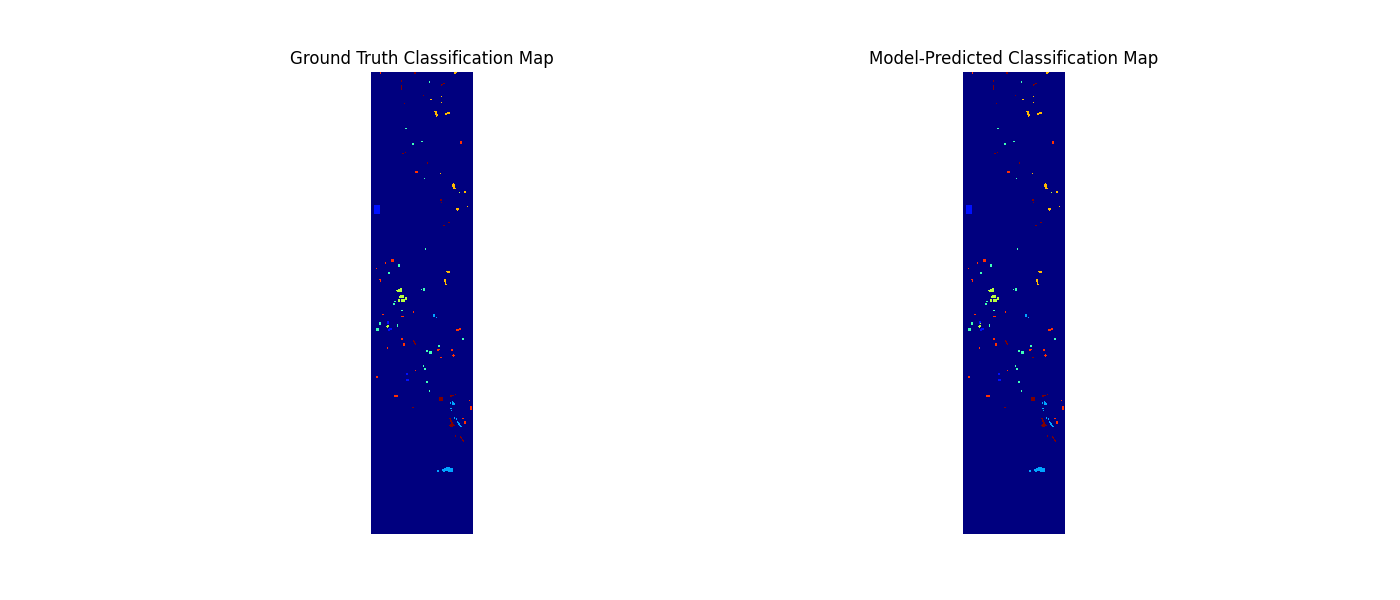}
        \caption{PWA}
    \end{subfigure}
    \begin{subfigure}[b]{0.3\textwidth}
        \includegraphics[width=0.18\linewidth, trim=700 50 250 60, clip, angle=90]{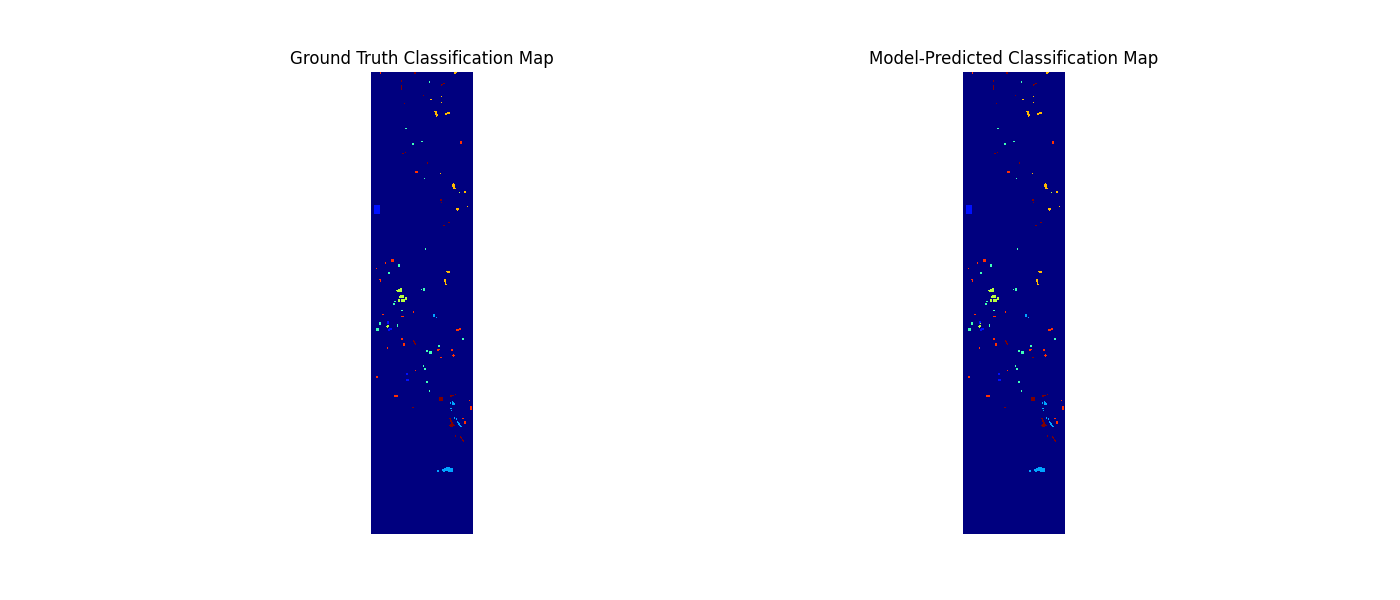}
        \caption{PWM}
    \end{subfigure}
        \begin{subfigure}[b]{0.3\textwidth}
         \includegraphics[width=0.18\linewidth, trim=275 50 675 60, clip, angle=90]{Output_Map/Houston13/3D_ConvSST-T5Encoder_Large/PWM_prediction_map_run1.png}
        \caption{GT}
    \end{subfigure}
    \caption{Comparison of classification maps for the 3D-ConvSST-T5 model on the Houston13 dataset, showing different fusion methods: Cross Attention (CA), Concatenation (CONCAT), Multi-Head Attention (MHA), Pixel-Wise Addition (PWA), Pixel-Wise Multiplication (PWM), and Ground Truth (GT).}
    \vspace{-3mm}
    \label{fig:3D-ConvSST-T5-Houston13}
\end{figure*}

%%%%%%%%%%%%%%%%%%%%%%%%%%%%%%%%%%%%%%%%%%%%% Houston13     3D_ConvSST-BertEncoder%%%%%%%%%%%%%%%%%%%%%%%%%%%%%%%%%%%%%%%%%%%%%%%%
\begin{figure*}[]
    \centering
    \begin{subfigure}[b]{0.3\textwidth}
         \includegraphics[width=0.18\linewidth, trim=700 50 250 60, clip, angle=90]{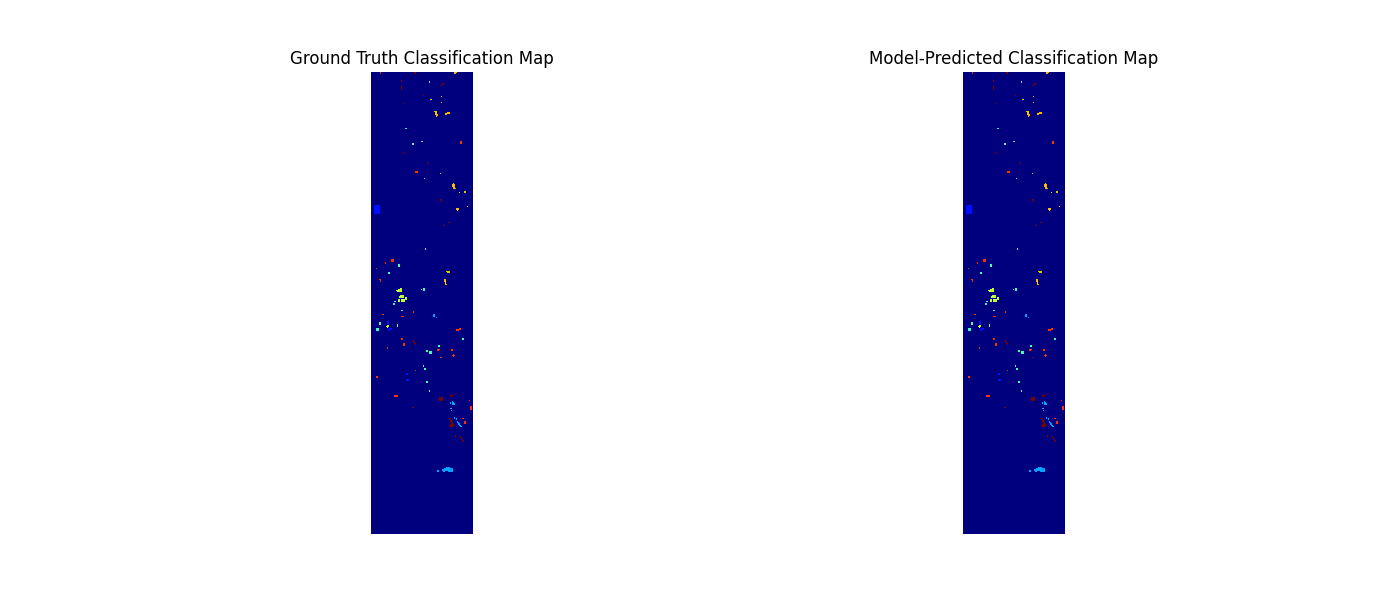}
        \caption{CA}
    \end{subfigure}
    \begin{subfigure}[b]{0.3\textwidth}
        \includegraphics[width=0.18\linewidth, trim=700 50 250 60, clip, angle=90]{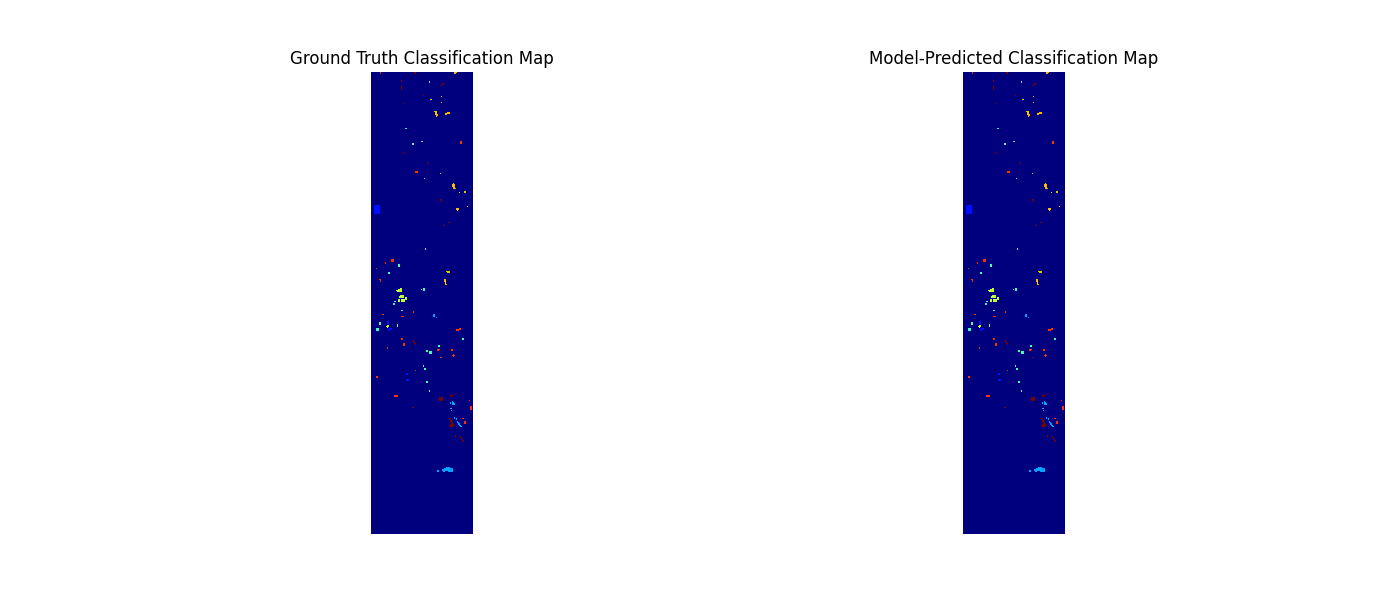}
        \caption{CONCAT}
    \end{subfigure}
    \begin{subfigure}[b]{0.3\textwidth}
        \includegraphics[width=0.18\linewidth, trim=700 50 250 60, clip, angle=90]{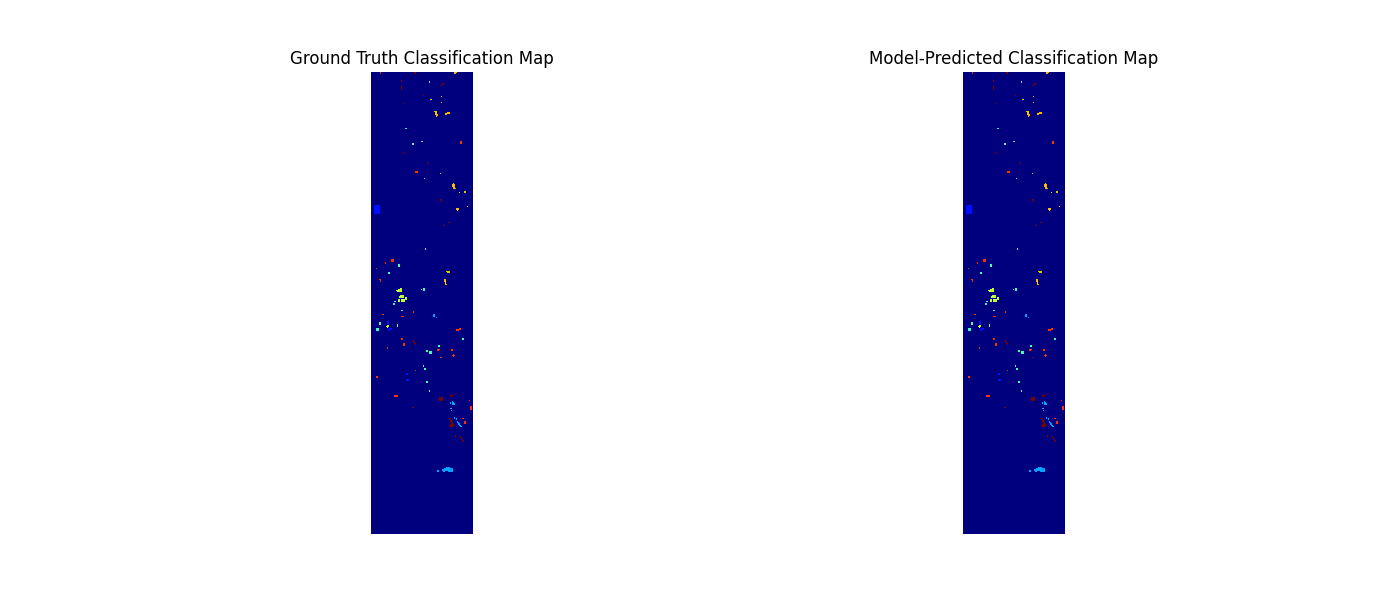}
        \caption{MHA}
    \end{subfigure}
    \begin{subfigure}[b]{0.3\textwidth}
        \includegraphics[width=0.18\linewidth, trim=700 50 250 60, clip, angle=90]{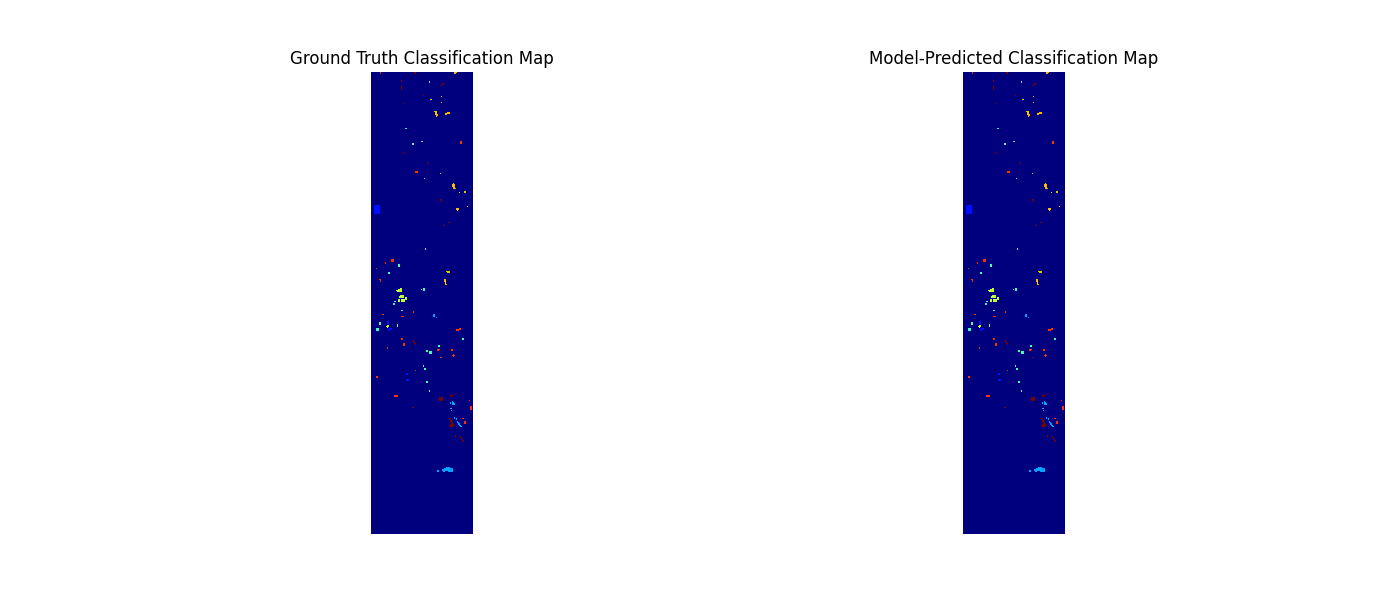}
        \caption{PWA}
    \end{subfigure}
    \begin{subfigure}[b]{0.3\textwidth}
        \includegraphics[width=0.18\linewidth, trim=700 50 250 60, clip, angle=90]{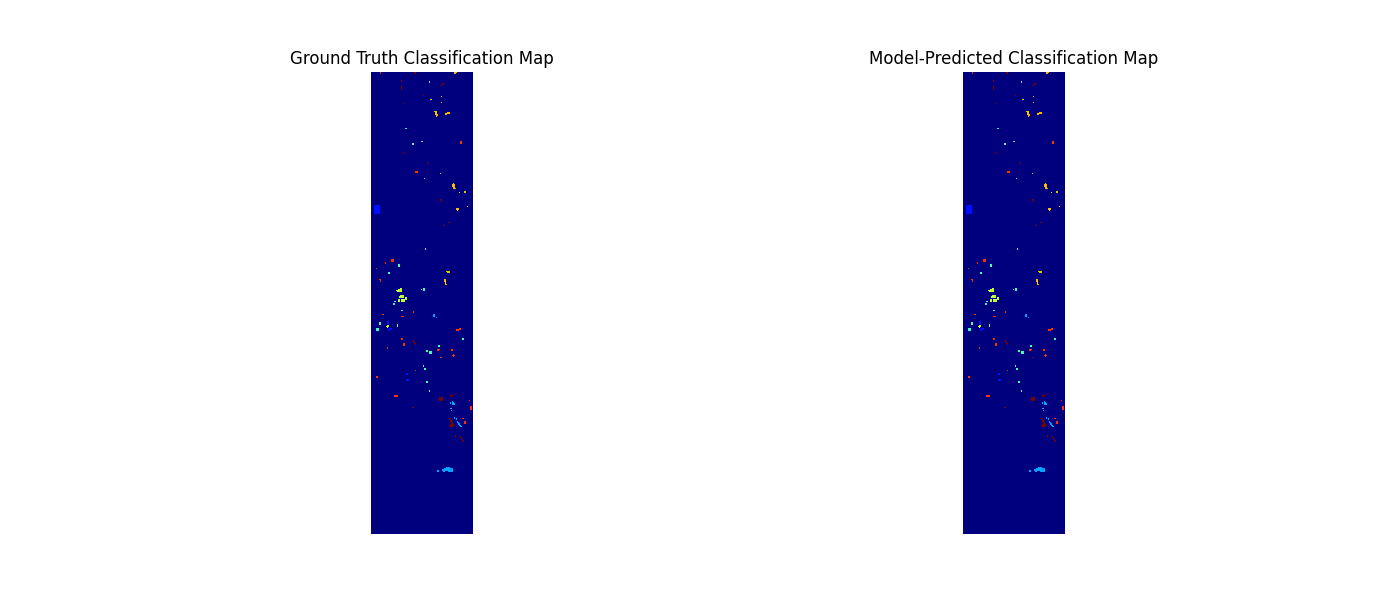}
        \caption{PWM}
    \end{subfigure}
        \begin{subfigure}[b]{0.3\textwidth}
         \includegraphics[width=0.18\linewidth, trim=275 50 675 60, clip, angle=90]{Output_Map/Houston13/3D_ConvSST-BertEncoder_Large/PWM_prediction_map_run1.png}
        \caption{GT}
    \end{subfigure}
    \caption{Comparison of classification maps for the 3D-ConvSST-BERT model on the Houston13 dataset, showing different fusion methods: Cross Attention (CA), Concatenation (CONCAT), Multi-Head Attention (MHA), Pixel-Wise Addition (PWA), Pixel-Wise Multiplication (PWM), and Ground Truth (GT).}
    \vspace{-3mm}
    \label{fig:3D-ConvSST-Bert-Houston13}
\end{figure*}

%%%%%%%%%%%%%%%%%%%%%%%%%%% Houston13     DBCTNet-T5Encoder%%%%%%%%%%%%%%%%%%%%%%%%%%%%%%%%%%%%%%%%%%%%%%%%
\begin{figure*}[]
    \centering
    \begin{subfigure}[b]{0.3\textwidth}
         \includegraphics[width=0.18\linewidth, trim=700 50 250 60, clip, angle=90]{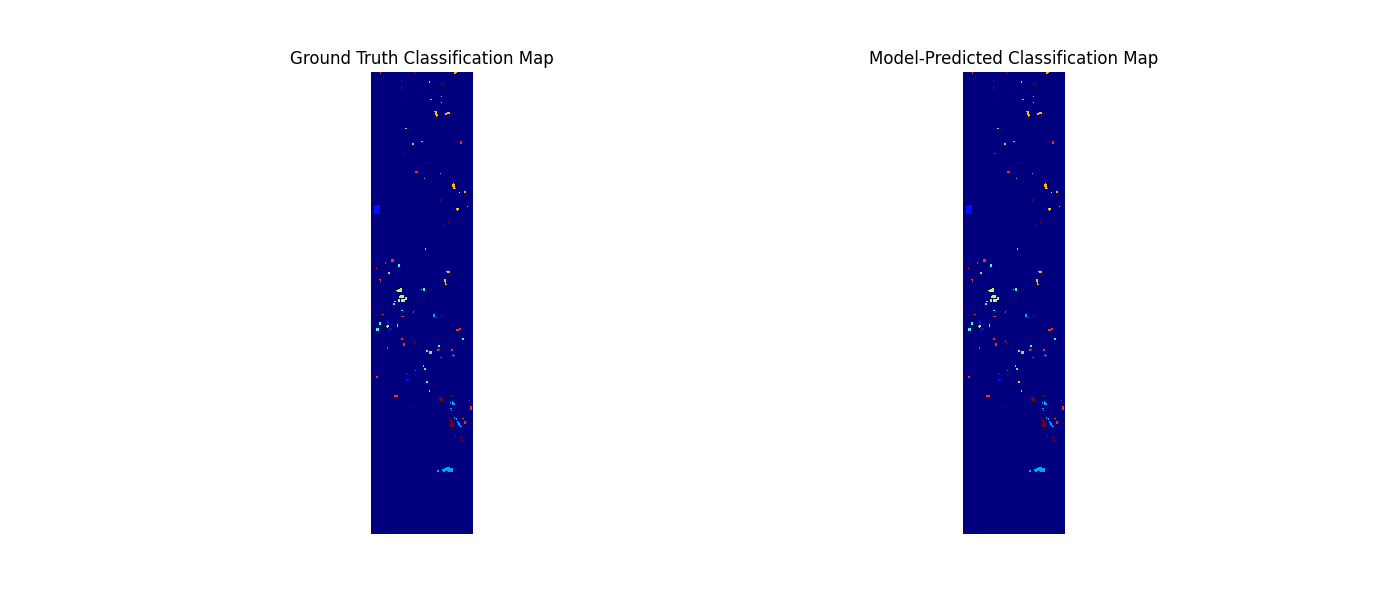}
        \caption{CA}
    \end{subfigure}
    \begin{subfigure}[b]{0.3\textwidth}
        \includegraphics[width=0.18\linewidth, trim=700 50 250 60, clip, angle=90]{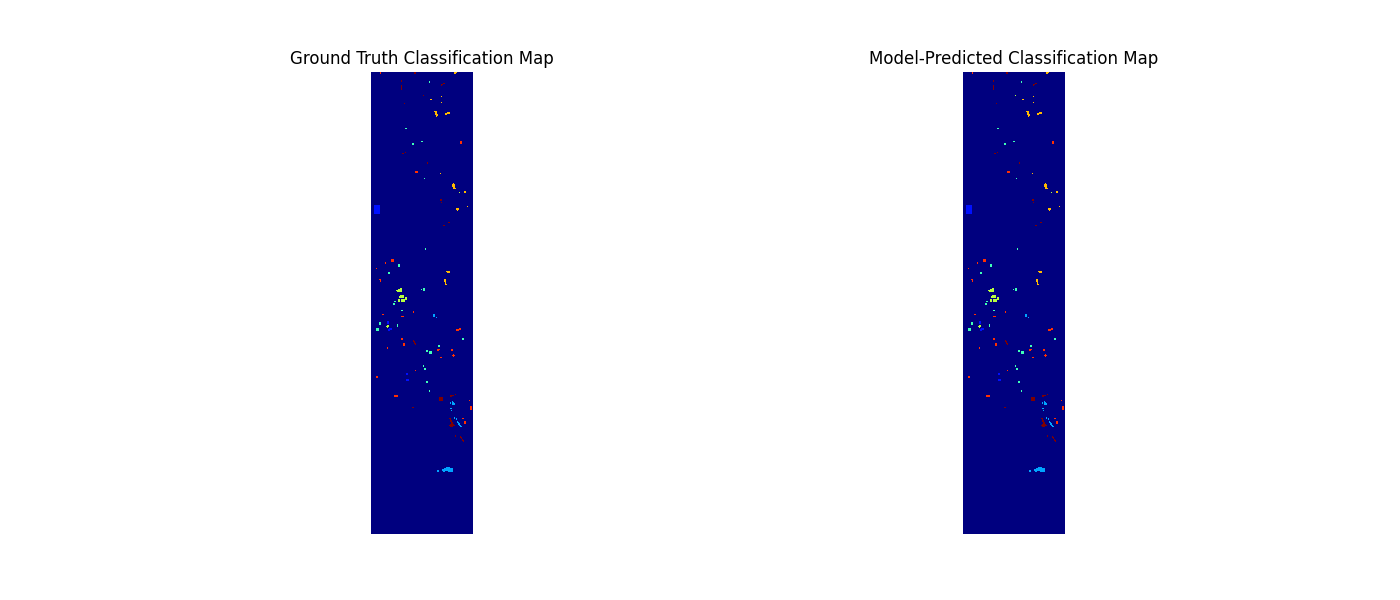}
        \caption{CONCAT}
    \end{subfigure}
    \begin{subfigure}[b]{0.3\textwidth}
        \includegraphics[width=0.18\linewidth, trim=700 50 250 60, clip, angle=90]{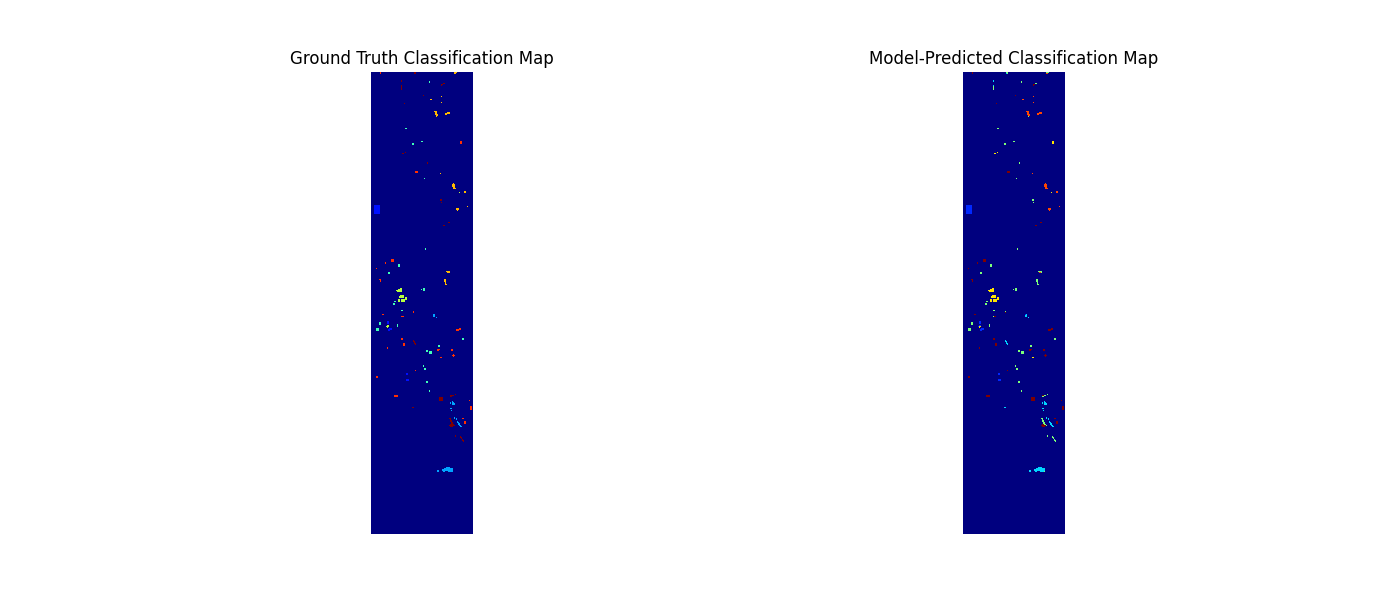}
        \caption{MHA}
    \end{subfigure}
    \begin{subfigure}[b]{0.3\textwidth}
        \includegraphics[width=0.18\linewidth, trim=700 50 250 60, clip, angle=90]{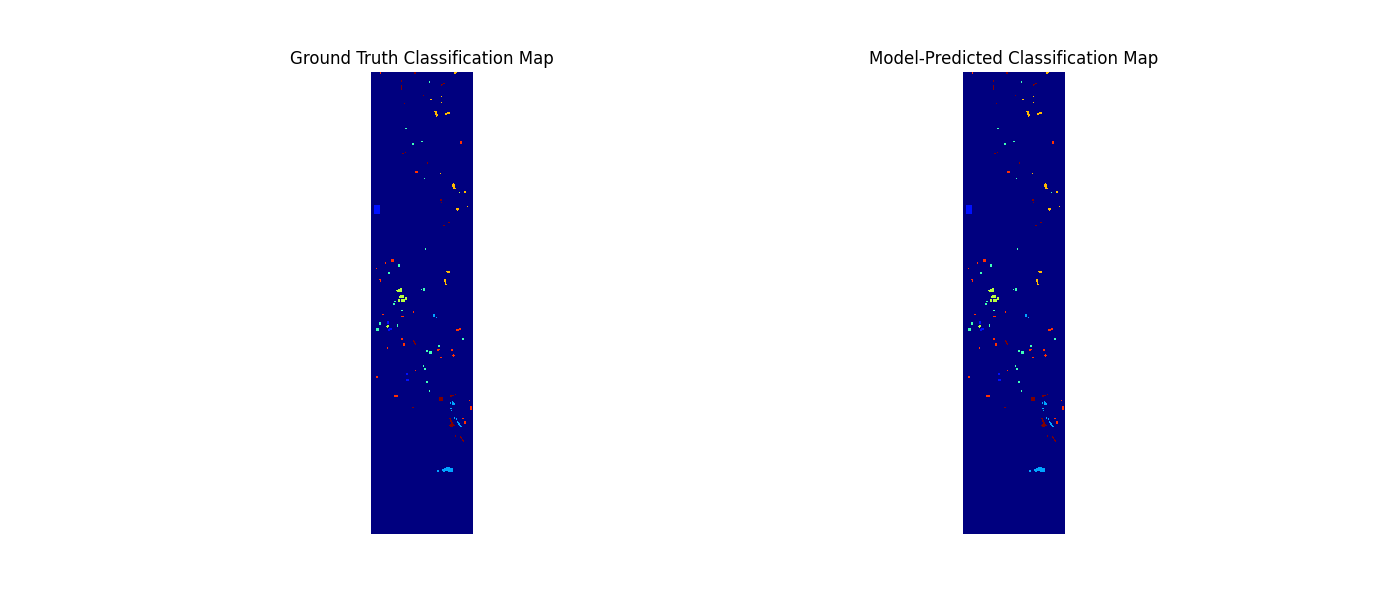}
        \caption{PWA}
    \end{subfigure}
    \begin{subfigure}[b]{0.3\textwidth}
        \includegraphics[width=0.18\linewidth, trim=700 50 250 60, clip, angle=90]{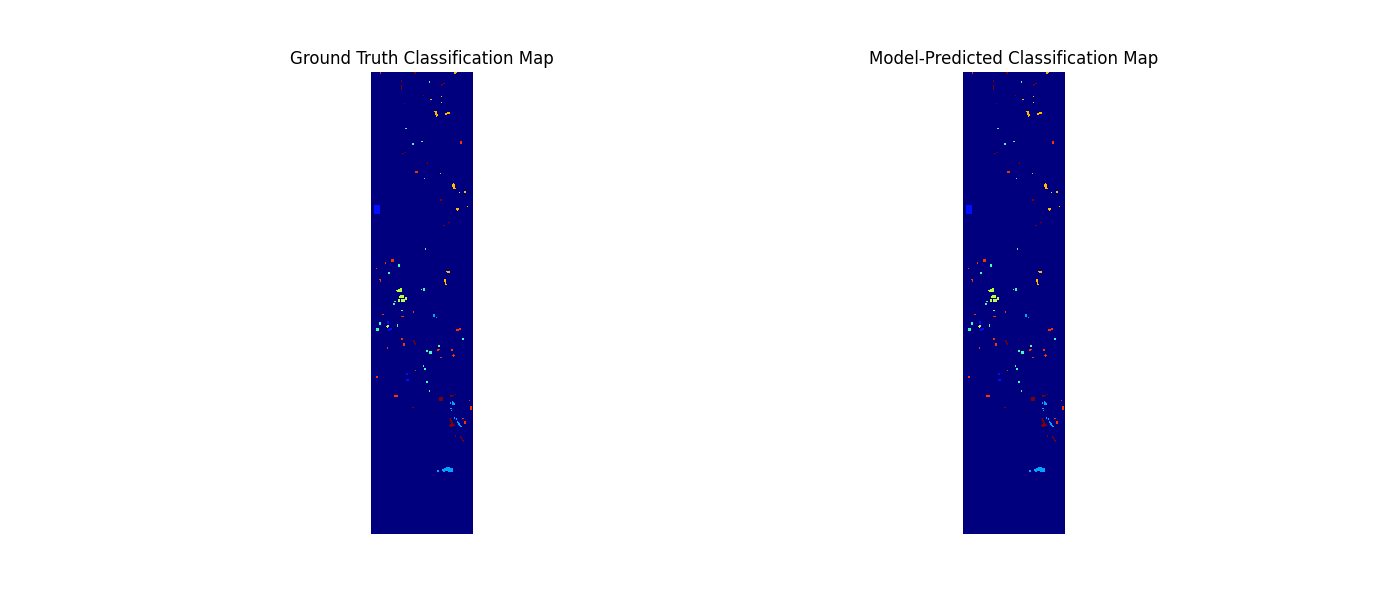}
        \caption{PWM}
    \end{subfigure}
    \begin{subfigure}[b]{0.3\textwidth}
         \includegraphics[width=0.18\linewidth, trim=275 50 675 60, clip, angle=90]{Output_Map/Houston13/DBCTNet-T5Encoder_Large/PWM_prediction_map_run1.png}
        \caption{GT}
    \end{subfigure}
    \caption{Comparison of classification maps for the DBCTNet-T5 model on the Houston13 dataset, showing different fusion methods: Cross Attention (CA), Concatenation (CONCAT), Multi-Head Attention (MHA), Pixel-Wise Addition (PWA), Pixel-Wise Multiplication (PWM), and Ground Truth (GT).}
    \vspace{-3mm}
    \label{fig:DBCTNet-T5-Houston13}
\end{figure*}

%%%%%%%%%%%%%%%%%%%%%%%%%%%%%%%%%%%%%%%%%%%%%%% Houston13     DBCNet-BertEncoder%%%%%%%%%%%%%%%%%%%%%%%%%%%%%%%%%%%%%%%%%%%%%%%%
\begin{figure*}[]
    \centering
    \begin{subfigure}[b]{0.3\textwidth}
         \includegraphics[width=0.18\linewidth, trim=700 50 250 60, clip, angle=90]{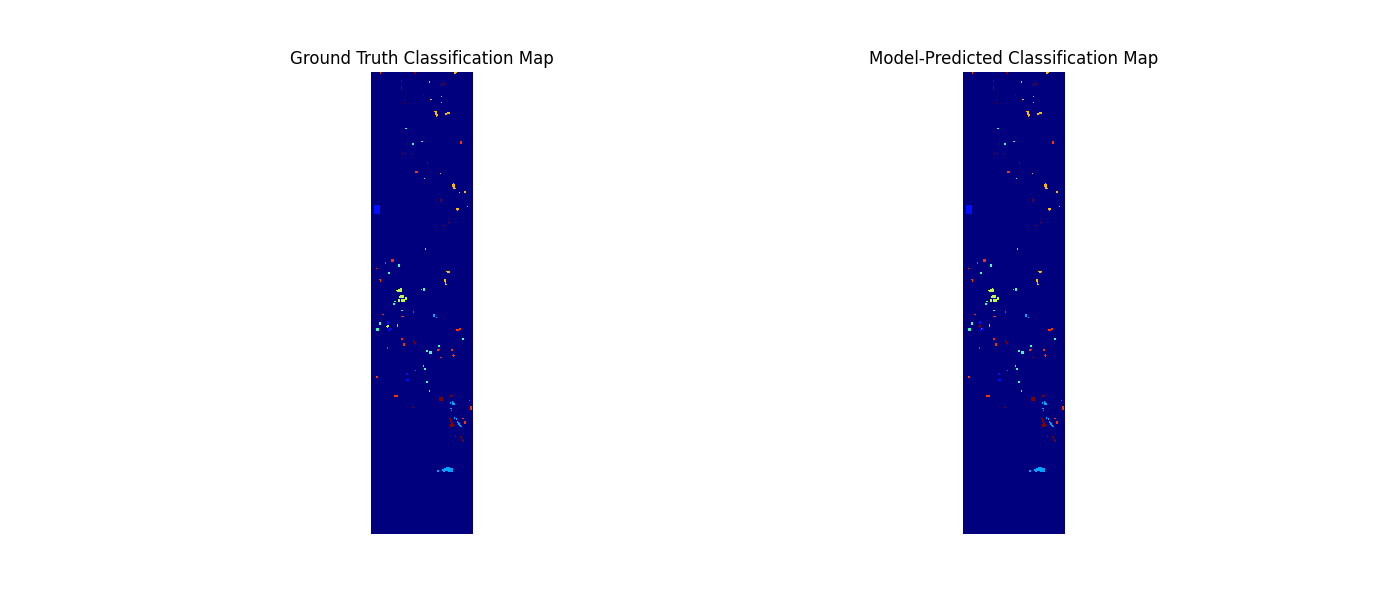}
        \caption{CA}
    \end{subfigure}
    \begin{subfigure}[b]{0.3\textwidth}
        \includegraphics[width=0.18\linewidth, trim=700 50 250 60, clip, angle=90]{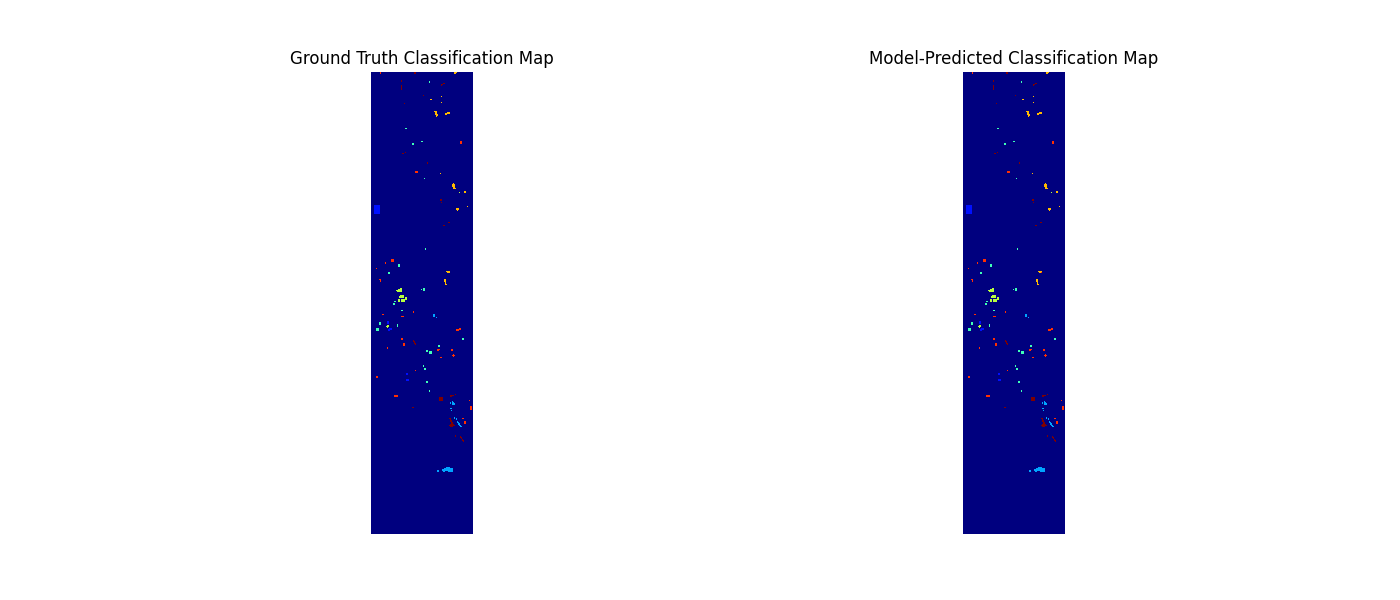}
        \caption{CONCAT}
    \end{subfigure}
    \begin{subfigure}[b]{0.3\textwidth}
        \includegraphics[width=0.18\linewidth, trim=700 50 250 60, clip, angle=90]{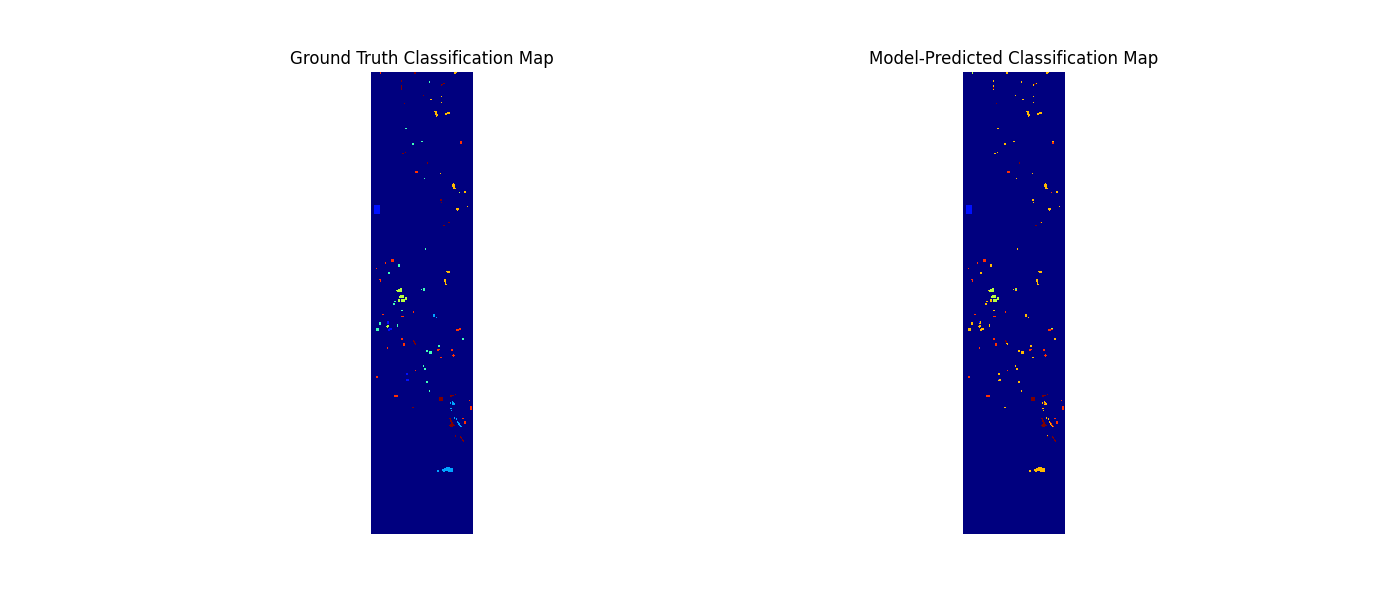}
        \caption{MHA}
    \end{subfigure}
    \begin{subfigure}[b]{0.3\textwidth}
        \includegraphics[width=0.18\linewidth, trim=700 50 250 60, clip, angle=90]{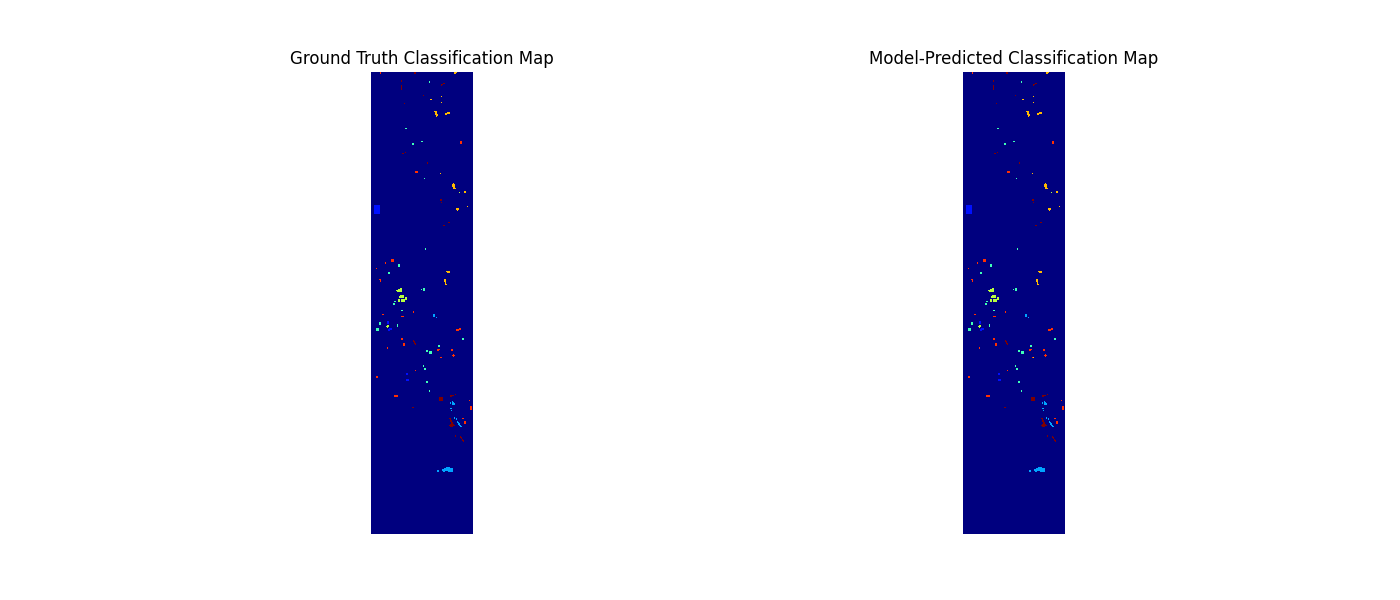}
        \caption{PWA}
    \end{subfigure}
    \begin{subfigure}[b]{0.3\textwidth}
        \includegraphics[width=0.18\linewidth, trim=700 50 250 60, clip, angle=90]{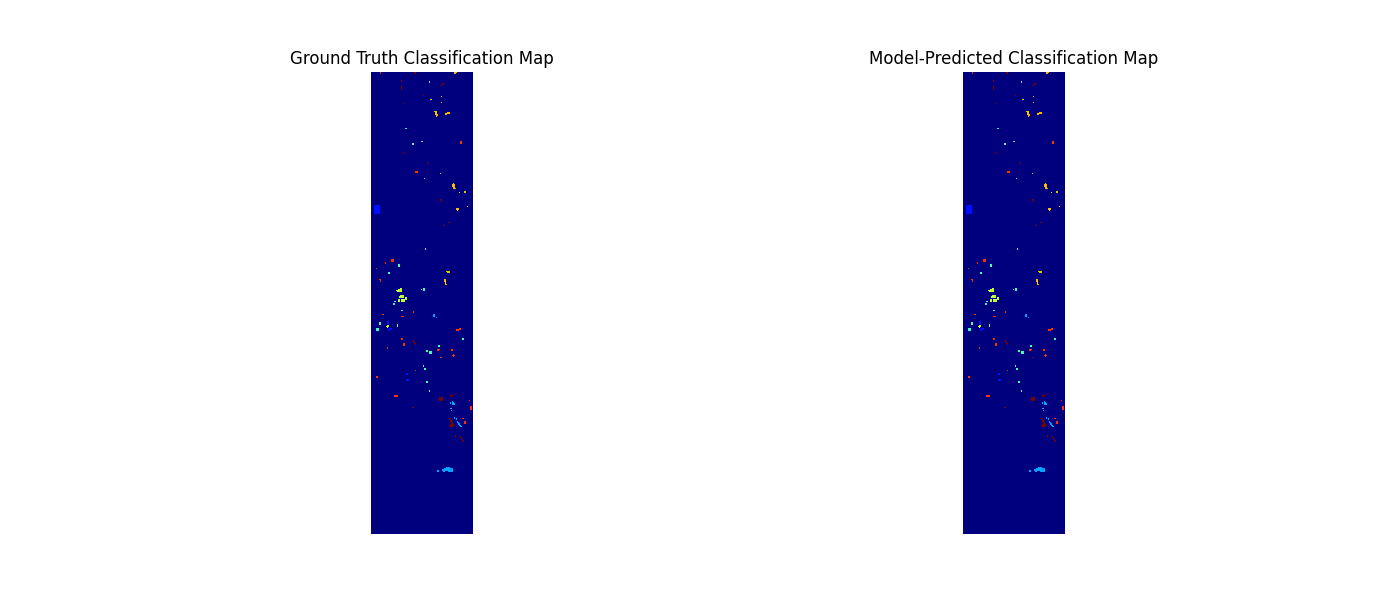}
        \caption{PWM}
    \end{subfigure}
        \begin{subfigure}[b]{0.3\textwidth}
         \includegraphics[width=0.18\linewidth, trim=275 50 675 60, clip, angle=90]{Output_Map/Houston13/DBCTNet-BertEncoder_Large/PWM_prediction_map_run1.png}
        \caption{GT}
    \end{subfigure}
    \caption{Comparison of classification maps for the DBCTNet-BERT model on the Houston13 dataset, showing different fusion methods: Cross Attention (CA), Concatenation (CONCAT), Multi-Head Attention (MHA), Pixel-Wise Addition (PWA), Pixel-Wise Multiplication (PWM), and Ground Truth (GT).}
    \vspace{-3mm}
    \label{fig:DBCTNet-Bert-Houston13}
\end{figure*}

%%%%%% Houston13
\begin{figure*}[!t]
    \centering
    \begin{subfigure}[b]{0.3\textwidth}
         \includegraphics[width=0.18\linewidth, trim=700 50 250 60, clip, angle=90]{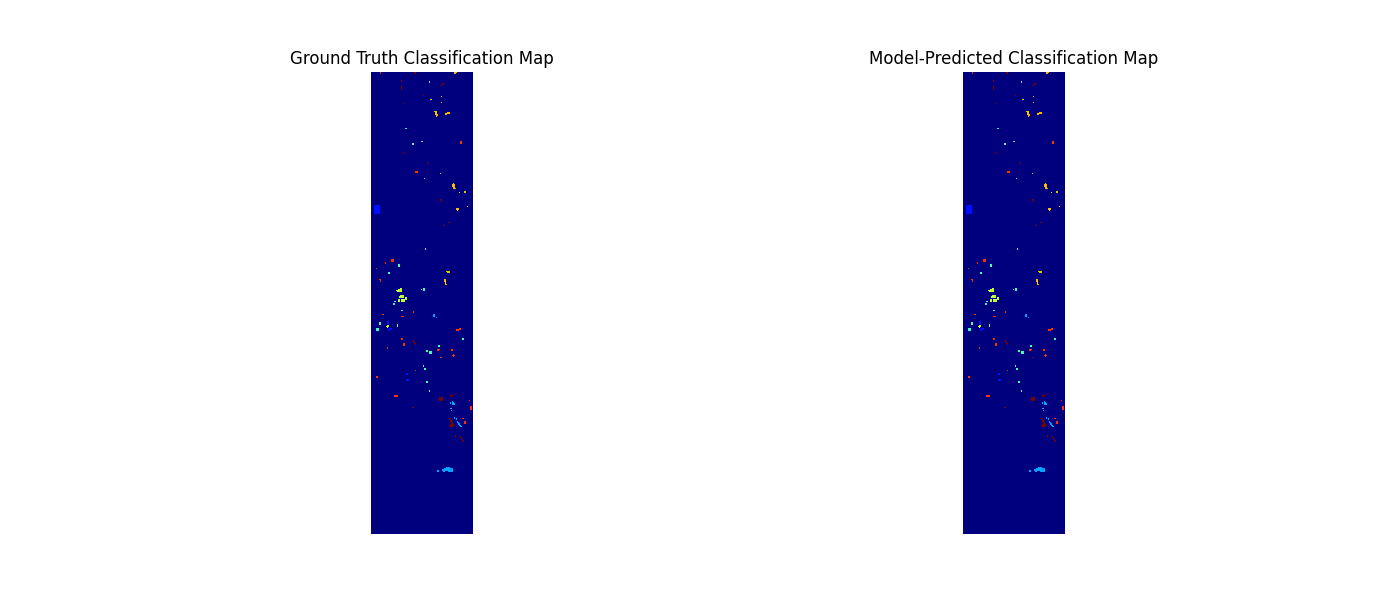}
        \caption{CA}
    \end{subfigure}
    \begin{subfigure}[b]{0.3\textwidth}
        \includegraphics[width=0.18\linewidth, trim=700 50 250 60, clip, angle=90]{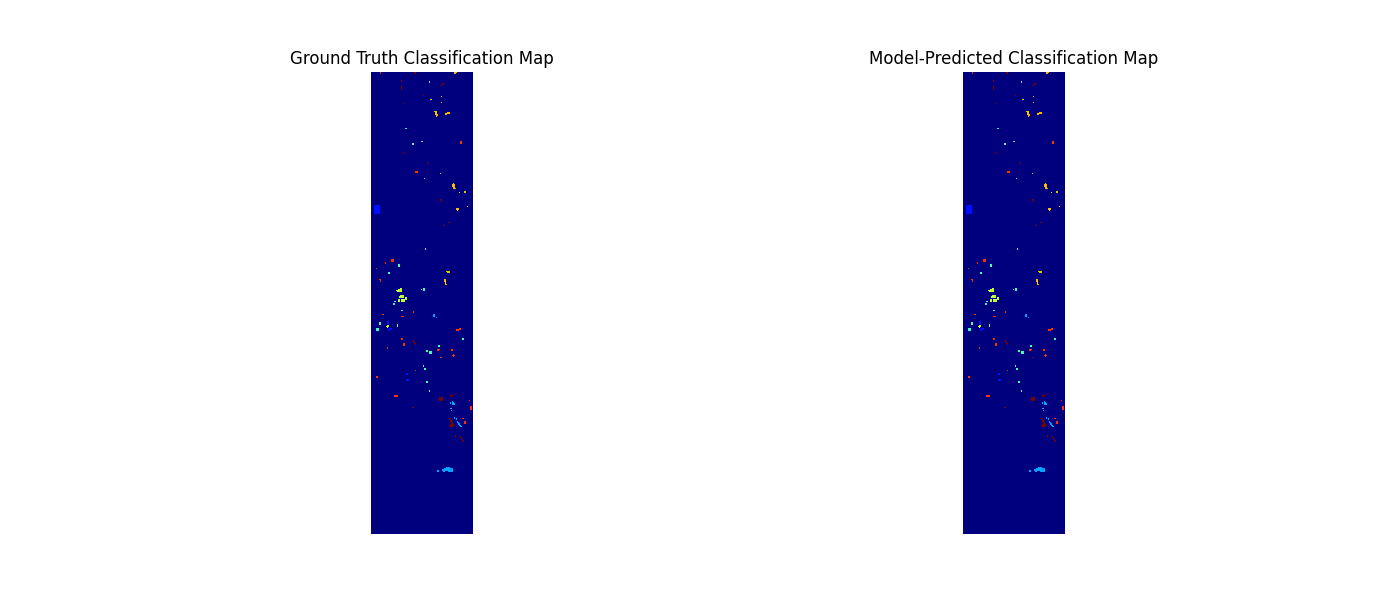}
        \caption{CONCAT}
    \end{subfigure}
    \begin{subfigure}[b]{0.3\textwidth}
        \includegraphics[width=0.18\linewidth, trim=700 50 250 60, clip, angle=90]{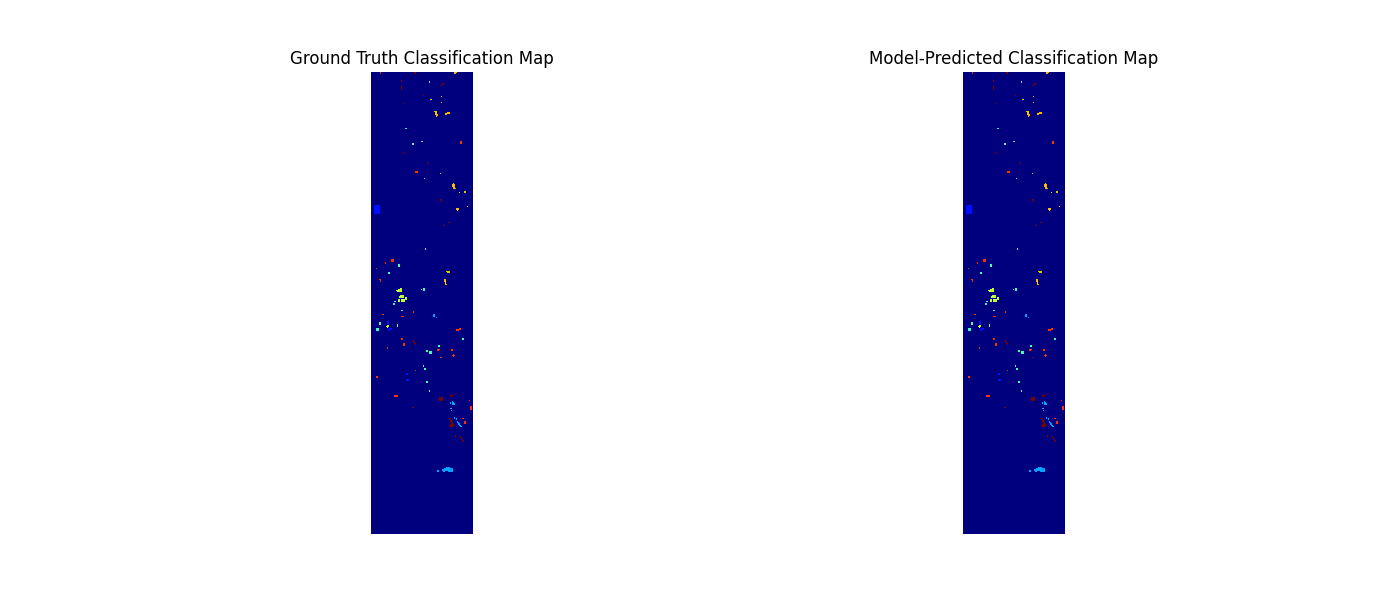}
        \caption{MHA}
    \end{subfigure}
    \begin{subfigure}[b]{0.3\textwidth}
        \includegraphics[width=0.18\linewidth, trim=700 50 250 60, clip, angle=90]{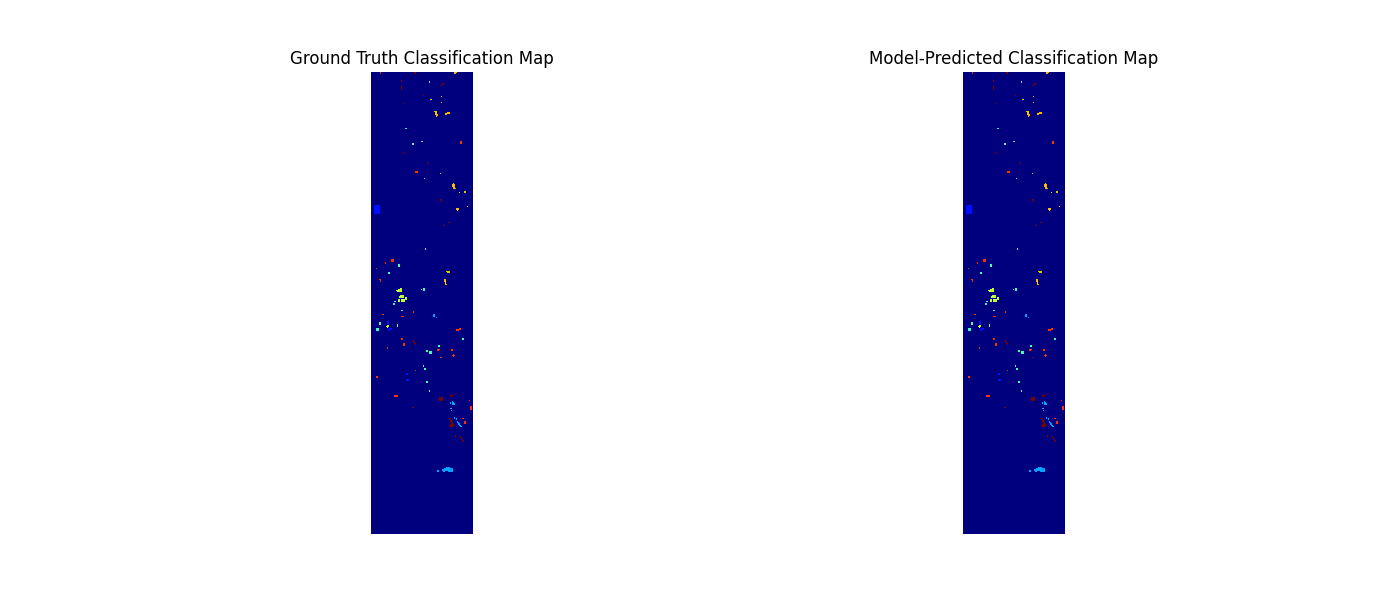}
        \caption{PWA}
    \end{subfigure}
    \begin{subfigure}[b]{0.3\textwidth}
        \includegraphics[width=0.18\linewidth, trim=700 50 250 60, clip, angle=90]{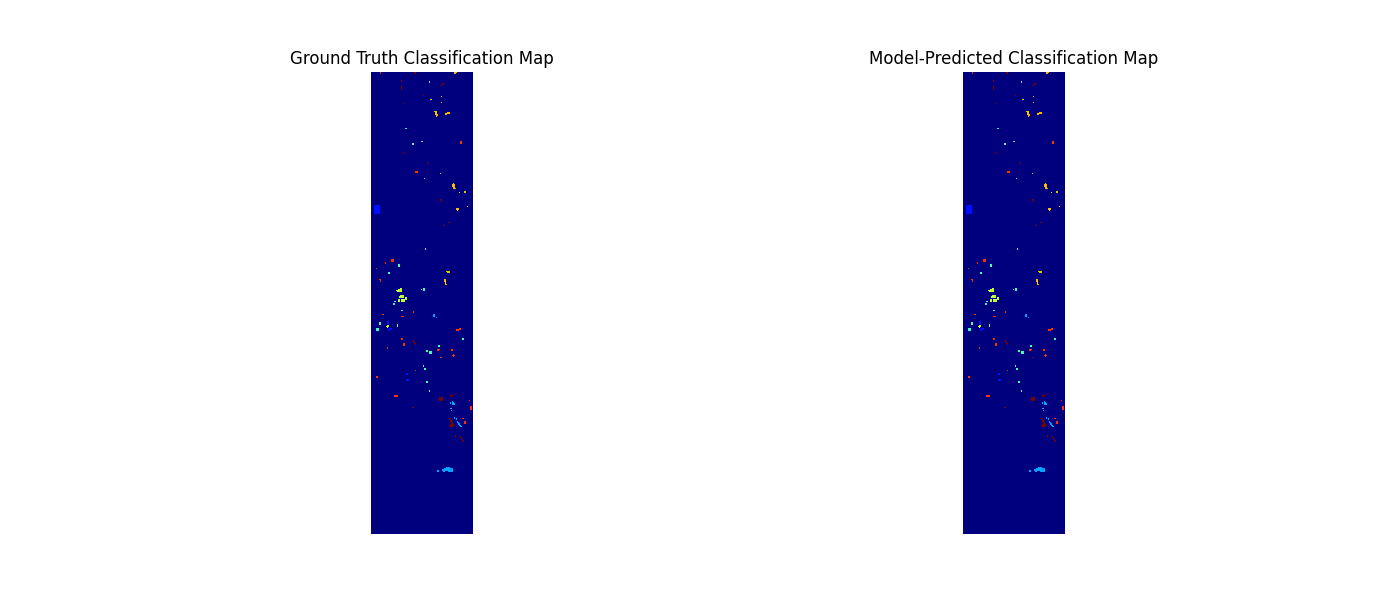}
        \caption{PWM}
    \end{subfigure}
        \begin{subfigure}[b]{0.3\textwidth}
         \includegraphics[width=0.18\linewidth, trim=275 50 675 60, clip, angle=90]{Output_Map/Houston13/FAHM-T5Encoder_Large/PWM_prediction_map_run1.png}
        \caption{GT}
    \end{subfigure}
    \caption{Comparison of classification maps for the FAHM-T5 model on the Houston13 dataset, showing different fusion methods: Cross Attention (CA), Concatenation (CONCAT), Multi-Head Attention (MHA), Pixel-Wise Addition (PWA), Pixel-Wise Multiplication (PWM), and Ground Truth (GT).}
    \vspace{-4mm}
    \label{fig:FAHM-T5-Houston13}
\end{figure*}

%%%%%%%%%%%%%%%%%%%%%%%%%%%%%% Houston13     FAHM-BertEncoder%%%%%%%%%%%%%%%%%%%%%%%%%%%%%%%%%%%%%%%%%%%%%%%%
\begin{figure*}[]
    \centering
    \begin{subfigure}[b]{0.3\textwidth}
         \includegraphics[width=0.18\linewidth, trim=700 50 250 60, clip, angle=90]{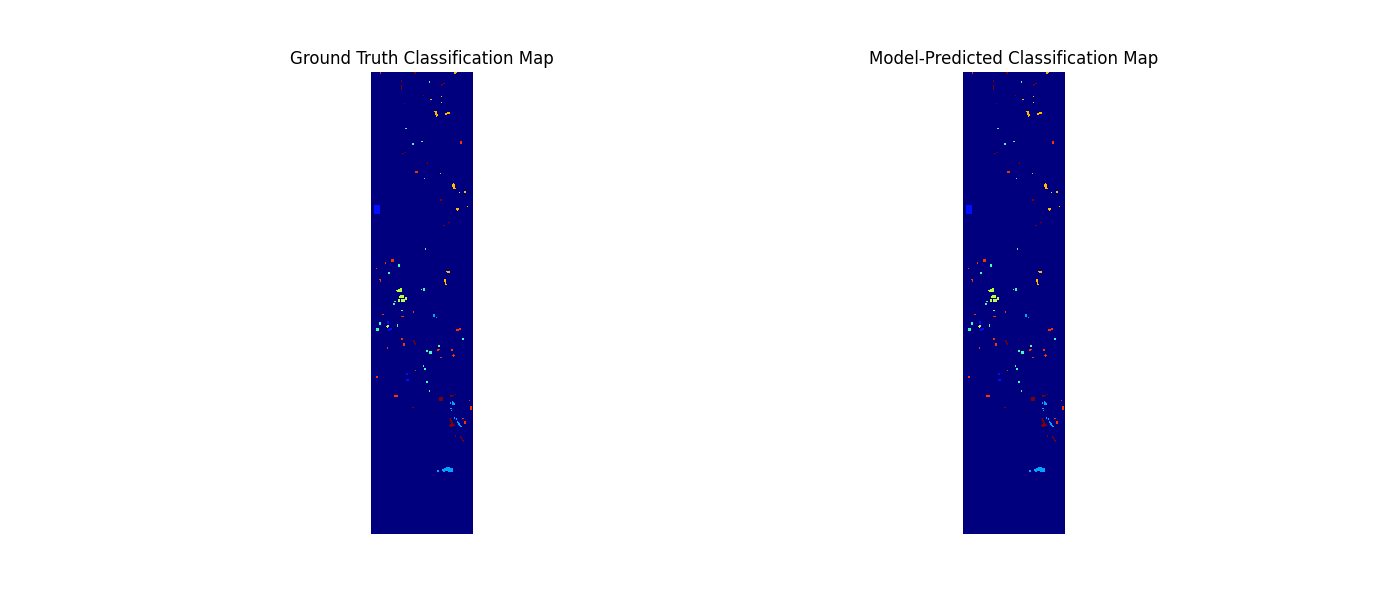}
        \caption{CA}
    \end{subfigure}
    \begin{subfigure}[b]{0.3\textwidth}
        \includegraphics[width=0.18\linewidth, trim=700 50 250 60, clip, angle=90]{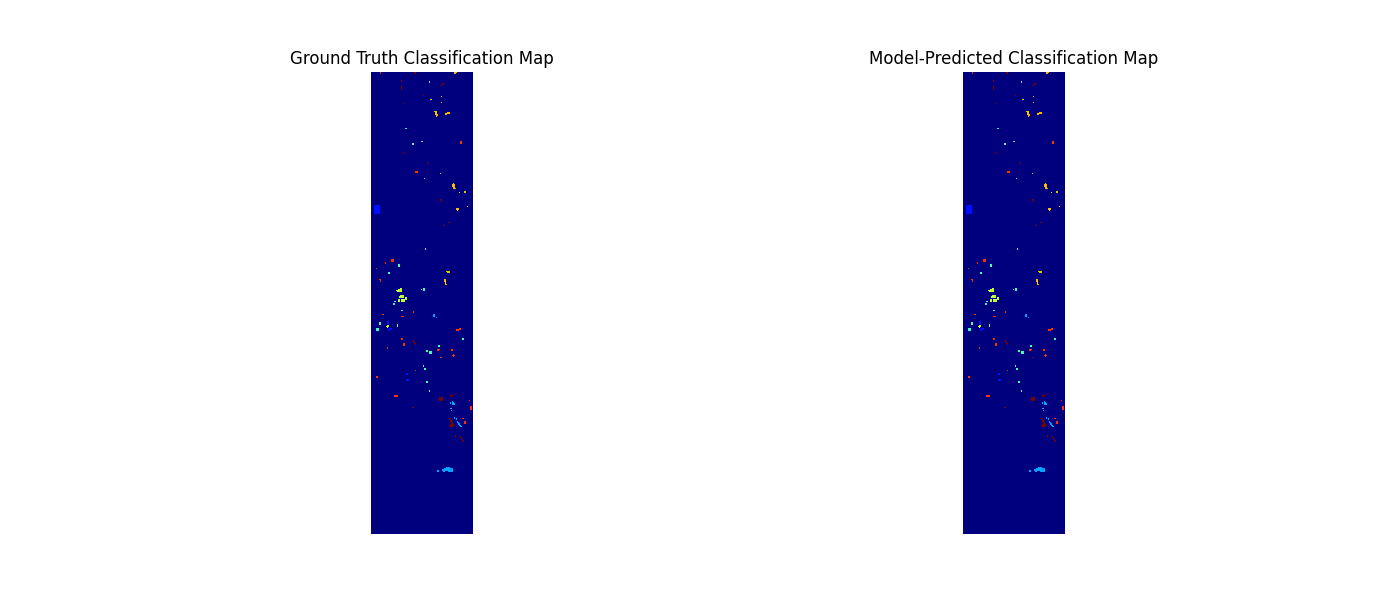}
        \caption{CONCAT}
    \end{subfigure}
    \begin{subfigure}[b]{0.3\textwidth}
        \includegraphics[width=0.18\linewidth, trim=700 50 250 60, clip, angle=90]{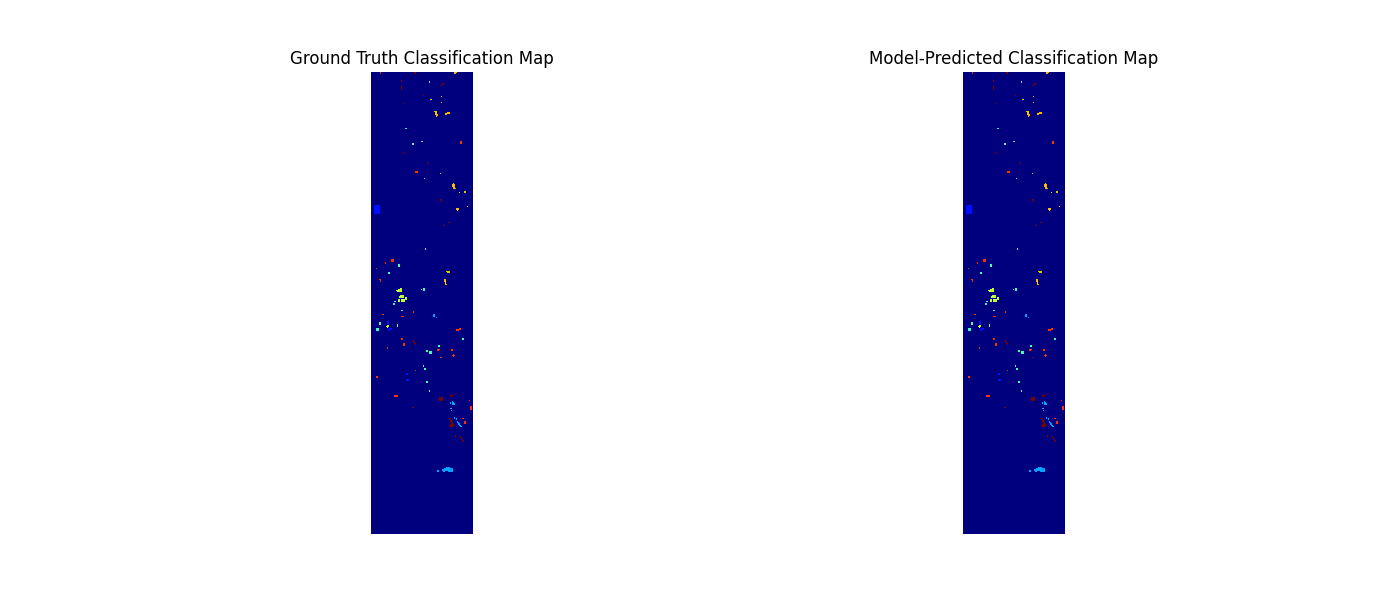}
        \caption{MHA}
    \end{subfigure}
    \begin{subfigure}[b]{0.3\textwidth}
        \includegraphics[width=0.18\linewidth, trim=700 50 250 60, clip, angle=90]{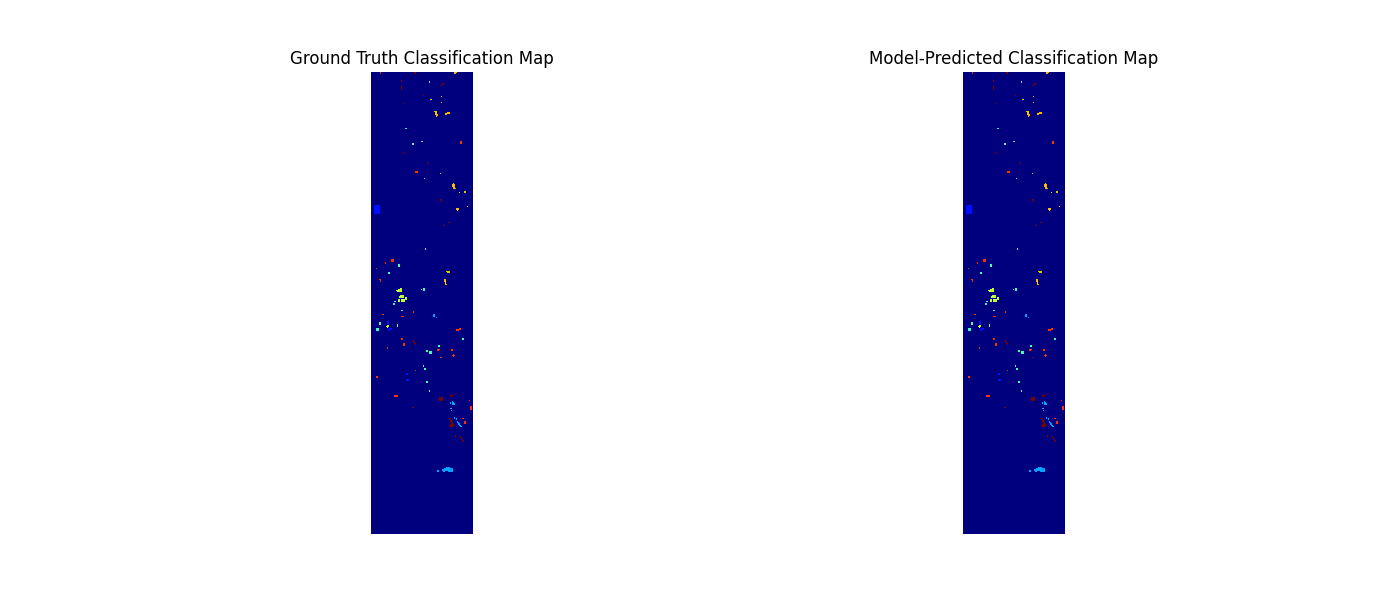}
        \caption{PWA}
    \end{subfigure}
    \begin{subfigure}[b]{0.3\textwidth}
        \includegraphics[width=0.18\linewidth, trim=700 50 250 60, clip, angle=90]{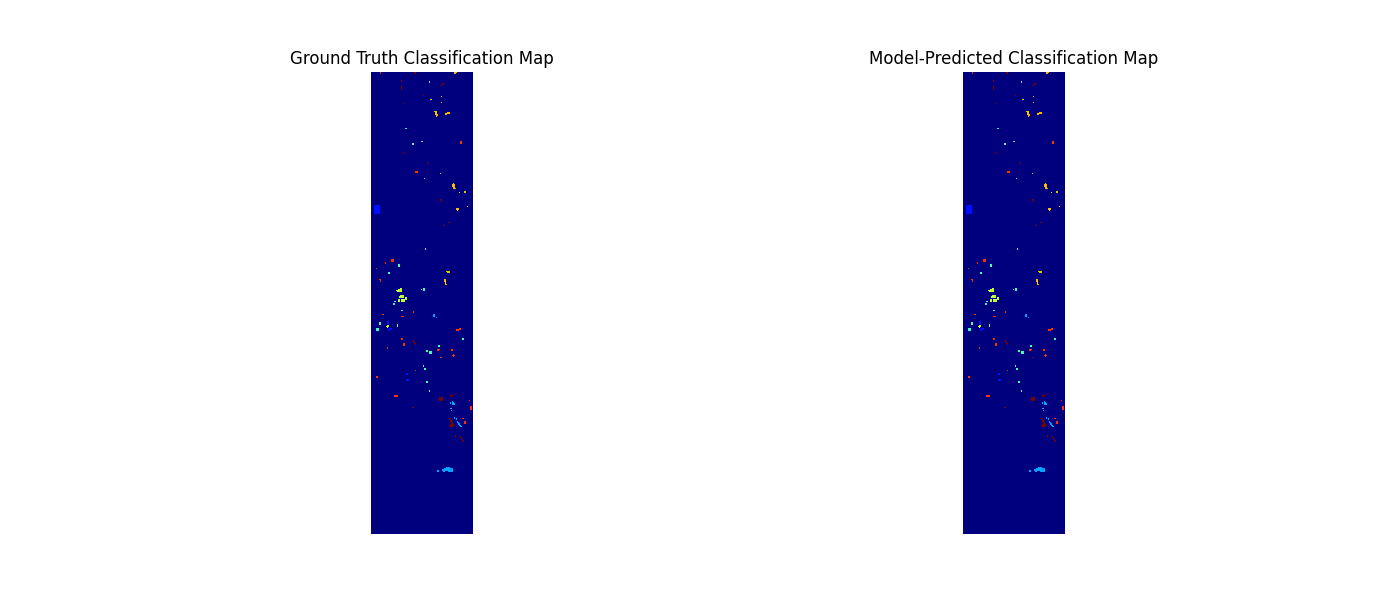}
        \caption{PWM}
    \end{subfigure}
        \begin{subfigure}[b]{0.3\textwidth}
         \includegraphics[width=0.18\linewidth, trim=275 50 675 60, clip, angle=90]{Output_Map/Houston13/FAHM-BertEncoder_Large/PWM_prediction_map_run1.png}
        \caption{GT}
    \end{subfigure}
    \caption{Comparison of classification maps for the FAHM-Bert model on the Houston13 dataset, showing different fusion methods: Cross Attention (CA), Concatenation (CONCAT), Multi-Head Attention (MHA), Pixel-Wise Addition (PWA), Pixel-Wise Multiplication (PWM), and Ground Truth (GT).}
    \vspace{-3mm}
    \label{fig:FAHM-Bert-Houston13}
\end{figure*}

The Houston13 visual results show noticeably crisper boundaries for buildings, pavements, and urban surfaces after incorporating language encoders. The original vision-only predictions have blurred, poorly separated regions, but after text-enhanced fusion, the segmentation maps show contiguous rooftops, clearer class delineation, and reduced confusion between vegetation and artificial materials. Fusion mechanisms like multi-head attention and concatenation produce the most visually appealing, ground-truth-like results, eliminating most misclassified speckles and elevating performance across all backbones. Figures (\ref{fig:3D-RCNet-T5-Houston13}, \ref{fig:3D-RCNet-Bert-Houston13}, \ref{fig:3D-ConvSST-T5-Houston13}, \ref{fig:3D-ConvSST-Bert-Houston13}, \ref{fig:DBCTNet-T5-Houston13}, \ref{fig:DBCTNet-Bert-Houston13}, \ref{fig:FAHM-T5-Houston13}, \ref{fig:FAHM-Bert-Houston13}) for 3D-ConvSST and FAHM stand out as nearly identical to the ground truth, confirming the efficacy of advanced fusion methods regardless of the text encoder.

% ============================================================================
% INDIAN PINES DATASET
% ============================================================================

%%%%%%%%%%%%%%%%%%%%%%%%%%%%%%%% Indian Pines 3DRCNet-T5Encoder %%%%%%%%%%%%%%%%%%%%%%%%%%%%%%%%%%%%%%%%%%%%%%%%
\begin{figure*}[]
    \centering
       \begin{subfigure}[b]{0.16\textwidth}
            \centering
          \includegraphics[width=1\linewidth, trim=565 50 120 60, clip, angle=90]{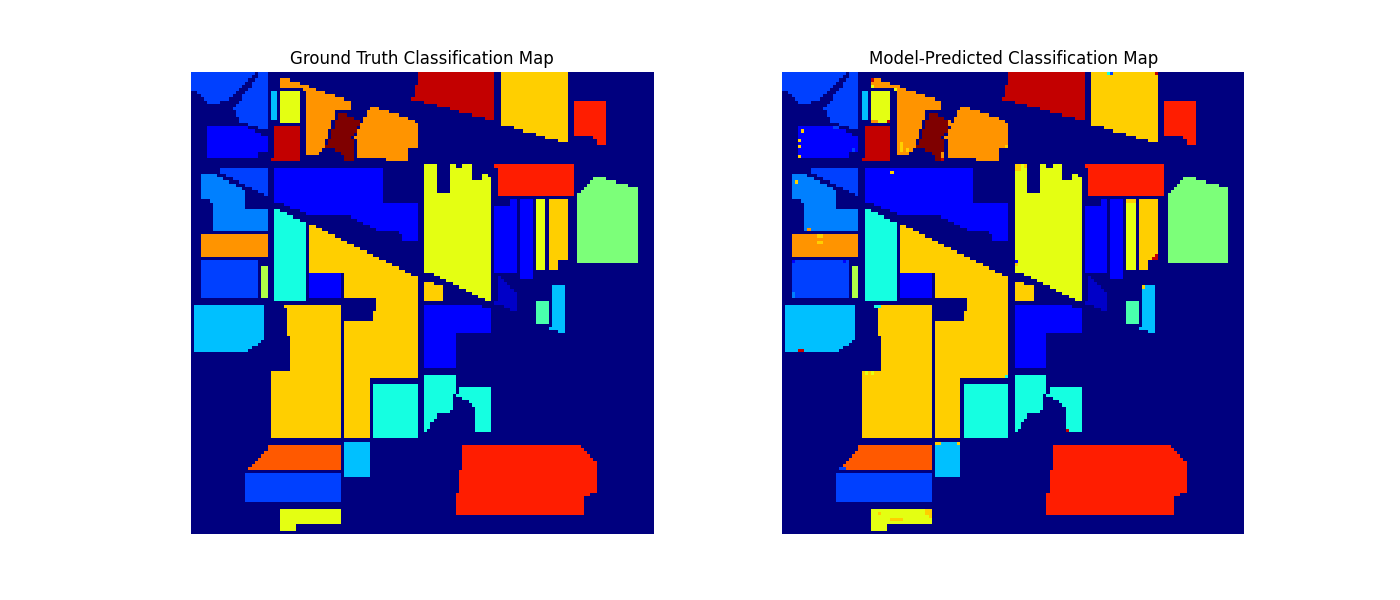}
        \caption{CA}
    \end{subfigure}
       \begin{subfigure}[b]{0.16\textwidth}
            \centering
          \includegraphics[width=1\linewidth,trim=565 50 120 60, clip, angle=90]{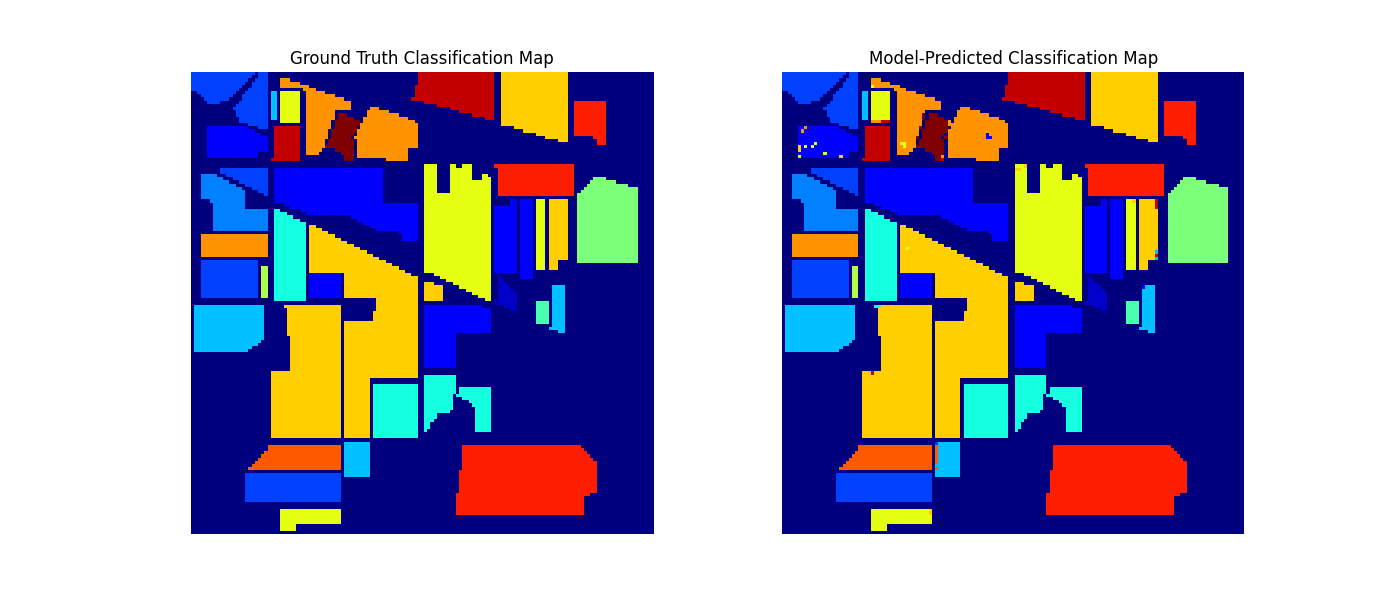}
        \caption{CONCAT}
    \end{subfigure}
       \begin{subfigure}[b]{0.16\textwidth}
            \centering
          \includegraphics[width=1\linewidth, trim=565 50 120 60, clip, angle=90]{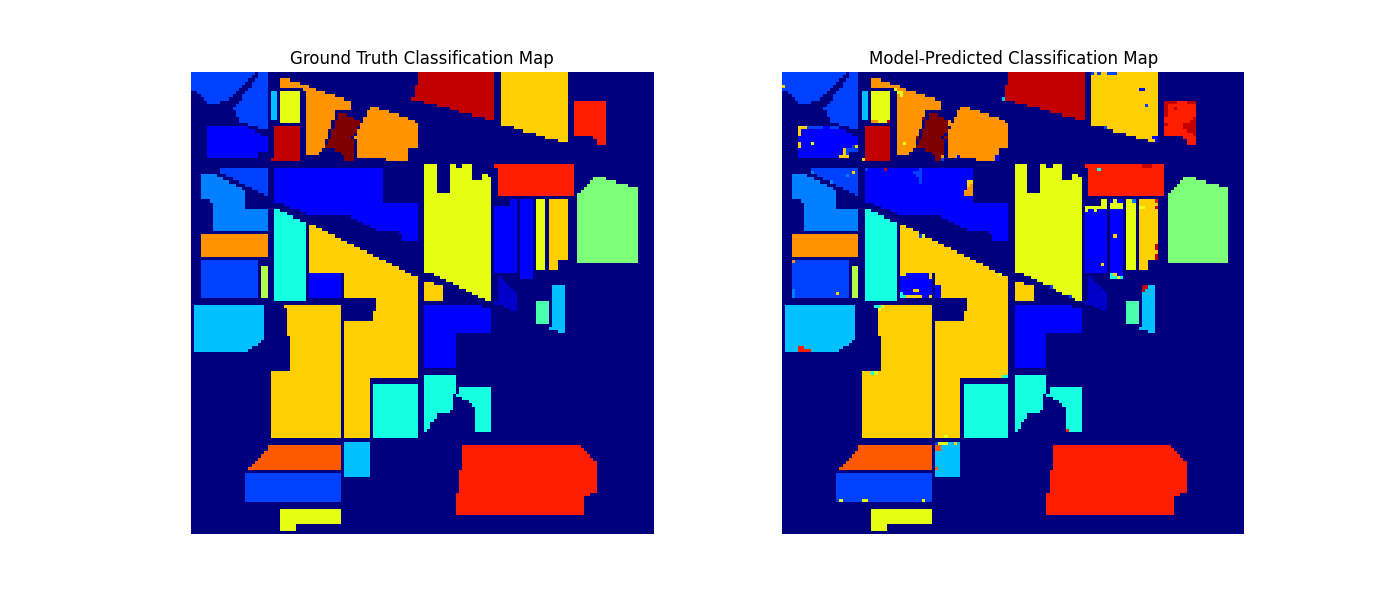}
        \caption{MHA}
    \end{subfigure}
       \begin{subfigure}[b]{0.16\textwidth}
            \centering
          \includegraphics[width=1\linewidth, trim=565 50 120 60, clip, angle=90]{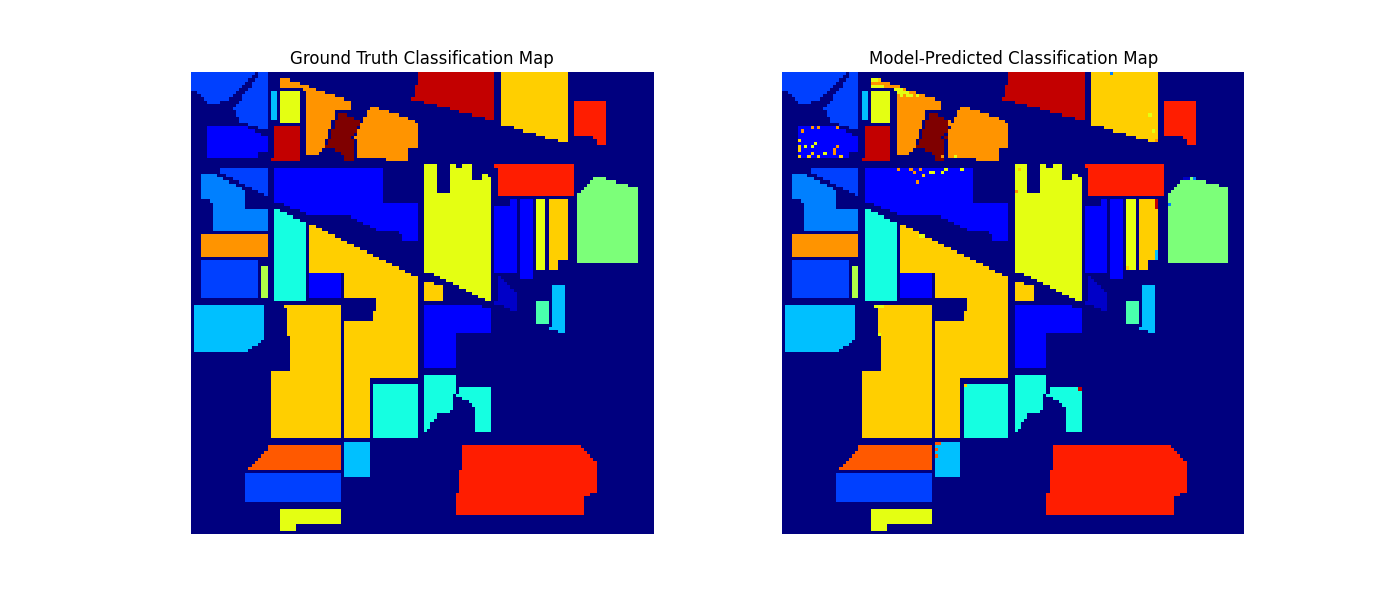}
        \caption{PWA}
    \end{subfigure}
       \begin{subfigure}[b]{0.16\textwidth}
            \centering
          \includegraphics[width=1\linewidth, trim=565 50 120 60, clip, angle=90]{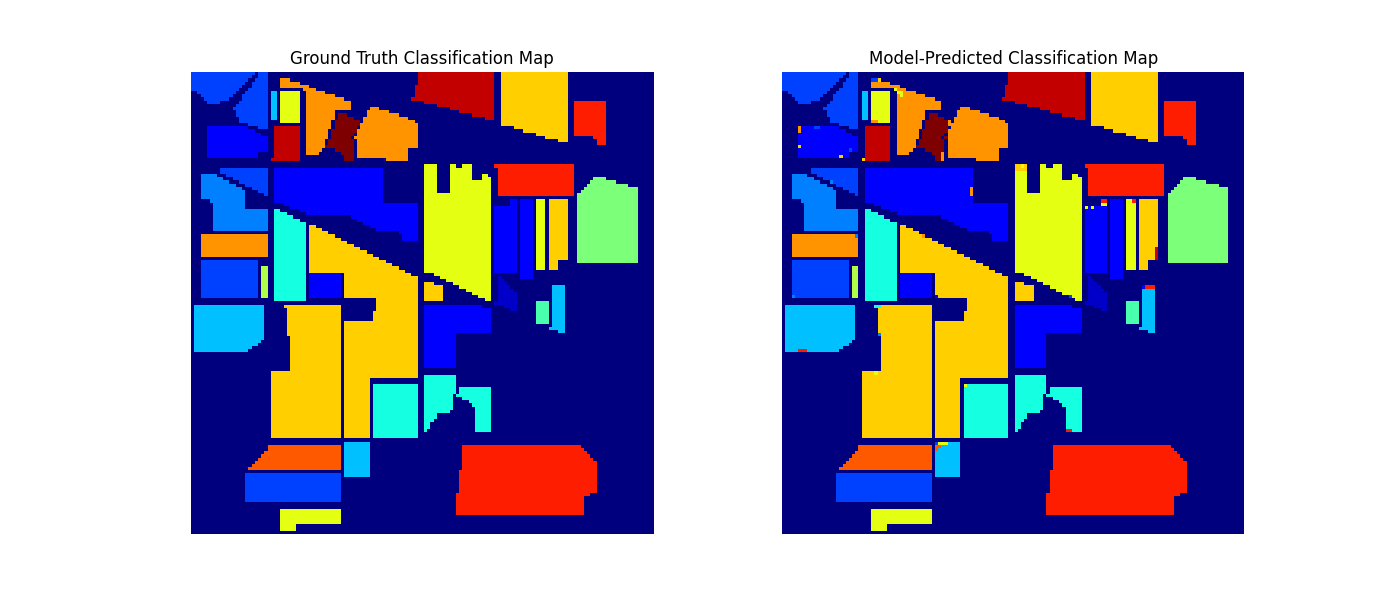}
        \caption{PWM}
    \end{subfigure}
       \begin{subfigure}[b]{0.16\textwidth}
            \centering
          \includegraphics[width=1\linewidth, trim=140 50 565 60, clip, angle=90]{Output_Map/Indian_Pines/3DRCNet-T5Encoder_Large/PWM_prediction_map_run1.png}
        \caption{GT}
    \end{subfigure}
    \caption{Comparison of classification maps for the 3D-RCNet-T5 model on the Indian Pines dataset, showing different fusion methods: Cross Attention (CA), Concatenation (CONCAT), Multi-Head Attention (MHA), Pixel-Wise Addition (PWA), Pixel-Wise Multiplication (PWM), and Ground Truth (GT).}
    \vspace{-3mm}
    \label{fig:3D-RCNet-T5-Indian Pines}
\end{figure*}

%%%%%%%%%%%%%%%%%%%%%%%%%%%%%%%%%%%%%%%%%%%%%%%%  Indian Pines  3DRCNet-BertEncoder   %%%%%%%%%%%%%%%%%%%%%%%%%%%%%%%%%%%%%%%%%%%%%%%%%%%%%%%%%%%%%%%%%%%%%
\begin{figure*}[]
    \centering
        \begin{subfigure}[b]{0.16\textwidth}
            \centering
          \includegraphics[width=1\linewidth, trim=565 50 120 60, clip, angle=90]{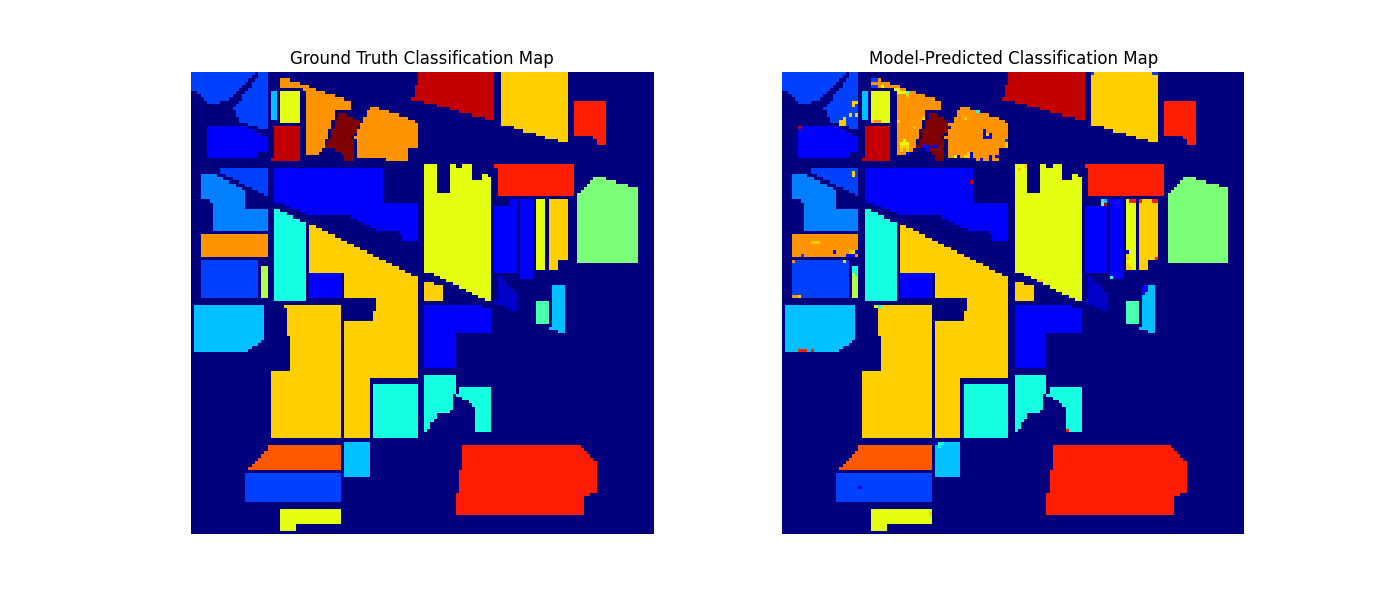}
        \caption{CA}
    \end{subfigure}
        \begin{subfigure}[b]{0.16\textwidth}
            \centering
         \includegraphics[width=1\linewidth, trim=565 50 120 60, clip, angle=90]{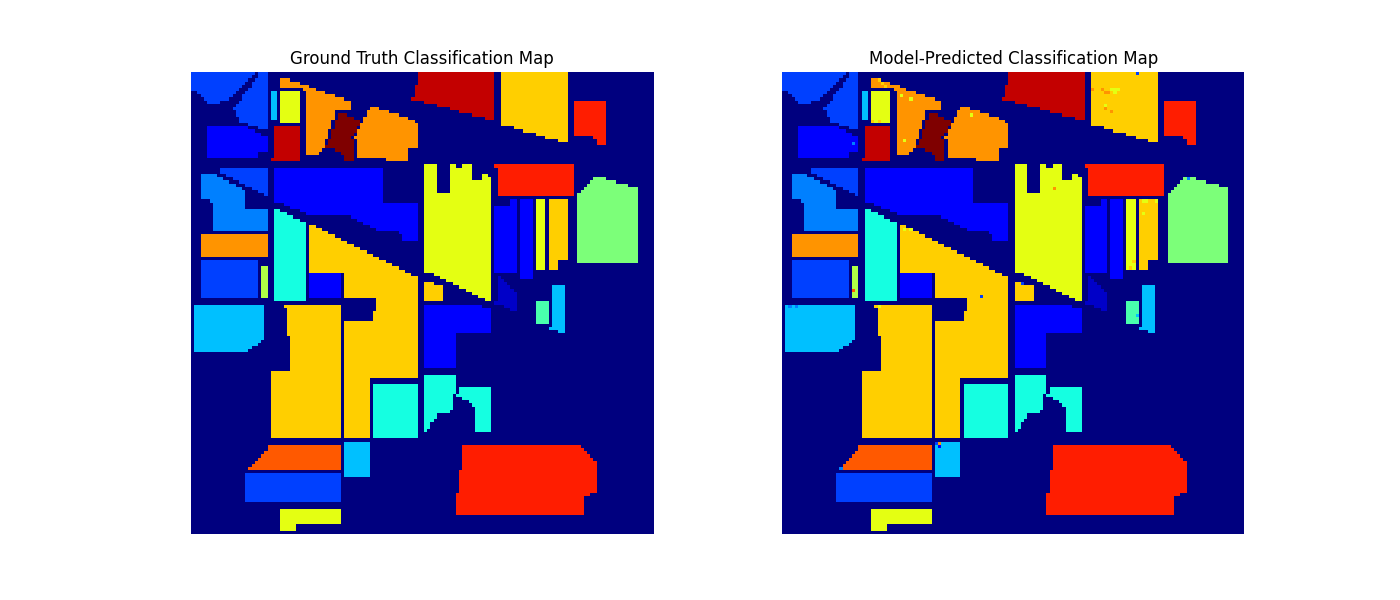}
        \caption{CONCAT}
    \end{subfigure}
        \begin{subfigure}[b]{0.16\textwidth}
            \centering
         \includegraphics[width=1\linewidth, trim=565 50 120 60, clip, angle=90]{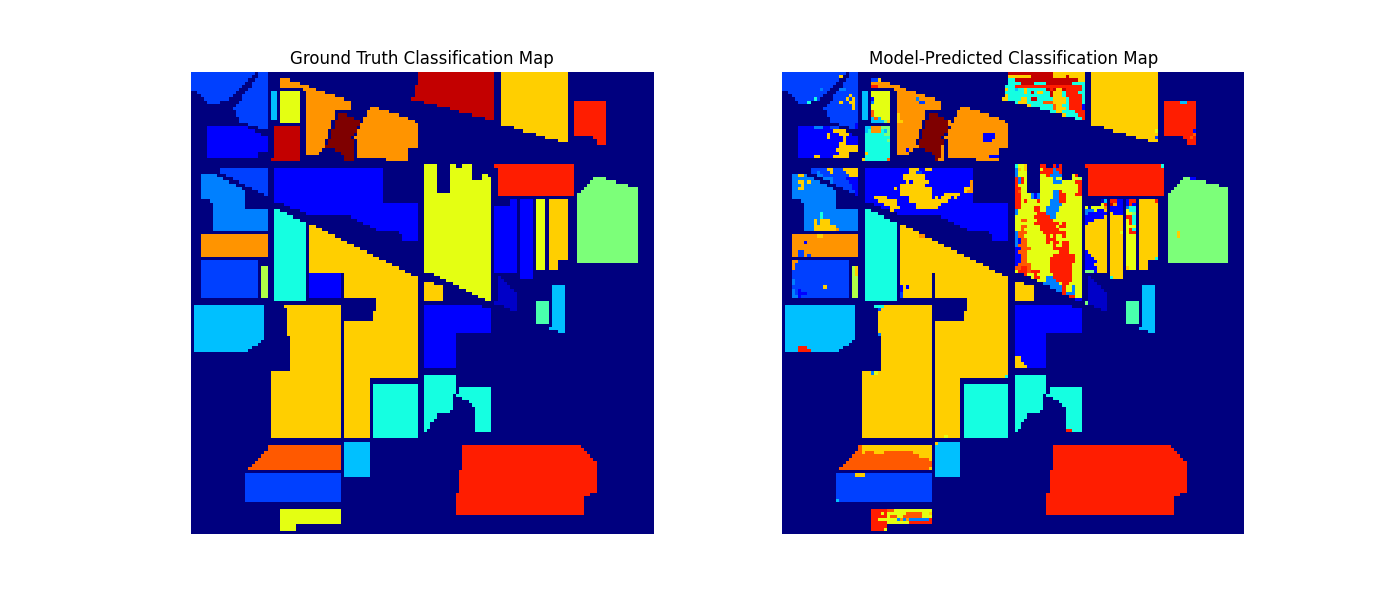}
        \caption{MHA}
    \end{subfigure}
        \begin{subfigure}[b]{0.16\textwidth}
            \centering
         \includegraphics[width=1\linewidth, trim=565 50 120 60, clip, angle=90]{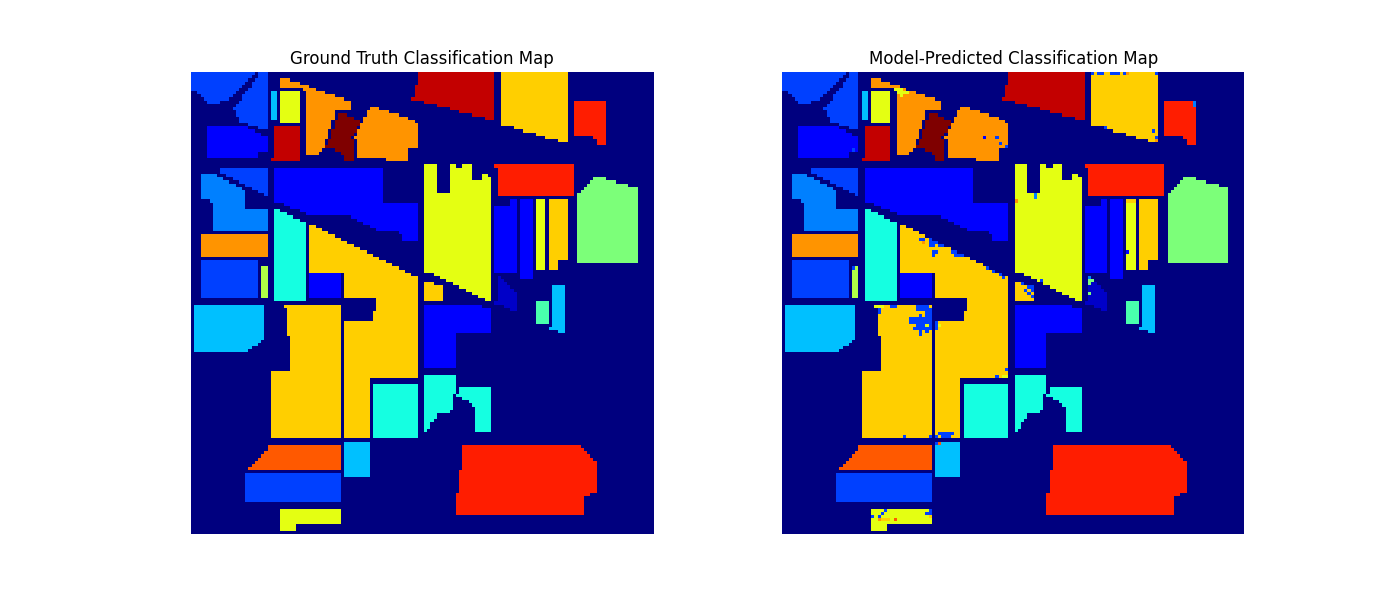}
        \caption{PWA}
    \end{subfigure}
        \begin{subfigure}[b]{0.16\textwidth}
            \centering
         \includegraphics[width=1\linewidth, trim=565 50 120 60, clip, angle=90]{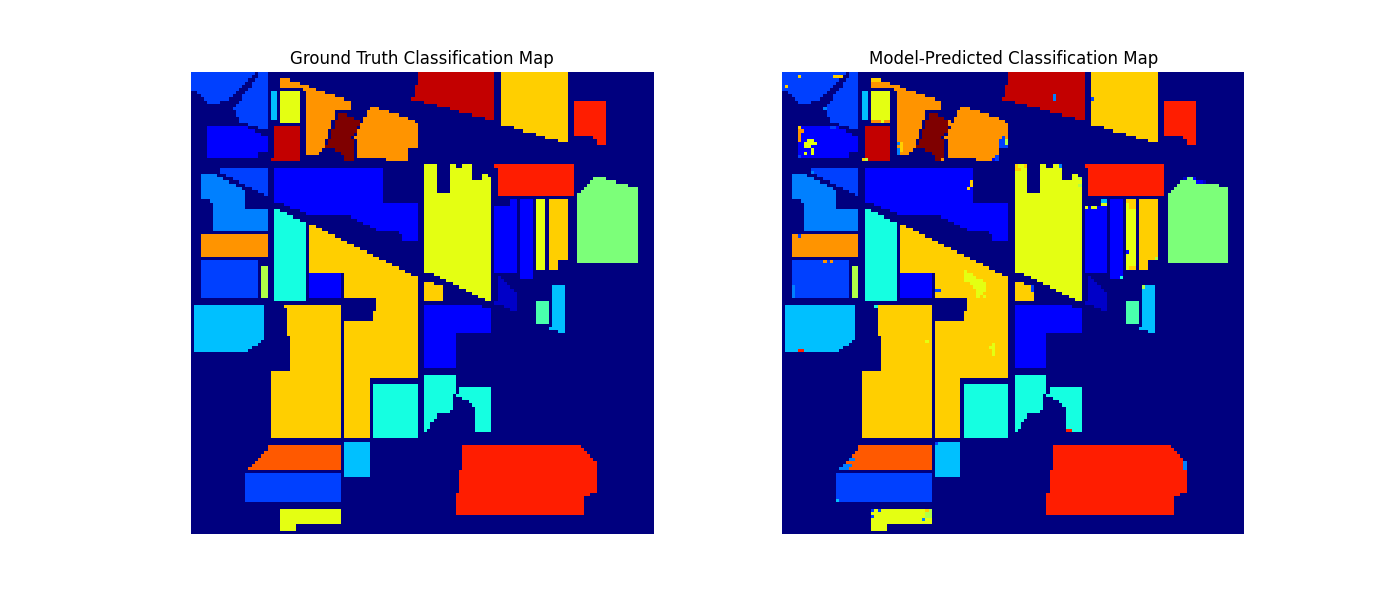}
        \caption{PWM}
    \end{subfigure}
        \begin{subfigure}[b]{0.16\textwidth}
            \centering
         \includegraphics[width=1\linewidth, trim=140 50 565 60, clip, angle=90]{Output_Map/Indian_Pines/3DRCNet-BertEncoder_Large/PWM_prediction_map_run1.png}
        \caption{GT}
    \end{subfigure}
    \caption{Comparison of classification maps for the 3D-RCNet-Bert model on the Indian Pines dataset, showing different fusion methods: Cross Attention (CA), Concatenation (CONCAT), Multi-Head Attention (MHA), Pixel-Wise Addition (PWA), Pixel-Wise Multiplication (PWM), and Ground Truth (GT).}
    \vspace{-3mm}
    \label{fig:3D-RCNet-Bert-Indian Pines}
\end{figure*}

%%%%%%%%%%%%%%%%%%%%%%%%%%%%%%%% Indian Pines 3D_ConvSST-T5Encoder %%%%%%%%%%%%%%%%%%%%%%%%%%%%%%%%%%%%%%%%%%%%%%%%
\begin{figure*}[]
    \centering
       \begin{subfigure}[b]{0.16\textwidth}
            \centering
          \includegraphics[width=1\linewidth, trim=565 50 120 60, clip, angle=90]{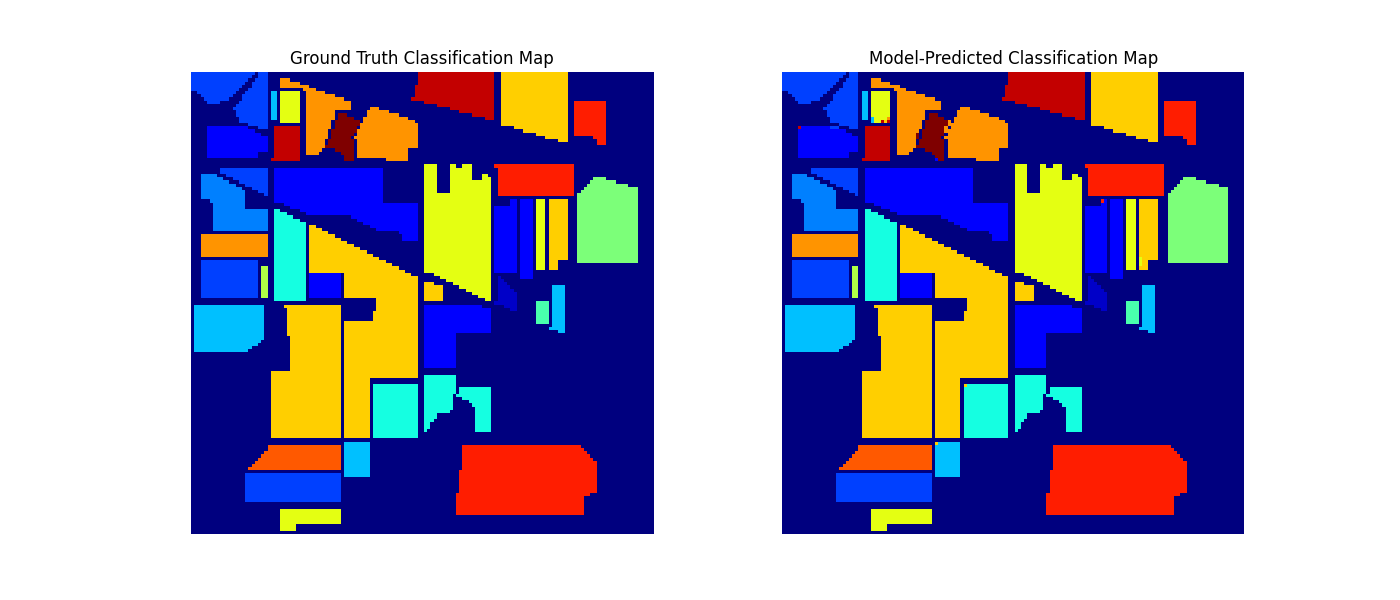}
        \caption{CA}
    \end{subfigure}
       \begin{subfigure}[b]{0.16\textwidth}
            \centering
          \includegraphics[width=1\linewidth, trim=565 50 120 60, clip, angle=90]{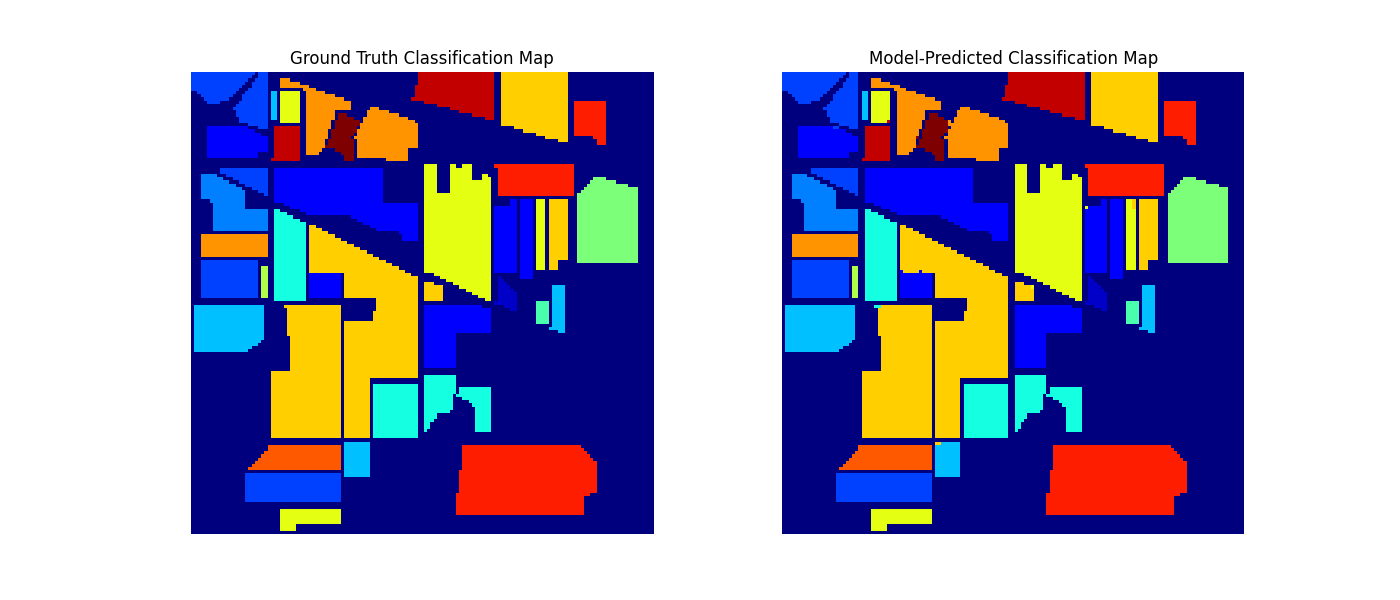}
        \caption{CONCAT}
    \end{subfigure}
       \begin{subfigure}[b]{0.16\textwidth}
            \centering
          \includegraphics[width=1\linewidth, trim=565 50 120 60, clip, angle=90]{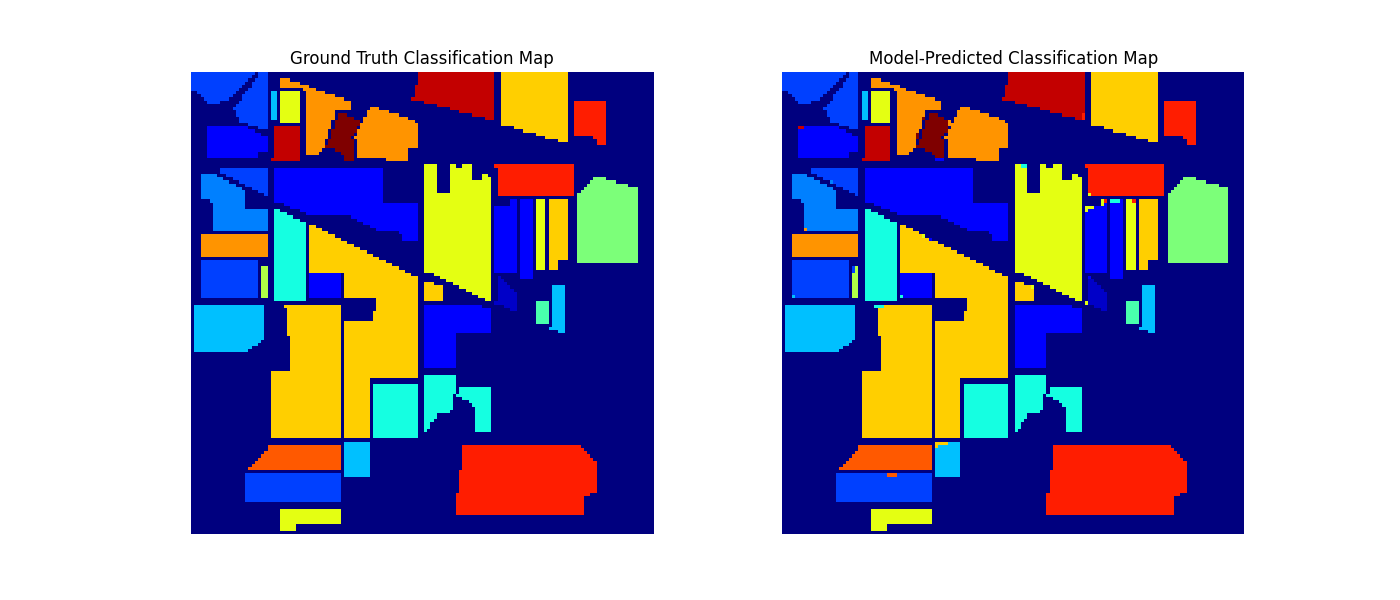}
        \caption{MHA}
    \end{subfigure}
       \begin{subfigure}[b]{0.16\textwidth}
            \centering
          \includegraphics[width=1\linewidth, trim=565 50 120 60, clip, angle=90]{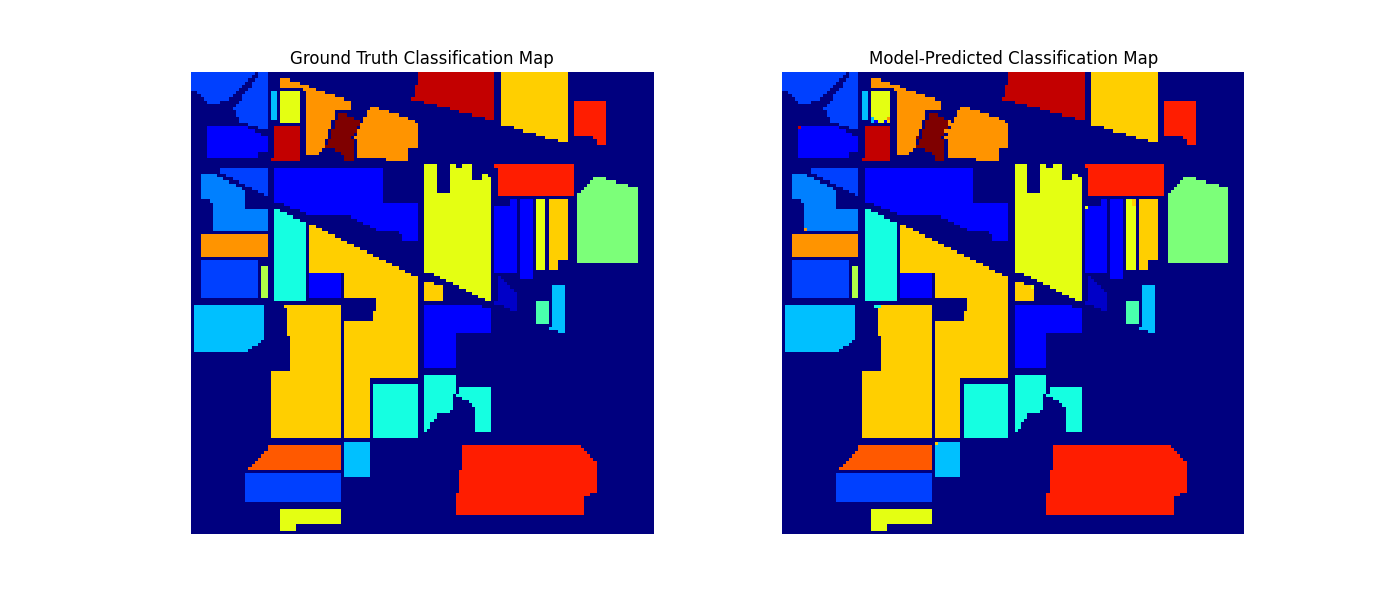}
        \caption{PWA}
    \end{subfigure}
       \begin{subfigure}[b]{0.16\textwidth}
            \centering
          \includegraphics[width=1\linewidth, trim=565 50 120 60, clip, angle=90]{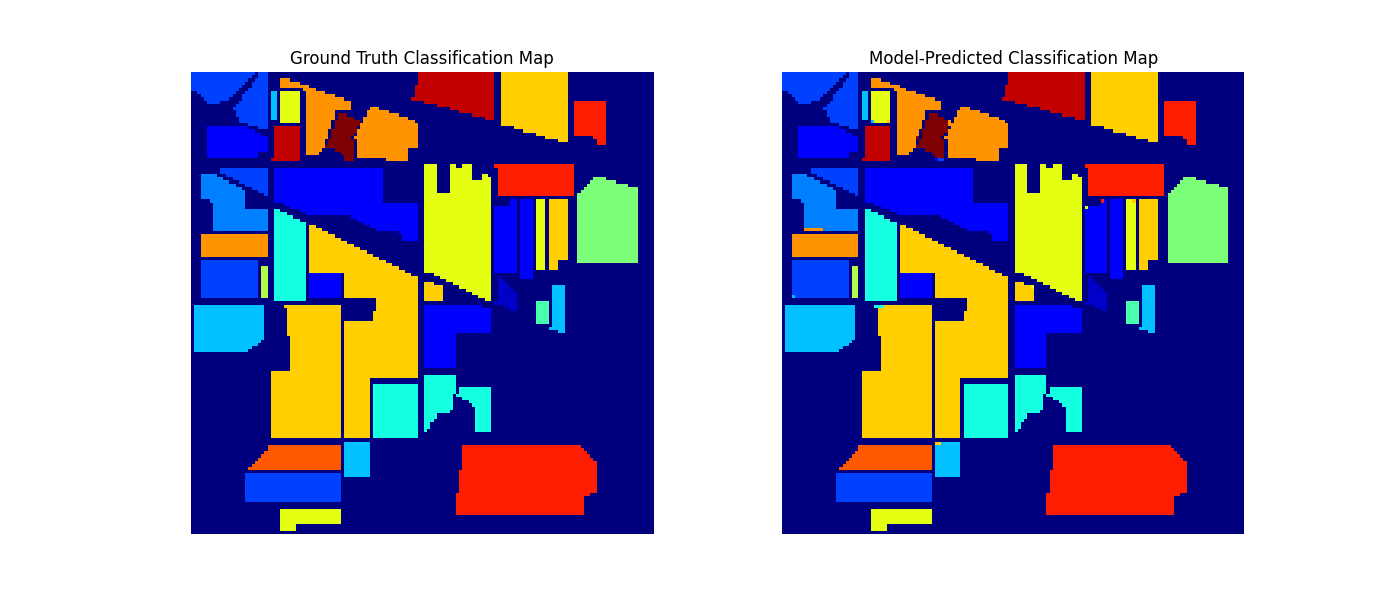}
        \caption{PWM}
    \end{subfigure}
       \begin{subfigure}[b]{0.16\textwidth}
            \centering
          \includegraphics[width=1\linewidth, trim=140 50 565 60, clip, angle=90]{Output_Map/Indian_Pines/3D_ConvSST-T5Encoder_Large/PWM_prediction_map_run1.png}
        \caption{GT}
    \end{subfigure}
    \caption{Comparison of classification maps for the 3D-ConvSST-T5 model on the Indian Pines dataset, showing different fusion methods: Cross Attention (CA), Concatenation (CONCAT), Multi-Head Attention (MHA), Pixel-Wise Addition (PWA), Pixel-Wise Multiplication (PWM), and Ground Truth (GT).}
    \vspace{-3mm}
    \label{fig:3D-ConvSST-T5-Indian Pines}
\end{figure*}

%%%%%%%%%%%%%%%%%%%%%%%%%%%%%%%%%%%%%%%%%%%%%%%%  Indian Pines  3D_ConvSST-BertEncoder   %%%%%%%%%%%%%%%%%%%%%%%%%%%%%%%%%%%%%%%%%%%%%%%%%%%%%%%%%%%%%%%%%%%%%
\begin{figure*}[]
    \centering
       \begin{subfigure}[b]{0.16\textwidth}
            \centering
          \includegraphics[width=1\linewidth, trim=565 50 120 60, clip, angle=90]{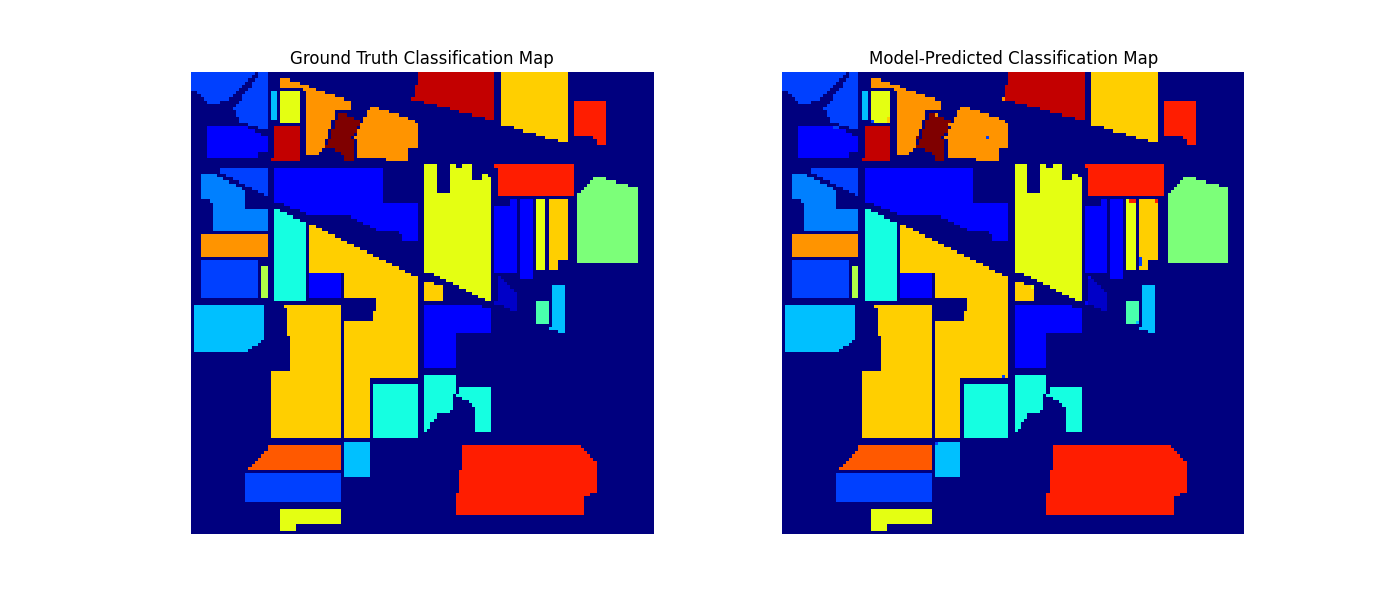}
        \caption{CA}
    \end{subfigure}
       \begin{subfigure}[b]{0.16\textwidth}
            \centering
          \includegraphics[width=1\linewidth, trim=565 50 120 60, clip, angle=90]{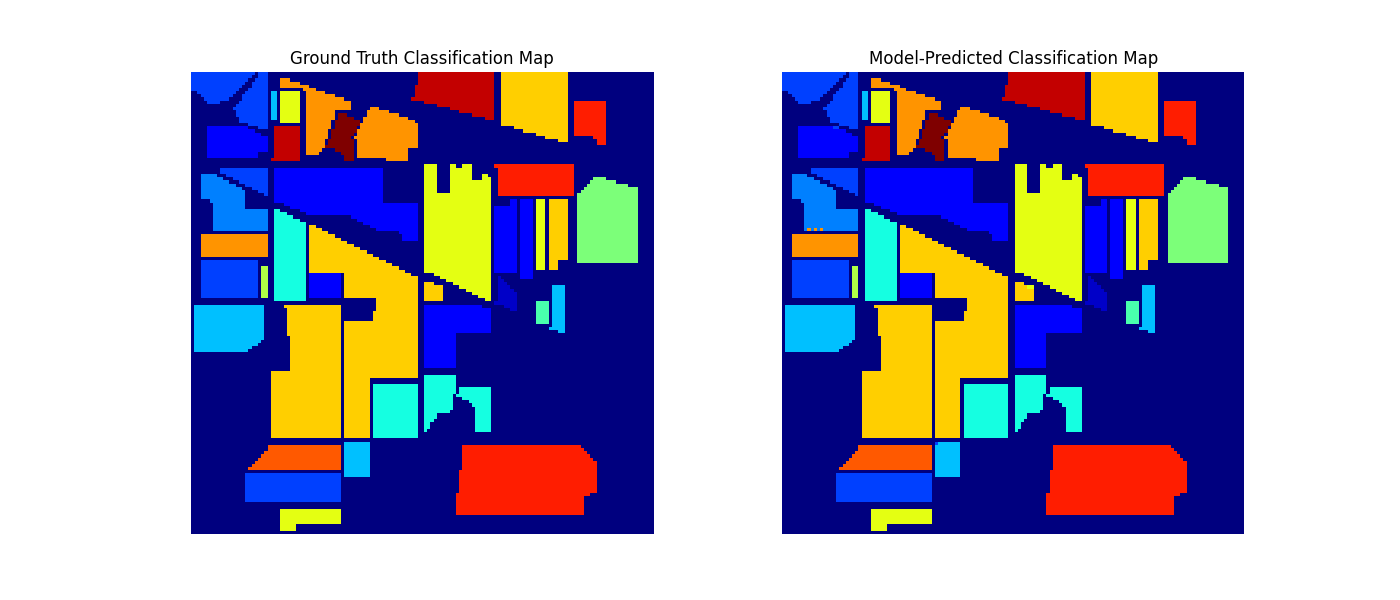}
        \caption{CONCAT}
    \end{subfigure}
       \begin{subfigure}[b]{0.16\textwidth}
            \centering
          \includegraphics[width=1\linewidth, trim=565 50 120 60, clip, angle=90]{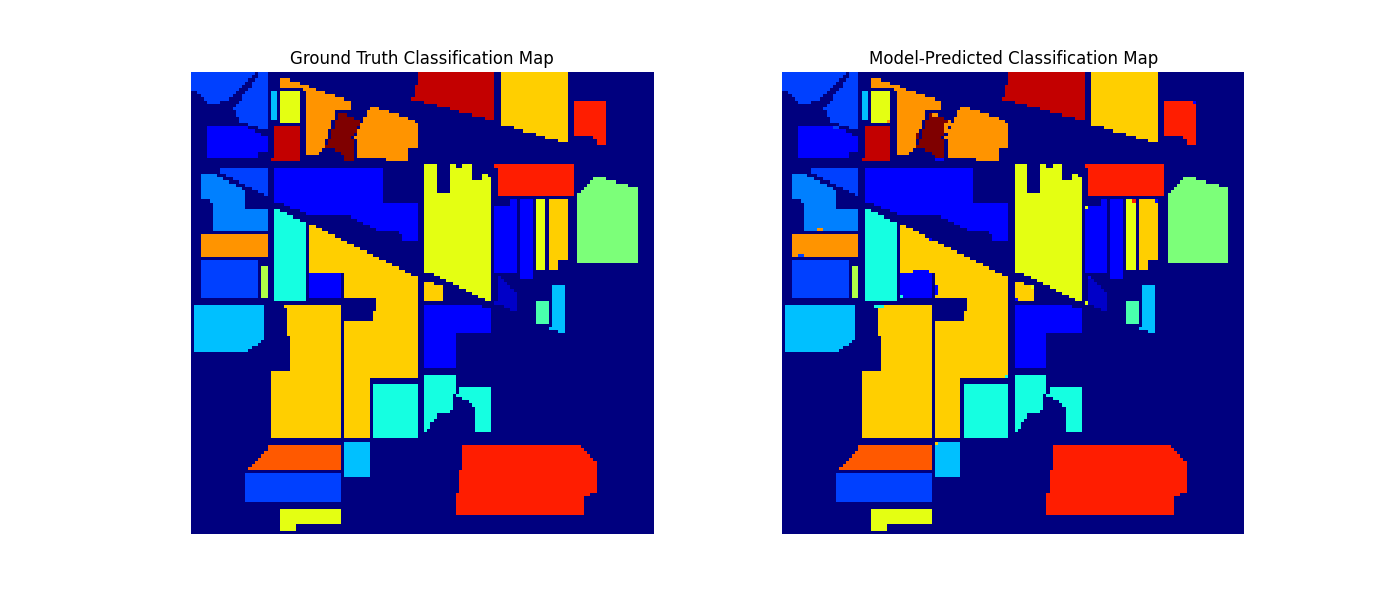}
        \caption{MHA}
    \end{subfigure}
       \begin{subfigure}[b]{0.16\textwidth}
            \centering
          \includegraphics[width=1\linewidth, trim=565 50 120 60, clip, angle=90]{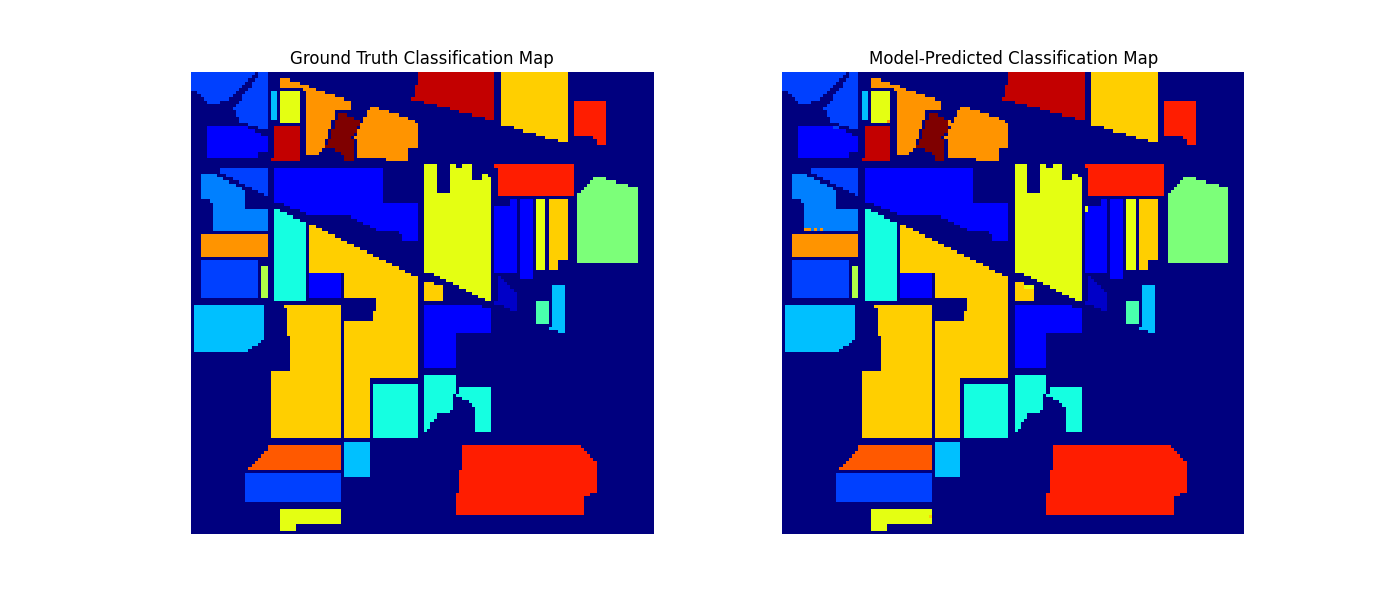}
        \caption{PWA}
    \end{subfigure}
       \begin{subfigure}[b]{0.16\textwidth}
            \centering
          \includegraphics[width=1\linewidth, trim=565 50 120 60, clip, angle=90]{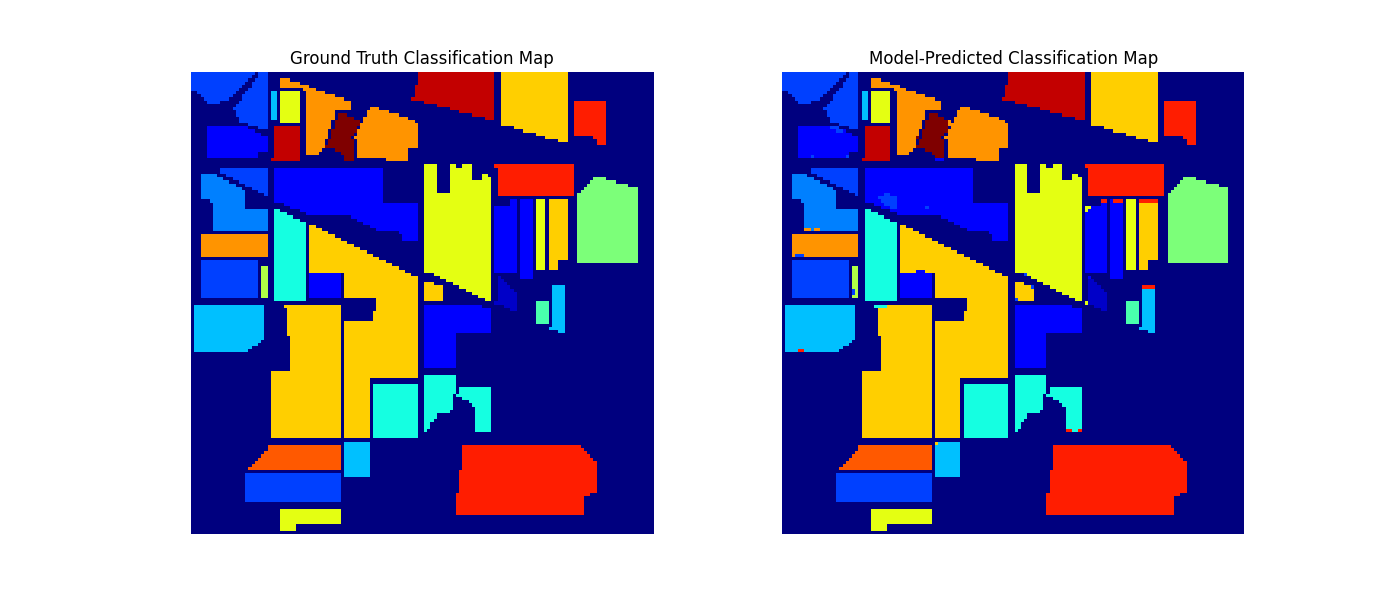}
        \caption{PWM}
    \end{subfigure}
       \begin{subfigure}[b]{0.16\textwidth}
            \centering
          \includegraphics[width=1\linewidth, trim=140 50 565 60, clip, angle=90]{Output_Map/Indian_Pines/3D_ConvSST-BertEncoder_Large/PWM_prediction_map_run1.png}
        \caption{GT}
    \end{subfigure}
    \caption{Comparison of classification maps for the 3D-ConvSST-Bert model on the Indian Pines dataset, showing different fusion methods: Cross Attention (CA), Concatenation (CONCAT), Multi-Head Attention (MHA), Pixel-Wise Addition (PWA), Pixel-Wise Multiplication (PWM), and Ground Truth (GT).}
    \vspace{-3mm}
    \label{fig:3D-ConvSST-Bert-Indian Pines}
\end{figure*}

%%%%%%%%%%%%%%%%%%%%%%%%%%%%%%%% Indian Pines DBCTNet-T5Encoder %%%%%%%%%%%%%%%%%%%%%%%%%%%%%%%%%%%%%%%%%%%%%%%%
\begin{figure*}[]
    \centering
       \begin{subfigure}[b]{0.16\textwidth}
            \centering
          \includegraphics[width=1\linewidth, trim=565 50 120 60, clip, angle=90]{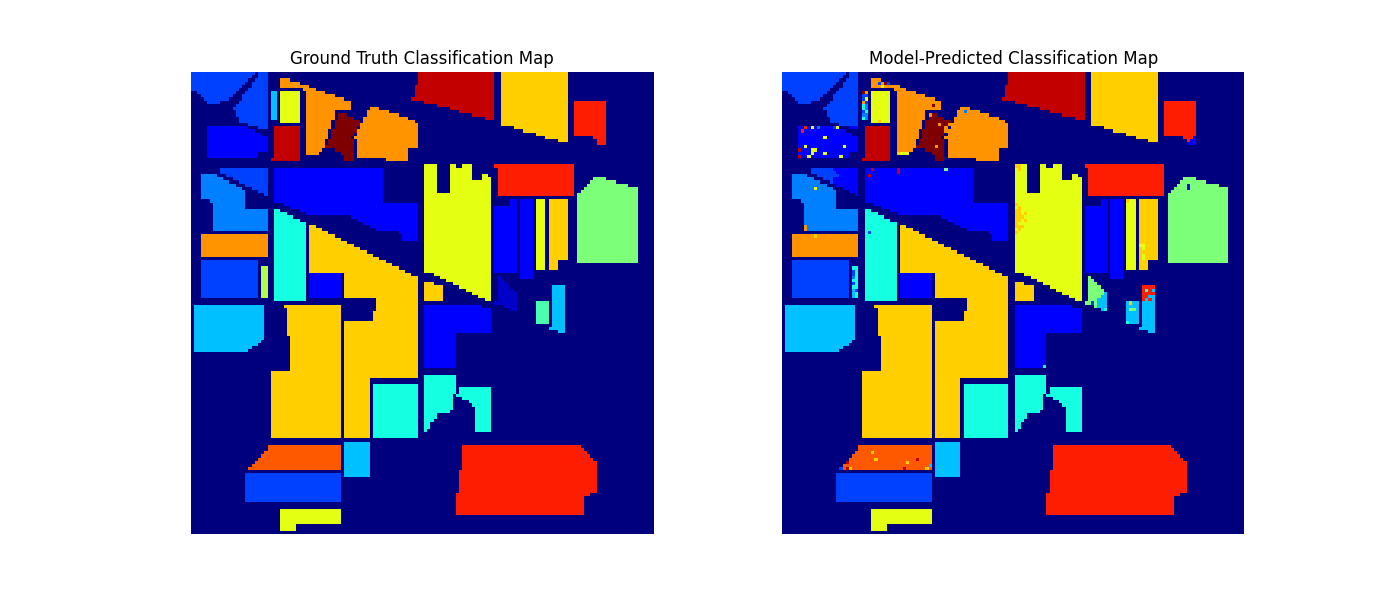}
        \caption{CA}
    \end{subfigure}
       \begin{subfigure}[b]{0.16\textwidth}
            \centering
          \includegraphics[width=1\linewidth, trim=565 50 120 60, clip, angle=90]{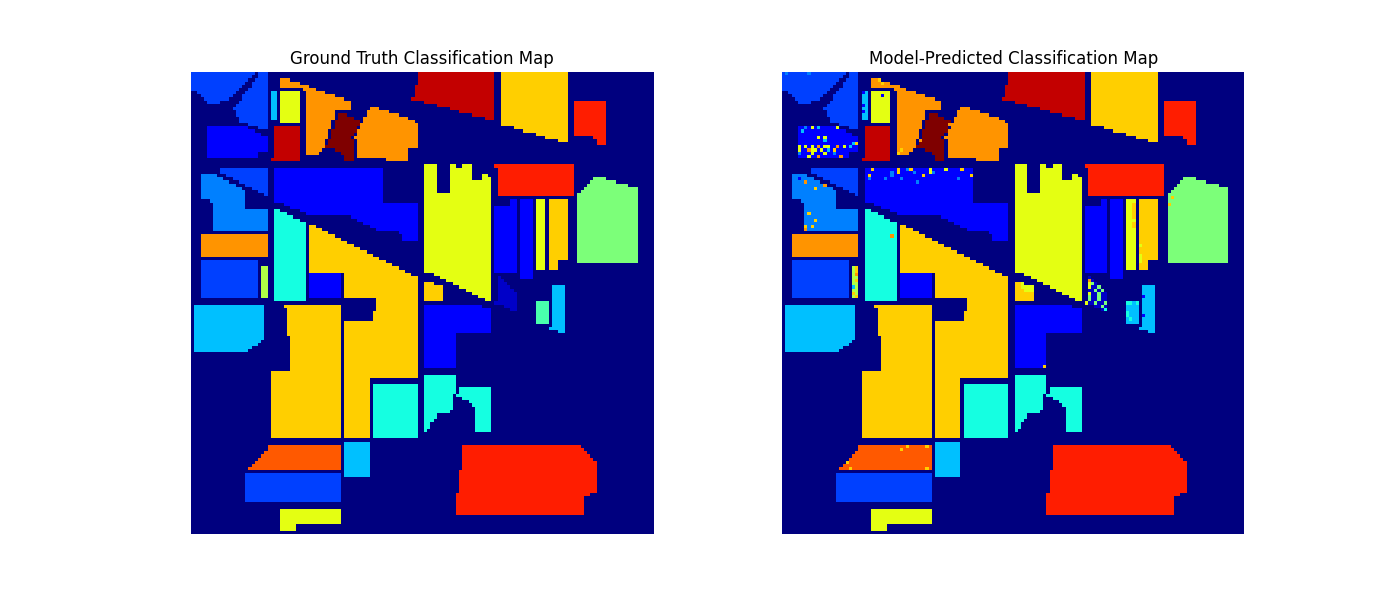}
        \caption{CONCAT}
    \end{subfigure}
       \begin{subfigure}[b]{0.16\textwidth}
            \centering
          \includegraphics[width=1\linewidth, trim=565 50 120 60, clip, angle=90]{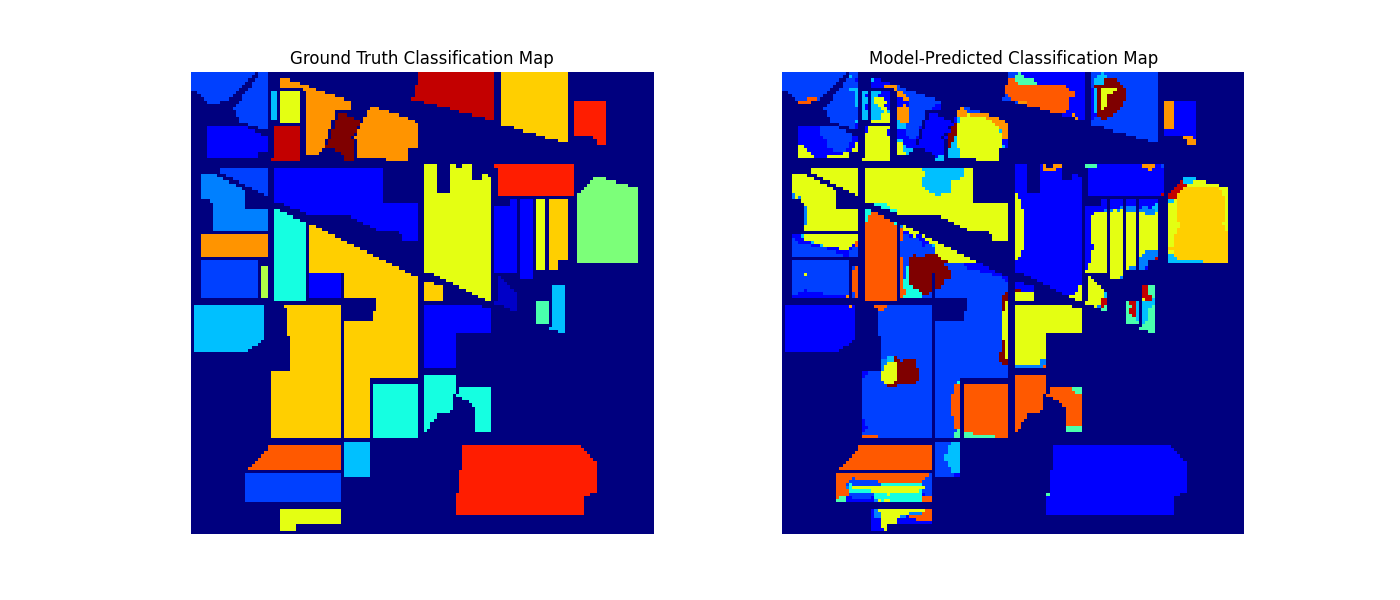}
        \caption{MHA}
    \end{subfigure}
       \begin{subfigure}[b]{0.16\textwidth}
            \centering
          \includegraphics[width=1\linewidth, trim=565 50 120 60, clip, angle=90]{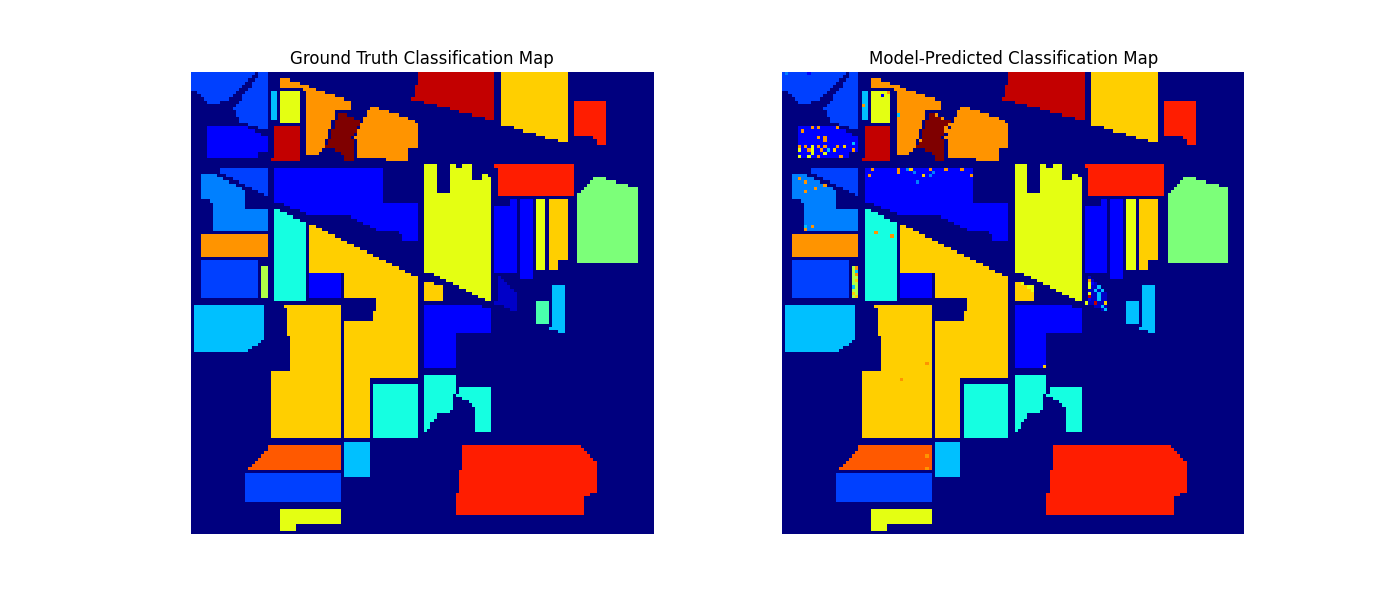}
        \caption{PWA}
    \end{subfigure}
       \begin{subfigure}[b]{0.16\textwidth}
            \centering
          \includegraphics[width=1\linewidth, trim=565 50 120 60, clip, angle=90]{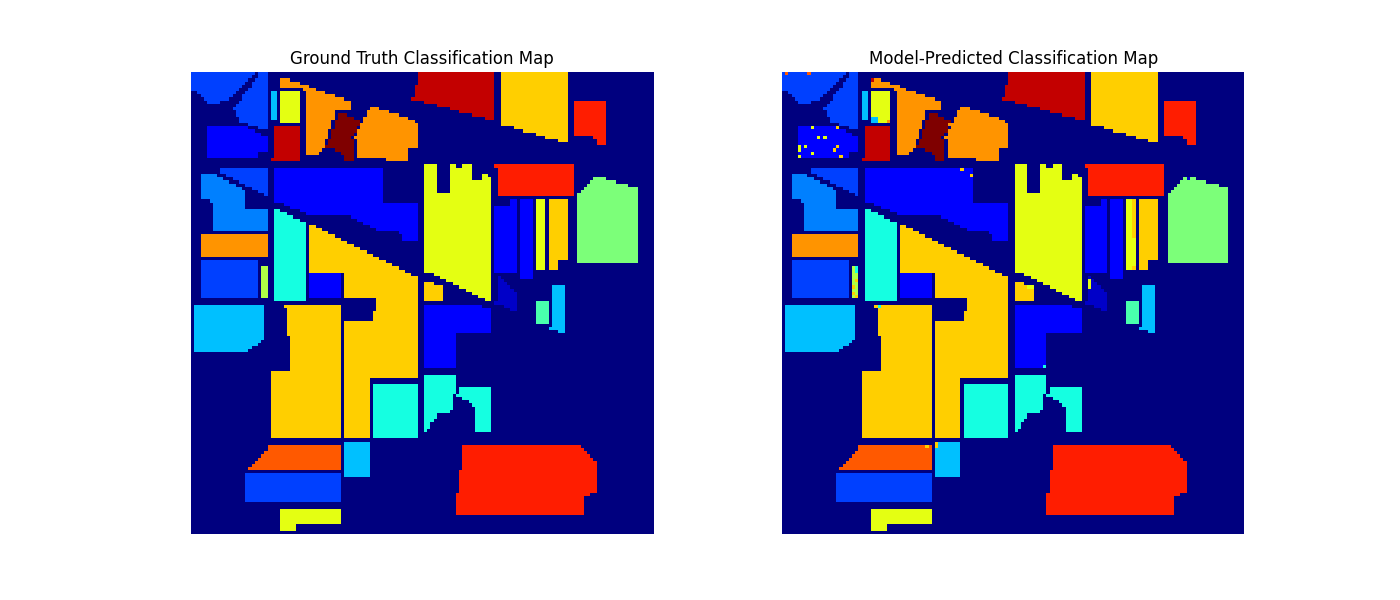}
        \caption{PWM}
    \end{subfigure}
       \begin{subfigure}[b]{0.16\textwidth}
            \centering
          \includegraphics[width=1\linewidth, trim=140 50 565 60, clip, angle=90]{Output_Map/Indian_Pines/DBCTNet-T5Encoder_Large/PWM_prediction_map_run1.png}
        \caption{GT}
    \end{subfigure}
    \caption{Comparison of classification maps for the DBCTNet-T5 model on the Indian Pines dataset, showing different fusion methods: Cross Attention (CA), Concatenation (CONCAT), Multi-Head Attention (MHA), Pixel-Wise Addition (PWA), Pixel-Wise Multiplication (PWM), and Ground Truth (GT).}
    \vspace{-3mm}
    \label{fig:DBCTNet-T5-Indian Pines}
\end{figure*}

%%%%%%%%%%%%%%%%%%%%%%%%%%%%%%%%%%%%%%%%%%%%%%%%  Indian Pines  DBCTNet-BertEncoder   %%%%%%%%%%%%%%%%%%%%%%%%%%%%%%%%%%%%%%%%%%%%%%%%%%%%%%%%%%%%%%%%%%%%%
\begin{figure*}[]
    \centering
       \begin{subfigure}[b]{0.16\textwidth}
            \centering
          \includegraphics[width=1\linewidth, trim=565 50 120 60, clip, angle=90]{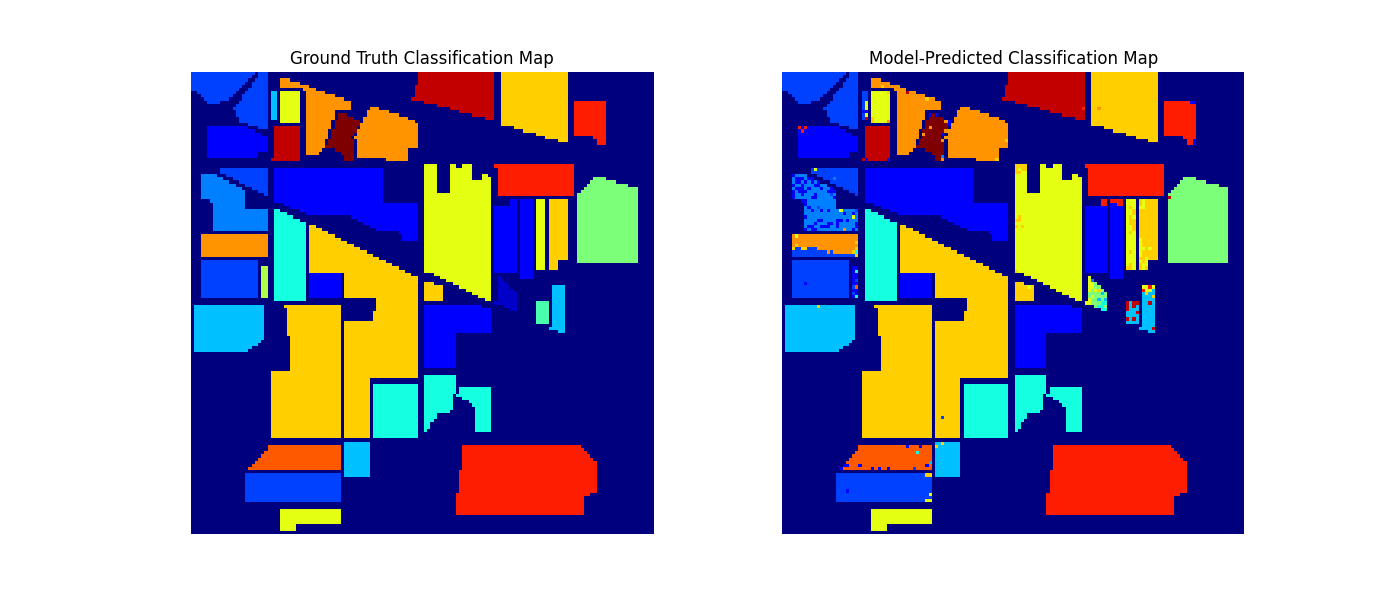}
        \caption{CA}
    \end{subfigure}
       \begin{subfigure}[b]{0.16\textwidth}
            \centering
          \includegraphics[width=1\linewidth, trim=565 50 120 60, clip, angle=90]{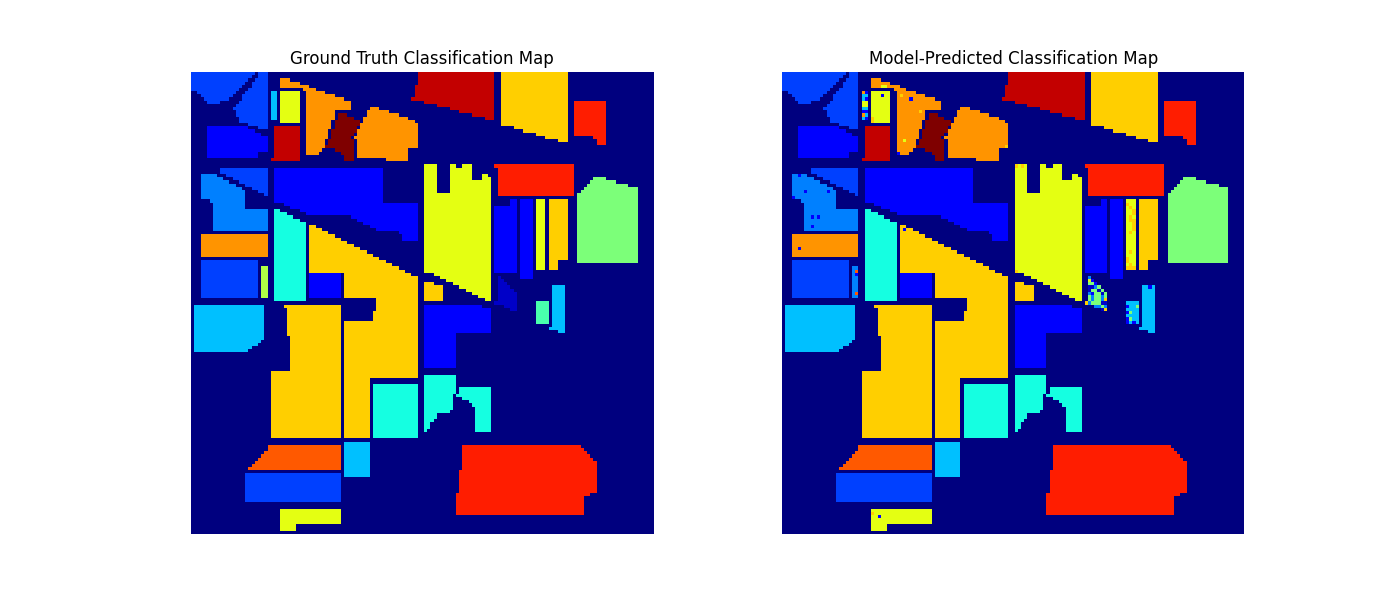}
        \caption{CONCAT}
    \end{subfigure}
       \begin{subfigure}[b]{0.16\textwidth}
            \centering
          \includegraphics[width=1\linewidth, trim=565 50 120 60, clip, angle=90]{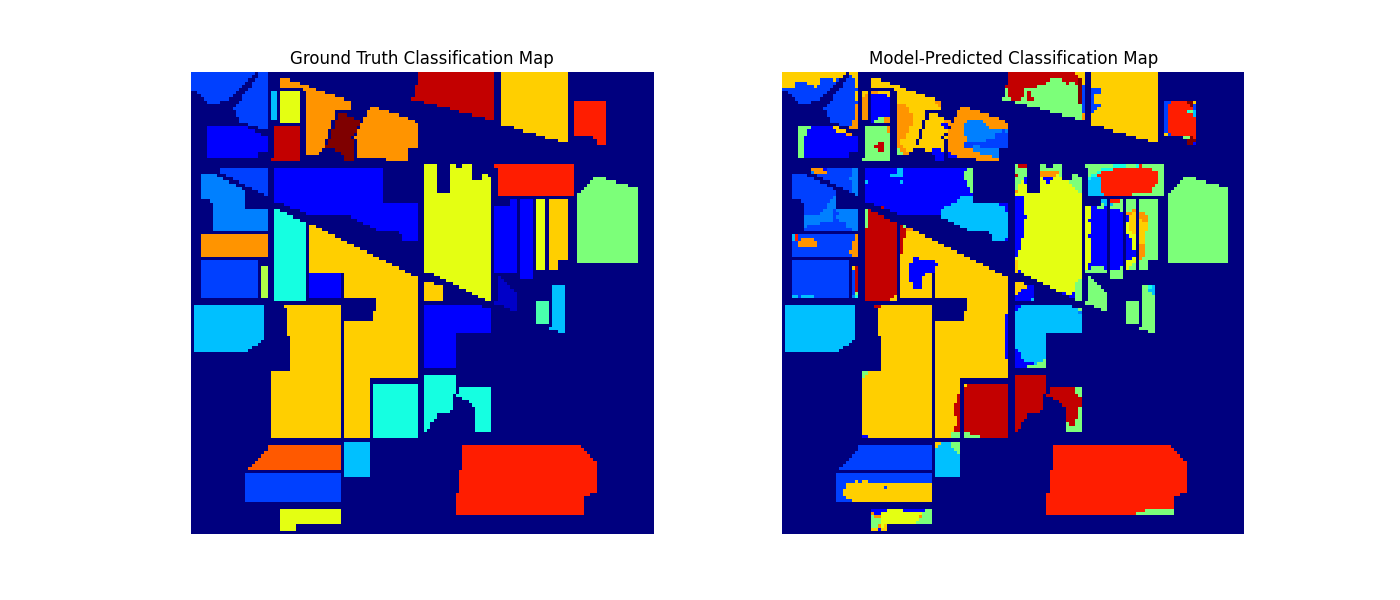}
        \caption{MHA}
    \end{subfigure}
       \begin{subfigure}[b]{0.16\textwidth}
            \centering
          \includegraphics[width=1\linewidth, trim=565 50 120 60, clip, angle=90]{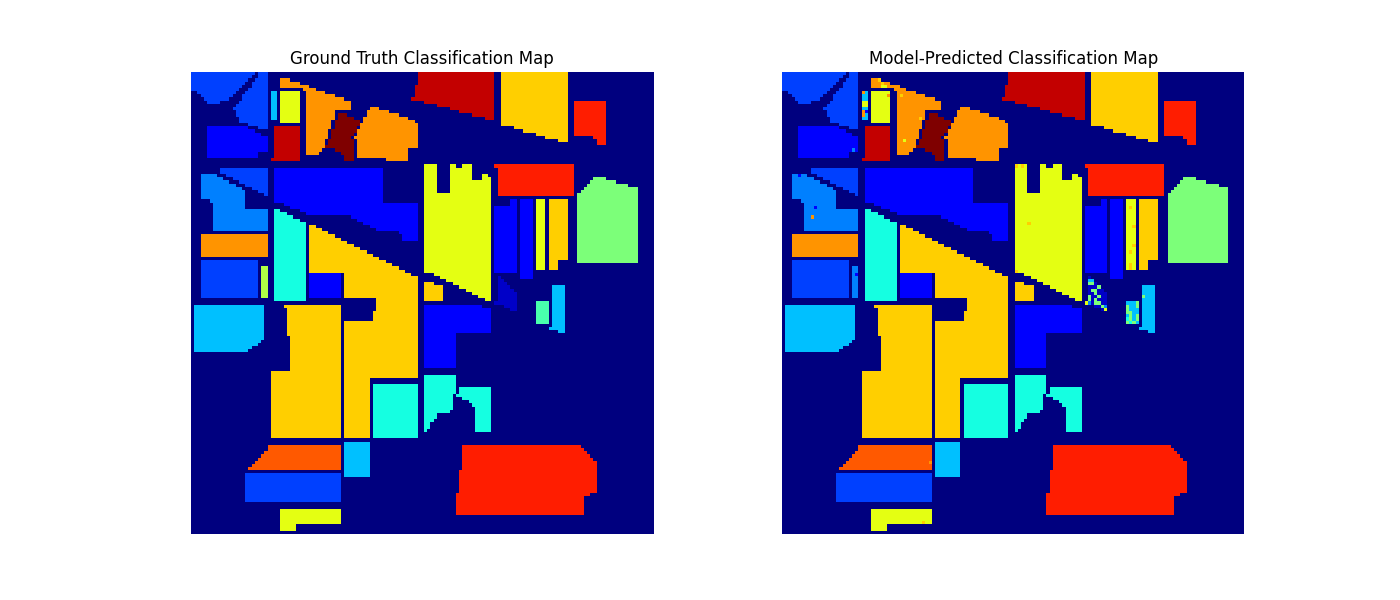}
        \caption{PWA}
    \end{subfigure}
       \begin{subfigure}[b]{0.16\textwidth}
            \centering
          \includegraphics[width=1\linewidth, trim=565 50 120 60, clip, angle=90]{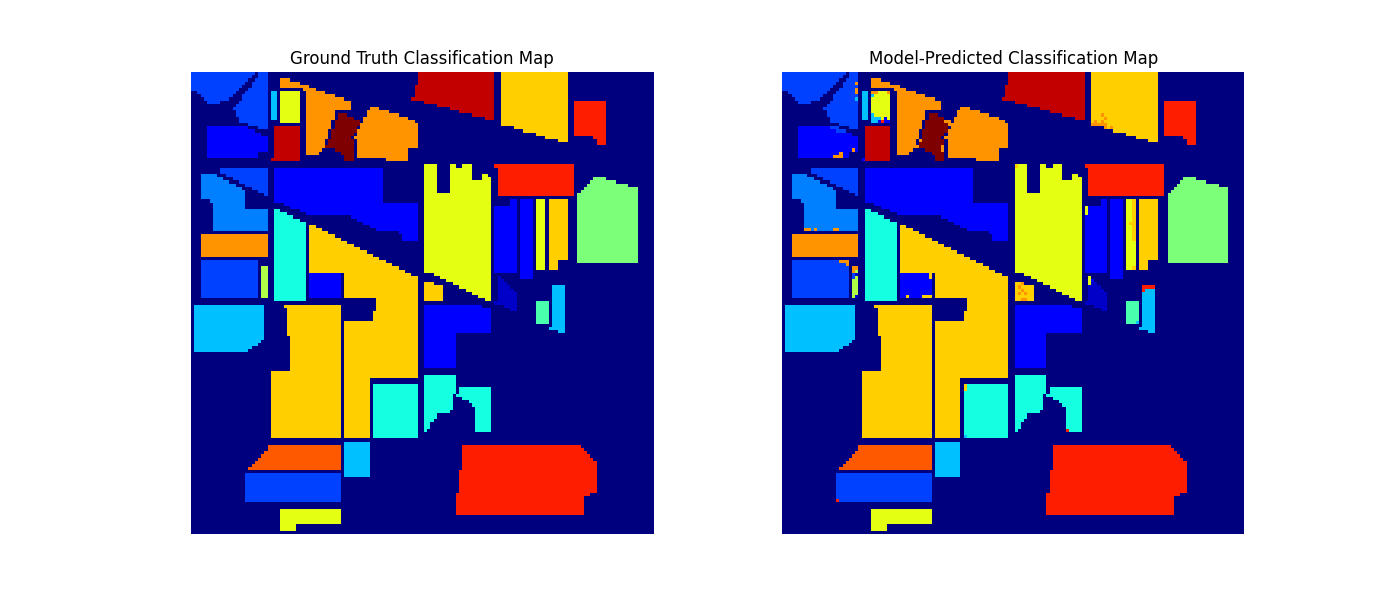}
        \caption{PWM}
    \end{subfigure}
       \begin{subfigure}[b]{0.16\textwidth}
            \centering
          \includegraphics[width=1\linewidth, trim=140 50 565 60, clip, angle=90]{Output_Map/Indian_Pines/DBCTNet-BertEncoder_Large/PWM_prediction_map_run1.png}
        \caption{GT}
    \end{subfigure}
    \caption{Comparison of classification maps for the DBCTNet-Bert model on the Indian Pines dataset, showing different fusion methods: Cross Attention (CA), Concatenation (CONCAT), Multi-Head Attention (MHA), Pixel-Wise Addition (PWA), Pixel-Wise Multiplication (PWM), and Ground Truth (GT).}
    \vspace{-3mm}
    \label{fig:DBCTNet-Bert-Indian Pines}
\end{figure*}

%%%%%% IP
\begin{figure*}[]
    \centering
       \begin{subfigure}[b]{0.16\textwidth}
            \centering
          \includegraphics[width=1\linewidth, trim=565 50 120 60, clip, angle=90]{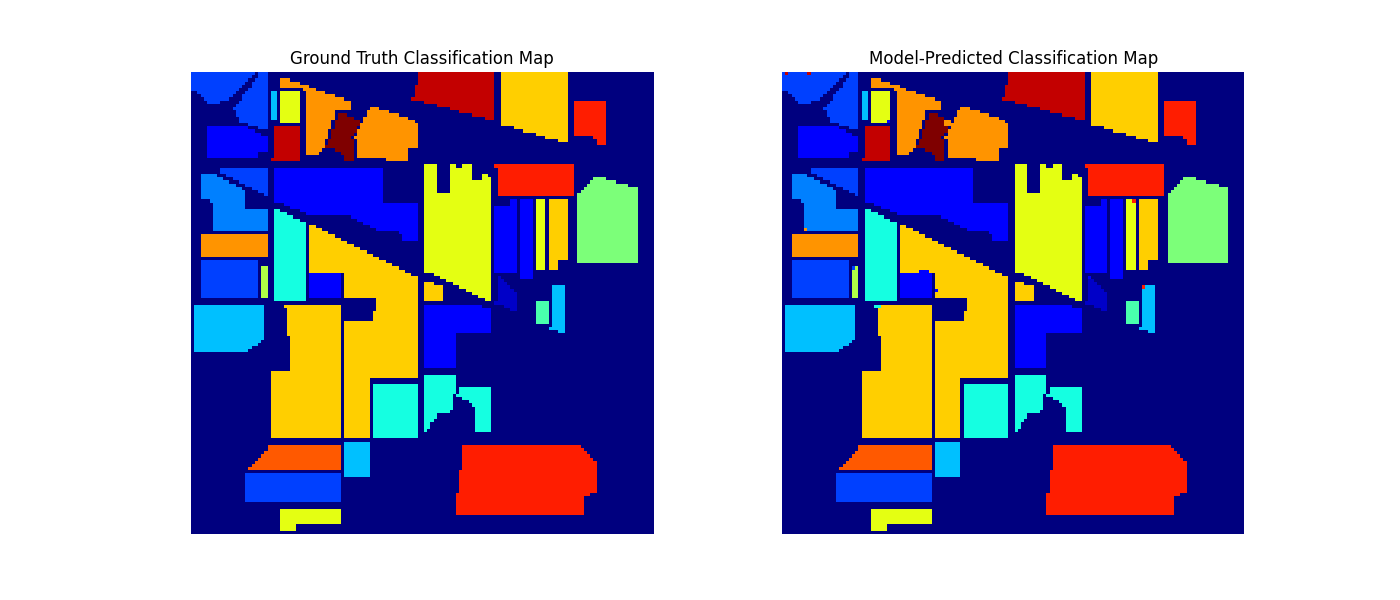}
        \caption{CA}
    \end{subfigure}
       \begin{subfigure}[b]{0.16\textwidth}
            \centering
          \includegraphics[width=1\linewidth, trim=565 50 120 60, clip, angle=90]{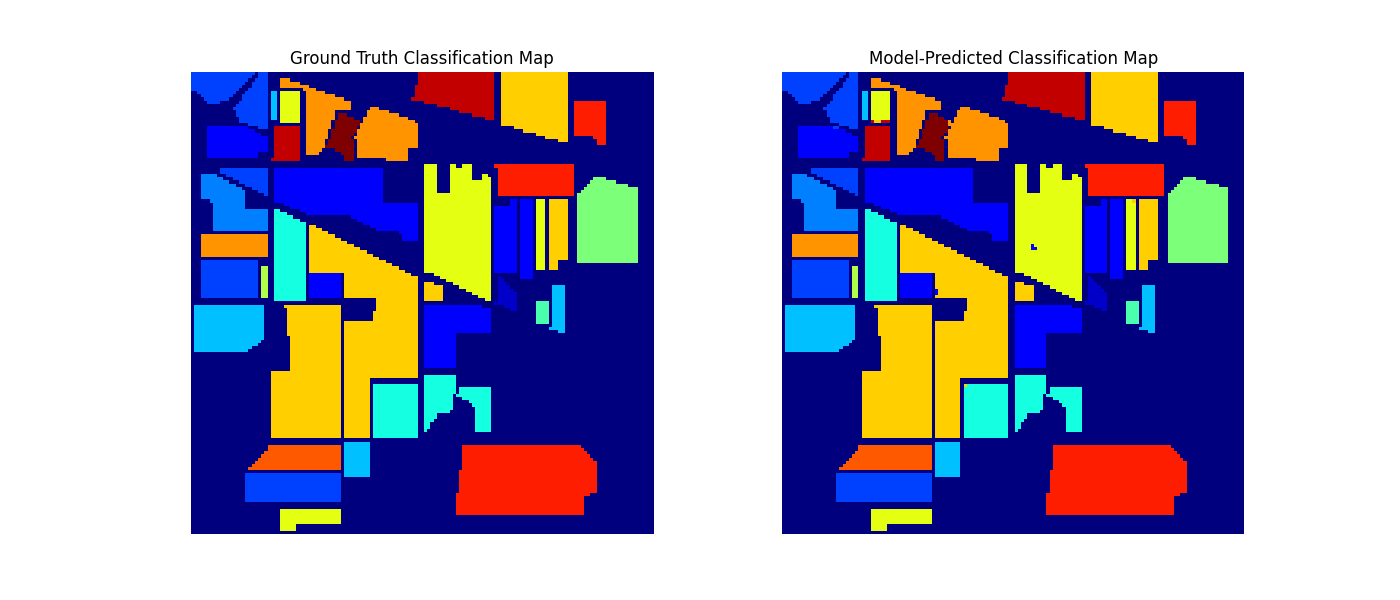}
        \caption{CONCAT}
    \end{subfigure}
       \begin{subfigure}[b]{0.16\textwidth}
            \centering
          \includegraphics[width=1\linewidth,  trim=565 50 120 60, clip, angle=90]{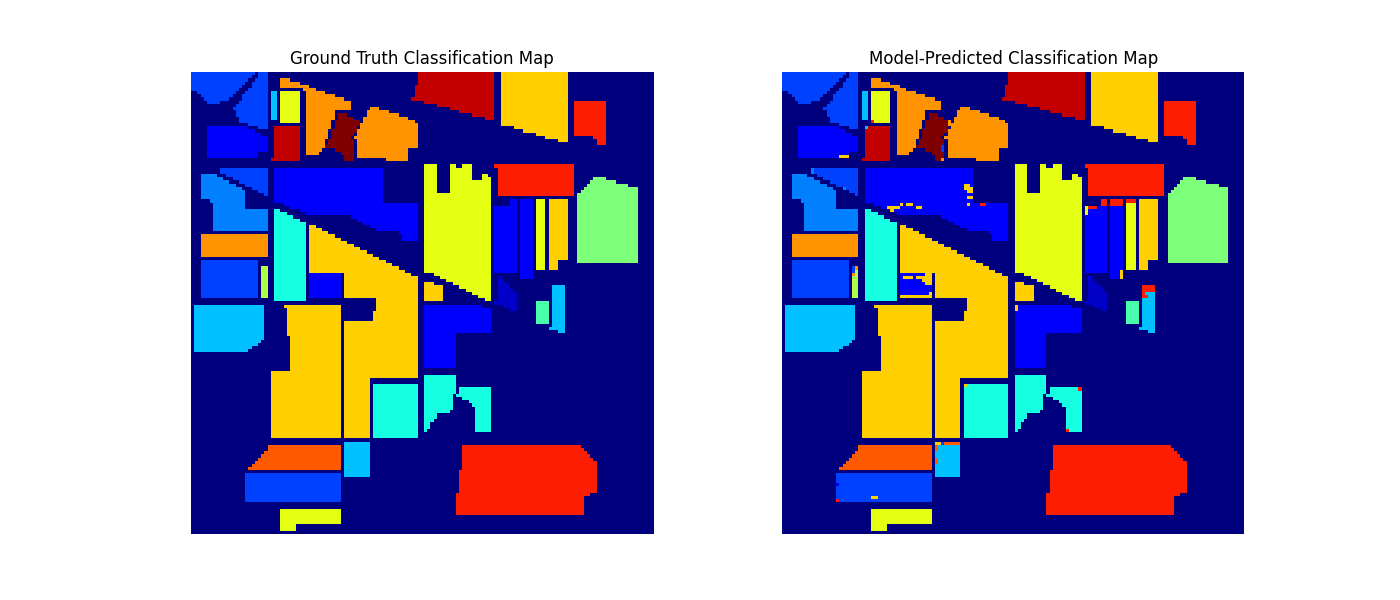}
        \caption{MHA}
    \end{subfigure}
       \begin{subfigure}[b]{0.16\textwidth}
            \centering
          \includegraphics[width=1\linewidth, trim=565 50 120 60, clip, angle=90]{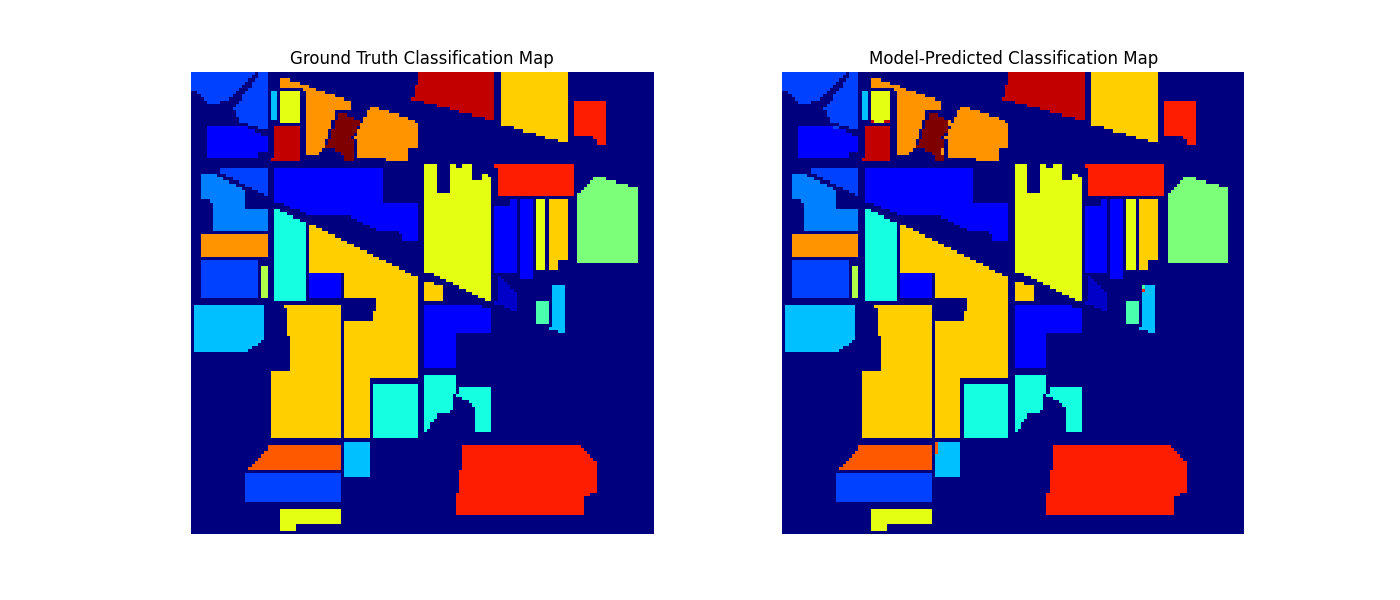}
        \caption{PWA}
    \end{subfigure}
       \begin{subfigure}[b]{0.16\textwidth}
            \centering
          \includegraphics[width=1\linewidth, trim=565 50 120 60, clip, angle=90]{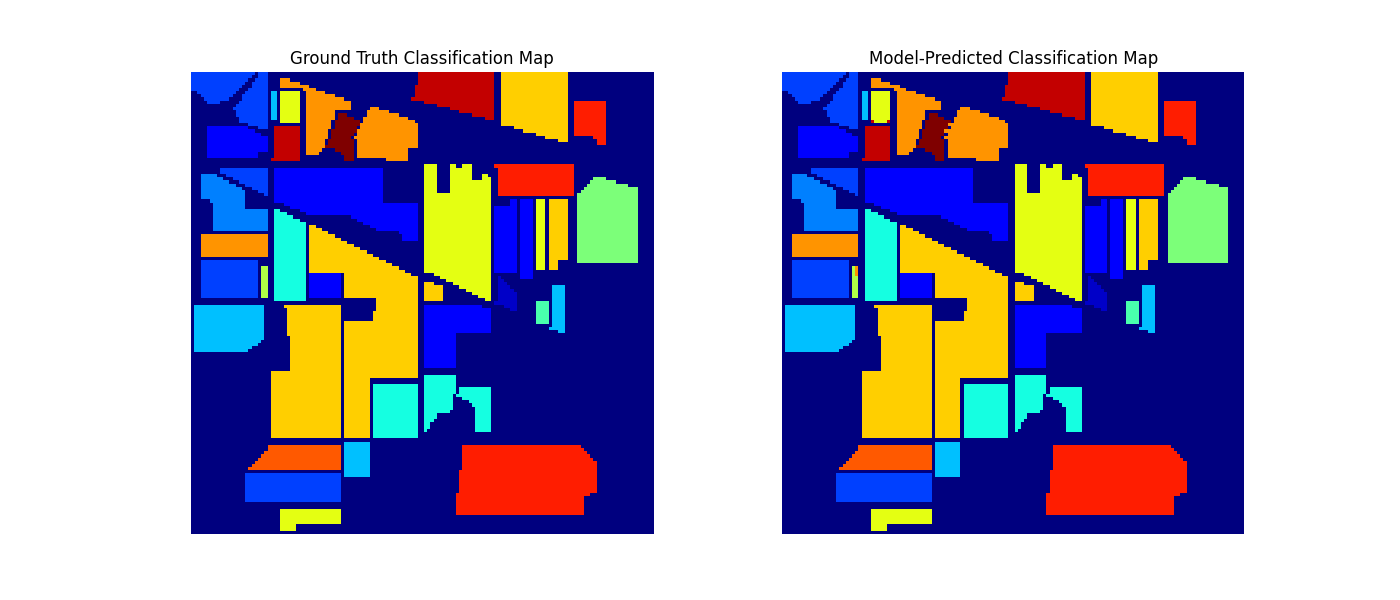}
        \caption{PWM}
    \end{subfigure}
       \begin{subfigure}[b]{0.16\textwidth}
            \centering
          \includegraphics[width=1\linewidth, trim=140 50 565 60, clip, angle=90]{Output_Map/Indian_Pines/FAHM-T5Encoder_Large/PWM_prediction_map_run1.png}
        \caption{GT}
    \end{subfigure}
    \caption{Comparison of classification maps for the FAHM-T5 model on the Indian Pines dataset, showing different fusion methods: Cross Attention (CA), Concatenation (CONCAT), Multi-Head Attention (MHA), Pixel-Wise Addition (PWA), Pixel-Wise Multiplication (PWM), and Ground Truth (GT).}
    \vspace{-3mm}
    \label{fig:FAHM-T5-Indian Pines}
\end{figure*}

%%%%%%%%%%%%%%%%%%%%%%%%%%%%%%%%%%%%%%%%%%%%%%%%%%%%%%%  Indian Pines  FAHM-BertEncoder   %%%%%%%%%%%%%%%%%%%%%%%%%%%%%%%%%%%%%%%%%%%%%%%%%%%%%%%%%%%%%%%%%%%%%
\begin{figure*}[]
    \centering
       \begin{subfigure}[b]{0.16\textwidth}
            \centering
          \includegraphics[width=1\linewidth, trim=565 50 120 60, clip, angle=90]{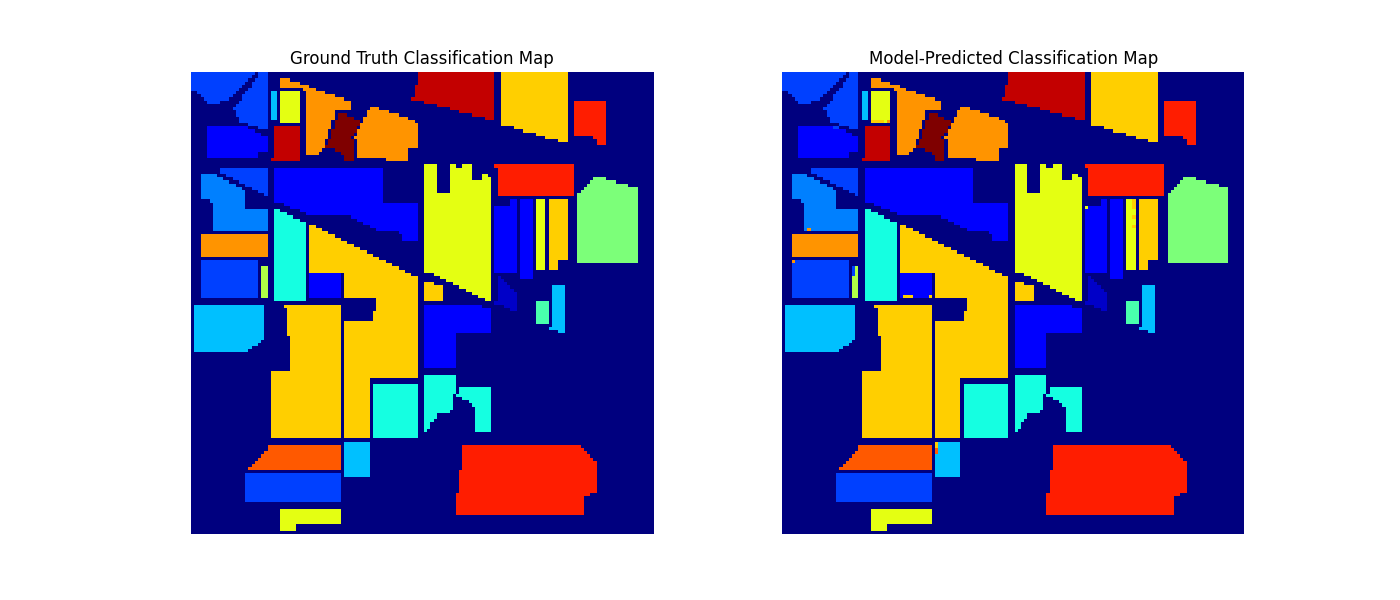}
        \caption{CA}
    \end{subfigure}
       \begin{subfigure}[b]{0.16\textwidth}
            \centering
          \includegraphics[width=1\linewidth, trim=565 50 120 60, clip, angle=90]{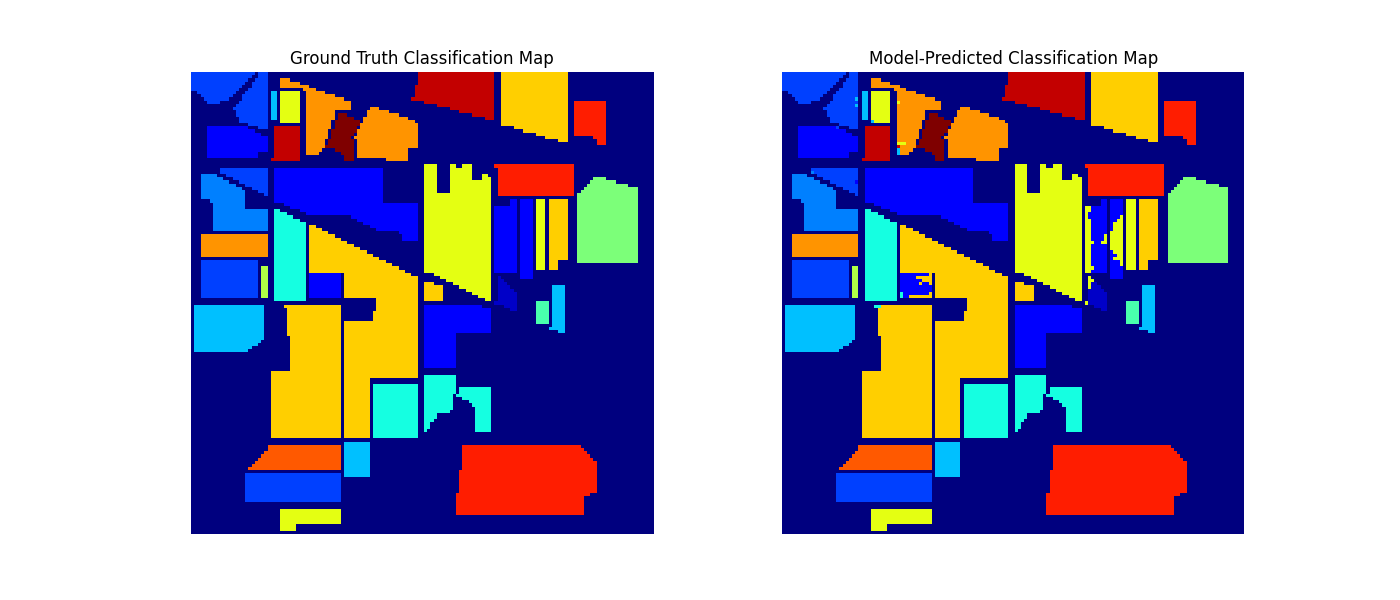}
        \caption{CONCAT}
    \end{subfigure}
       \begin{subfigure}[b]{0.16\textwidth}
            \centering
          \includegraphics[width=1\linewidth, trim=565 50 120 60, clip, angle=90]{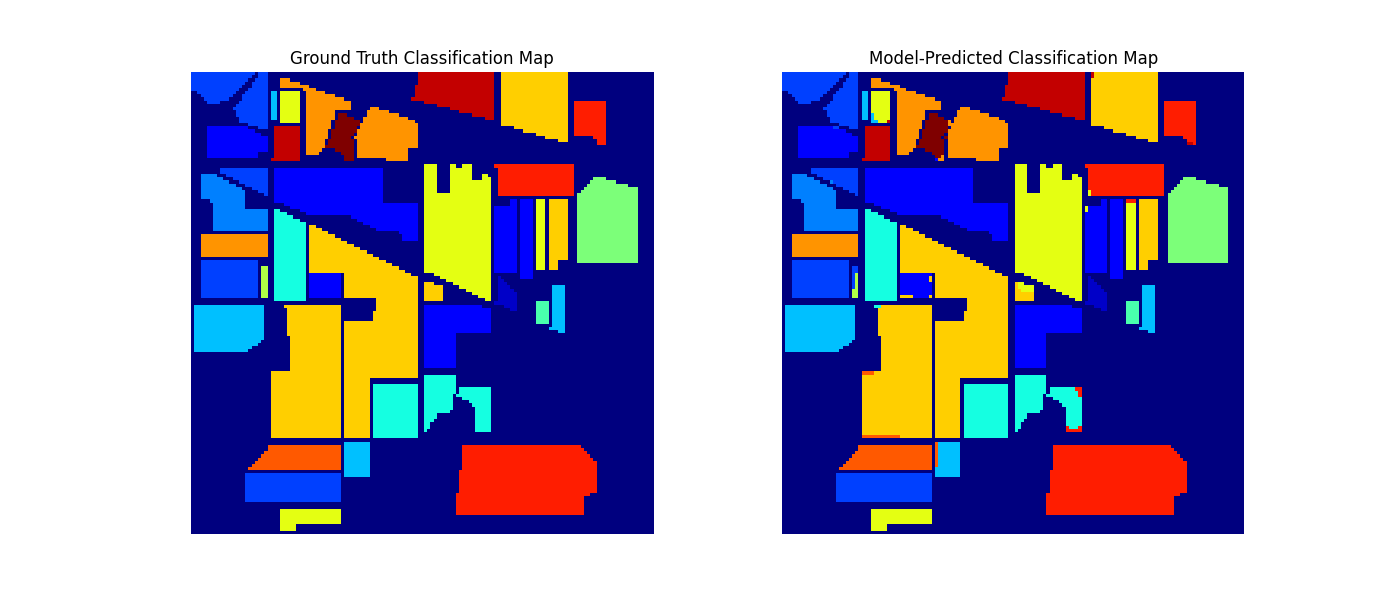}
        \caption{MHA}
    \end{subfigure}
       \begin{subfigure}[b]{0.16\textwidth}
            \centering
          \includegraphics[width=1\linewidth, trim=565 50 120 60, clip, angle=90]{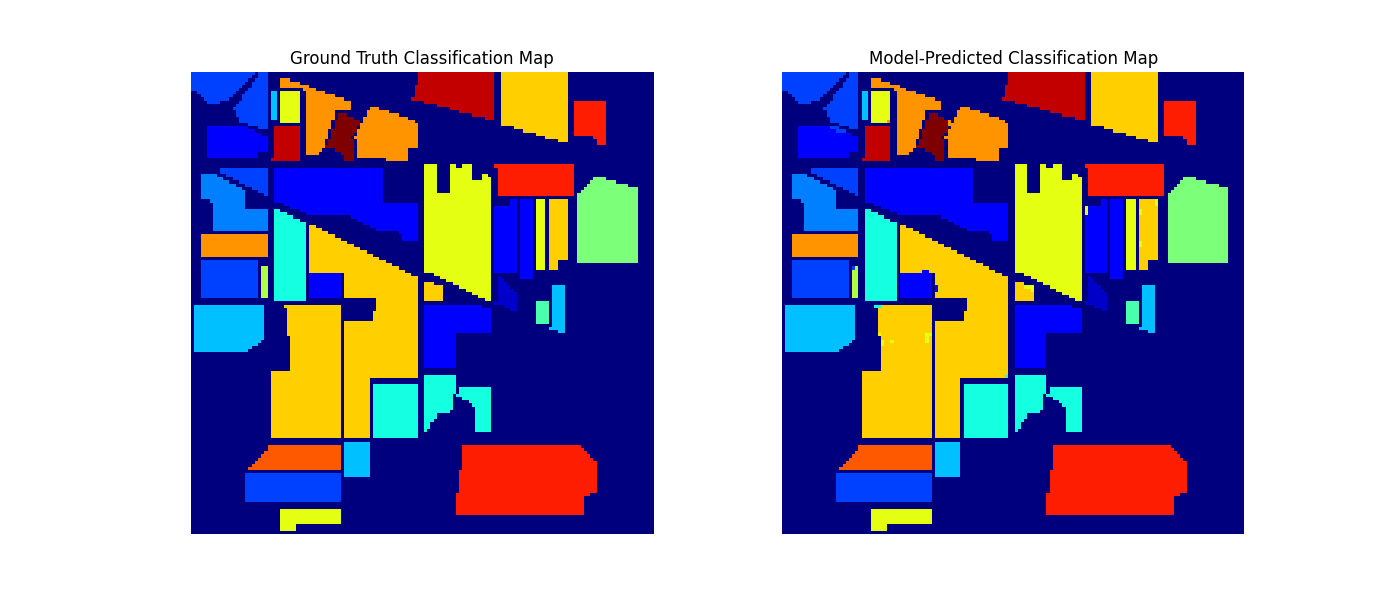}
        \caption{PWA}
    \end{subfigure}
       \begin{subfigure}[b]{0.16\textwidth}
            \centering
          \includegraphics[width=1\linewidth, trim=565 50 120 60, clip, angle=90]{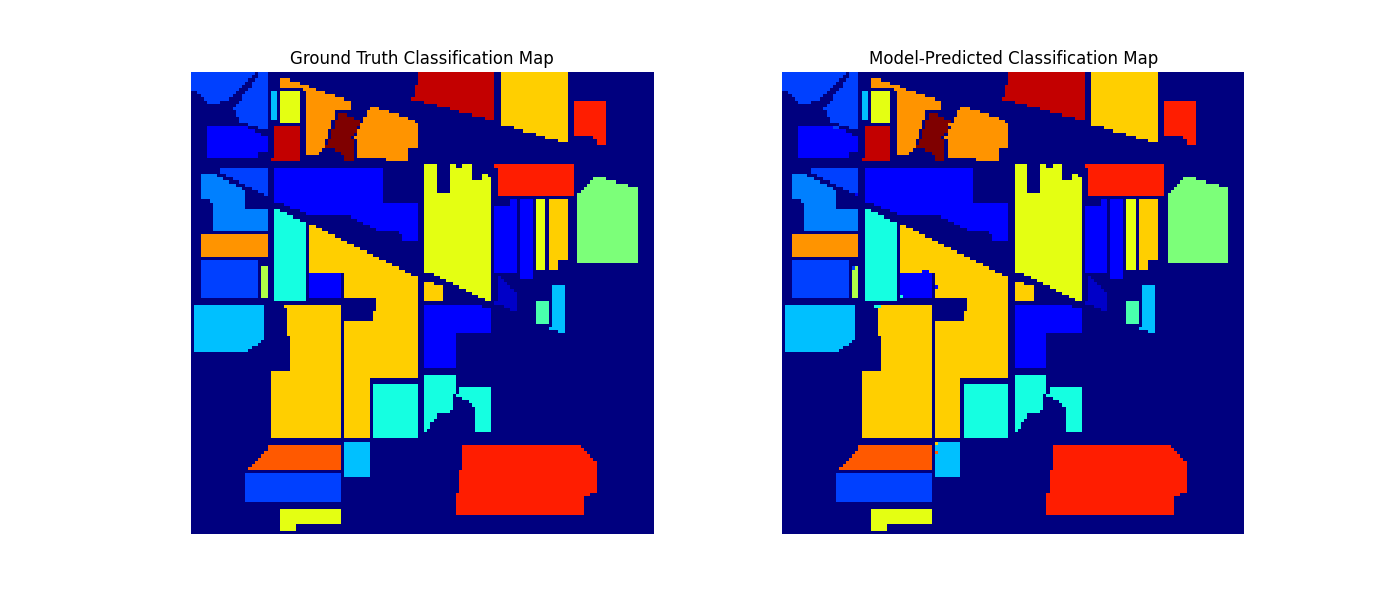}
        \caption{PWM}
    \end{subfigure}
       \begin{subfigure}[b]{0.16\textwidth}
            \centering
          \includegraphics[width=1\linewidth, trim=140 50 565 60, clip, angle=90]{Output_Map/Indian_Pines/FAHM-BertEncoder_Large/PWM_prediction_map_run1.png}
        \caption{GT}
    \end{subfigure}
    \caption{Comparison of classification maps for the FAHM-Bert model on the Indian Pines dataset, showing different fusion methods: Cross Attention (CA), Concatenation (CONCAT), Multi-Head Attention (MHA), Pixel-Wise Addition (PWA), Pixel-Wise Multiplication (PWM), and Ground Truth (GT).}
    \vspace{-3mm}
    \label{fig:FAHM-Bert-Indian Pines}
\end{figure*}

Indian Pines is where weaker models (3D-RCNet, DBCTNet) visually benefit the most from language encoder fusion: vision-only maps are very speckled, but text fusion results in clear, accurate class boundaries and large uniform regions. The best backbones (3D-ConvSST, FAHM), when paired with strong fusions, produce visuals with high fidelity to ground truth, even at thin field boundaries and tiny land fragments. Attention-based fusions and concatenation are more robust, controlling mixed-pixel noise and preserving fine map details. Occasionally, pixel-wise fusions may rival top methods for certain architectures, but the figures (\ref{fig:3D-RCNet-T5-Indian Pines}, \ref{fig:3D-RCNet-Bert-Indian Pines}, \ref{fig:3D-ConvSST-T5-Indian Pines}, \ref{fig:3D-ConvSST-Bert-Indian Pines}, \ref{fig:DBCTNet-T5-Indian Pines}, \ref{fig:DBCTNet-Bert-Indian Pines}, \ref{fig:FAHM-T5-Indian Pines}, \ref{fig:FAHM-Bert-Indian Pines}) show that attention and concat fusion are more universally reliable for visual clarity.

% ============================================================================
% KSC DATASET
% ============================================================================

%%%%%%%%%%%%%%%%%%%%%%%%%%%%%%%% KSC 3DRCNet-T5Encoder %%%%%%%%%%%%%%%%%%%%%%%%%%%%%%%%%%%%%%%%%%%%%%%%
\begin{figure*}[]
    \centering
       \begin{subfigure}[b]{0.16\textwidth}
            \centering
          \includegraphics[width=1\linewidth, trim=575 70 120 70, clip, angle=90, clip, angle=90]{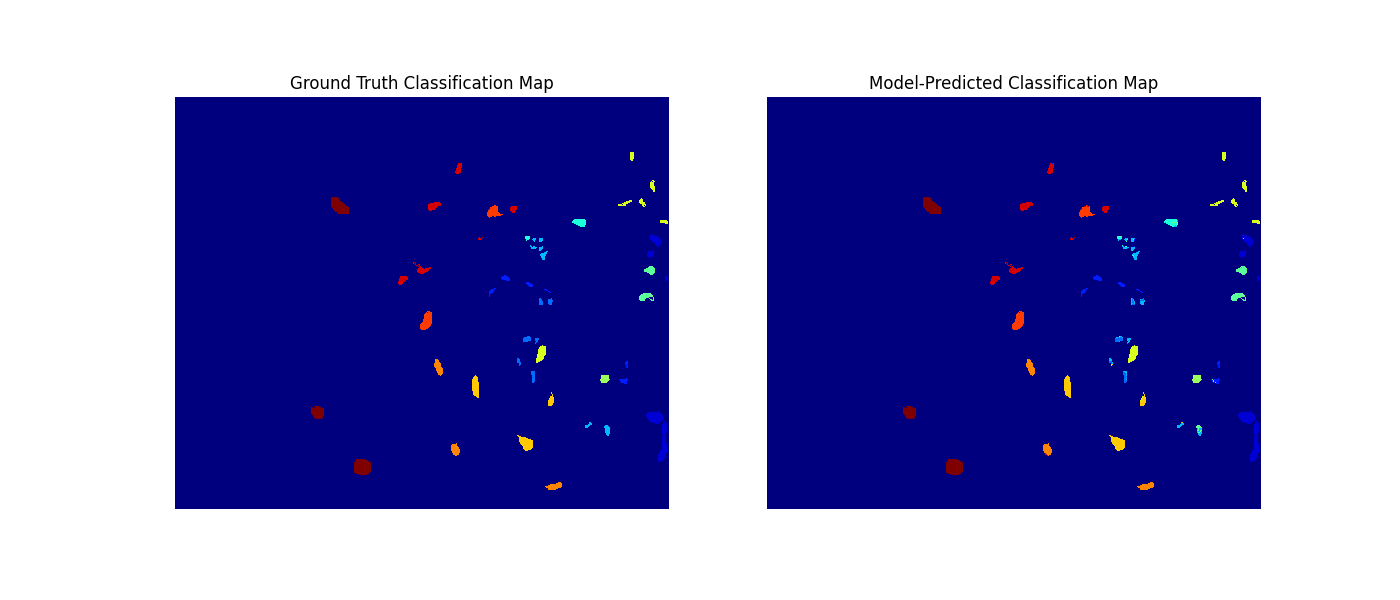}
        \caption{CA}
    \end{subfigure}
       \begin{subfigure}[b]{0.16\textwidth}
            \centering
          \includegraphics[width=1\linewidth, trim=575 70 120 70, clip, angle=90, clip, angle=90]{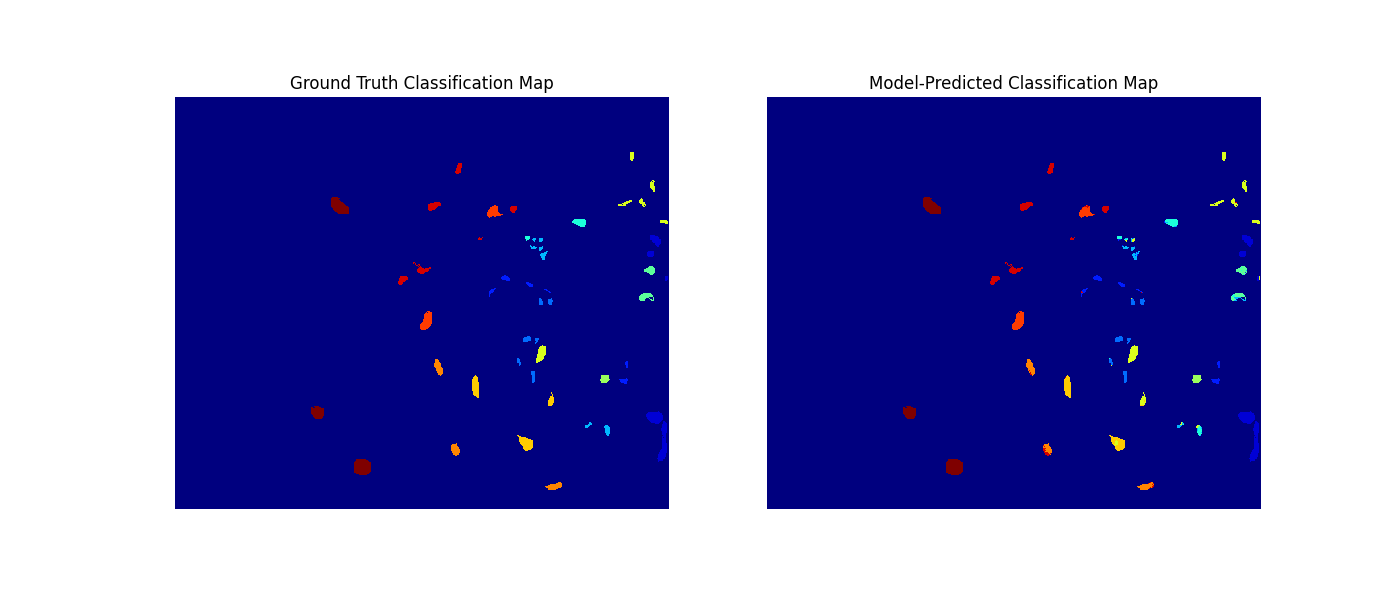}
        \caption{CONCAT}
    \end{subfigure}
       \begin{subfigure}[b]{0.16\textwidth}
            \centering
          \includegraphics[width=1\linewidth, trim=575 70 120 70, clip, angle=90, clip, angle=90]{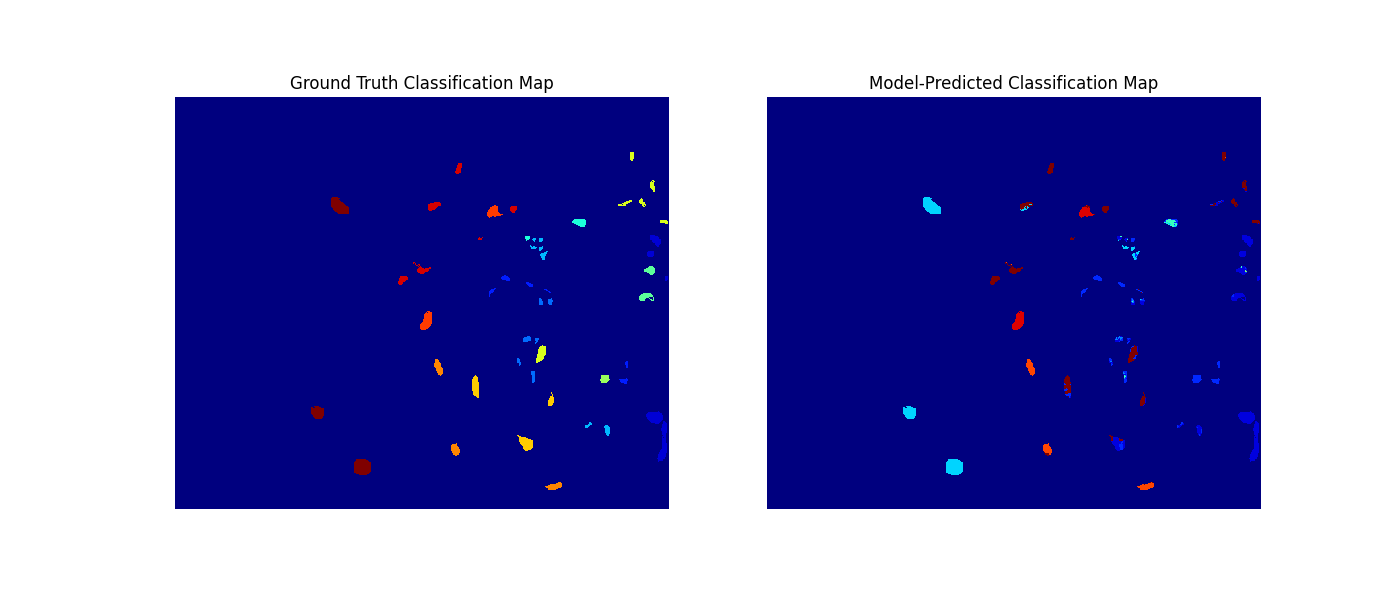}
        \caption{MHA}
    \end{subfigure}
       \begin{subfigure}[b]{0.16\textwidth}
            \centering
          \includegraphics[width=1\linewidth, trim=575 70 120 70, clip, angle=90, clip, angle=90]{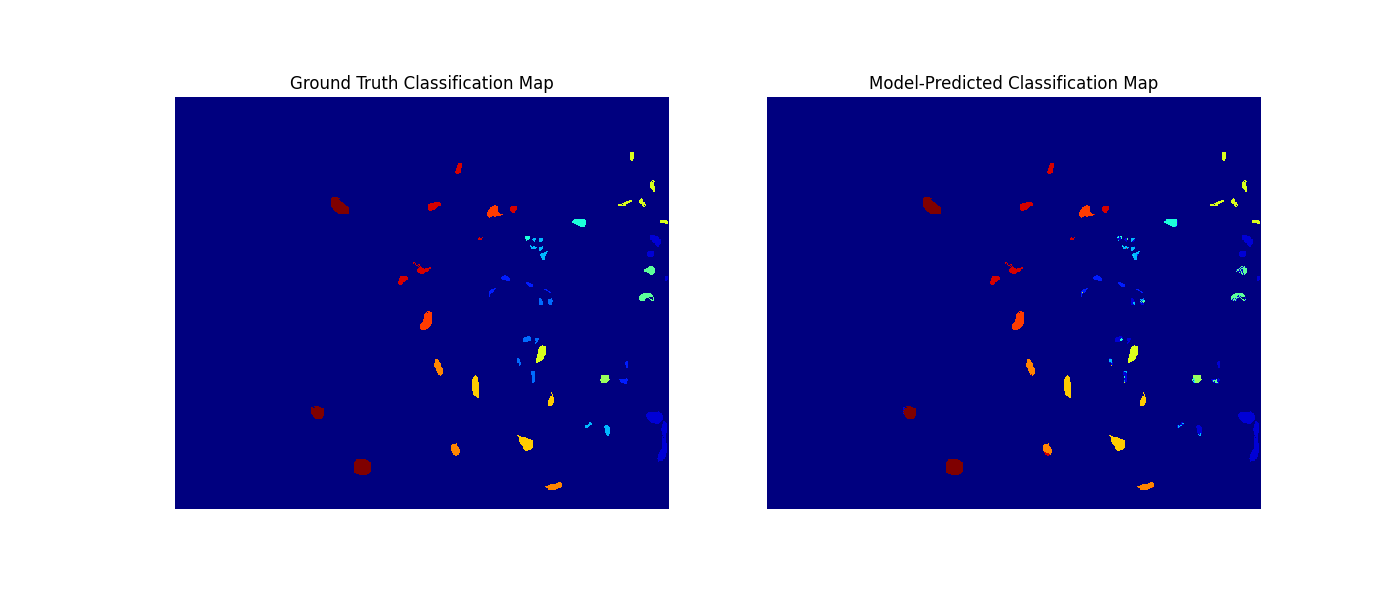}
        \caption{PWA}
    \end{subfigure}
       \begin{subfigure}[b]{0.16\textwidth}
            \centering
          \includegraphics[width=1\linewidth, trim=575 70 120 70, clip, angle=90, clip, angle=90]{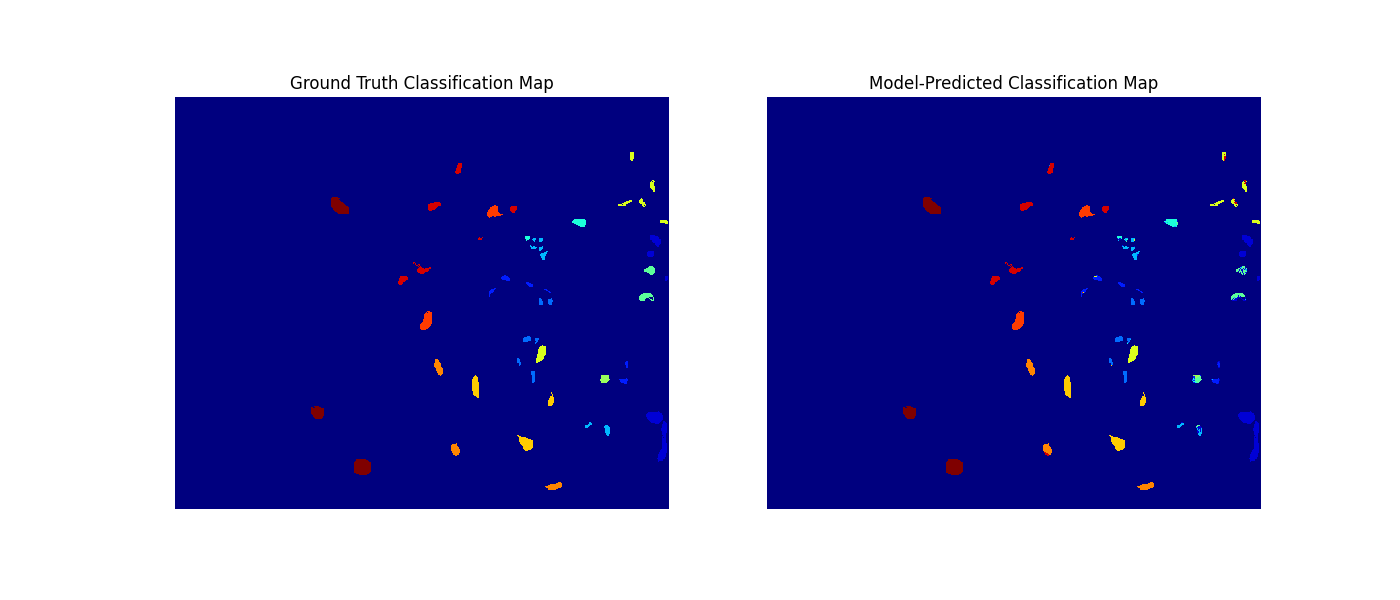}
        \caption{PWM}
    \end{subfigure}
       \begin{subfigure}[b]{0.16\textwidth}
            \centering
          \includegraphics[height=0.94\textwidth, width=1\linewidth,trim=130 70 540 70, clip, angle=180]{Output_Map/KSC/3DRCNet-T5Encoder_Large/PWM_prediction_map_run1.png}
        \caption{GT}
    \end{subfigure}
    \caption{Comparison of classification maps for the 3D-RCNet-T5 model on the KSC dataset, showing different fusion methods: Cross Attention (CA), Concatenation (CONCAT), Multi-Head Attention (MHA), Pixel-Wise Addition (PWA), Pixel-Wise Multiplication (PWM), and Ground Truth (GT).}
    \vspace{-3mm}
    \label{fig:3D-RCNet-T5-KSC}
\end{figure*}

%%%%%%%%%%%%%%%%%%%%%%%%%%%%%%%% KSC 3DRCNet-BertEncoder %%%%%%%%%%%%%%%%%%%%%%%%%%%%%%%%%%%%%%%%%%%%%%%%
\begin{figure*}[]
    \centering
       \begin{subfigure}[b]{0.16\textwidth}
            \centering
          \includegraphics[width=1\linewidth, trim=575 70 120 70, clip, angle=90, clip, angle=90]{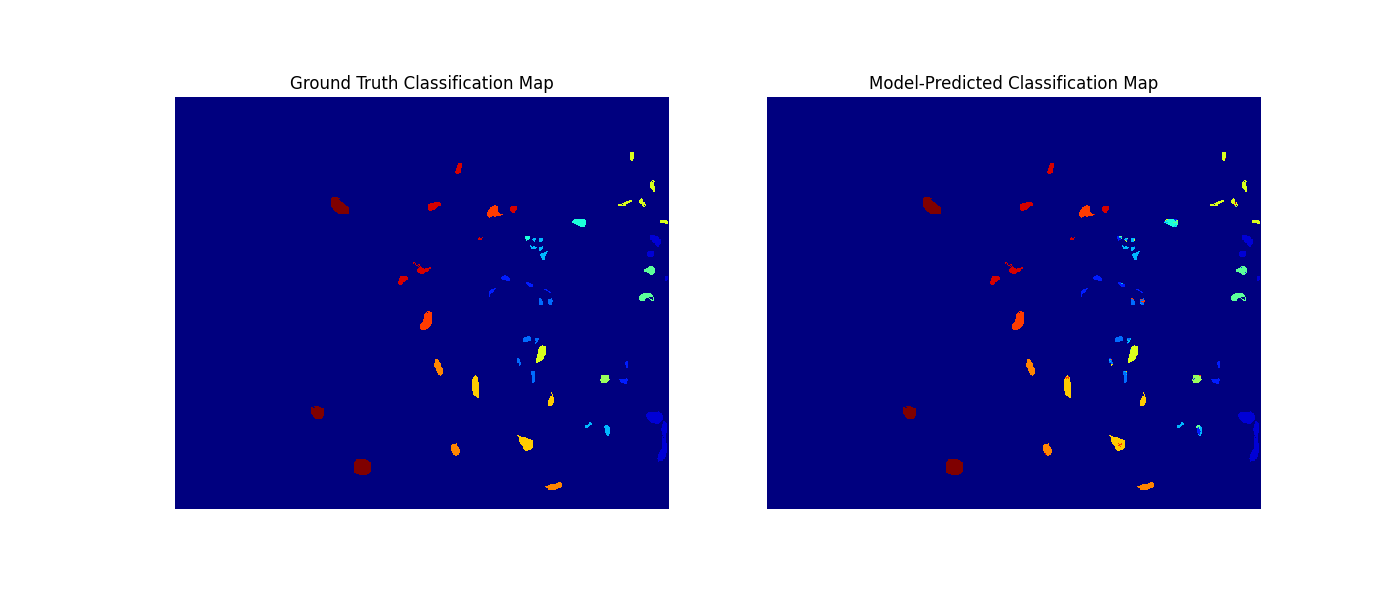}
        \caption{CA}
    \end{subfigure}
       \begin{subfigure}[b]{0.16\textwidth}
            \centering
          \includegraphics[width=1\linewidth, trim=575 70 120 70, clip, angle=90, clip, angle=90]{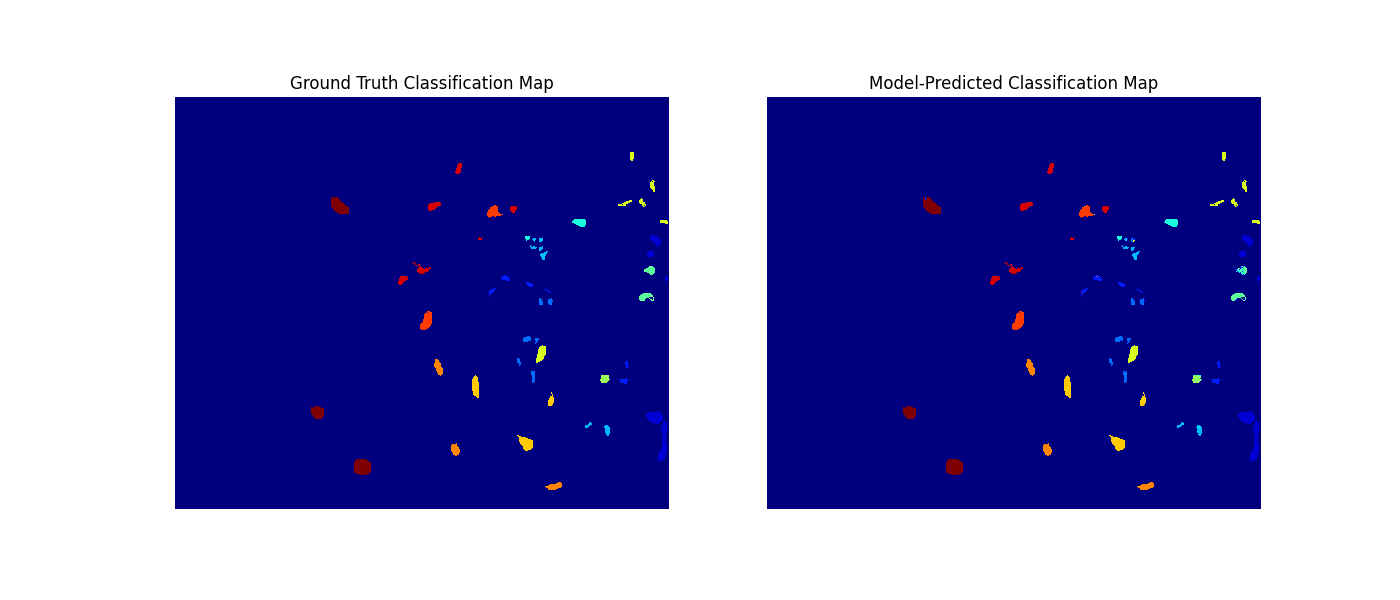}
        \caption{CONCAT}
    \end{subfigure}
       \begin{subfigure}[b]{0.16\textwidth}
            \centering
          \includegraphics[width=1\linewidth, trim=575 70 120 70, clip, angle=90, clip, angle=90]{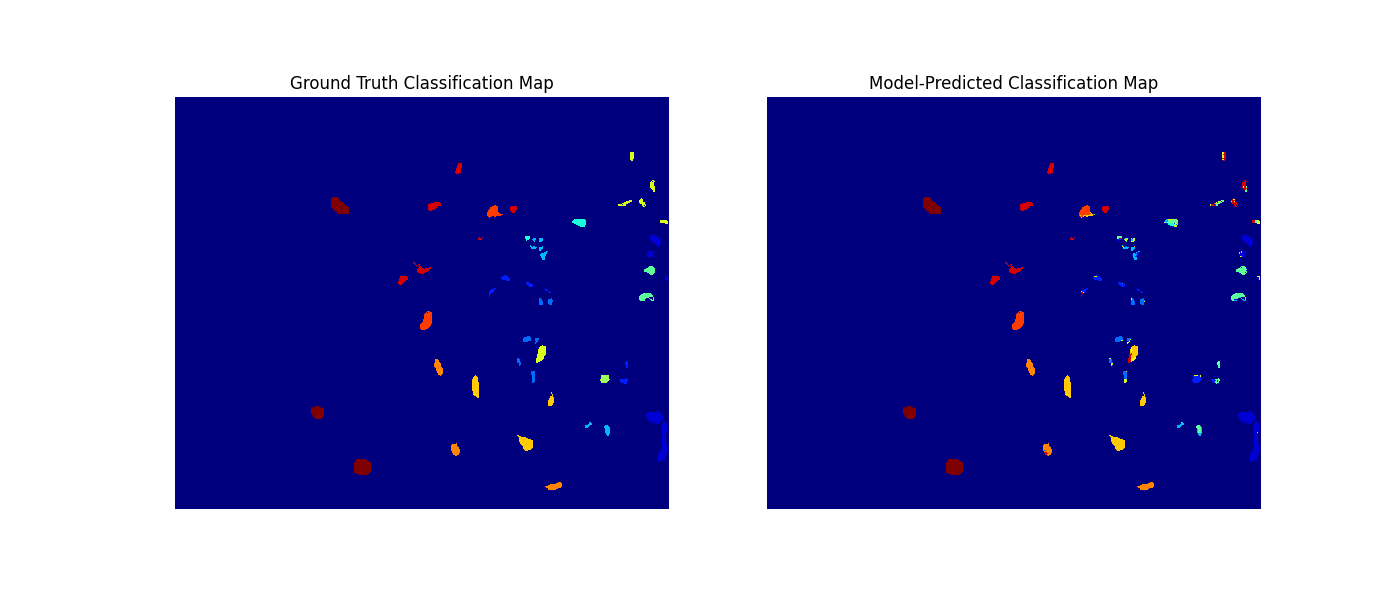}
        \caption{MHA}
    \end{subfigure}
       \begin{subfigure}[b]{0.16\textwidth}
            \centering
          \includegraphics[width=1\linewidth, trim=575 70 120 70, clip, angle=90, clip, angle=90]{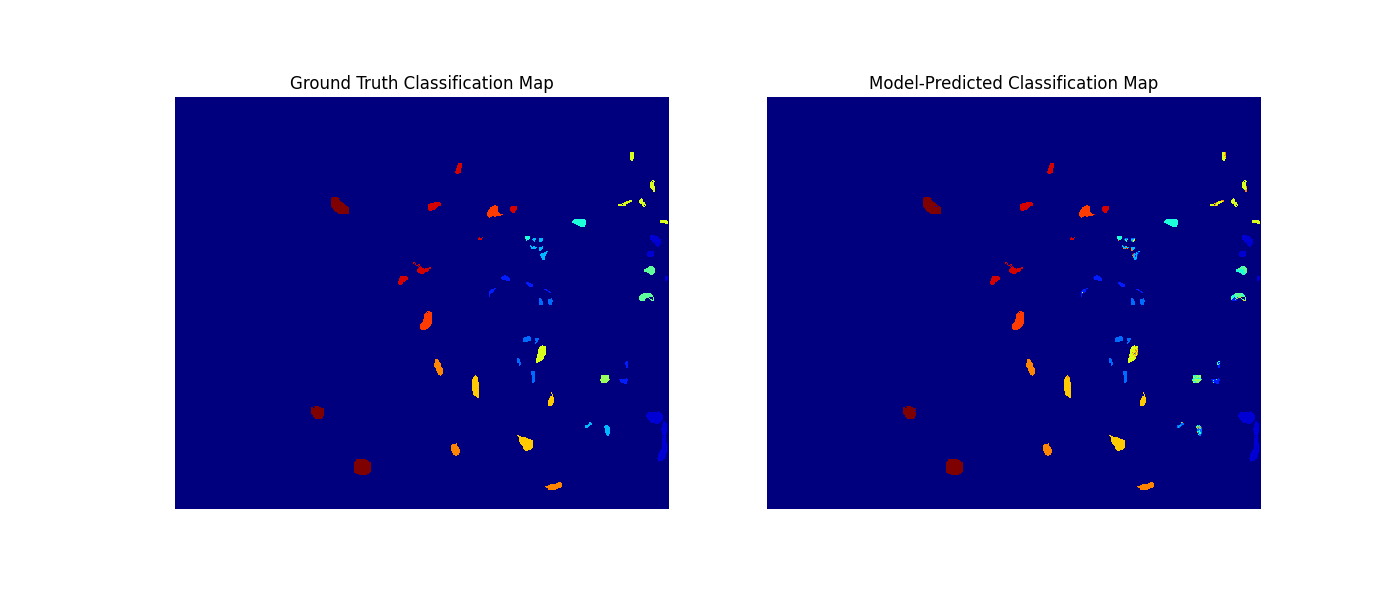}
        \caption{PWA}
    \end{subfigure}
       \begin{subfigure}[b]{0.16\textwidth}
            \centering
          \includegraphics[width=1\linewidth, trim=575 70 120 70, clip, angle=90, clip, angle=90]{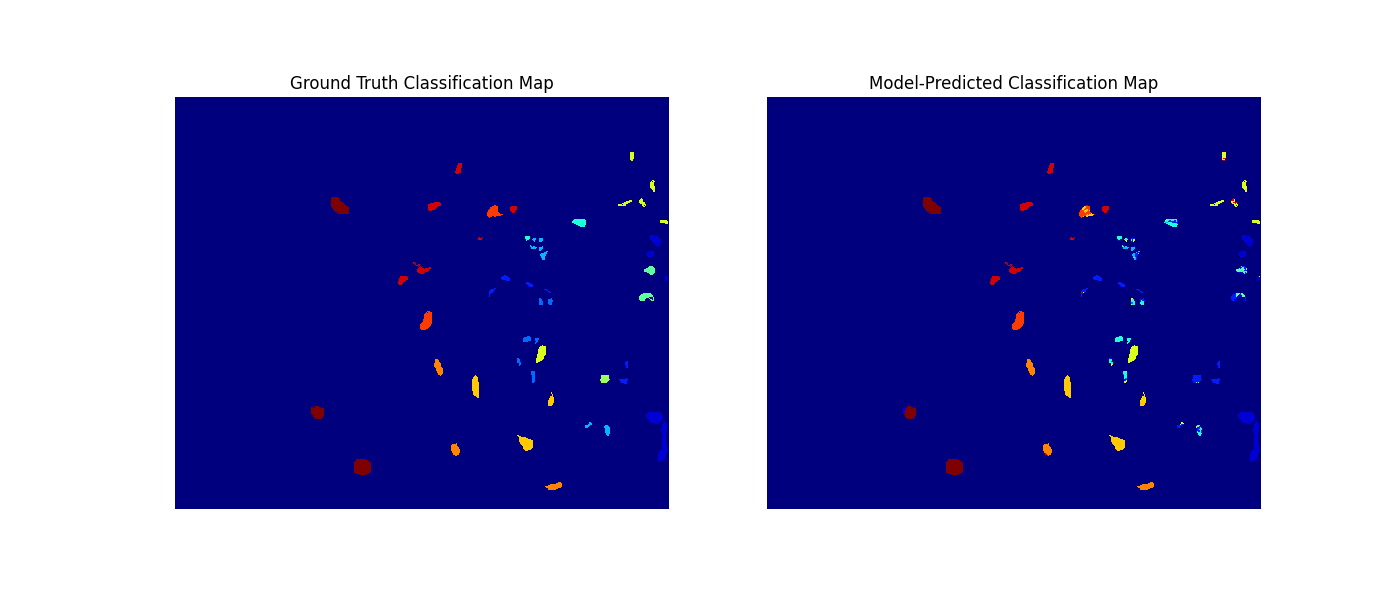}
        \caption{PWM}
    \end{subfigure}
       \begin{subfigure}[b]{0.16\textwidth}
            \centering
          \includegraphics[height=0.94\textwidth, width=1\linewidth, trim=130 70 540 70, clip, angle=180]{Output_Map/KSC/3DRCNet-BertEncoder_Large/PWM_prediction_map_run1.png}
        \caption{GT}
    \end{subfigure}
    \caption{Comparison of classification maps for the 3D-RCNet-Bert model on the KSC dataset, showing different fusion methods: Cross Attention (CA), Concatenation (CONCAT), Multi-Head Attention (MHA), Pixel-Wise Addition (PWA), Pixel-Wise Multiplication (PWM), and Ground Truth (GT).}
    \vspace{-3mm}
    \label{fig:3D-RCNet-Bert-KSC}
\end{figure*}

%%%%%%%%%%%%%%%%%%%%%%%%%%%%%%%% KSC 3D_ConvSST-T5Encoder %%%%%%%%%%%%%%%%%%%%%%%%%%%%%%%%%%%%%%%%%%%%%%%%
\begin{figure*}[t]
    \centering
       \begin{subfigure}[b]{0.16\textwidth}
            \centering
          \includegraphics[width=1\linewidth, trim=575 70 120 70, clip, angle=90, clip, angle=90]{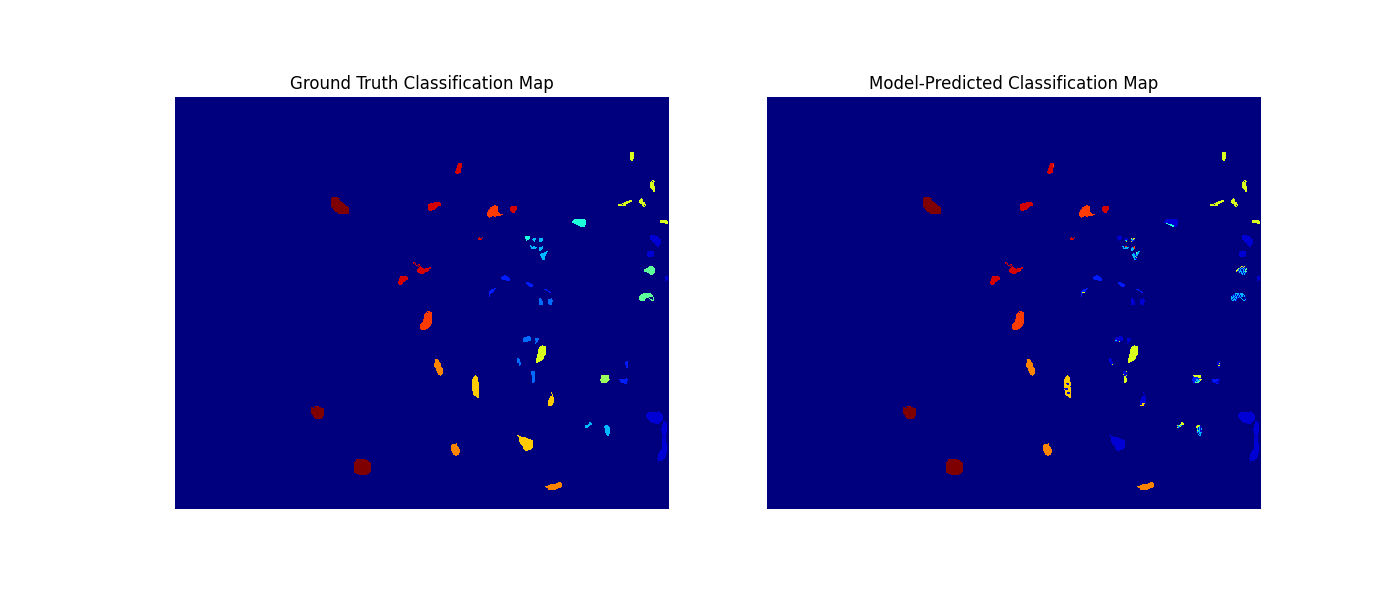}
        \caption{CA}
    \end{subfigure}
       \begin{subfigure}[b]{0.16\textwidth}
            \centering
          \includegraphics[width=1\linewidth,trim=575 70 120 70, clip, angle=90, clip, angle=90]{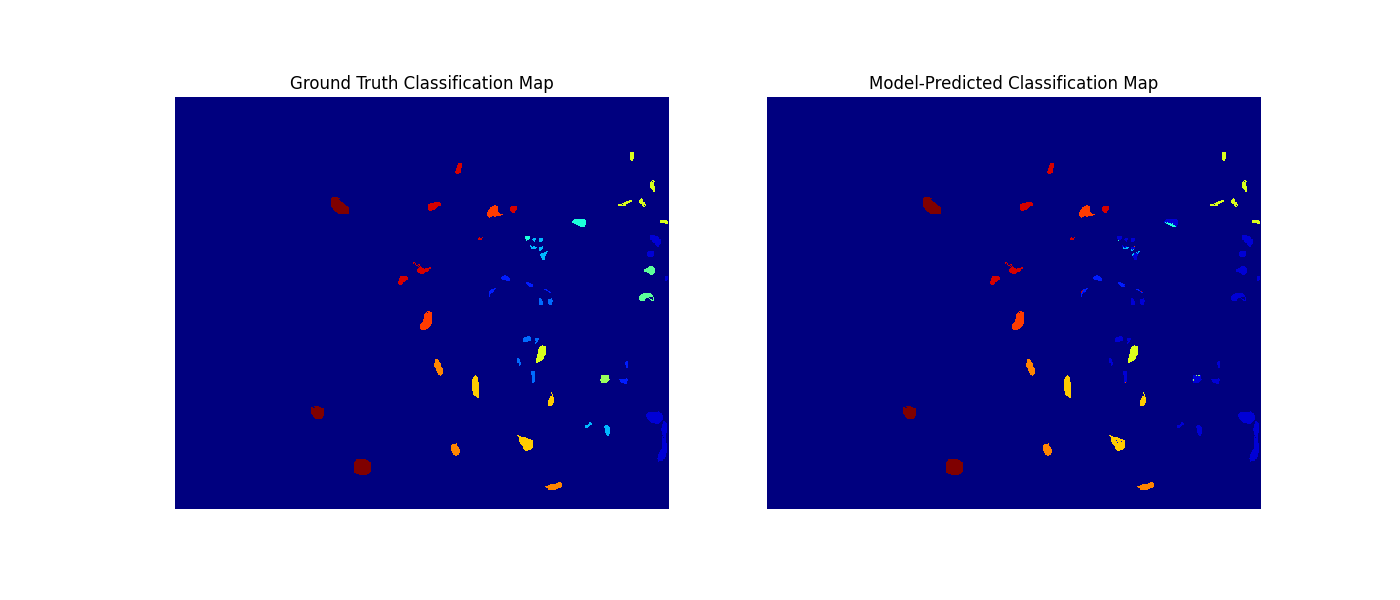}
        \caption{CONCAT}
    \end{subfigure}
       \begin{subfigure}[b]{0.16\textwidth}
            \centering
          \includegraphics[width=1\linewidth, trim=575 70 120 70, clip, angle=90, clip, angle=90]{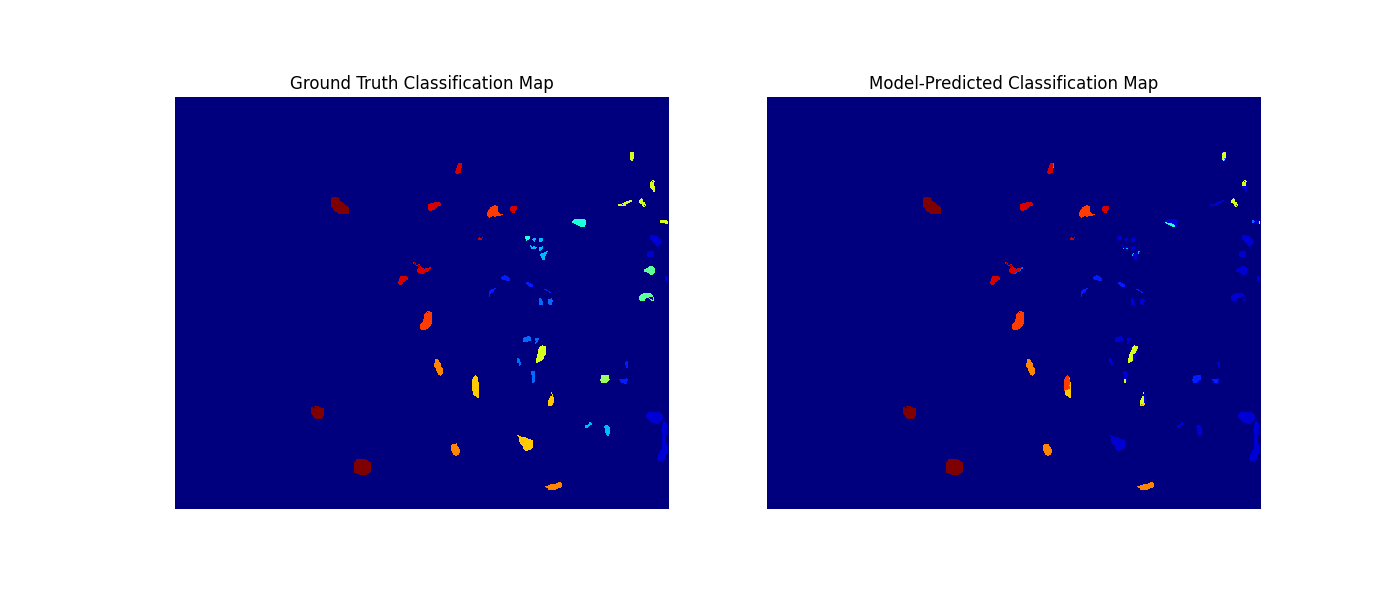}
        \caption{MHA}
    \end{subfigure}
       \begin{subfigure}[b]{0.16\textwidth}
            \centering
          \includegraphics[width=1\linewidth, trim=575 70 120 70, clip, angle=90, clip, angle=90]{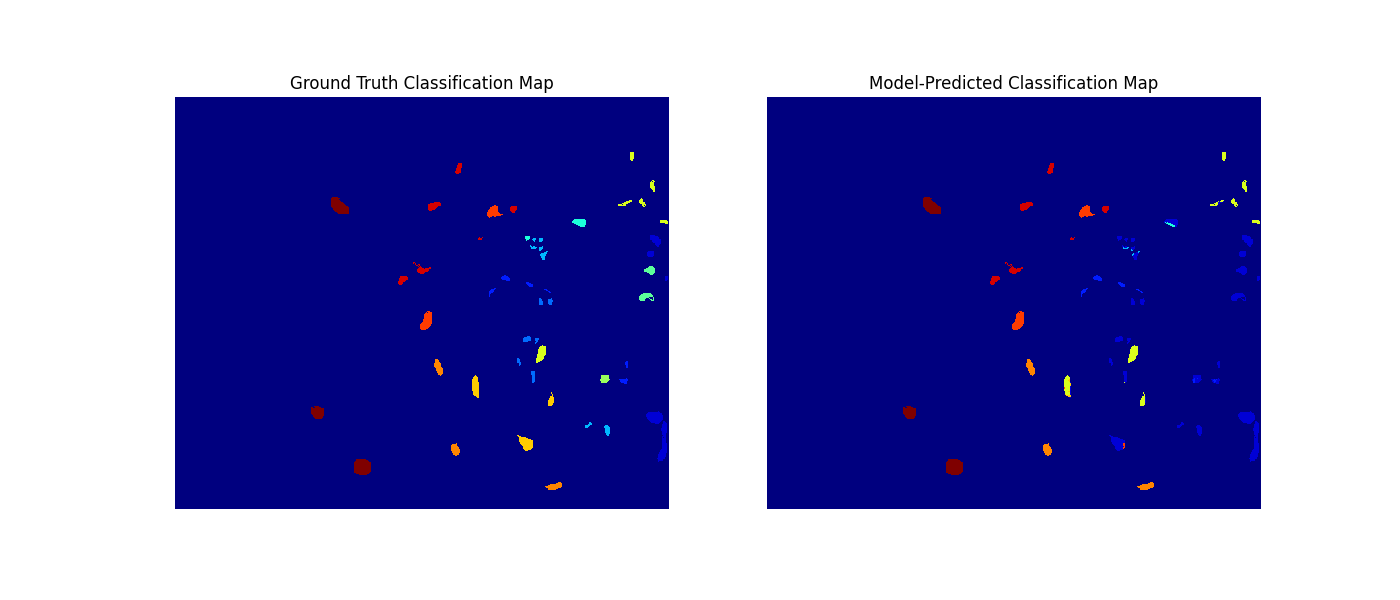}
        \caption{PWA}
    \end{subfigure}
       \begin{subfigure}[b]{0.16\textwidth}
            \centering
          \includegraphics[width=1\linewidth, trim=575 70 120 70, clip, angle=90, clip, angle=90]{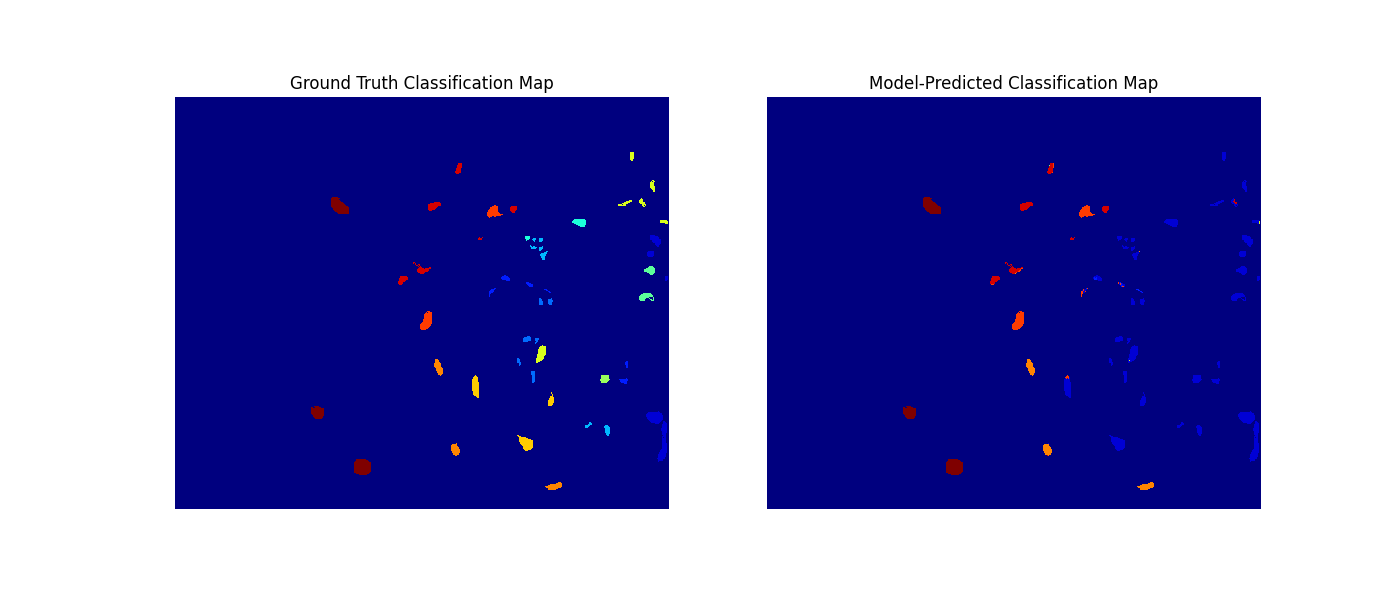}
        \caption{PWM}
    \end{subfigure}
       \begin{subfigure}[b]{0.16\textwidth}
            \centering
          \includegraphics[height=0.94\textwidth, width=1\linewidth, trim=130 70 540 70, clip, angle=180]{Output_Map/KSC/3D_ConvSST-T5Encoder_Large/PWM_prediction_map_run1.png}
        \caption{GT}
    \end{subfigure}
    \caption{Comparison of classification maps for the 3D-ConvSST-T5 model on the KSC dataset, showing different fusion methods: Cross Attention (CA), Concatenation (CONCAT), Multi-Head Attention (MHA), Pixel-Wise Addition (PWA), Pixel-Wise Multiplication (PWM), and Ground Truth (GT).}
    \vspace{-3mm}
    \label{fig:3D-ConvSST-T5-KSC}
\end{figure*}

%%%%%%%%%%%%%%%%%%%%%%%%%%%%%%%% KSC 3D_ConvSST-BertEncoder %%%%%%%%%%%%%%%%%%%%%%%%%%%%%%%%%%%%%%%%%%%%%%%%
\begin{figure*}[]
    \centering
       \begin{subfigure}[b]{0.16\textwidth}
            \centering
          \includegraphics[width=1\linewidth, trim=575 70 120 70, clip, angle=90, clip, angle=90]{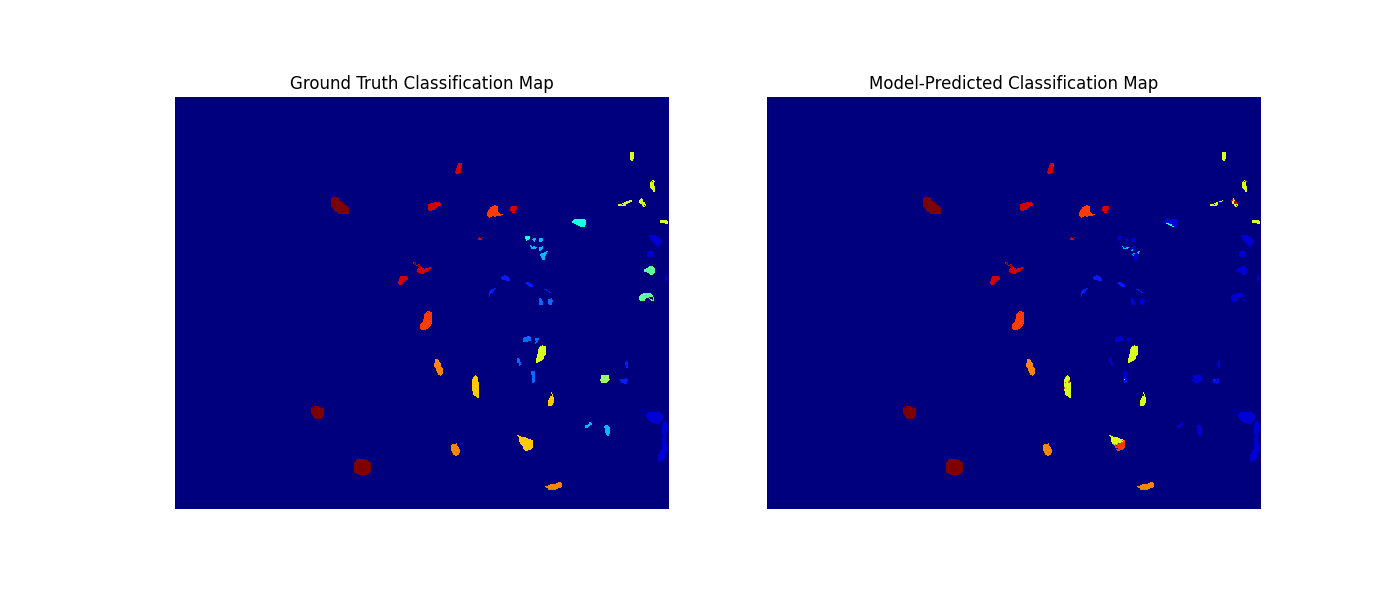}
        \caption{CA}
    \end{subfigure}
       \begin{subfigure}[b]{0.16\textwidth}
            \centering
          \includegraphics[width=1\linewidth, trim=575 70 120 70, clip, angle=90, clip, angle=90]{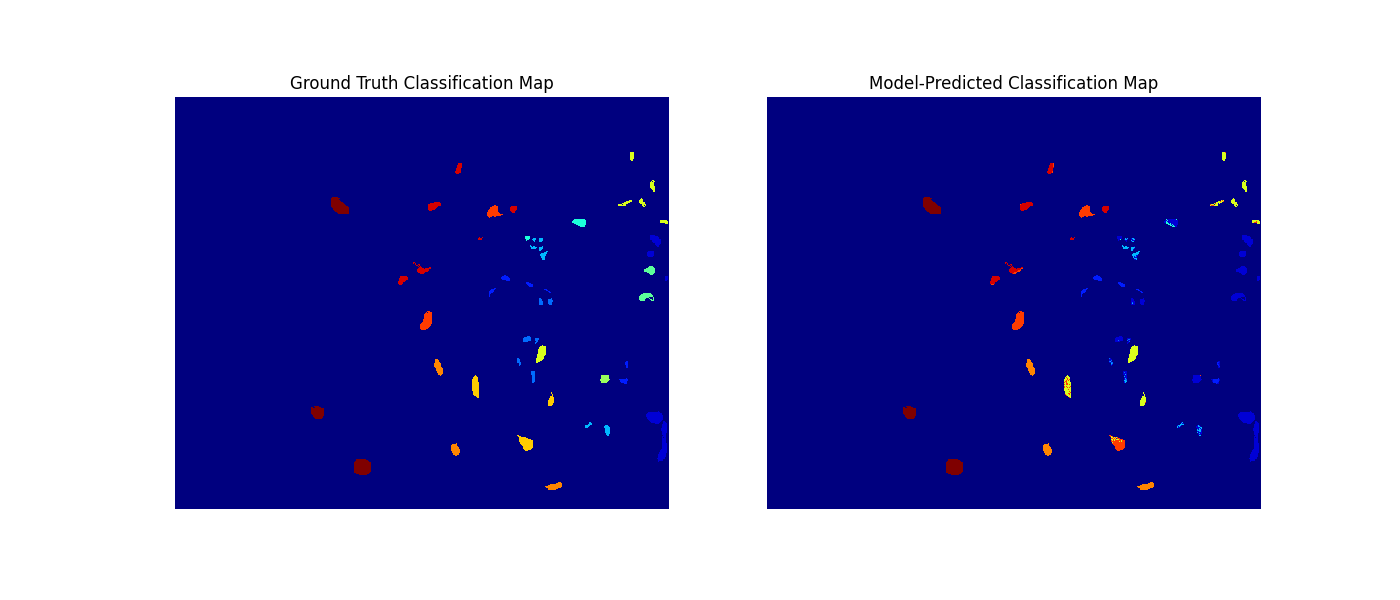}
        \caption{CONCAT}
    \end{subfigure}
       \begin{subfigure}[b]{0.16\textwidth}
            \centering
          \includegraphics[width=1\linewidth, trim=575 70 120 70, clip, angle=90, clip, angle=90]{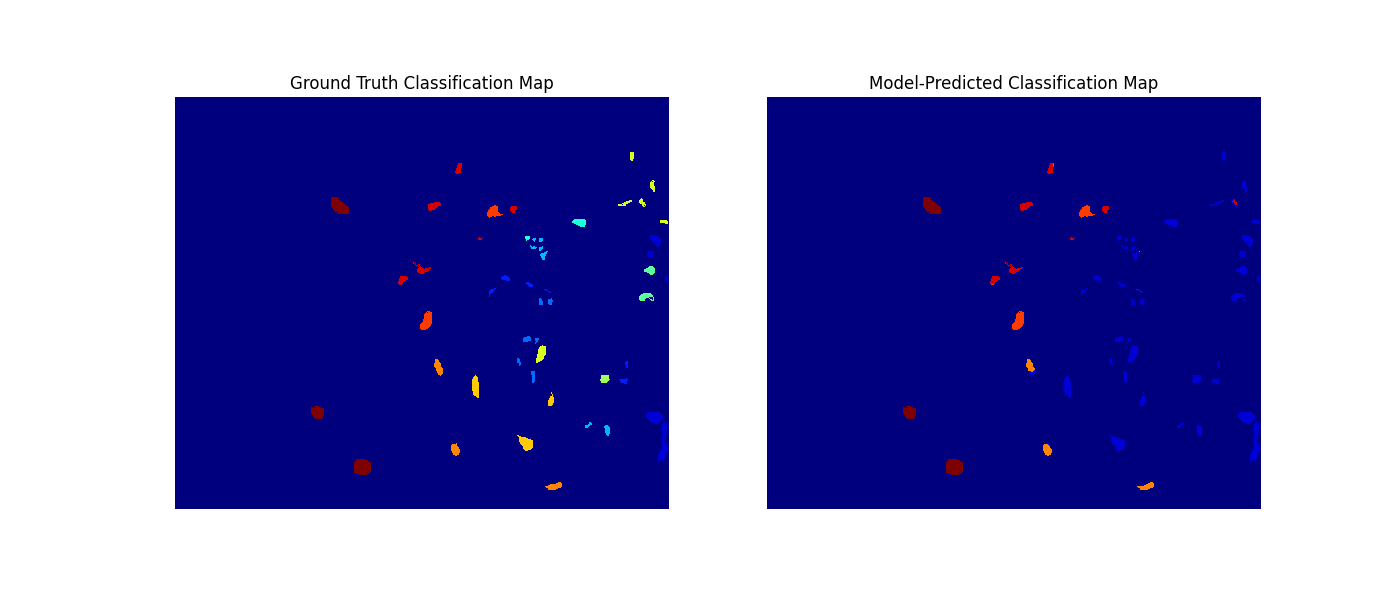}
        \caption{MHA}
    \end{subfigure}
       \begin{subfigure}[b]{0.16\textwidth}
            \centering
          \includegraphics[width=1\linewidth, trim=575 70 120 70, clip, angle=90, clip, angle=90]{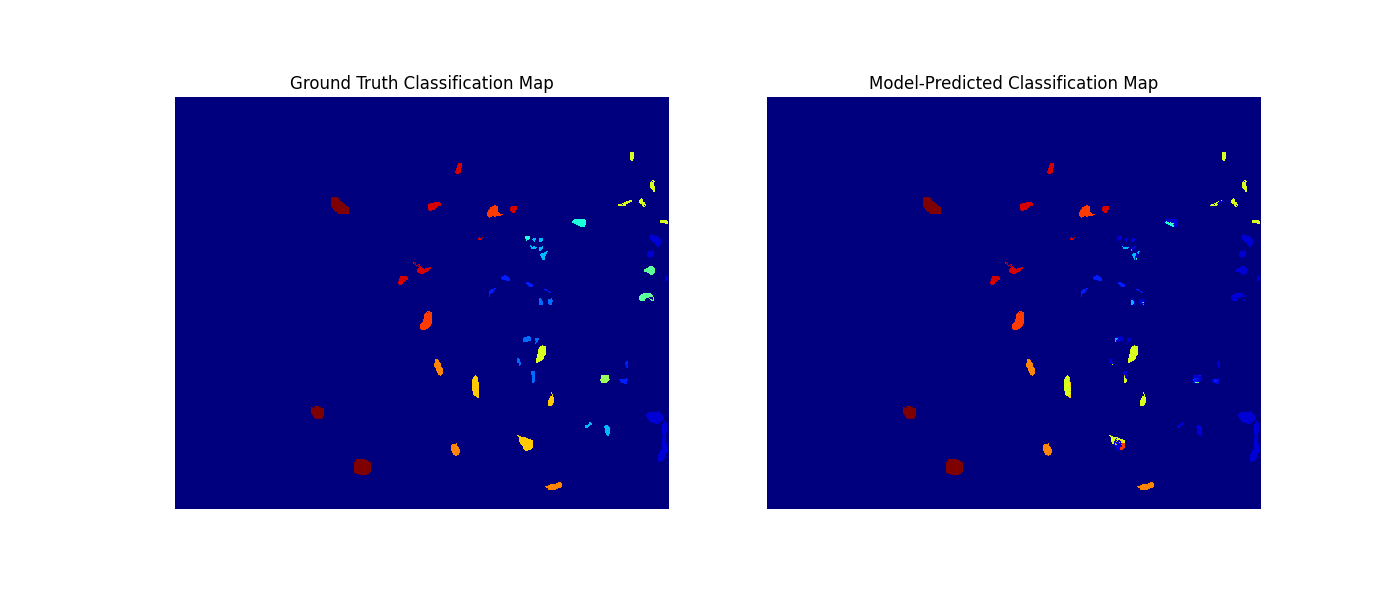}
        \caption{PWA}
    \end{subfigure}
       \begin{subfigure}[b]{0.16\textwidth}
            \centering
          \includegraphics[width=1\linewidth, trim=575 70 120 70, clip, angle=90, clip, angle=90]{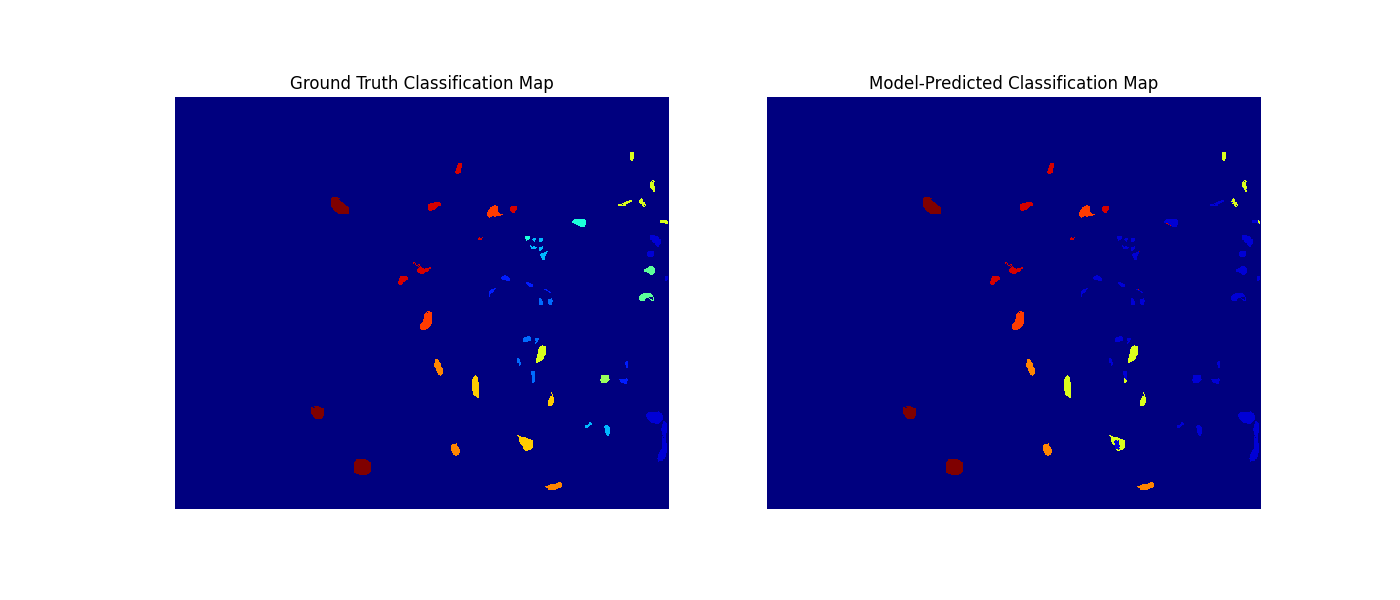}
        \caption{PWM}
    \end{subfigure}
       \begin{subfigure}[b]{0.16\textwidth}
            \centering
          \includegraphics[height=0.94\textwidth, width=1\linewidth, trim=130 70 540 70, clip, angle=180]{Output_Map/KSC/3D_ConvSST-BertEncoder_Large/PWM_prediction_map_run1.png}
        \caption{GT}
    \end{subfigure}
    \caption{Comparison of classification maps for the 3D-ConvSST-Bert model on the KSC dataset, showing different fusion methods: Cross Attention (CA), Concatenation (CONCAT), Multi-Head Attention (MHA), Pixel-Wise Addition (PWA), Pixel-Wise Multiplication (PWM), and Ground Truth (GT).}
    \vspace{-3mm}
    \label{fig:3D-ConvSST-Bert-KSC}
\end{figure*}

%%%%%%%%%%%%%%%%%%%%%%%%%%%%%%%% KSC DBCTNet-T5Encoder %%%%%%%%%%%%%%%%%%%%%%%%%%%%%%%%%%%%%%%%%%%%%%%%
\begin{figure*}[]
    \centering
       \begin{subfigure}[b]{0.16\textwidth}
            \centering
          \includegraphics[width=1\linewidth, trim=575 70 120 70, clip, angle=90, clip, angle=90]{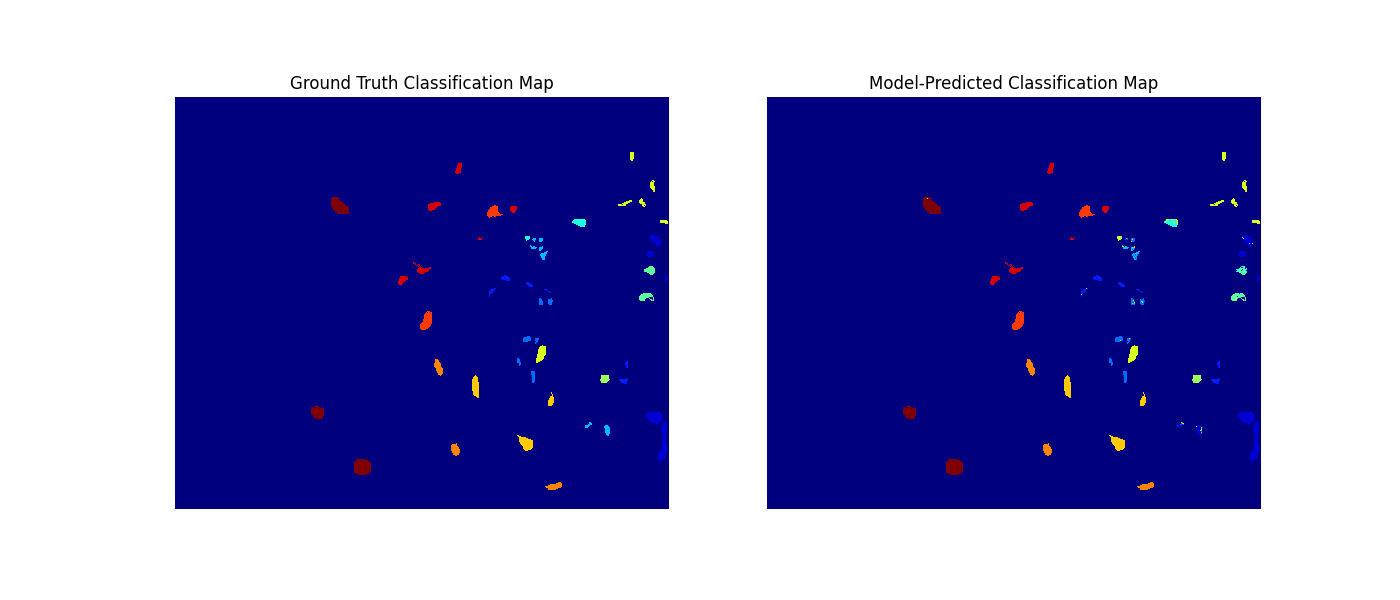}
        \caption{CA}
    \end{subfigure}
       \begin{subfigure}[b]{0.16\textwidth}
            \centering
          \includegraphics[width=1\linewidth, trim=575 70 120 70, clip, angle=90, clip, angle=90]{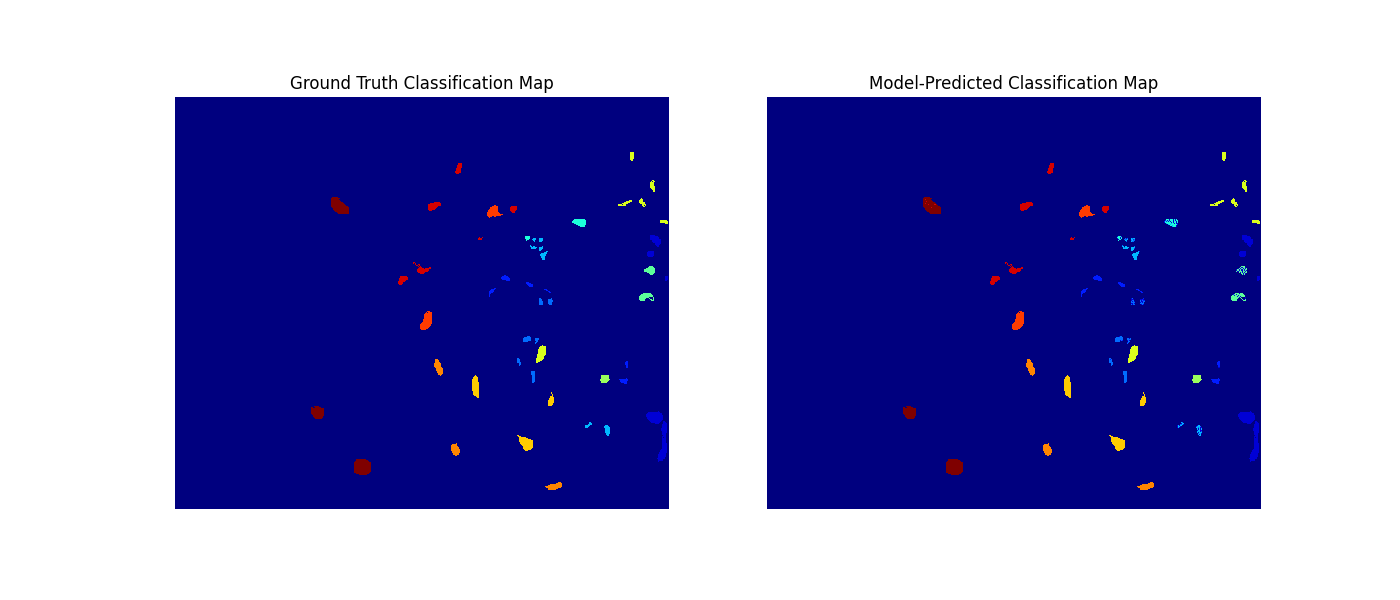}
        \caption{CONCAT}
    \end{subfigure}
       \begin{subfigure}[b]{0.16\textwidth}
            \centering
          \includegraphics[width=1\linewidth, trim=575 70 120 70, clip, angle=90, clip, angle=90]{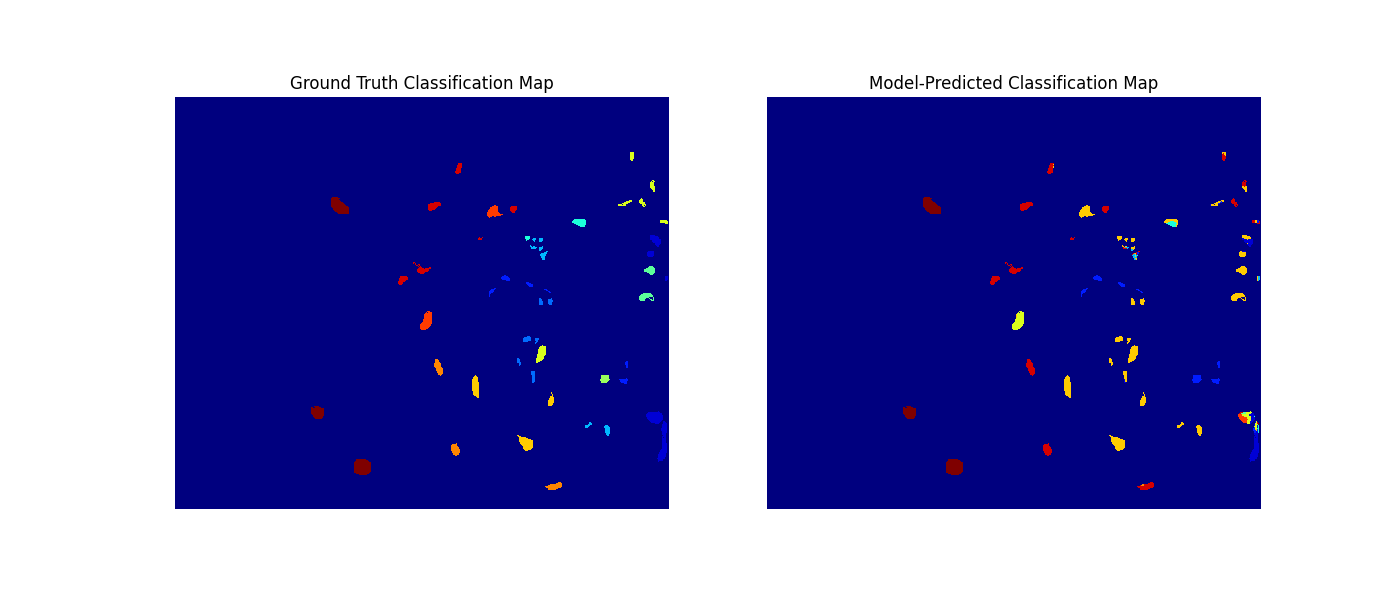}
        \caption{MHA}
    \end{subfigure}
       \begin{subfigure}[b]{0.16\textwidth}
            \centering
          \includegraphics[width=1\linewidth, trim=575 70 120 70, clip, angle=90, clip, angle=90]{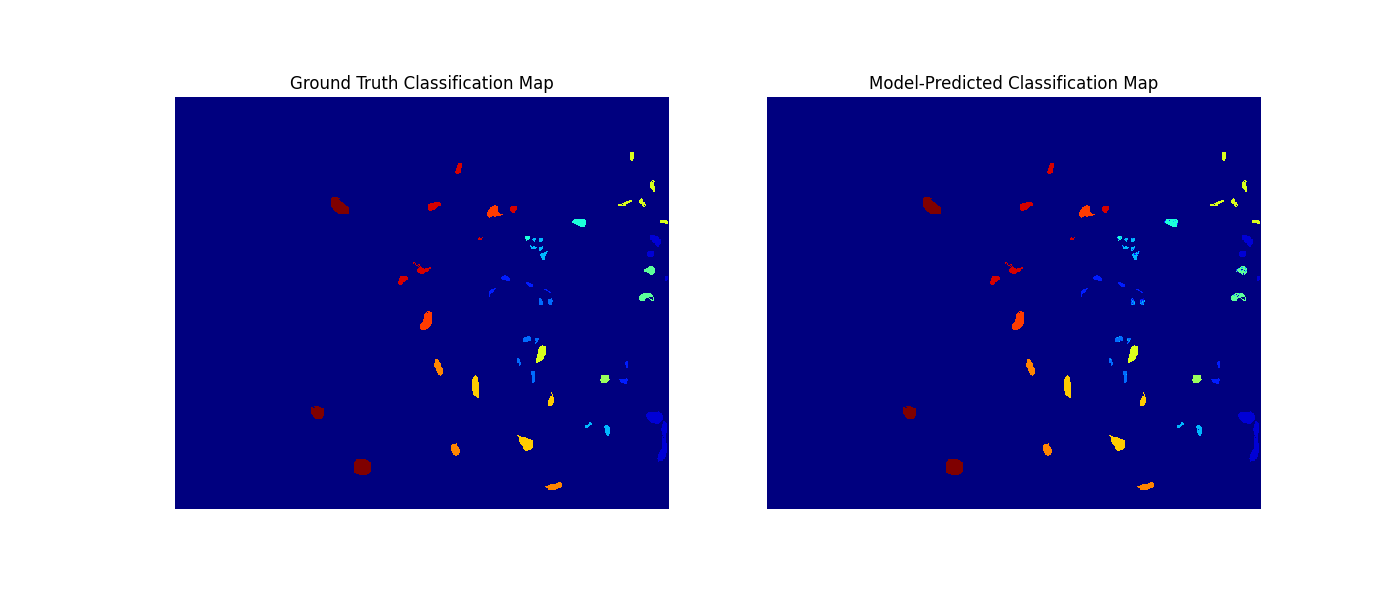}
        \caption{PWA}
    \end{subfigure}
       \begin{subfigure}[b]{0.16\textwidth}
            \centering
          \includegraphics[width=1\linewidth, trim=575 70 120 70, clip, angle=90, clip, angle=90]{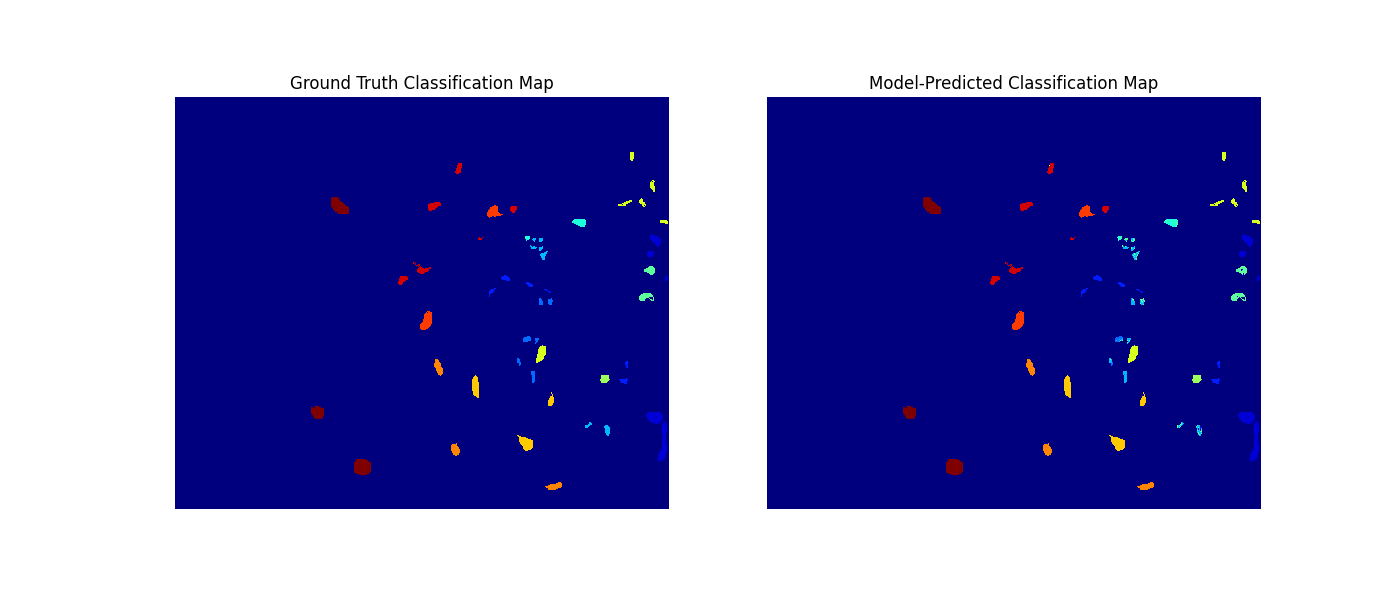}
        \caption{PWM}
    \end{subfigure}
       \begin{subfigure}[b]{0.16\textwidth}
            \centering
          \includegraphics[height=0.94\textwidth, width=1\linewidth, trim=130 70 540 70, clip, angle=180]{Output_Map/KSC/DBCTNet-T5Encoder_Large/PWM_prediction_map_run1.png}
        \caption{GT}
    \end{subfigure}
    \caption{Comparison of classification maps for the DBCTNet-T5 model on the KSC dataset, showing different fusion methods: Cross Attention (CA), Concatenation (CONCAT), Multi-Head Attention (MHA), Pixel-Wise Addition (PWA), Pixel-Wise Multiplication (PWM), and Ground Truth (GT).}
    \vspace{-3mm}
    \label{fig:DBCTNet-T5-KSC}
\end{figure*}

%%%%%%%%%%%%%%%%%%%%%%%%%%%%%%%% KSC DBCTNet-BertEncoder %%%%%%%%%%%%%%%%%%%%%%%%%%%%%%%%%%%%%%%%%%%%%%%%
\begin{figure*}[]
    \centering
       \begin{subfigure}[b]{0.16\textwidth}
            \centering
          \includegraphics[width=1\linewidth, trim=575 70 120 70, clip, angle=90, clip, angle=90]{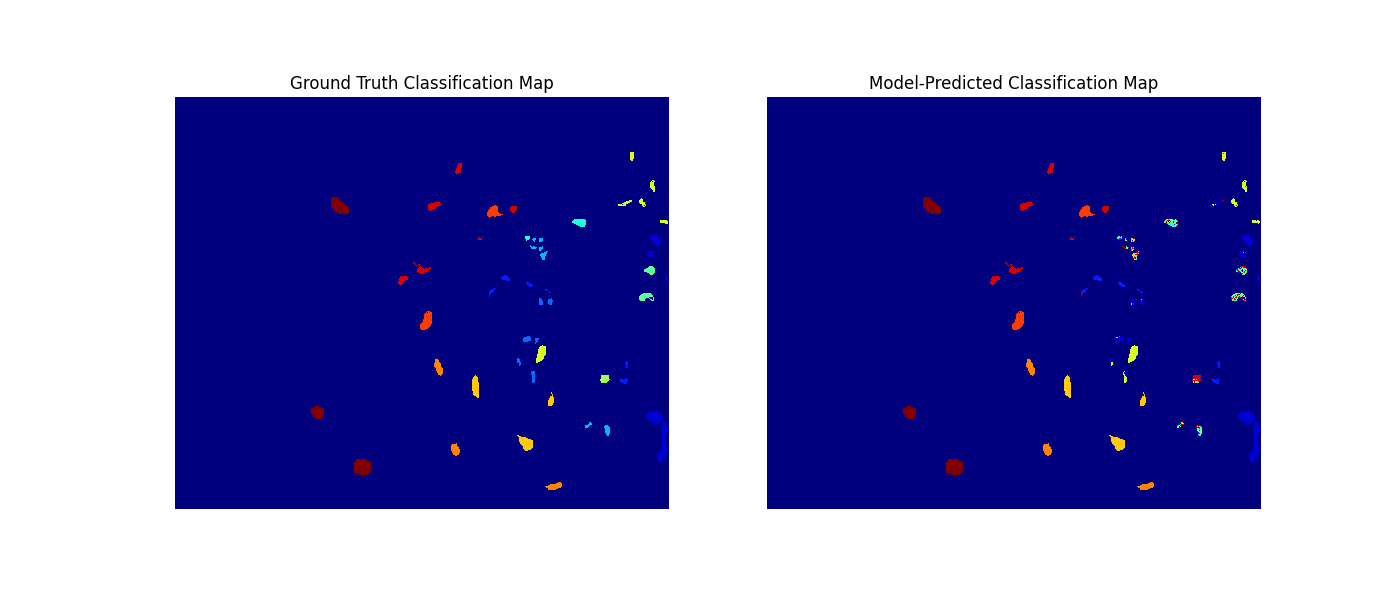}
        \caption{CA}
    \end{subfigure}
       \begin{subfigure}[b]{0.16\textwidth}
            \centering
          \includegraphics[width=1\linewidth, trim=575 70 120 70, clip, angle=90, clip, angle=90]{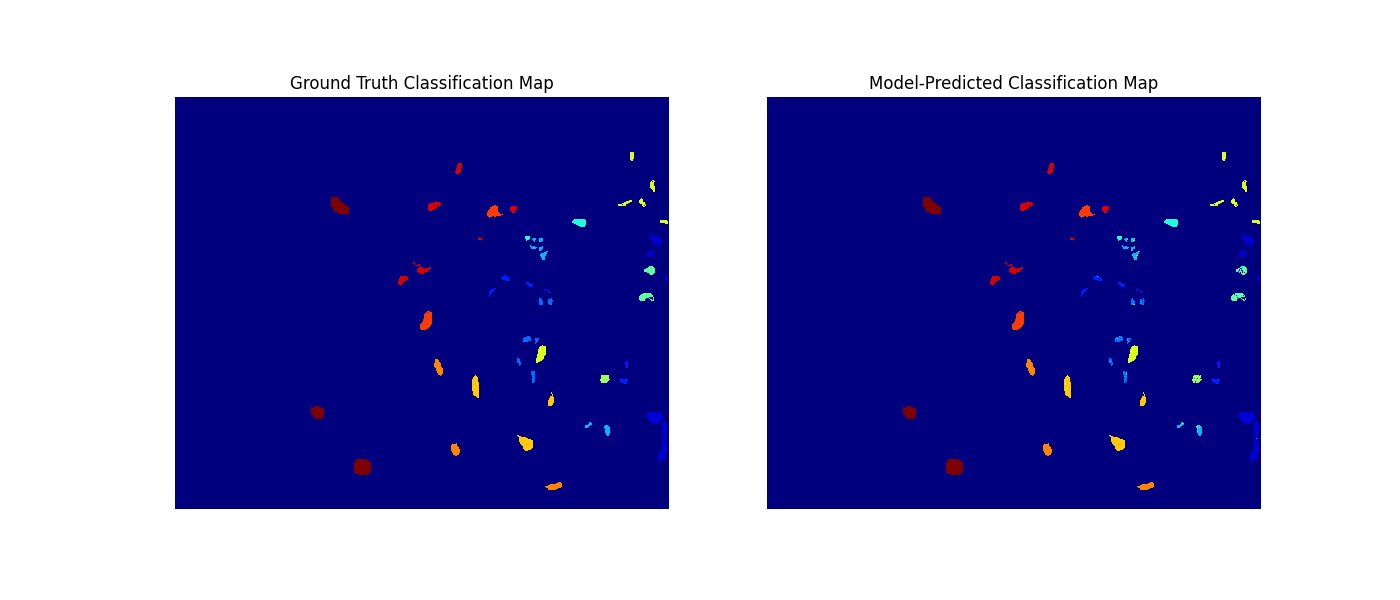}
        \caption{CONCAT}
    \end{subfigure}
       \begin{subfigure}[b]{0.16\textwidth}
            \centering
          \includegraphics[width=1\linewidth, trim=575 70 120 70, clip, angle=90, clip, angle=90]{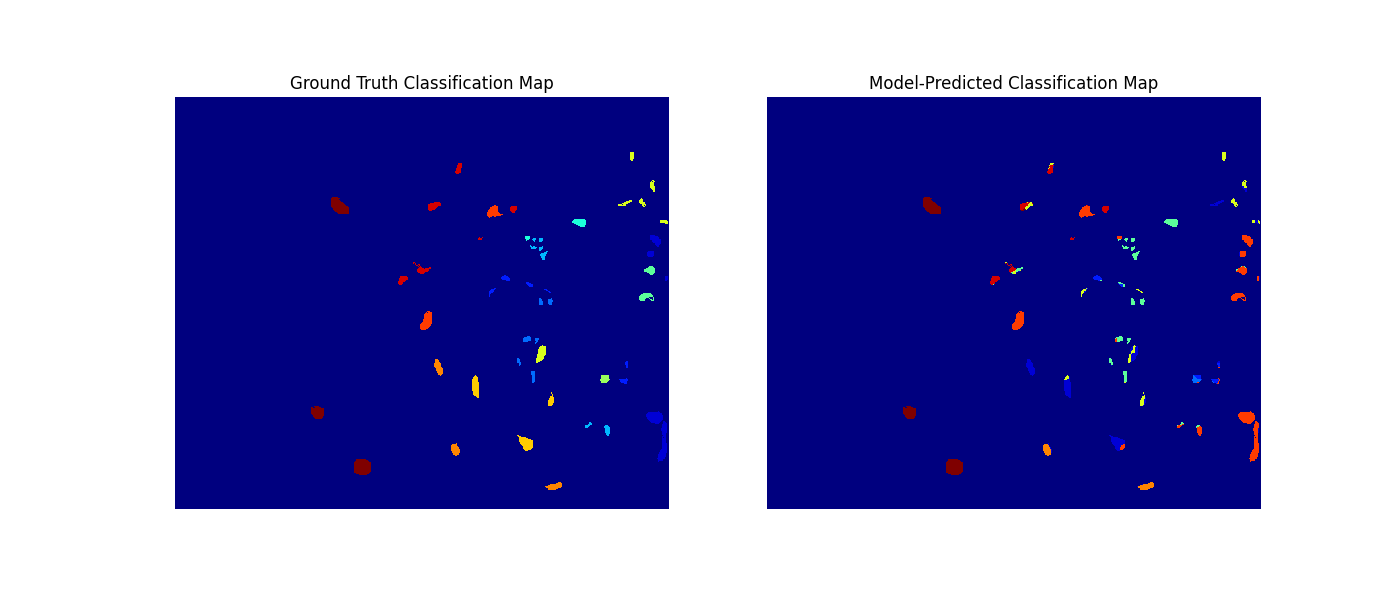}
        \caption{MHA}
    \end{subfigure}
       \begin{subfigure}[b]{0.16\textwidth}
            \centering
          \includegraphics[width=1\linewidth, trim=575 70 120 70, clip, angle=90, clip, angle=90]{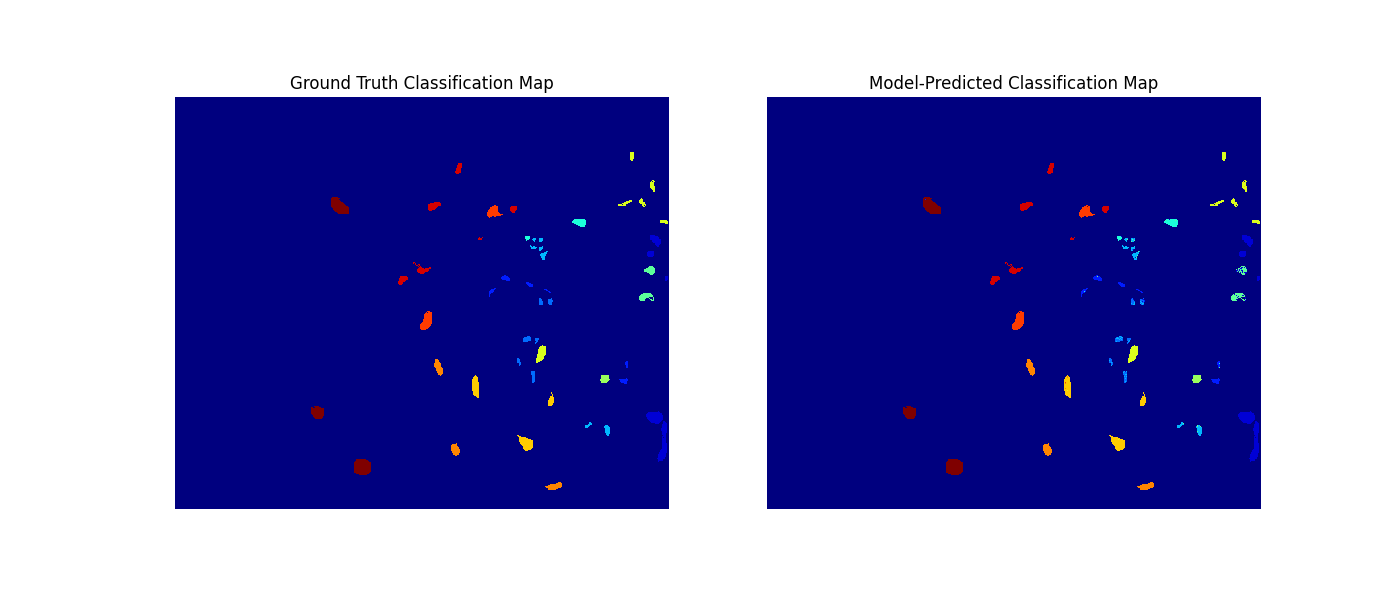}
        \caption{PWA}
    \end{subfigure}
       \begin{subfigure}[b]{0.16\textwidth}
            \centering
          \includegraphics[width=1\linewidth, trim=575 70 120 70, clip, angle=90, clip, angle=90]{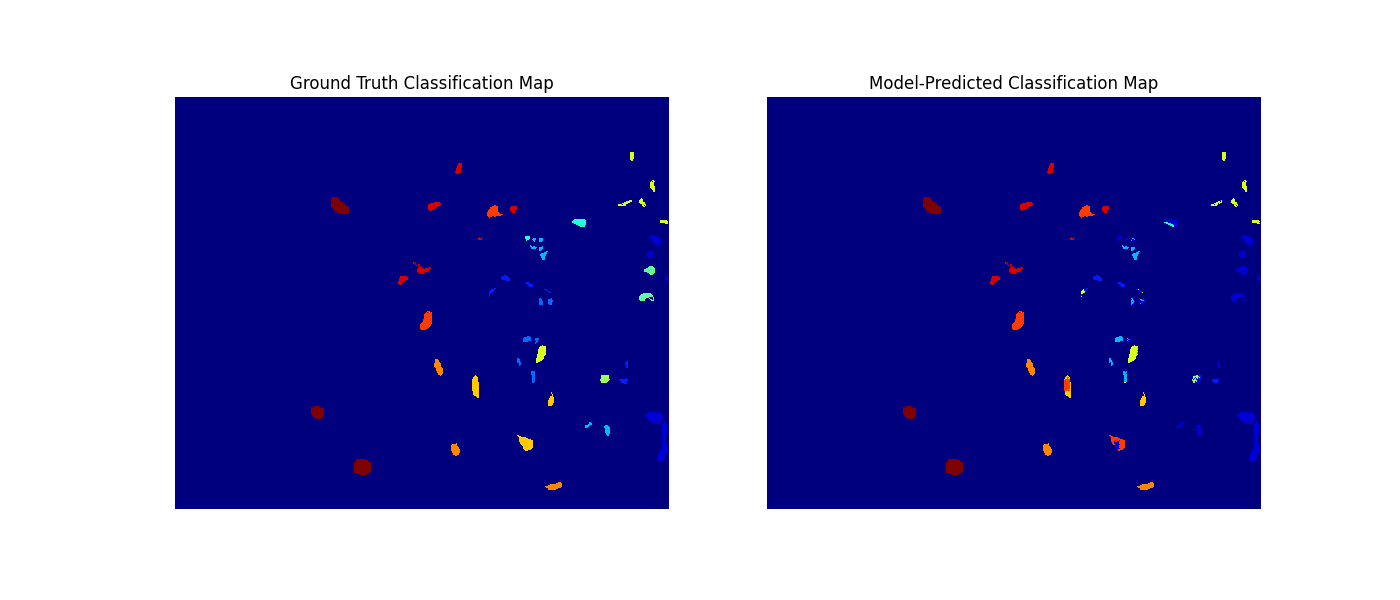}
        \caption{PWM}
    \end{subfigure}
       \begin{subfigure}[b]{0.16\textwidth}
            \centering
          \includegraphics[height=0.94\textwidth, width=1\linewidth, trim=130 70 540 70, clip, angle=180]{Output_Map/KSC/DBCTNet-BertEncoder_Large/PWM_prediction_map_run1.png}
        \caption{GT}
    \end{subfigure}
    \caption{Comparison of classification maps for the DBCTNet-Bert model on the KSC dataset, showing different fusion methods: Cross Attention (CA), Concatenation (CONCAT), Multi-Head Attention (MHA), Pixel-Wise Addition (PWA), Pixel-Wise Multiplication (PWM), and Ground Truth (GT).}
    \vspace{-3mm}
    \label{fig:DBCTNet-Bert-KSC}
\end{figure*}

%%%%%% KSC
\begin{figure*}[]
    \centering
       \begin{subfigure}[b]{0.16\textwidth}
            \centering
          \includegraphics[width=1\linewidth, trim=575 70 120 70, clip, angle=90, clip, angle=90]{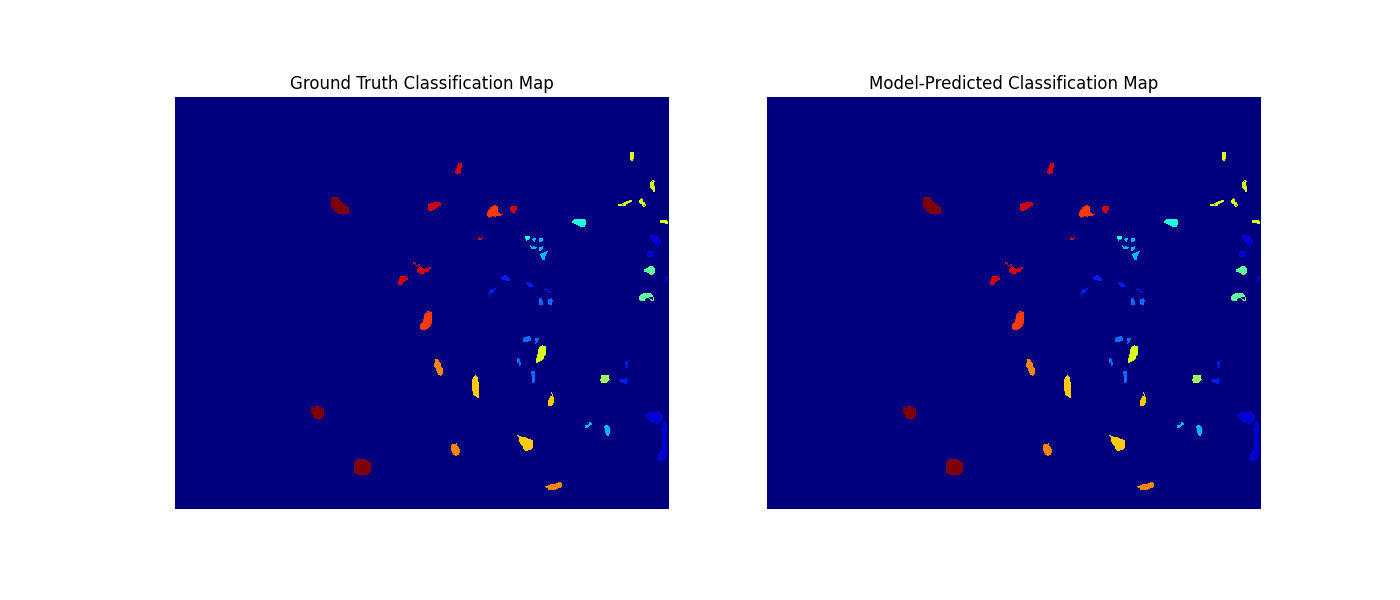}
        \caption{CA}
    \end{subfigure}
       \begin{subfigure}[b]{0.16\textwidth}
            \centering
          \includegraphics[width=1\linewidth, trim=575 70 120 70, clip, angle=90, clip, angle=90]{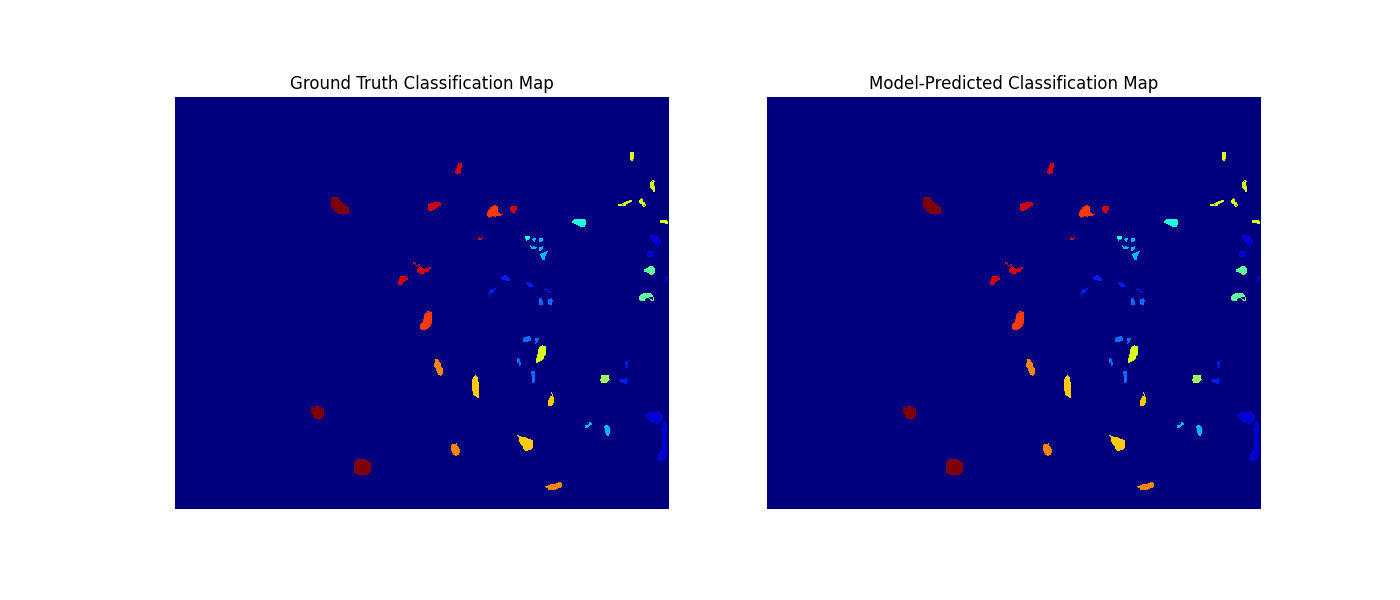}
        \caption{CONCAT}
    \end{subfigure}
       \begin{subfigure}[b]{0.16\textwidth}
            \centering
          \includegraphics[width=1\linewidth, trim=575 70 120 70, clip, angle=90, clip, angle=90]{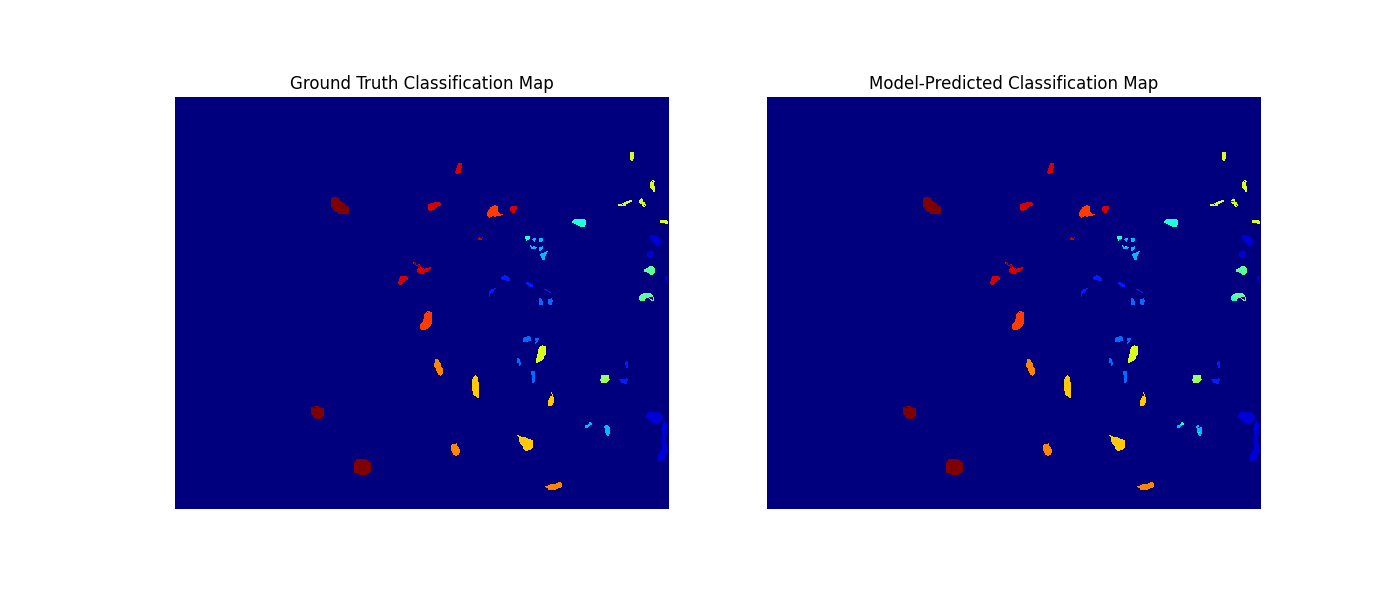}
        \caption{MHA}
    \end{subfigure}
       \begin{subfigure}[b]{0.16\textwidth}
            \centering
          \includegraphics[width=1\linewidth, trim=575 70 120 70, clip, angle=90, clip, angle=90]{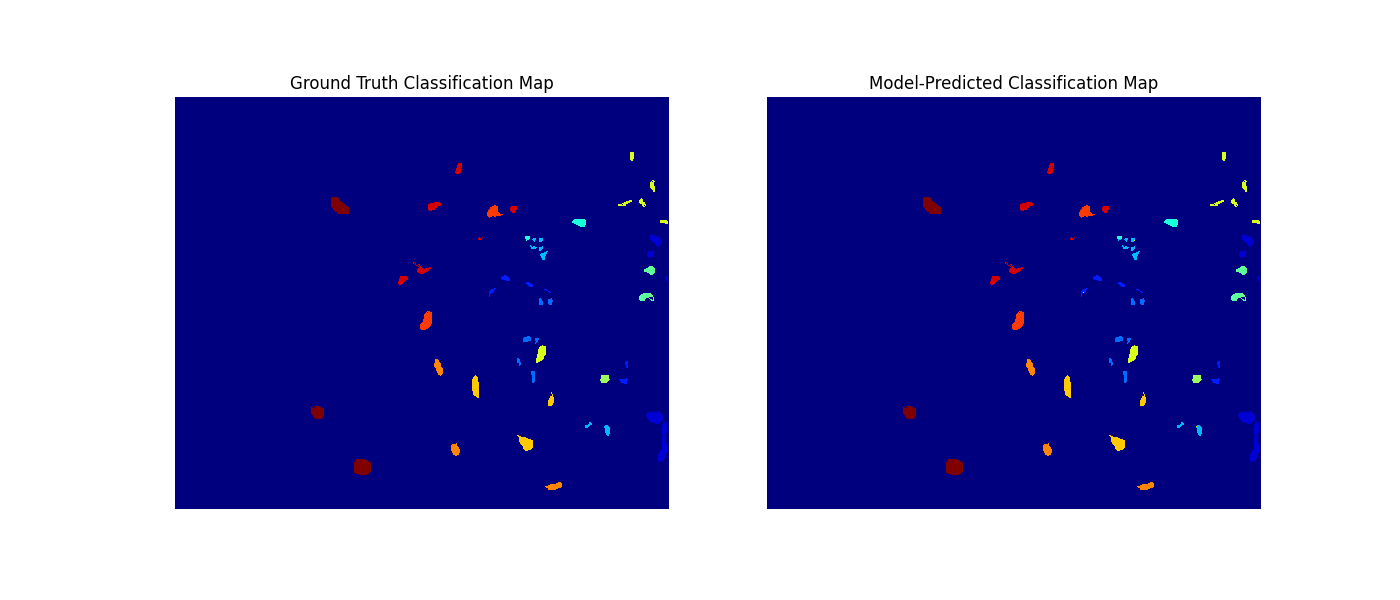}
        \caption{PWA}
    \end{subfigure}
       \begin{subfigure}[b]{0.16\textwidth}
            \centering
          \includegraphics[width=1\linewidth, trim=575 70 120 70, clip, angle=90, clip, angle=90]{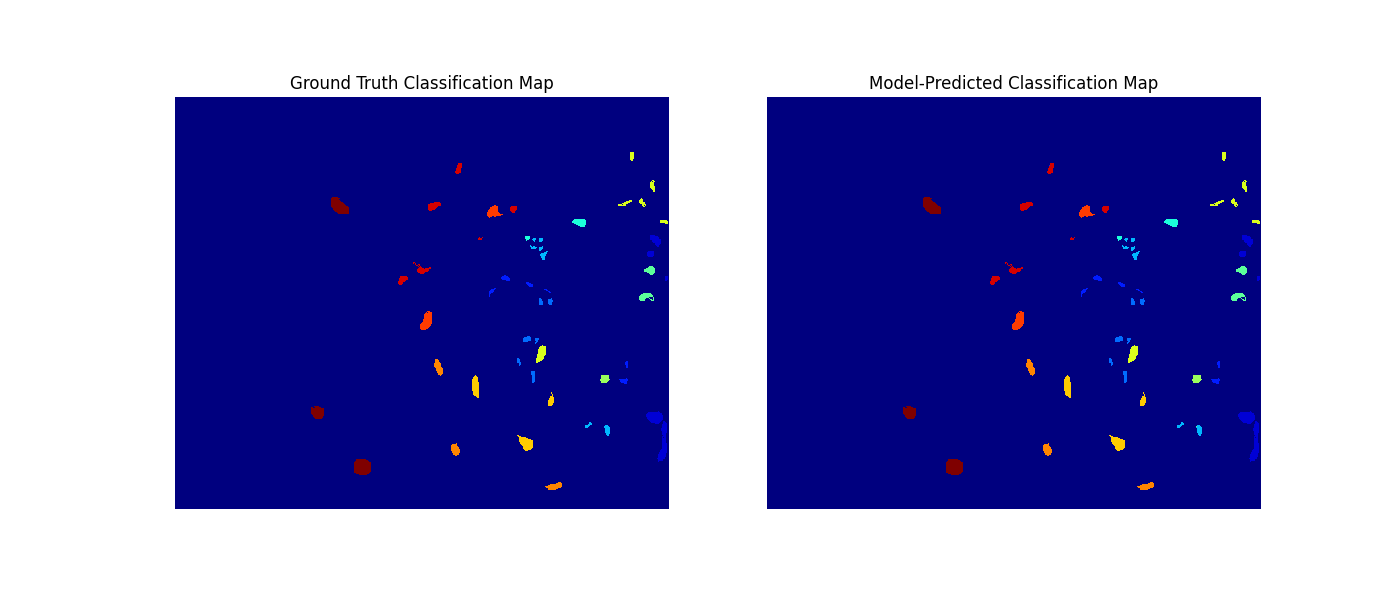}
        \caption{PWM}
    \end{subfigure}
       \begin{subfigure}[b]{0.16\textwidth}
            \centering
          \includegraphics[height=0.94\textwidth, width=1\linewidth, trim=130 70 540 70, clip, angle=180]{Output_Map/KSC/FAHM-T5Encoder_Large/PWM_prediction_map_run1.png}
        \caption{GT}
    \end{subfigure}
    \caption{Comparison of classification maps for the FAHM-T5 model on the Kennedy Space Center dataset, showing different fusion methods: Cross Attention (CA), Concatenation (CONCAT), Multi-Head Attention (MHA), Pixel-Wise Addition (PWA), Pixel-Wise Multiplication (PWM), and Ground Truth (GT).}
    \vspace{-3mm}
    \label{fig:FAHM-T5-KSC}
\end{figure*}

%%%%%%%%%%%%%%%%%%%%%%%%%%%%%%%%%%%%%%%%%%%%%%%%%%%%%%%%%%%%%%%%%%%%%%%%%%%%%%%%% KSC FAHM-BertEncoder %%%%%%%%%%%%%%%%%%%%%%%%%%%%%%%%%%%%%%%%%%%%%%%%
\begin{figure*}[]
    % \vspace{-30mm}
    \centering
       \begin{subfigure}[b]{0.16\textwidth}
            \centering
          \includegraphics[width=1\linewidth, trim=575 70 120 70, clip, angle=90, clip, angle=90]{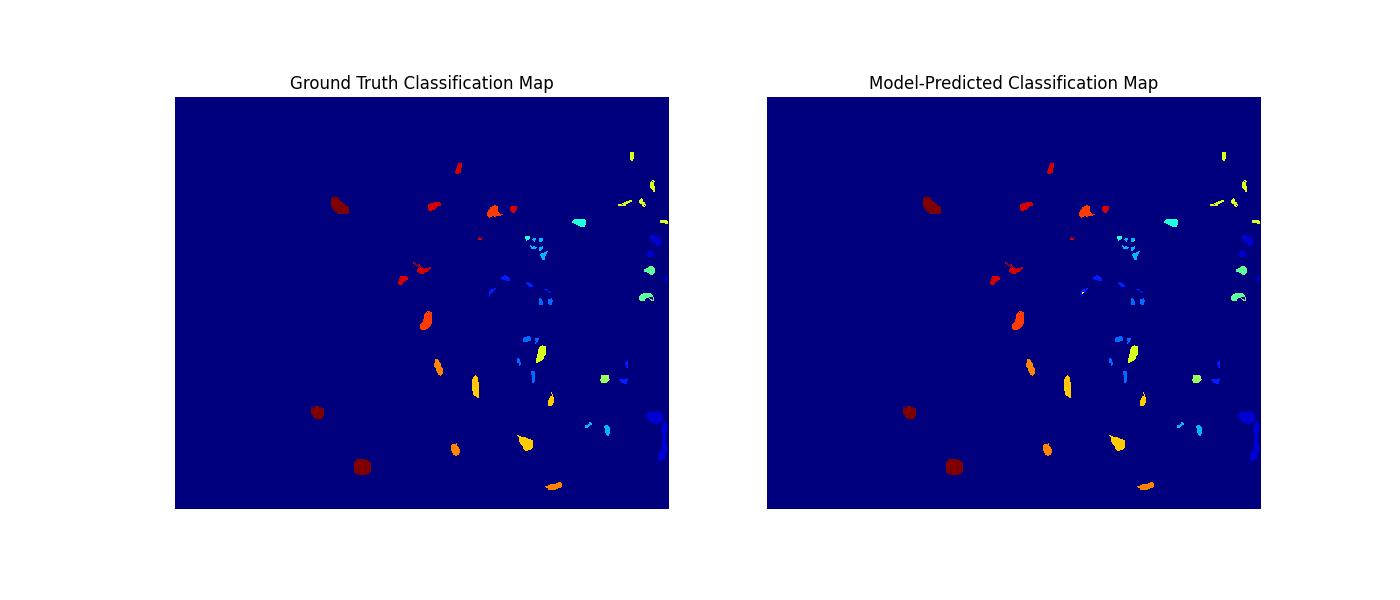}
        \caption{CA}
    \end{subfigure}
       \begin{subfigure}[b]{0.16\textwidth}
            \centering
          \includegraphics[width=1\linewidth, trim=575 70 120 70, clip, angle=90, clip, angle=90]{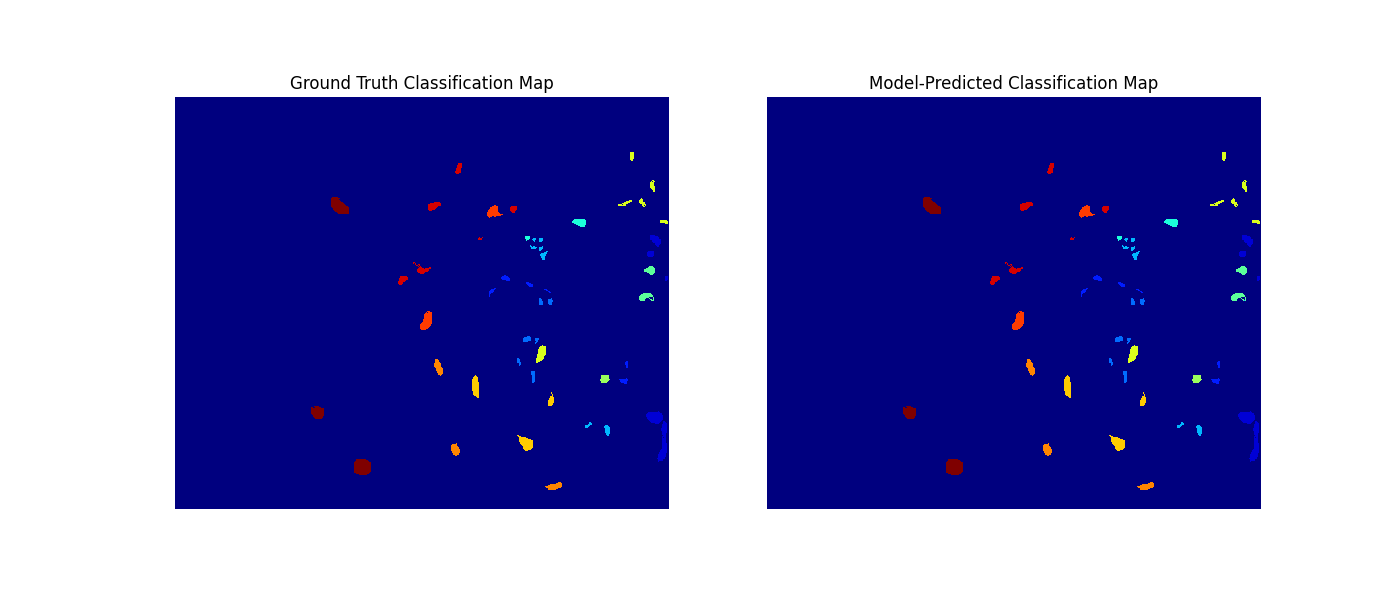}
        \caption{CONCAT}
    \end{subfigure}
       \begin{subfigure}[b]{0.16\textwidth}
            \centering
          \includegraphics[width=1\linewidth, trim=575 70 120 70, clip, angle=90, clip, angle=90]{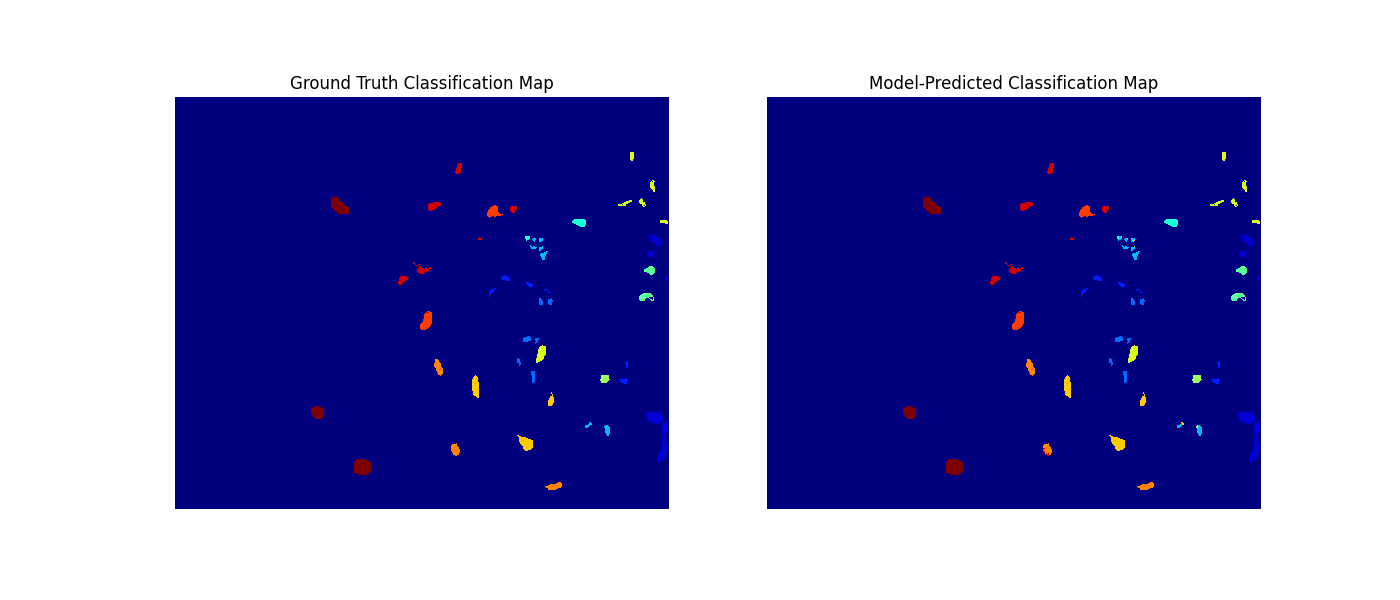}
        \caption{MHA}
    \end{subfigure}
       \begin{subfigure}[b]{0.16\textwidth}
            \centering
          \includegraphics[width=1\linewidth, trim=575 70 120 70, clip, angle=90, clip, angle=90]{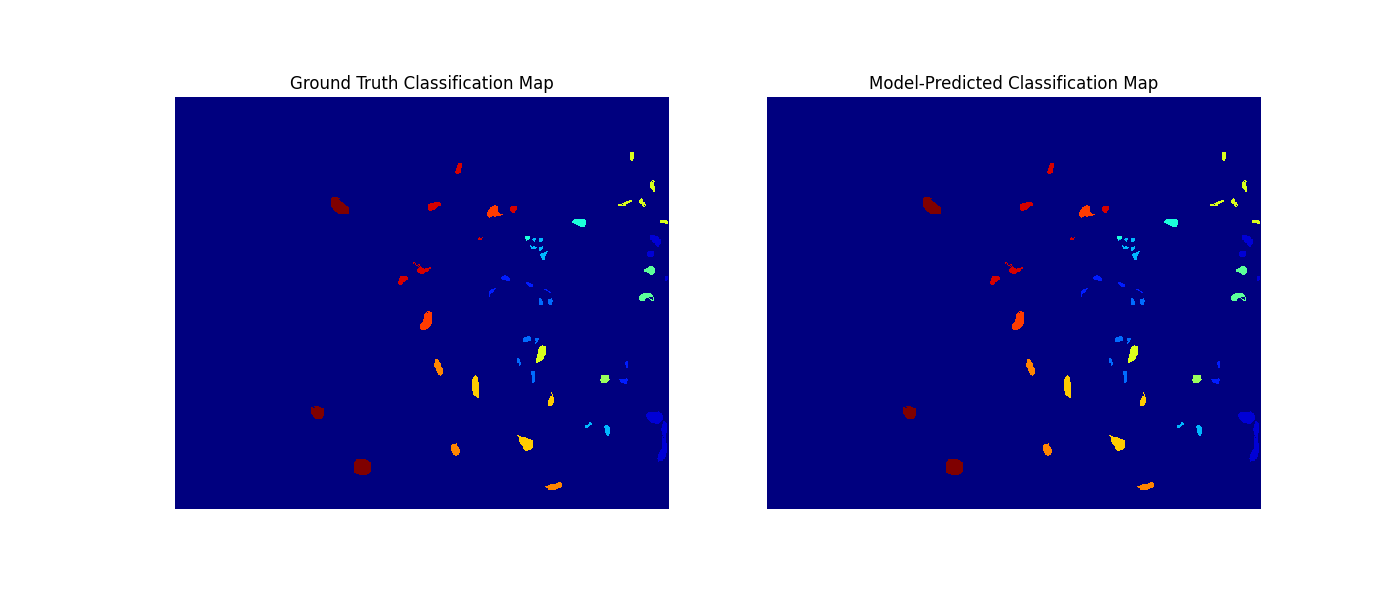}
        \caption{PWA}
    \end{subfigure}
       \begin{subfigure}[b]{0.16\textwidth}
            \centering
          \includegraphics[width=1\linewidth, trim=575 70 120 70, clip, angle=90, clip, angle=90]{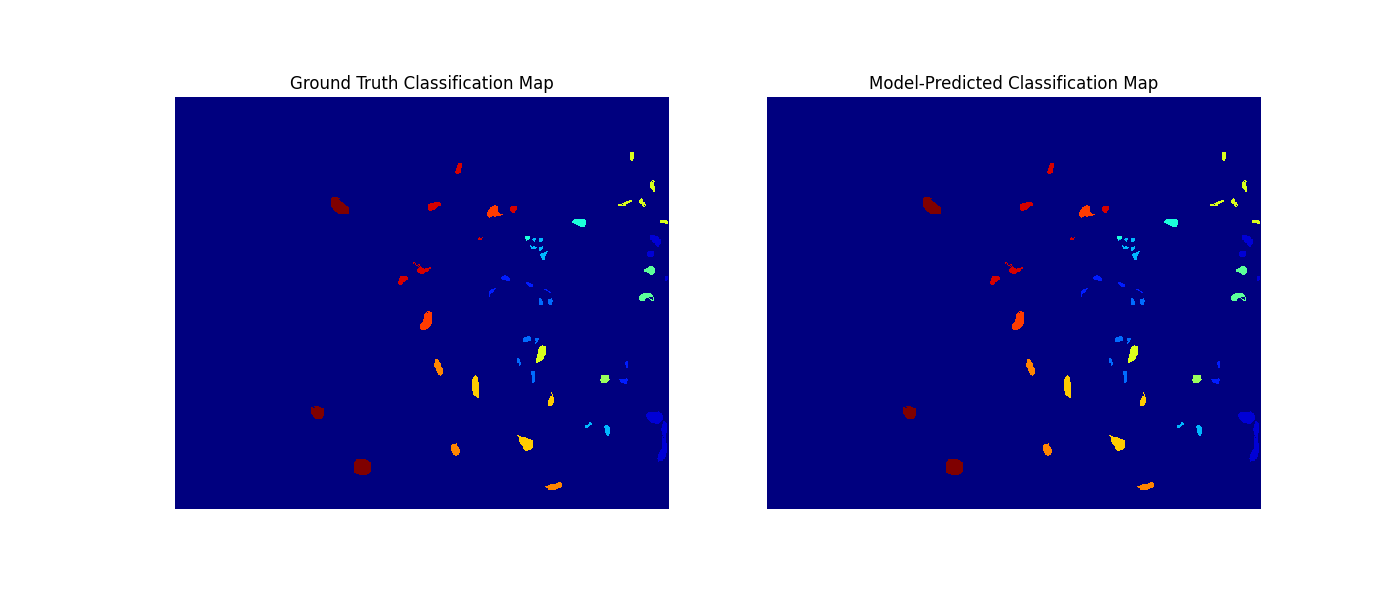}
        \caption{PWM}
    \end{subfigure}
       \begin{subfigure}[b]{0.16\textwidth}
            \centering
          \includegraphics[height=0.94\textwidth, width=1\linewidth, trim=130 70 540 70, clip, angle=180]{Output_Map/KSC/FAHM-BertEncoder_Large/PWM_prediction_map_run1.png}
        \caption{GT}
    \end{subfigure}
    \caption{Comparison of classification maps for the FAHM-Bert model on the KSC dataset, showing different fusion methods: Cross Attention (CA), Concatenation (CONCAT), Multi-Head Attention (MHA), Pixel-Wise Addition (PWA), Pixel-Wise Multiplication (PWM), and Ground Truth (GT).}
    \vspace{-3mm}
    \label{fig:FAHM-Bert-KSC}
\end{figure*}

KSC is the most challenging, with original vision-only maps displaying severe noise and large segmentation mistakes—especially in fine or marshy classes. Introducing text encoders noticeably improves recovery of rare land types and sharpens complex borders in visuals across all models, as seen in the referenced figures (\ref{fig:3D-RCNet-T5-KSC}, \ref{fig:3D-RCNet-Bert-KSC}, \ref{fig:3D-ConvSST-T5-KSC}, \ref{fig:3D-ConvSST-Bert-KSC}, \ref{fig:DBCTNet-T5-KSC}, \ref{fig:DBCTNet-Bert-KSC}, \ref{fig:FAHM-T5-KSC}, \ref{fig:FAHM-Bert-KSC}). Attention mechanisms (MHA, CA) and concatenation provide the most stable noise suppression, with visual results approaching ground-truth quality for FAHM and robust improvements for DBCTNet and 3D-RCNet. Pixel-wise addition/multiplication sometimes peak in visual fidelity for DBCTNet, but are more variable overall. Figures for FAHM (Bert/T5) demonstrate near-saturation, echoing the strongest quantitative results with visually smooth, sharply segmented land cover maps.

% ============================================================================
% CONCLUSION
% ============================================================================

Overall, the maps confirm three practical takeaways. First, adding text encoders is the main driver of qualitative improvement, turning noisy vision-only outputs into clean, GT-like maps. Second, attention-based (CA/MHA) and concatenation fusion are the most consistently strong choices across datasets and backbones; they reliably reduce speckle and preserve boundaries. Third, pixel-wise fusions (PWA/PWM) are not universally inferior—on some backbone–dataset pairs (notably DBCTNet on Indian Pines and KSC), they can be the top performer—but their success is more configuration-dependent. In short, for dependable map quality across settings, prefer CA/MHA or CONCAT

\end{document}